\documentclass{siamart1116}

\def\bbbr{{\mathbb R}} 

\newcommand{\argmaxlocal}{\operatorname{argmaxlocal}}

\usepackage{amsfonts}
\usepackage{graphicx}
\ifpdf
  \DeclareGraphicsExtensions{.eps,.pdf,.png,.jpg}
\else
  \DeclareGraphicsExtensions{.eps}
\fi

\newcommand{\TheTitle}{Dense scale selection over space, time and space-time} 
\newcommand{\TheAuthors}{T. Lindeberg}

\headers{\TheTitle}{\TheAuthors}

\title{{\TheTitle}\thanks{To appear in SIAM Journal on Imaging
    Sciences. Submitted to the editors Sep 22,
    2017. Revised Nov 17, 2017. Accepted Nov 27, 2017.
\funding{This work was funded by the Swedish Research Council 
              under contract no.~2014-4083.}}}

\author{
  Tony Lindeberg\thanks{Computational Brain Science Lab,
        Department of Computational Science and Technology,
        School of Computer Science and Communication,
        KTH Royal Institute of Technology,
        SE-100 44 Stockholm, Sweden 
    (\email{tony@kth.se}, \url{https://www.kth.se/profile/tony}).}
}

\begin{document}

\maketitle

% REQUIRED
\begin{abstract}
Scale selection methods based on local extrema over scale of
scale-normalized derivatives have been primarily developed 
to be applied sparsely --- at image points where the magnitude of a
scale-normalized differential expression additionally assumes local
extrema over the domain where the data are defined.
This paper presents a methodology for performing dense scale
selection, so that hypotheses about local characteristic scales
in images, temporal signals and video can be computed
at every image point and every time moment.
A critical problem when designing mechanisms for dense scale selection
is that the scale at which scale-normalized differential entities
assume local extrema over scale can be strongly dependent on the local
order of the locally dominant differential structure.
To address this problem, we propose a methodology where local extrema
over scale are detected of a quasi quadrature measure involving
scale-space derivatives up to order two
and propose two independent mechanisms to reduce the phase dependency
of the local scale estimates by:
(i)~introducing a second layer of post-smoothing prior to the
detection of local extrema over scale and
(ii)~performing local phase compensation based on a model
of the phase dependency of the local scale estimates depending on the
relative strengths between first- {\em vs.\/}\ second-order differential structure.
This general methodology is applied over three types of domains:
(i)~spatial images, (ii)~temporal signals and (iii)~spatio-temporal video.
Experiments demonstrate that the proposed methodology leads to intuitively
reasonable results with local scale estimates that reflect variations
in the characteristic scales of locally dominant structures over space
and time.
\end{abstract}

% REQUIRED

% REQUIRED

\section{Introduction}

The notion of scale is essential when computing features from image
data for purposes in biological or artificial visual perception.
Results from biological research regarding the early visual areas in
the LGN and V1 (Hubel and Wiesel \cite{HubWie59-Phys,HubWie62-Phys,HubWie05-book};
DeAngelis {\em et al.\/}\
\cite{DeAngOhzFre95-TINS,deAngAnz04-VisNeuroSci})
as well as theoretical results from normative theory of visual
operations (Lindeberg \cite{Lin13-BICY,Lin16-JMIV}) 
based on scale-space theory
(Iijima \cite{Iij62}; Witkin \cite{Wit83}; Koenderink \cite{Koe84-BC,Koe88-BC};
 Koenderink and van Doorn \cite{KoeDoo87-BC,KoeDoo92-PAMI};
 Lindeberg \cite{Lin93-Dis,Lin10-JMIV};
 Florack \cite{Flo97-book}; 
 Sporring {\em et al.\/}\ \cite{SpoNieFloJoh96-SCSPTH};
 Weickert {\em et al.\/}\ \cite{WeiIshImi99-JMIV};
 ter Haar Romeny {\em et al.\/}\ \cite{RomFloNie01-SCSP,Haa04-book}) state that local image
 measurements in terms of receptive fields constitute a both natural
 and efficient model for expressing early visual operations.

When applying such spatial or spatio-temporal receptive fields at
multiple spatial and temporal scales, a basic observation that one can
make is that the responses that are obtained from the receptive field
operators can be strongly dependent on the scale levels at which they
are applied.
Thus, one may raise the problem whether it is possible from the data
itself to generate hypotheses about local appropriate scales in the
image data, so as to adapt subsequent processing to the local image structures.
Initially, this problem could possibly be seen as intractable. 
Would it at all be possible to generate hypotheses about interest
scale levels before recognizing the objects we are interested in or
defining the specific purpose for which the scale estimates are to be used?
Research in (Lindeberg \cite{Lin93-Dis,Lin97-IJCV,Lin98-IJCV}) with
follow-up work by several authors 
(Bretzner {\em et al.\/}\ \cite{BL97-CVIU,BreLapLin02-FG};
Chomat {\em et al.\/}\ \cite{ChoVerHalCro00-ECCV};
Lowe \cite{Low04-IJCV};
Mikolajczyk and Schmid \cite{MikSch04-IJCV};
Lazebnik {\em et al.\/}\ \cite{LazSchPon05-PAMI};
Rothganger {\em et al.\/}\ \cite{RotLazSchPon06-IJCV};
Bay {\em et al.\/}\ \cite{BayEssTuyGoo08-CVIU};  
Tuytelaars and Mikolajczyk \cite{TuyMik08-Book};
Negre {\em et al.\/}\ \cite{NegBraCroLau08-ExpRob};
Lindeberg \cite{Lin12-JMIV,Lin15-JMIV})
has, however, demonstrated that such an approach is feasible
(see \cite{Lin99-CVHB,Lin14-EncCompVis} for overviews).
A general framework for 
automatic selection of local characteristic scales can be formulated based on the detection of
local extrema over scale of scale-normalized feature responses.
Specifically, such scale estimates transform in a scale-covariant way
under spatial scaling transformations of the image domain, which is a
highly desirable property of a scale selection mechanism, since it
implies that the scale estimates will automatically follow local scale
variations in the image data.
% selection leads to scale invariance in
%the sense that the local scale estimates transform in a
%scale-covariant way under spatial scaling transformations of the image
%domain.
Corresponding local scale selection mechanisms can also be expressed
over temporal and spatio-temporal domains (Lindeberg \cite{Lin17-JMIV,Lin17-SSVM,Lin17-JMIV-subm}).

A common property of most of the successful applications of scale
selection to computer vision applications, however, is that the scale
selection method is applied sparsely over the image domain, most
commonly at interest points.
% \cite{Lin97-IJCV,BreLapLin02-FG,Low04-IJCV,MikSch04-IJCV,LazSchPon05-PAMI,RotLazSchPon06-IJCV,BayEssTuyGoo08-CVIU,TuyMik08-Book,NegBraCroLau08-ExpRob,Lin12-JMIV,Lin15-JMIV}. 
If attempting to perform dense scale selection based on the two most
common rotationally invariant differential invariants for interest
point detection, the spatial Laplacian or the determinant of the
spatial Hessian, then the results of scale selection will usually not
be stable or useful far away from the interest points.

To address this problem, an initial mechanism for dense scale
selection was proposed in (Lindeberg \cite{Lin97-IJCV}) based on a
the detection of local extrema over scale of a 
spatial quasi quadrature measure that constitutes a rotationally
invariant measure of the amount of energy in the first- and
second-order differential structure of a spatial image.
Modifications of this approach were used by
Almansa and Lindeberg \cite{AL00-IP}  for estimating the local scale
of fingerprint patterns for fingerprint recognition and were
specifically shown to improve the quality of minutiae extraction.
Related methods for scale selection have been developed by  
Kadir and Brady \cite{KadBra01-IJCV} and 
Sporring {\em et al.\/}\ \cite{SpoCoilTra00-ICIP})
by detecting peaks of weighted entropy measures or Lyaponov
functionals over scale,
by minimizing normalized error measures over scale 
(Lindeberg \cite{Lin97-IVC}),
by determining local scales for variable bandwidth mean
shift from the scale bandwidth that maximizes the norm of the
normalized mean shift vector
(Comaniciu {\em et al.\/}\ \cite{ComRamMee01-ICCV}),
by detecting maxima of steered energy responses over scales 
(Ng and Bharath \cite{NgBha04-ECCV}),
by comparing reliability measures from statistical classifiers for texture analysis at multiple
scales (Kang {\em et al.\/}\ \cite{KanMorNag05-ScSp}),
by measuring the size variations of the regions that pixels belong
to under total variation (TV) flow (Brox and Weickert \cite{BroWei06-JVCIR}),
by measuring local oscillations in signals
(Jones and Le \cite{JonLe09-ApplCompHarmAnal}) or
by computing image segmentations from the scales at which a
supervised classifier delivers class labels with the highest reliability measure
(Loog {\em et al.\/}\ \cite{LooLiTax09-LNCS}; Li {\em et al.\/} \cite{LiTaxLoo12-IVC}). 
Specifically, a more algorithmic way of generating scale estimates for
image matching away
from the locations of interest points was recently proposed by 
Hassner {\em et al.\/}\ \cite{HasFilMayZel17-PAMI}
and Tau and Hassner \cite{TauHas16-PAMI} by considering subspaces generated by local image descriptors
computed over multiple scales to improve the performance of stereo matching.

Closely related issues of estimating dominant scales in signals have been studied in
wavelet theory and local frequency analysis (Cohen \cite{Coh95-book}). 
For example, a local Gaussian-weighted windowed Fourier transform
of a 1-D signal corresponds to filtering with Gabor functions \cite{Gab46}. 
Mallat and Hwang \cite{MalHwa92-IT} proposed to characterize singularities in terms of
Lipshitz exponents and detected maxima in the wavelet transform.
Pure wavelet and/or local frequency based methods have, however, been
less developed for 2-D image data or 2+1-D video data.

The subject of this article is to perform a deeper study into the
problem of dense scale selection for images and video. A basic problem that can be observed
if performing dense scale selection based on the basic quasi
quadrature measure in \cite{Lin97-IJCV} is that the local scale
estimates can be strongly phase dependent. %For example, 
If applied to a sine wave in one or two dimensions, the scale estimates can be
biased depending on the relative strength of first-order {\em vs.\/}\
second-order differential structure. To reduce this phase dependency,
we will consider two independent mechanisms in terms of (i)~spatial
smoothing and (ii)~local phase compensation. We will specifically
analyze the properties of these mechanisms, determine free parameters in
the corresponding methods and show that these mechanisms may
reduce the local phase dependency substantially.  We will also
generalize this dense scale selection mechanism to the spatio-temporal
domain, to perform simultaneous dense scale selection of both local
spatial and temporal scales. Compared to wavelet-based approaches for
local frequency analysis, the methods that we propose are invariant to
rotations over the spatial domain. Additionally, we prove that both
the spatial and the temporal scale estimates are provably covariant
under independent scaling transformations of the spatial and the
temporal domains, implying that the local scale estimates are
guaranteed to automatically follow local variations in the spatial
extent and the temporal duration of spatial, temporal or
spatio-temporal image structures, which is a highly desirable property
of a scale selection mechanism. Experiments on different types of
spatial images, temporal signals and spatio-temporal video demonstrate
that the proposed theory leads to dense spatial and temporal scale maps with
intuitively reasonable properties.

\section{Dense spatial scale selection over a purely spatial domain}
\label{sec-spat-dense-scsel}

\begin{figure*}[hbt]
  \begin{center}
    \begin{tabular}{cccccc}
     \hspace{-3mm} {\footnotesize\em original image} 
       & {\footnotesize $|\nabla^2 L|$} 
       &  {\footnotesize $\sqrt{|\det {\cal H} L|}$} 
       & {\footnotesize $\sqrt{{\cal Q}_{(x,y),1} L}$} 
       & {\footnotesize $\sqrt{{\cal Q}_{(x,y),2} L}$}
       & {\footnotesize $\sqrt{{\cal Q}_{(x,y)} L}$} \\
    \hspace{-3mm} \includegraphics[width=0.13\textwidth]{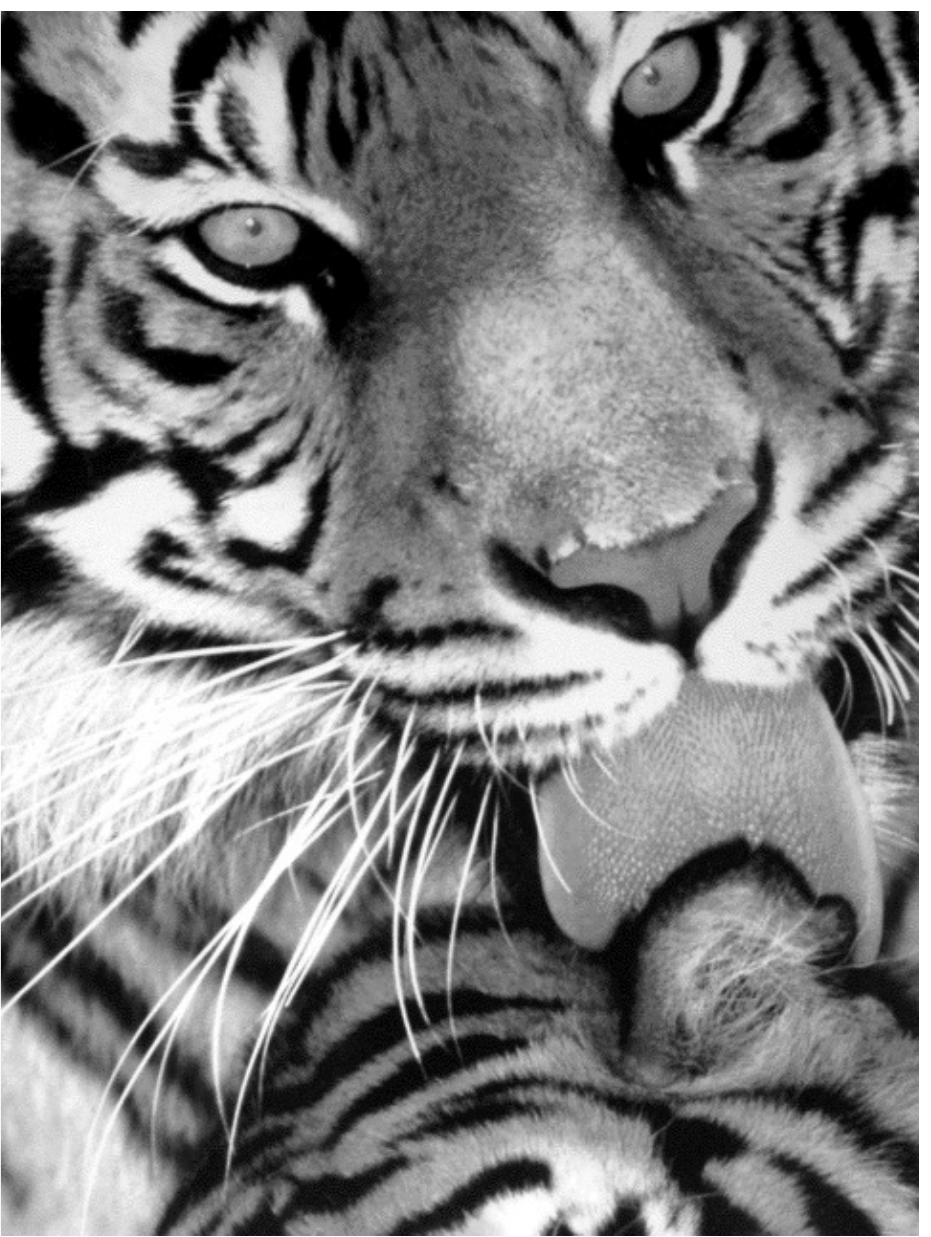} 
       & \includegraphics[width=0.13\textwidth]{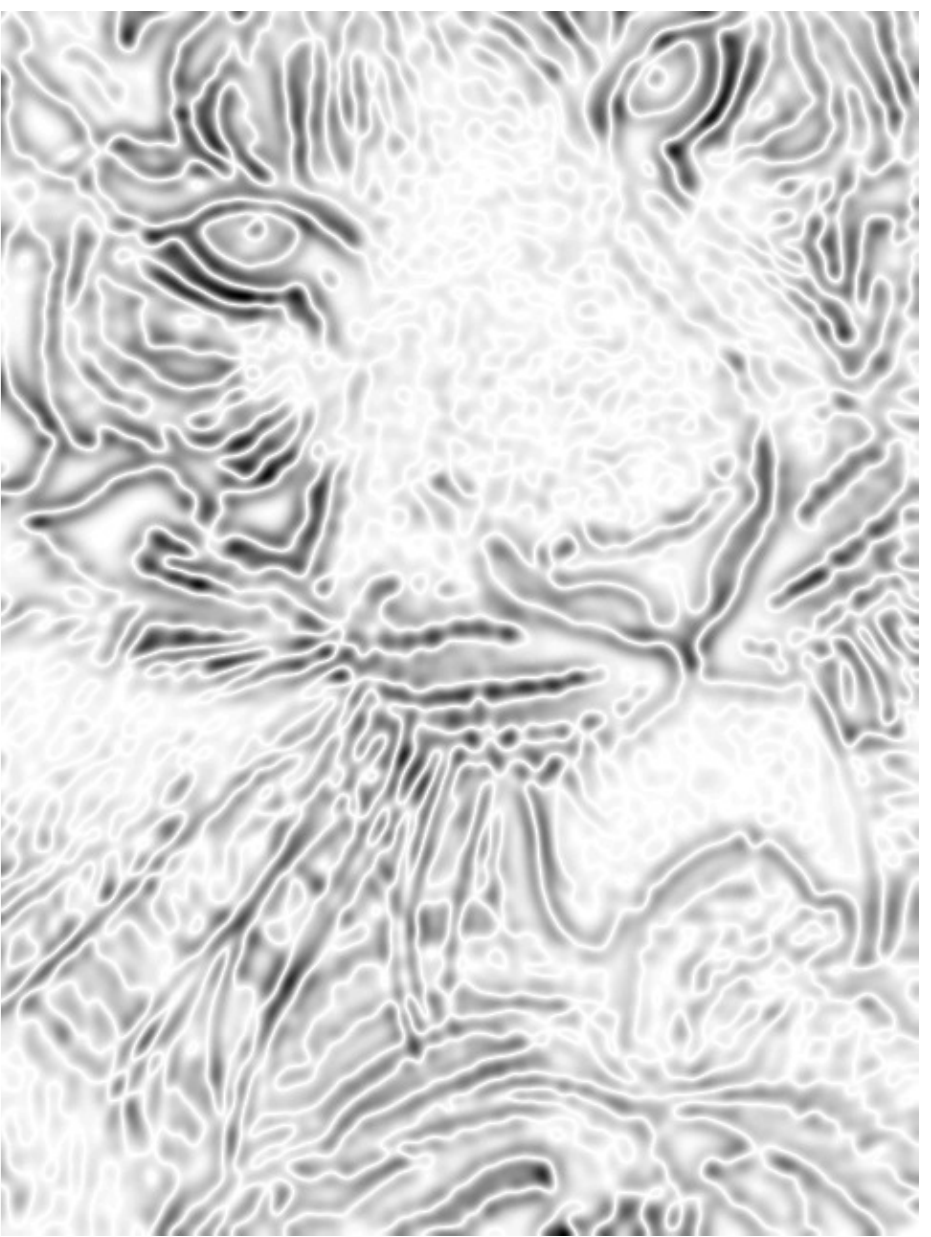} 
       & \includegraphics[width=0.13\textwidth]{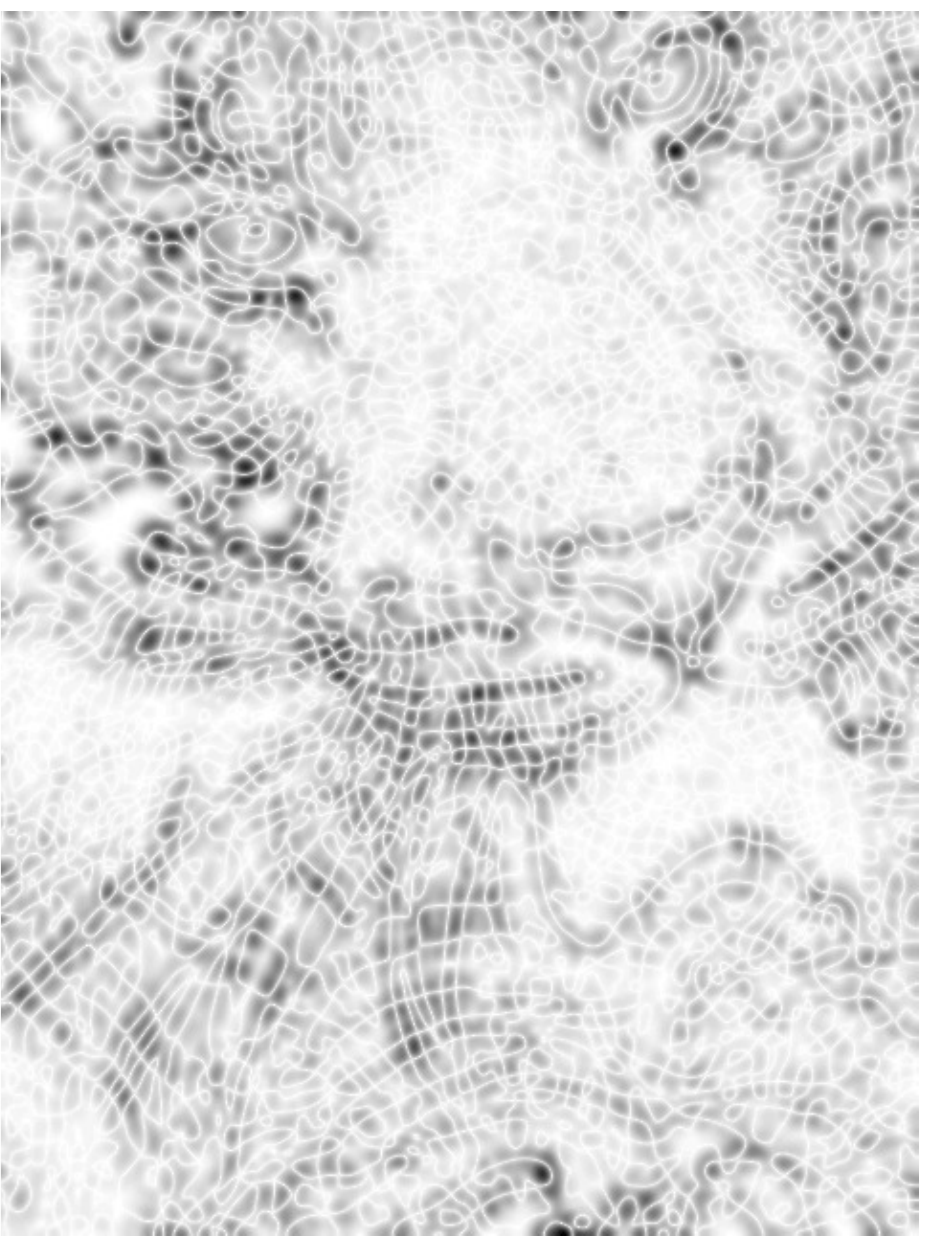} 
       & \includegraphics[width=0.13\textwidth]{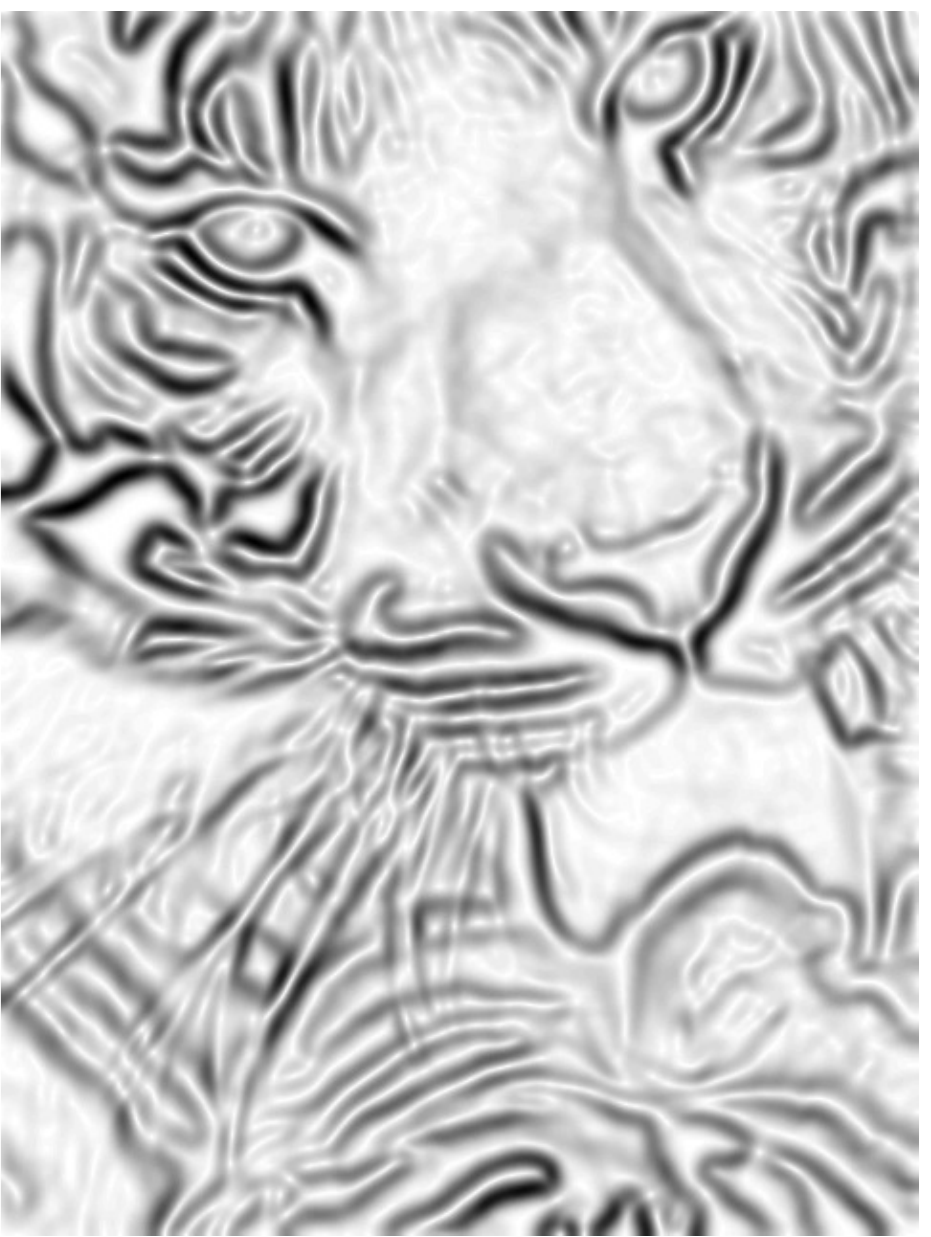} 
       & \includegraphics[width=0.13\textwidth]{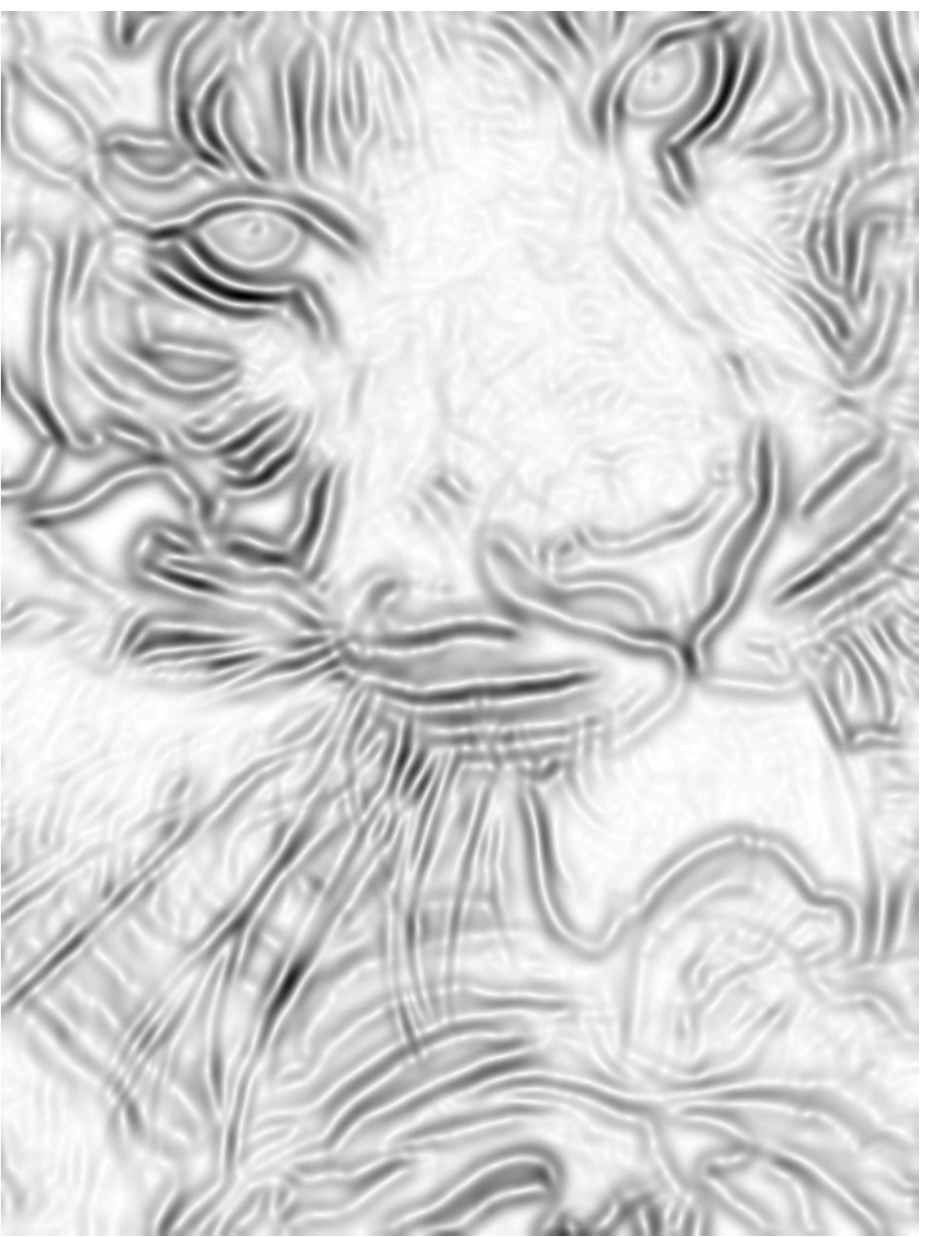} 
       & \includegraphics[width=0.13\textwidth]{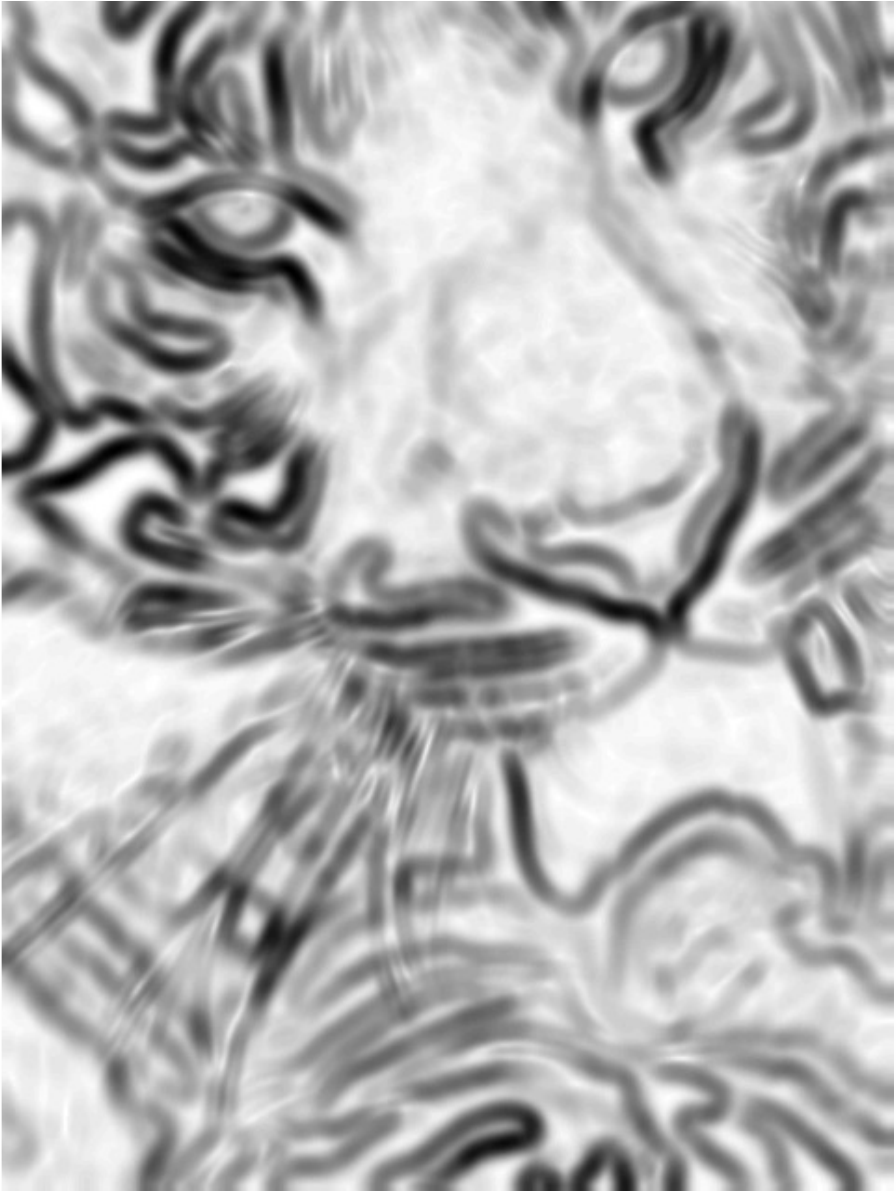} \\
    & {\footnotesize $|\overline{\nabla^2} L|$} 
       &  {\footnotesize $\sqrt{|\overline{\det {\cal H}} L|}$} 
       & {\footnotesize $\sqrt{\overline{\cal Q}_{(x,y),1} L}$} 
       & {\footnotesize $\sqrt{\overline{\cal Q}_{(x,y),2} L}$}
       & {\footnotesize $\sqrt{\overline{\cal Q}_{(x,y)} L}$} \\
       & \includegraphics[width=0.13\textwidth]{tiger2-Laplace-t16-eps-converted-to} 
       & \includegraphics[width=0.13\textwidth]{tiger2-detHessian-t16-eps-converted-to} 
       & \includegraphics[width=0.13\textwidth]{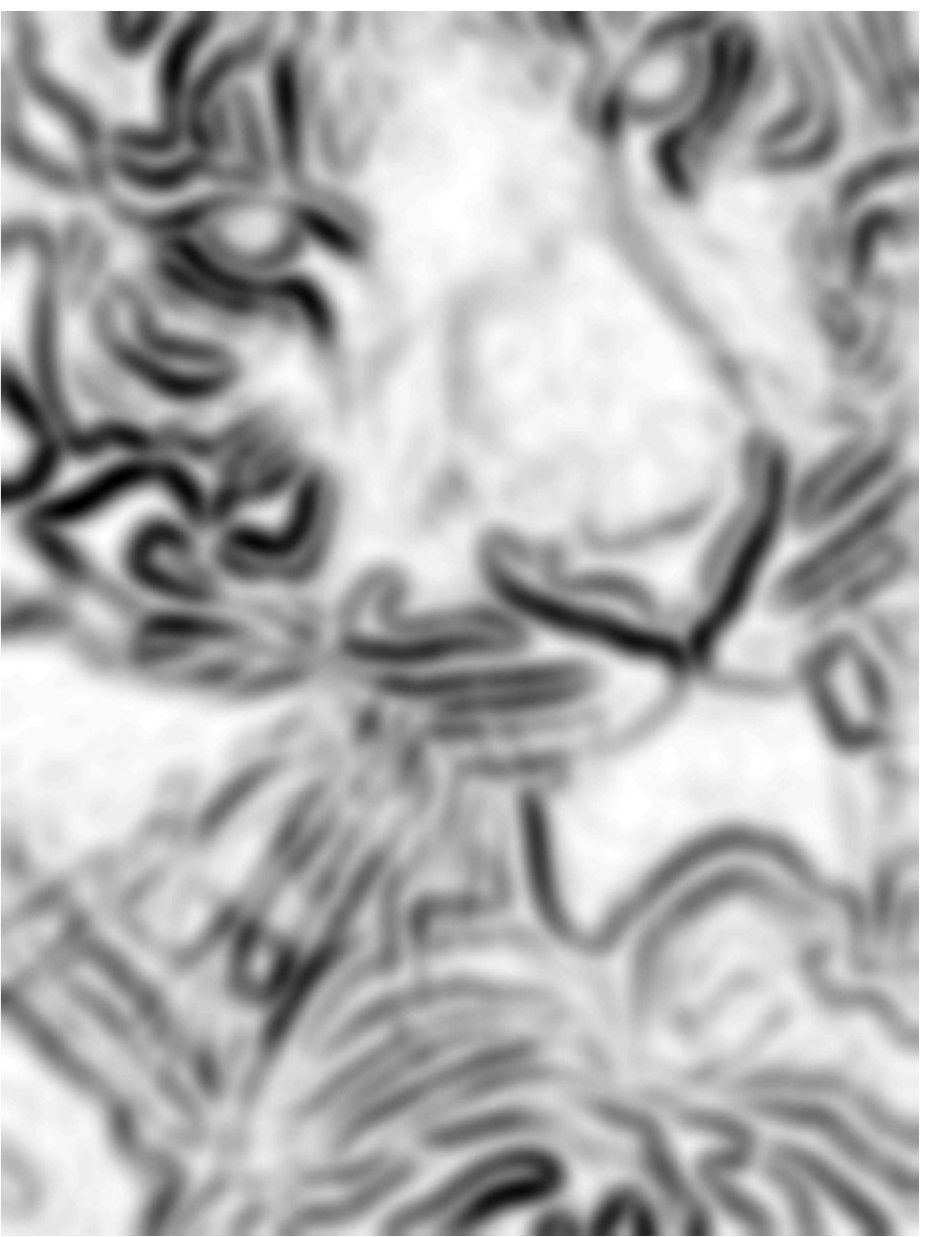} 
       & \includegraphics[width=0.13\textwidth]{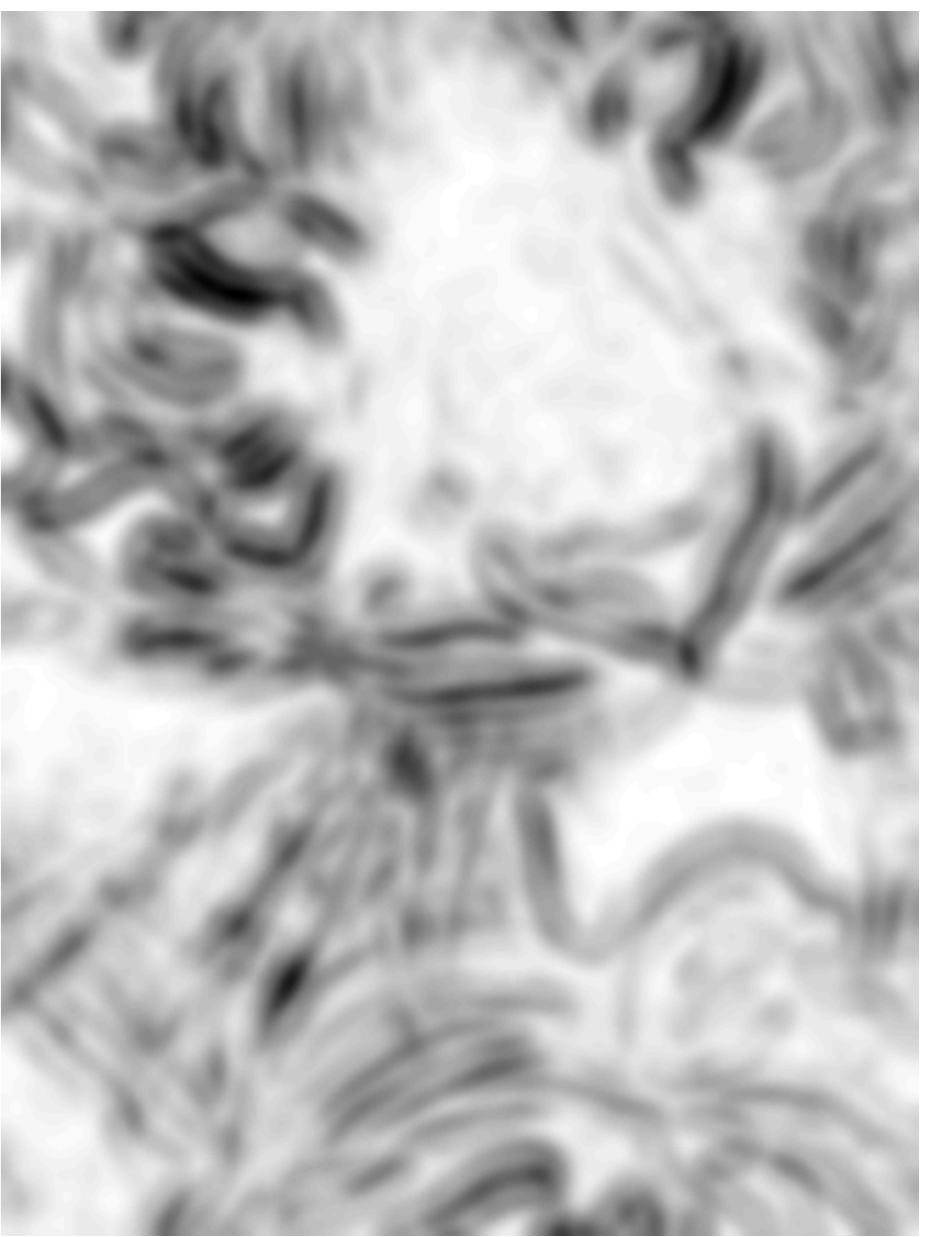} 
       & \includegraphics[width=0.13\textwidth]{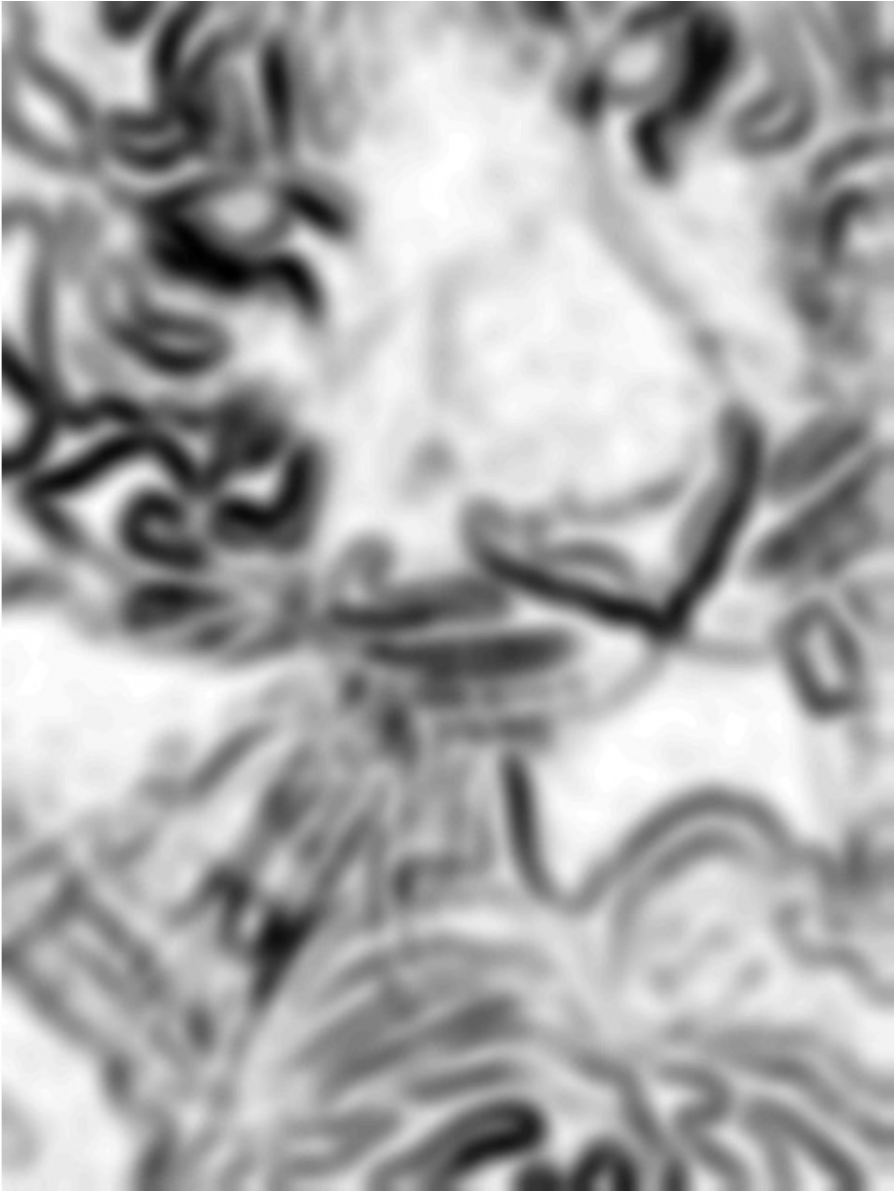} 
     \end{tabular}
  \end{center}
\caption{The result of computing (top row) the unsmoothed quasi quadrature measure
${\cal Q}_{(x,y),norm} L$ and (bottom row) the post-smoothed quasi quadrature entity 
$\overline{\cal Q}_{(x,y),norm} L$ with their underlying first- and second-order
  components at scale $t = 16$ for $C_s = 1/\sqrt{2}$. 
  For comparison, (top row) Laplacian and determinant of the
  Hessian responses as well as (bottom row) the result of applying
  spatial smoothing to these differential operators for relative post-smoothing scale $c =
  1$ are also shown at the same scale to illustrate
  how Laplacian and determinant of the Hessian responses will be limited to specific aspects of local
  image structures and thereby the limitations of using the Laplacian
  or the determinant of the Hessian 
  operators only for dense scale selection. (Image size: $480 \times 640$.)}
  \label{fig-tiger-quasi-quad-postsmoothed-c1}
\end{figure*}

The context we consider %for dense spatial scale selection 
is a spatial
scale-space representation $L(x, y;\; s)$ 
defined from any 2-D image $f(x, y)$ by convolution with Gaussian kernels
\begin{equation}
  g(x, y;\; s) = \frac{1}{2 \pi s} e^{-(x^2+y^2)/2s}
\end{equation}
at different spatial
scales $s$
(Iijima \cite{Iij62}; Witkin \cite{Wit83}; Koenderink \cite{Koe84-BC};
 Koenderink and van Doorn \cite{KoeDoo87-BC,KoeDoo92-PAMI};
 Lindeberg \cite{Lin93-Dis,Lin10-JMIV};
 Florack \cite{Flo97-book}; 
 Sporring {\em et al.\/}\ \cite{SpoNieFloJoh96-SCSPTH};
 Weickert {\em et al.\/}\ \cite{WeiIshImi99-JMIV};
 ter Haar Romeny \cite{Haa04-book})
\begin{equation}
  L(\cdot, \cdot;\; s) = g(\cdot, \cdot;\; s) * f(\cdot, \cdot) 
\end{equation}
and with
$\gamma$-normalized derivatives defined at any scale $s$ according to (Lindeberg \cite{Lin97-IJCV})
\begin{equation}
  \partial_{\xi} = s^{\gamma_s/2} \, \partial_x, \quad
  \partial_{\eta} = s^{\gamma_s/2} \, \partial_y.
\end{equation}
If attempting to perform dense local scale selection 
in the possibly most straightforward manner,
by detecting local extrema of the 
scale-normalized Laplacian 
$\nabla_{norm}^2 L = s \, (L_{xx} + L_{yy})$ 
or the scale-normalized determinant of the Hessian
$\det {\cal H}_{norm} L = s^2 \, (L_{xx} L_{yy} - L_{xy}^2)$
at points that are not interest points, one will soon find out that 
the resulting scale estimates will be strongly dependent on the points
at which they are computed. The reason for this is that the underlying interest point
detectors primarily respond to very specific aspects of the second-order
differential structure, see
Figure~\ref{fig-tiger-quasi-quad-postsmoothed-c1} for an illustration.
Scale selection by the scale-normalized Laplacian or the
scale-normalized determinant of the Hessian
operators is therefore primarily intended for image structures that lead to
strong responses for these differential operators, such as spatial
interest points (Lindeberg \cite{Lin97-IJCV,Lin12-JMIV,Lin15-JMIV}).

For the performing dense scale selection, it is therefore more natural to
seek a differential expression that responds to wider classes of
image structures, comprising both first- and second-order differential
structures, and without bias towards primarily specific aspects of the 
second-order differential image structure.

\subsection{A spatial quasi quadrature measure}

By combining the rotationally
invariant differential invariants
(i)~the {\em gradient magnitude\/}
\begin{equation}
  \label{eq-gradmagn}
  |\nabla L |^2 = L_x^2 + L_y^2
\end{equation}
as a measure of the amount of first-order
structure and
(ii)~the {\em Frobenius norm of the Hessian matrix\/}
\begin{equation}
  \| {\cal H} L \|_F^2 = L_{xx}^2 + 2 L_{xy}^2 + L_{yy}^2
\end{equation}
as a measure of the amount of second-order structure,
we will consider extensions of the following {\em quasi quadrature\/}
measure (Lindeberg \cite[Eq.~(63)]{Lin97-IJCV})
% \footnote{The notation used for scale-normalized differential
%   expressions in this paper is as follows: 
%    (i)~differential
%    expressions expressed in terms of scale-normalized
%    derivatives based on the specific choice of $\gamma_s = 1$ 
%    (such as equation (\ref{eq-def-quasi-quad})) are subscripted by ``norm'',
%    (ii)~differential
%    expressions expressed in terms of scale-normalized
%    derivatives for more general values of $\gamma_s \neq 1$ 
%    (such as equation (\ref{eq-quasi-quad-scale-mod-gamma1-gamma2-space})) are subscripted by
%    ``$\gamma$-norm'',
%    (iii)~differential expressions in terms of scale-normalized
%    derivatives for $\gamma_s = 1$ and additionally complemented
%    by complementary $\Gamma$-normalization (such as equation~(\ref{eq-quasi-quad-scale-mod})) are
%    subscripted by ``$\Gamma$-norm'', whereas 
%    (v)~differential expressions
%    not involving any scale normalization (such as equation (\ref{eq-gradmagn})) are not subscripted by any
%    scale normalization attribute.}
\begin{align}
  \begin{split}
    \label{eq-def-quasi-quad}
    {\cal Q}_{(x,y),norm} L 
    & = s \, (L_x^2 + L_y^2) + C_s \, s^2 \, (L_{xx}^2 + 2 L_{xy}^2 + L_{yy}^2)
  \end{split}
\end{align}
based on scale-normalized derivatives for $\gamma_s = 1$.
This differential entity can be seen as an approximation of 
the notion of 
a quadrature pair of an odd and even filter (Gabor \cite{Gab46})
as more traditionally formulated based on a Hilbert transform
(Bracewell \cite[p.~267-272]{Bra99}) and then extended to 2-D image
space, while being 
confined within the family of differential
expressions based on Gaussian derivatives and additionally 
being rotationally invariant.

If complemented by spatial integration, the components of this quasi
quadrature measure are specifically related to the following
class of energy measures over the frequency domain
(Lindeberg \cite[App.~A.3]{Lin97-IJCV})
(here expressed in terms of multi-index notation for the partial derivatives 
$x^{\alpha} = x_1^{\alpha_1} \dots x_D^{\alpha_D}$ with 
$|\alpha| = \alpha_1 + \dots + \alpha_D$):
\begin{equation}
  E_{m,\gamma-norm} 
  = \int_{x \in \bbbr^D} \sum_{|\alpha| = m} s^{m\gamma_s} \, L_{x^{\alpha}}^2 \, dx
  = \frac{s^{m\gamma_s}}{(2 \pi)^D}
      \int_{\omega \in \bbbr^D} | \omega|^{2 m} \, \hat{g}^2(\omega;\; s)
      \, d\omega.
\end{equation}
For the specific choice of $C_s = 1/2$, the quasi quadrature measure
(\ref{eq-def-quasi-quad}) coincides
with the proposals by Loog \cite{Loo07-SSVM} and Griffin
\cite{Gri07-PAMI} to define a metric of the $N$-jet in scale space.

\subsubsection{Complementary scale normalization}

To allow for richer degrees of freedom regarding the scale selection
properties, we allow for complementary scale
normalization of the form
\begin{align}
  \begin{split}
    \label{eq-quasi-quad-scale-mod}
    {\cal Q}_{(x,y),\Gamma-norm} L 
    & = \frac{s \, (L_x^2 + L_y^2) + C_s \, s^2 \, (L_{xx}^2 + 2 L_{xy}^2 + L_{yy}^2)}{s^{\Gamma_s}},
  \end{split}
\end{align}
which is still within the class of scale-normalized
differential expressions as obtained from $\gamma$-normalized
derivatives 
\begin{align}
  \begin{split}
    \label{eq-quasi-quad-scale-mod-gamma1-gamma2-space}
    {\cal Q}_{(x,y),\gamma-norm} L 
   =  s^{\gamma_1} \, (L_x^2 + L_y^2)
   + C_s \,  s^{2 \gamma_2} \, (L_{xx}^2 + 2 L_{xy}^2 + L_{yy}^2)
 \end{split}
\end{align}
for $\gamma_1 = 1 - \Gamma_s$ and $\gamma_2 = 1 - \frac{\Gamma_s}{2}$.
A major motivation for introducing the parameter $\Gamma_s$ in 
equation~(\ref{eq-quasi-quad-scale-mod}) is that if 
we would use $\Gamma_s = 0$ then it can be shown (Lindeberg \cite{Lin98-IJCV}) that the selected
scale would be infinite for any diffuse step edge, which is not a
desirable property for a dense scale selection mechanism, while if
using a value of $\Gamma_s > 0$, the selected scale for a diffuse edge
will be finite \cite[Equation~(23)]{Lin98-IJCV}
\begin{equation}
  \hat{s} 
  = \frac{\gamma_1}{1-\gamma_1} \, s_0 
  = \frac{1-\Gamma_s}{\Gamma_s} \, s_0.
\end{equation}
Concerning the choice of $\Gamma_s$, it can be observed
that setting $\Gamma_s = 1/2$ leads to $\gamma_1 = 1/2$ and 
$\gamma_2 = 3/4$, which are the values derived for edge detection and
ridge detection respectively to make the scale
estimate for a diffuse step edge reflect the diffuseness of the edge
and the scale estimate for a Gaussian ridge reflect the width of
the ridge \cite{Lin98-IJCV}.
For blob detection based on second-order derivatives, $\gamma_s=1$
corresponding to $\Gamma_s = 0$ is on
the other hand the preferred choice to ensure that the scale level for
a rotationally symmetric or affine deformed Gaussian blob reflects the
scale of the blob \cite{Lin97-IJCV,Lin12-JMIV}.
From these indications, we could expect to choose the complementary
scale normalization parameter $\Gamma_s$ in the range $\Gamma_s \in ]0, \tfrac{1}{2}]$.

\subsubsection{Complementary spatial post-smoothing}
\label{sec-compl-spat-postsmooth}

While the combination of first- and second-order information in
(\ref{eq-def-quasi-quad}) and (\ref{eq-quasi-quad-scale-mod}) 
will decrease the spatial dependency of the differential expression
compared to using only either first- or second-order information, 
the resulting differential expressions will not produce a constant
scale estimate for a sine wave, unless the computations are performed at a
scale level perfectly adapted to the wavelength of the signal.
To reduce the local ripples caused by this 
{\em phase dependency\/}, we introduce complementary smoothing of
the quadrature entity ${\cal Q}_{(x,y),\Gamma-norm}$ using an integration
scale parameter $s_{int}$ proportional to the local scale parameter
$s$ used for computing the spatial derivatives
\begin{equation}
    \label{eq-quasi-quad-scale-mod-post-smooth}
    \overline{\cal Q}_{(x,y),\Gamma-norm} L
    = {\cal E}_{s_{int}}({\cal Q}_{(x,y),\Gamma-norm})
\end{equation}
where ${\cal E}_{s_{int}}$ denotes a Gaussian averaging operation with scale
parameter $s_{int} = c^2 \, s$.
We also define the first- and second-order components of this entity as
\begin{align}
  \begin{split}
    \label{eq-Q1bar-spatial}
    \overline{\cal Q}_{(x,y),1,\Gamma-norm} L
     = s^{-\Gamma_s} \, {\cal E}_{s_{int}}(|\nabla L |^2),  %L_{\xi}^2 + L_{\eta}^2), 
\end{split}\\
\begin{split}
    \label{eq-Q2bar-spatial}
    \overline{\cal Q}_{(x,y),2,\Gamma-norm} L
     = C_s \, s^{-\Gamma_s} \, {\cal E}_{s_{int}}(\| {\cal H} L \|_F^2). %L_{\xi\xi}^2 + 2 L_{\xi\eta}^2 + L_{\eta\eta}^2).
 \end{split}
\end{align}
Figure~\ref{fig-tiger-quasi-quad-postsmoothed-c1}
shows the result of computing these quasi quadrature measures as well as 
the underlying first- and second-order components for a grey-level 
image that contains image textures at different scales.
As can be seen from the results: (i)~the quasi quadrature is less sensitive to the
spatial variability in a dense textured image pattern compared to the
Laplacian or the determinant of the Hessian operators and (ii)~the
post-smoothing operation decreases the sensitivity further.

\begin{figure}[hbtp]
  \begin{center}
    \begin{tabular}{ccc}
    \hspace{-2mm} {\footnotesize\em original image} 
       &  {\footnotesize\em $\sqrt{{\cal Q}_{(x,y),\Gamma-norm} L}$}
       &  {\footnotesize\em $\sqrt{\overline{\cal Q}_{(x,y),\Gamma-norm}
           L}$}\\
    \hspace{-2mm} \includegraphics[height=0.12\textheight]{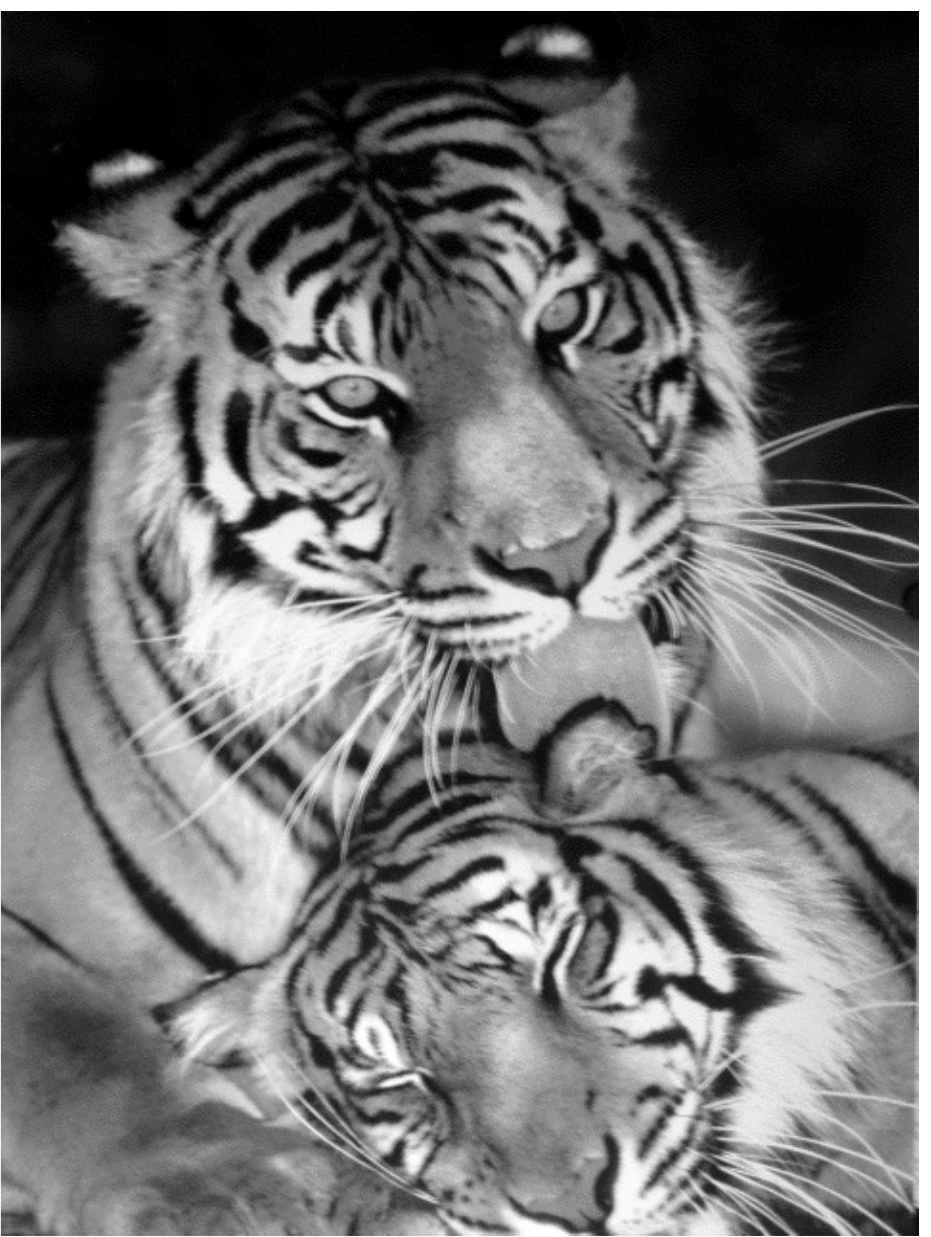} 
       & \includegraphics[height=0.12\textheight]{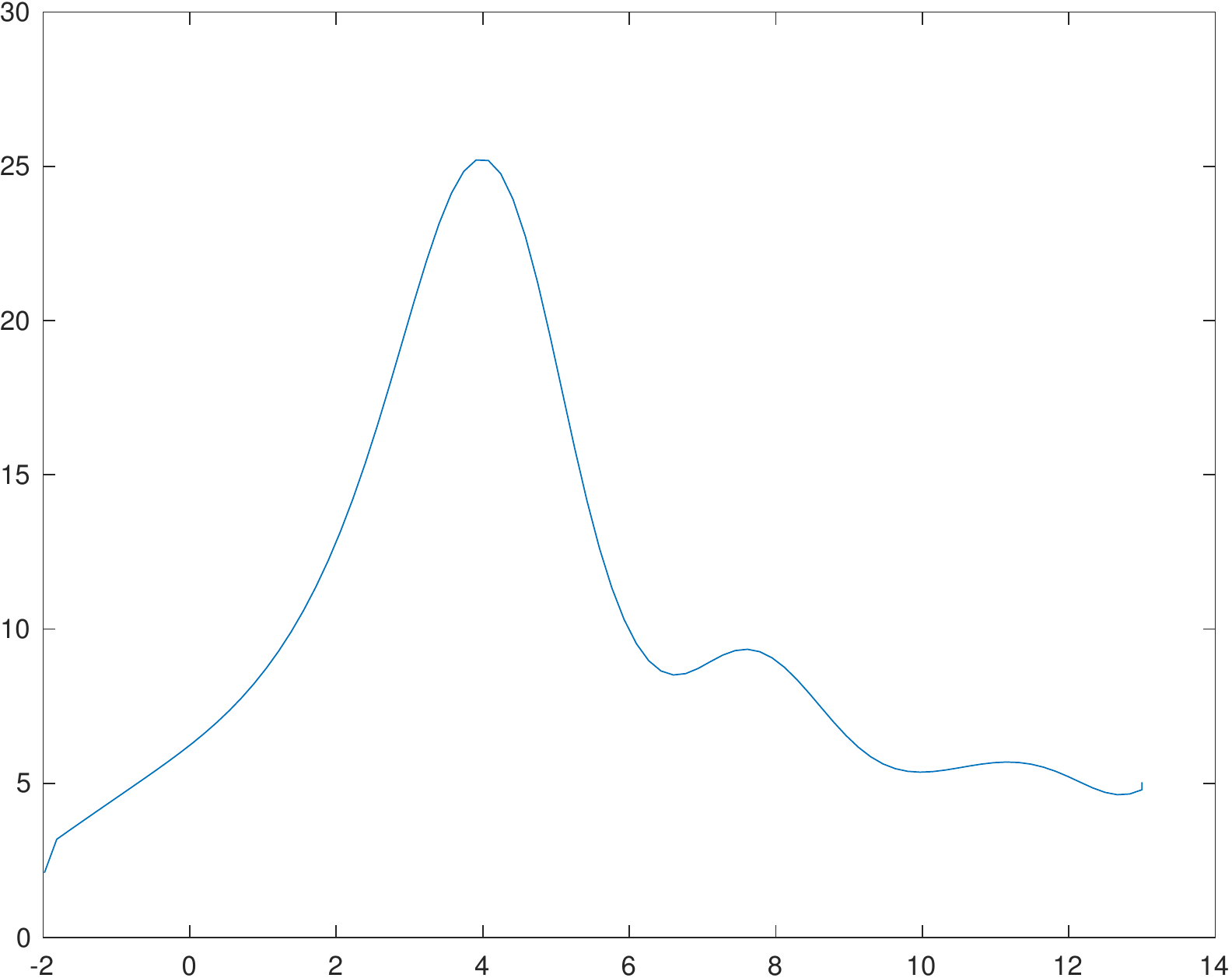} 
       & \includegraphics[height=0.12\textheight]{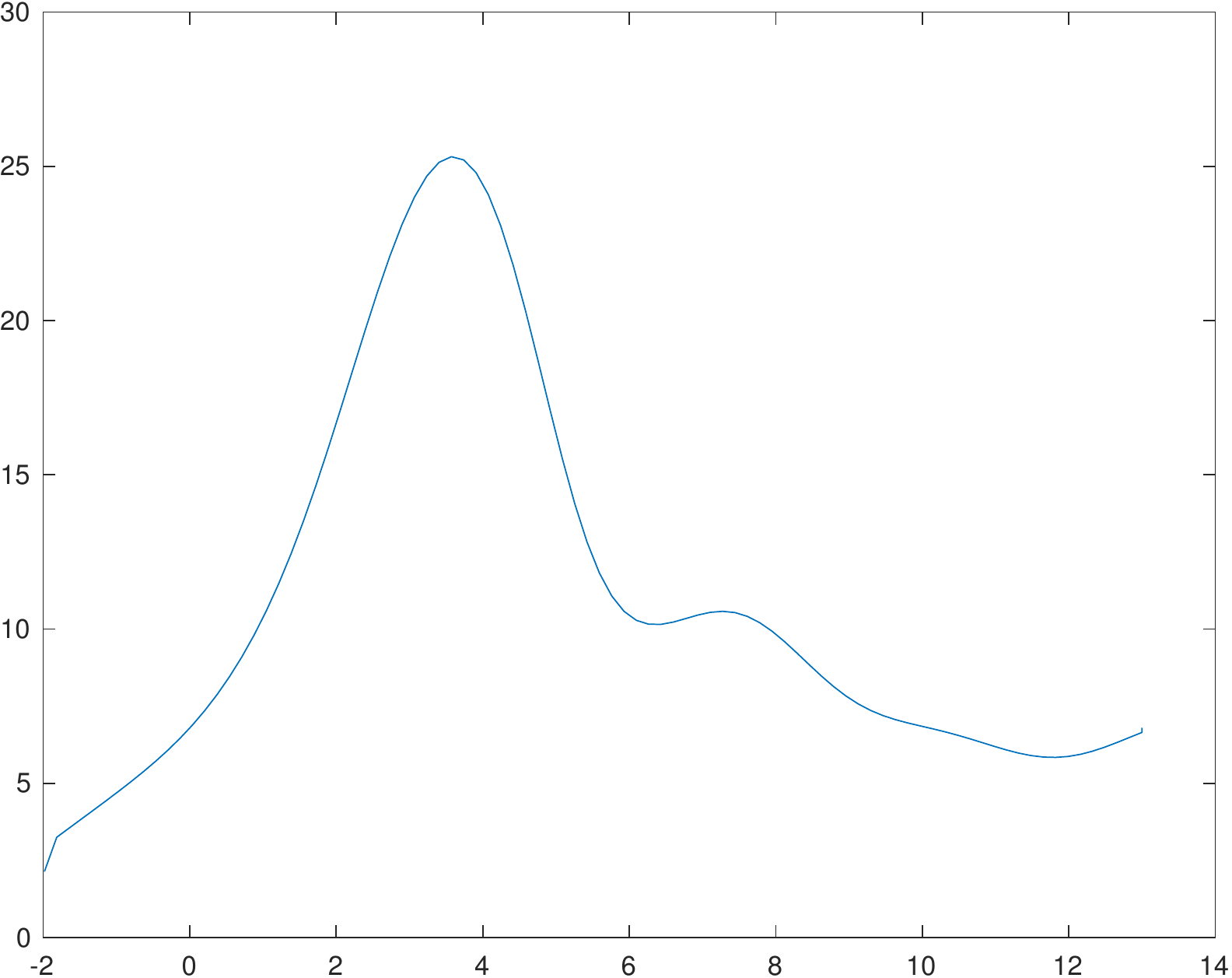} \\
   \hspace{-2mm} \includegraphics[height=0.12\textheight]{tiger2-vert-eps-converted-to} 
       & \includegraphics[height=0.12\textheight]{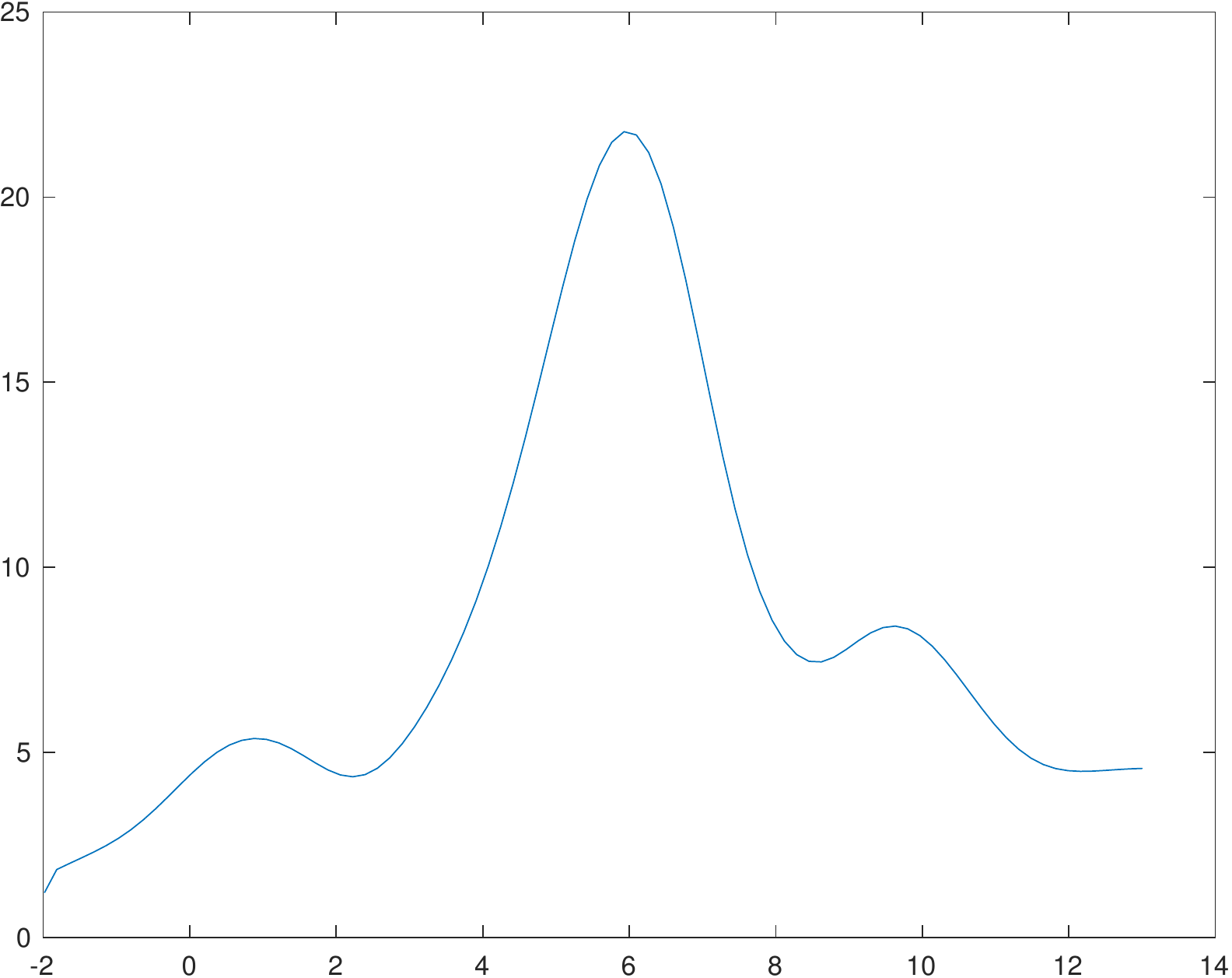} 
       & \includegraphics[height=0.12\textheight]{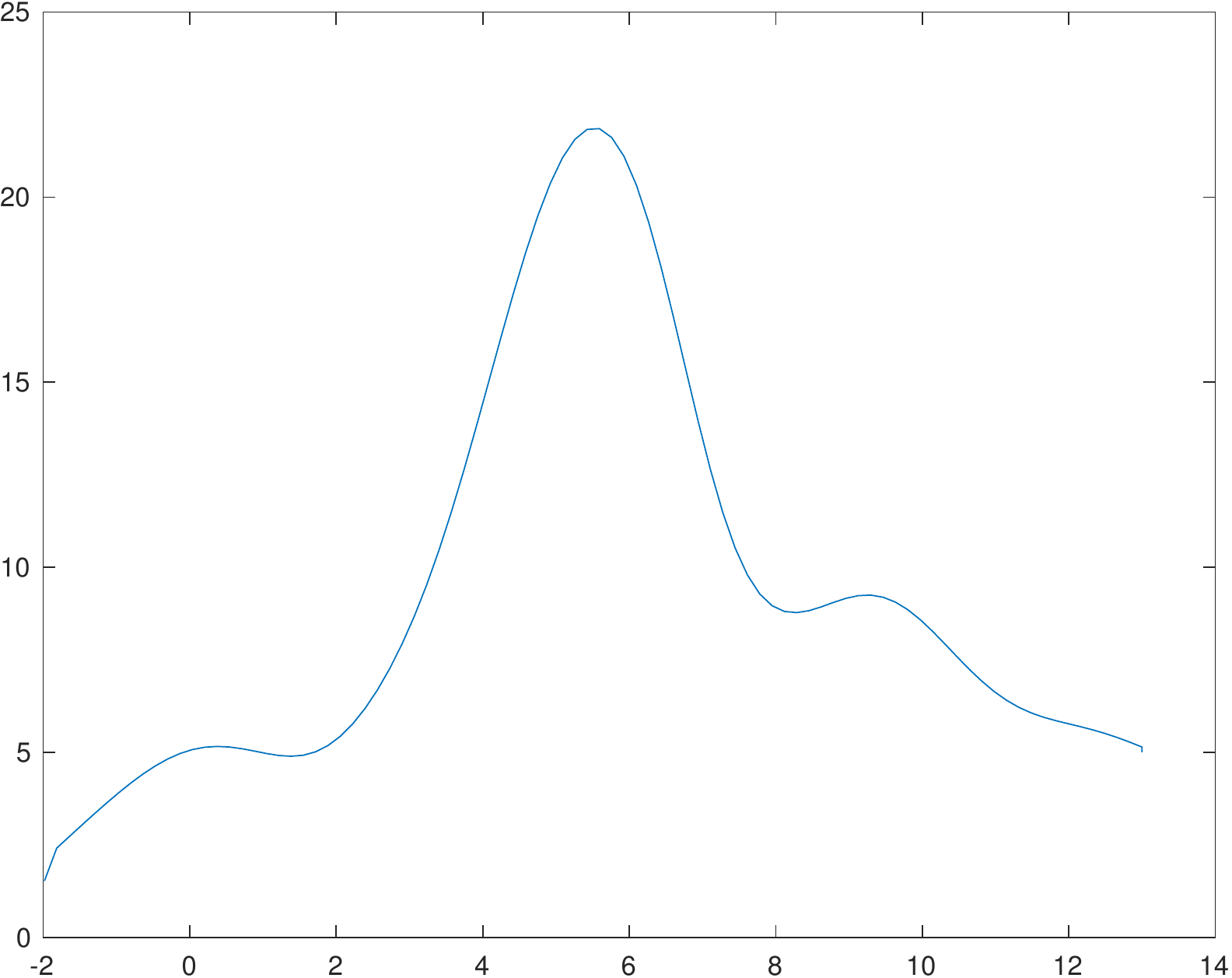} \\
   \hspace{-2mm}  \includegraphics[height=0.12\textheight]{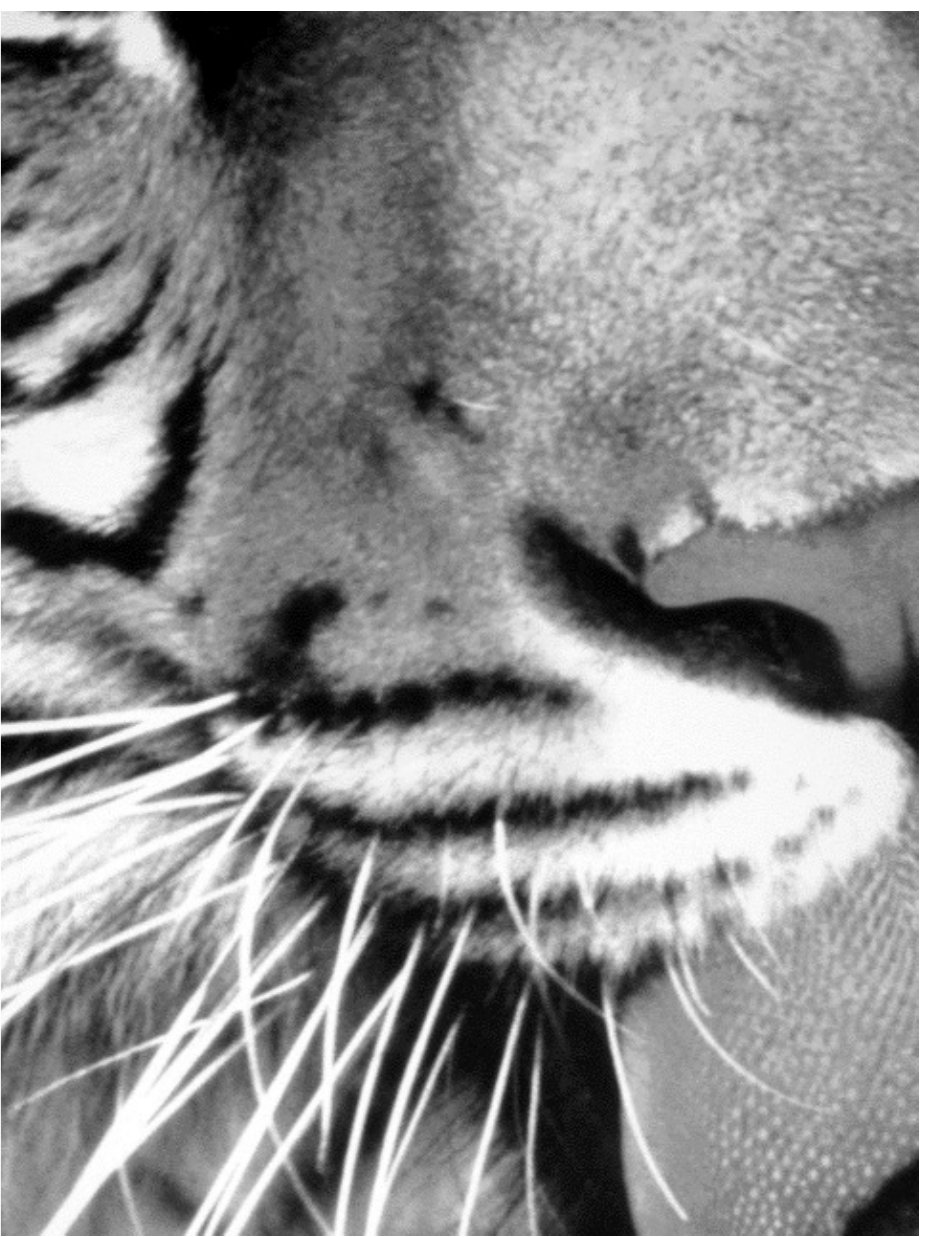} 
       & \includegraphics[height=0.12\textheight]{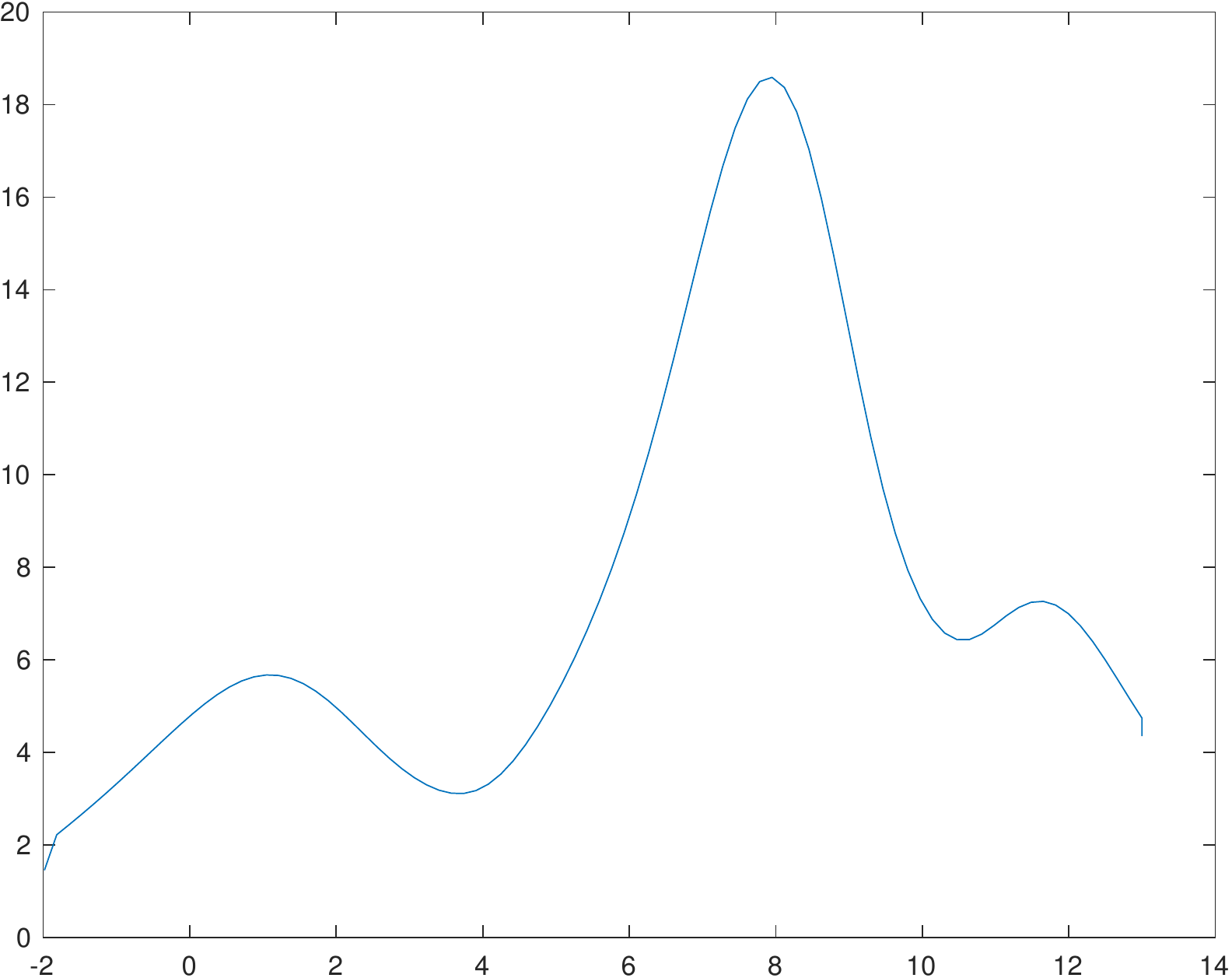} 
       & \includegraphics[height=0.12\textheight]{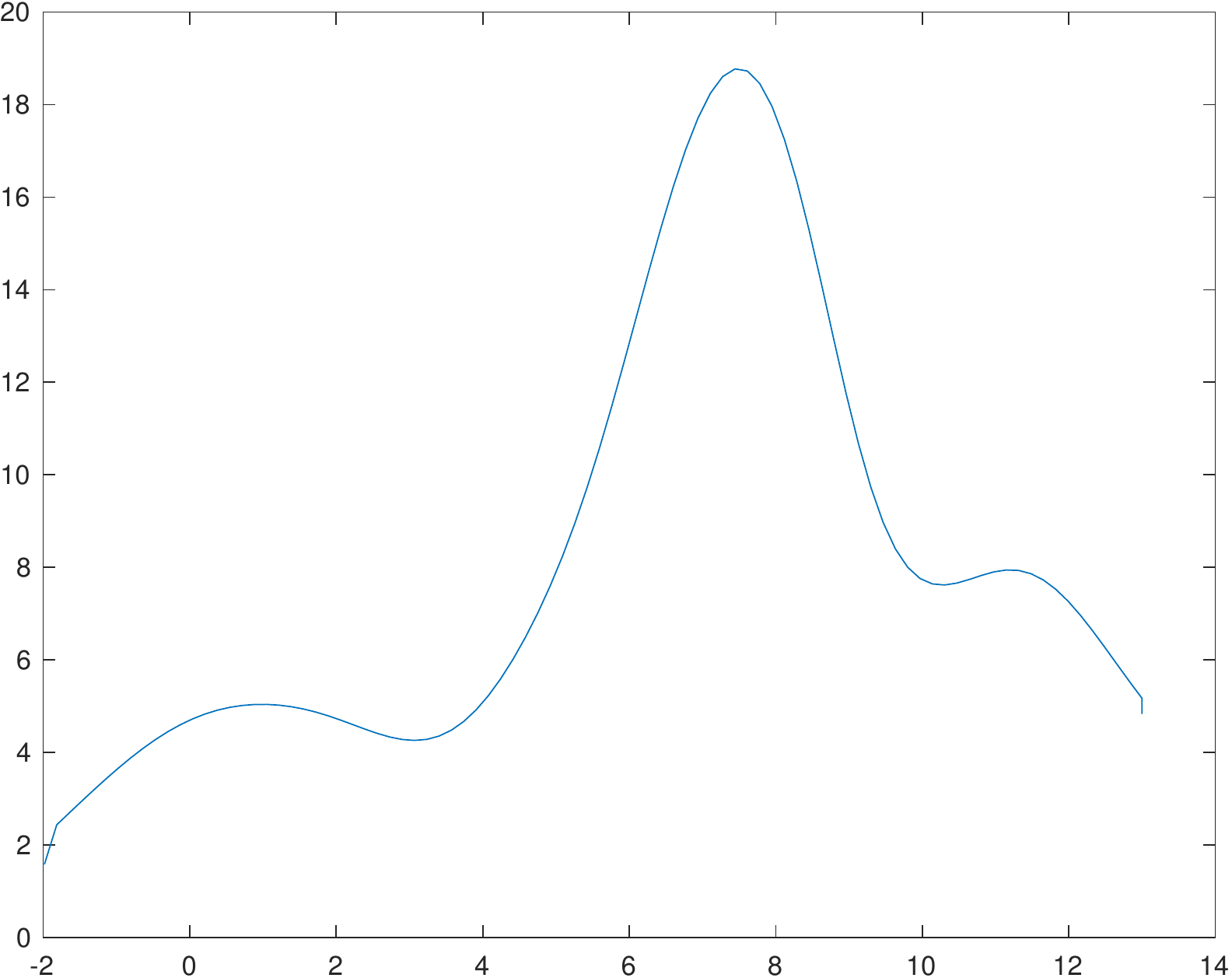}

    \end{tabular}
  \end{center}
\caption{Scale-space signatures computed at  corresponding central points for
    three images of a poster taken from different distances.
    Each scale-space signature shows the variation over scale of the
    scale-normalized quasi quadrature measure at the center as function of effective
    scale approximated by $s_{eff} = \log_2 (s_0 + s)$ for $s_0 = 1/8$.
   (middle column) the unsmoothed quasi
   quadrature measure ${\cal Q}_{(x,y),\Gamma-norm} L$ using 
   $\Gamma_s = 1/4$.
   (right column) the post-smoothed quasi
   quadrature measure $\overline{\cal Q}_{(x,y),\Gamma-norm} L$ using
   $\Gamma_s = 1/4$ and $c = 1$.
   Note how the scale levels at which the local maxima over scale are
   assumed follow the size variations in the image domain caused by
   varying the distance between the camera and the poster.
   All results have been computed using $C_s = 1/\sqrt{(1 - \Gamma_s)(2 - \Gamma_s)}$.
  (Scale range: $s \in [0.1, 8192]$, Image size: $480 \times 640$ pixels.)}
  \label{fig-tiger-scspsig-quasi-quad-Gamma0p25}
\end{figure}

\subsubsection{Basic scale selection method}

A basic method for dense scale selection does therefore consist of detecting
local extrema over scale of either the pointwise scale-normalized quasi quadrature
entity (\ref{eq-quasi-quad-scale-mod})
\begin{equation}
  \label{eq-scale-sel-quasi-quad}
     \hat{s}_{{\cal Q} L} = \argmaxlocal_{s > 0} {\cal Q}_{(x,y),\Gamma-norm} L
\end{equation}
or the corresponding post-smoothed entity (\ref{eq-quasi-quad-scale-mod-post-smooth})
\begin{equation}
   \label{eq-scale-sel-quasi-quad-post-smooth}
   \hat{s}_{\overline{\cal Q} L} = \argmaxlocal_{s > 0} \overline{\cal Q}_{(x,y),\Gamma-norm} L.
\end{equation}
Figure~\ref{fig-tiger-scspsig-quasi-quad-Gamma0p25} illustrates the
effects of these operations for three different images of the same
poster taken for different distances between the poster and the
camera. As can be seen from the graphs, the scale values at which the
local extrema over scale are assumed adapt to the size variations in
the image domain and are assumed at coarser scales relative to the
fixed image resolution as the camera approaches the object.

\subsection{Scale selection properties}
\label{eq-sc-sel-props}

When applying this methodology in practice, there are a
number of additional issues that need to be considered, which we will illustrate by
closed form theoretical analysis using idealized image models representing a dense
texture pattern or sparse image features respectively.

\subsubsection{Two-dimensional sine wave}

For a two-dimensional sine wave
\begin{equation}
  \label{eq-2D-sine-wave}
  f(x, y) = \sin (\omega_0 x) + \sin (\omega_0 y),
\end{equation}
the scale-space representation can be computed in closed form
\begin{equation}
  L(x, y;\; t) = e^{-\omega_0^2 s/2} (\sin (\omega_0 x) + \sin (\omega_0 y)).
\end{equation}
If we disregard the spatial post-smoothing step by setting the
proportionality parameter $c$ between the local scale parameter and
the integration scale parameter to $c = 0$, then the quasi quadrature entity assumes the form
\begin{align}
  \begin{split}
     {\cal Q}_{(x,y),\Gamma-norm} L =
     &  \omega_0^2 e^{-\omega_0^2 s} s^{1-\Gamma_s} \times
  \end{split}\nonumber\\
  \begin{split}
     \label{eq-Qnormbar-2D-sine-wave-general}
      &  \left(
              \cos ^2(\omega_0 x)+\cos ^2(\omega_0 y) 
              + C_s \omega_0^2 s \left(\sin ^2(\omega_0 x)+\sin^2(\omega_0 y)\right)
           \right).
  \end{split}
\end{align}
By differentiating this expression with respect to scale $s$, it follows
that for the image points $(x, y) = (m \pi/\omega_0, n \pi/\omega_0)$ 
at which only the first-order component 
\begin{equation}
  {\cal Q}_{(x,y),1,\Gamma-norm} L 
  = \frac{| \nabla_{norm} L |^2}{s^{\Gamma_s}}
  = \frac{s \, (L_x^2 + L_y^2)}{s^{\Gamma_s}}
\end{equation}
responds, the local extrema over scales are assumed at
\begin{equation}
  \label{eq-t1-1D-sine-wave}
  \hat{s}_1 = \frac{1-\Gamma_s}{\omega_0^2}.
\end{equation}
Correspondingly, for the spatial positions 
$(x, y) = ((\pi/2 + m \pi)/\omega_0, (\pi/2 + n \pi)/\omega_0)$ 
at which only the second-order component 
\begin{equation}
  {\cal Q}_{(x,y),2,\Gamma-norm} L 
  = \frac{C_s \, \| {\cal H}_{norm} L \|_F^2}{s^{\Gamma_s}}
  = \frac{C_s \,  s^2 \; (L_{xx}^2 + 2 L_{xy}^2 + L_{yy}^2)}{s^{\Gamma_s}}
\end{equation}
responds, the local extrema are assumed at
\begin{equation}
  \label{eq-t2-1D-sine-wave}
  \hat{s}_2 = \frac{2 - \Gamma_s}{\omega_0^2}.
\end{equation}
In this respect, the combination of first- and second-order
derivatives in the quasi quadrature entity will lead to a strong
{\em phase dependency in the scale estimates\/}.

At the intermediate points 
$(x, y) = ((\pi/4 + m \pi/2)/\omega_0, (\pi/4 + n \pi/2)/\omega_0)$,
the scale estimate is given by
\begin{equation}
  \label{eq-phase-indep-sc-est-1D-sine-wave}
  \hat{s}_{intermed} = 
  \frac{\sqrt{C_s^2 (\Gamma_s-2)^2-2 C_s \Gamma_s+1}-C_s (\Gamma_s-2)-1}
         {2 C_s \omega_0^2}.
\end{equation}
If we require the scale estimates at the intermediate
points to be equal to the geometric average of the extreme cases
\begin{equation}
  \label{eq-mean-scale-est-from-t1-t2-geom}
  \hat{s}_{intermed} 
  = \sqrt{\hat{s}_1 \, \hat{s}_2} 
  = \frac{\sqrt{(1 - \Gamma_s) (2 - \Gamma_s)}}{\omega_0^2},
\end{equation}
then this implies that $C_s$ should be chosen as
\begin{equation}
  \label{eq-C-from-Gamma-when-c-is-zero-geom}
  C_s = \frac{1}{2 - \Gamma_s}.
\end{equation}
Alternatively, if we
determine the weighting parameter $C_s$ such that the relative
strengths of the first- and second-order components become 
equal at the midpoints $(x, y) = (\pi/4 + m \pi/2)/\omega_0, \pi/4 + m \pi/2)/\omega_0)$
between the extreme points 
for the scale corresponding to the geometric average
$\sqrt{\hat{s}_1 \, \hat{s}_2}$ of the extreme values
\begin{equation}
  \label{eq-determ-C-balanced-first-second-order-responses}
  \left. {\cal Q}_{(x,y),1,\Gamma-norm} L \right|_{x = \frac{\pi}{4 \omega_0}, y = \frac{\pi}{4 \omega_0}, s= \sqrt{\hat{s}_1 \, \hat{s}_2}}
  = \left. {\cal Q}_{(x,y),2,\Gamma-norm} L \right|_{x = \frac{\pi}{4 \omega_0}, y = \frac{\pi}{4 \omega_0}, s= \sqrt{\hat{s}_1 \, \hat{s}_2}},
\end{equation}
then this implies that the relative weighting factor $C_s$ between the first-
and second-order derivative responses should be chosen as
\begin{equation}
  \label{eq-C-from-Gamma-when-c-is-zero-geom-temp}
  C_s = \frac{1}{\sqrt{(1 - \Gamma_s)(2 - \Gamma_s)}}. 
\end{equation}

\begin{figure}[hbtp]
  \begin{center}
    \begin{tabular}{cccc}
     & \multicolumn{3}{c}{\small\em Dense scale selection without post-smoothing} \\ \\
     & {\small $\Gamma_s = 0$} 
     & {\small $\Gamma_s = \frac{1}{4}$} 
     & {\small $\Gamma_s = \frac{1}{2}$} \\ \\
      & \includegraphics[width=0.25\textwidth]{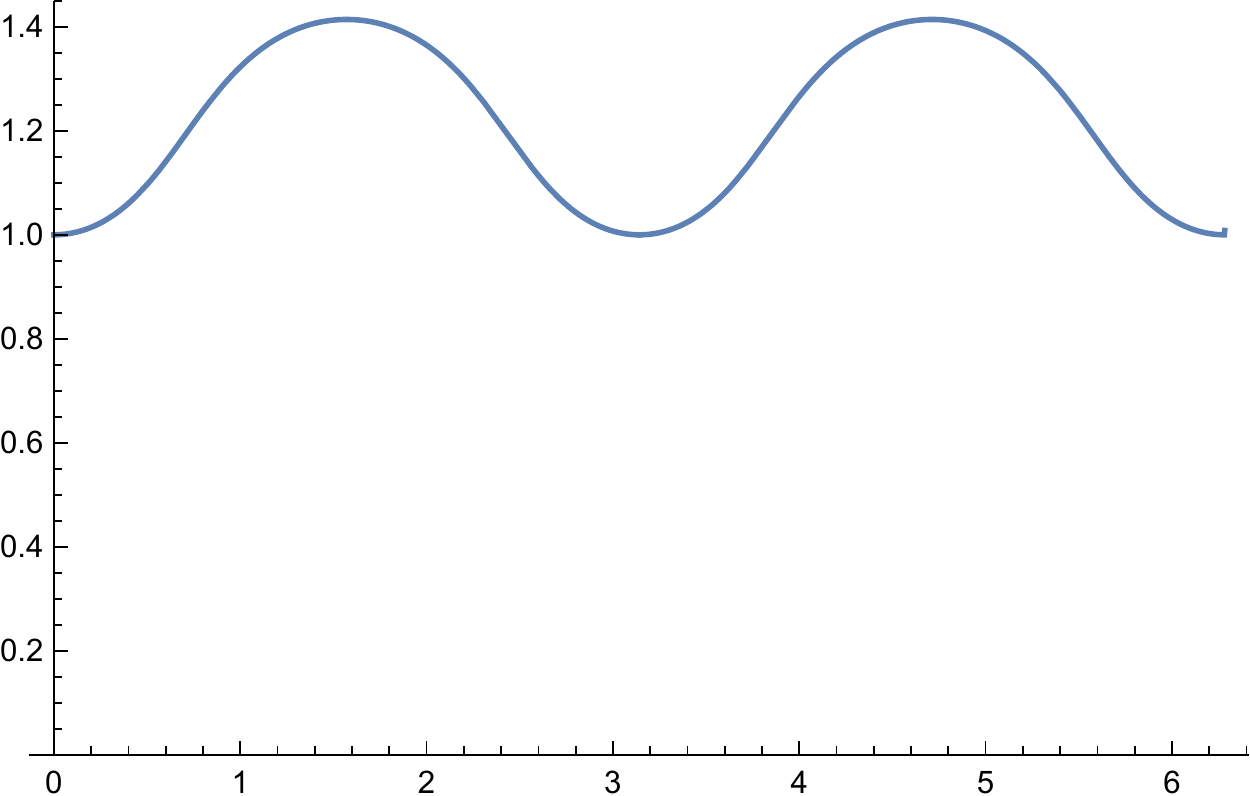}
       & \includegraphics[width=0.25\textwidth]{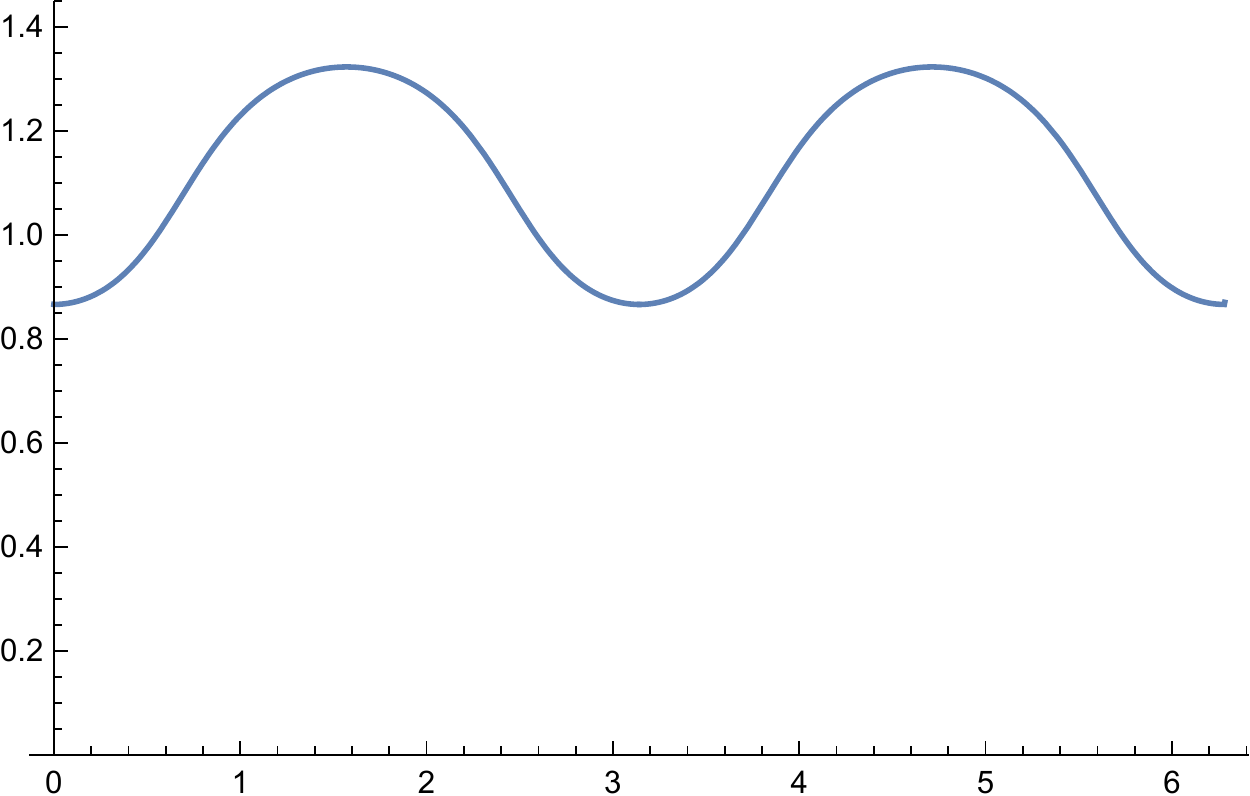}
       & \includegraphics[width=0.25\textwidth]{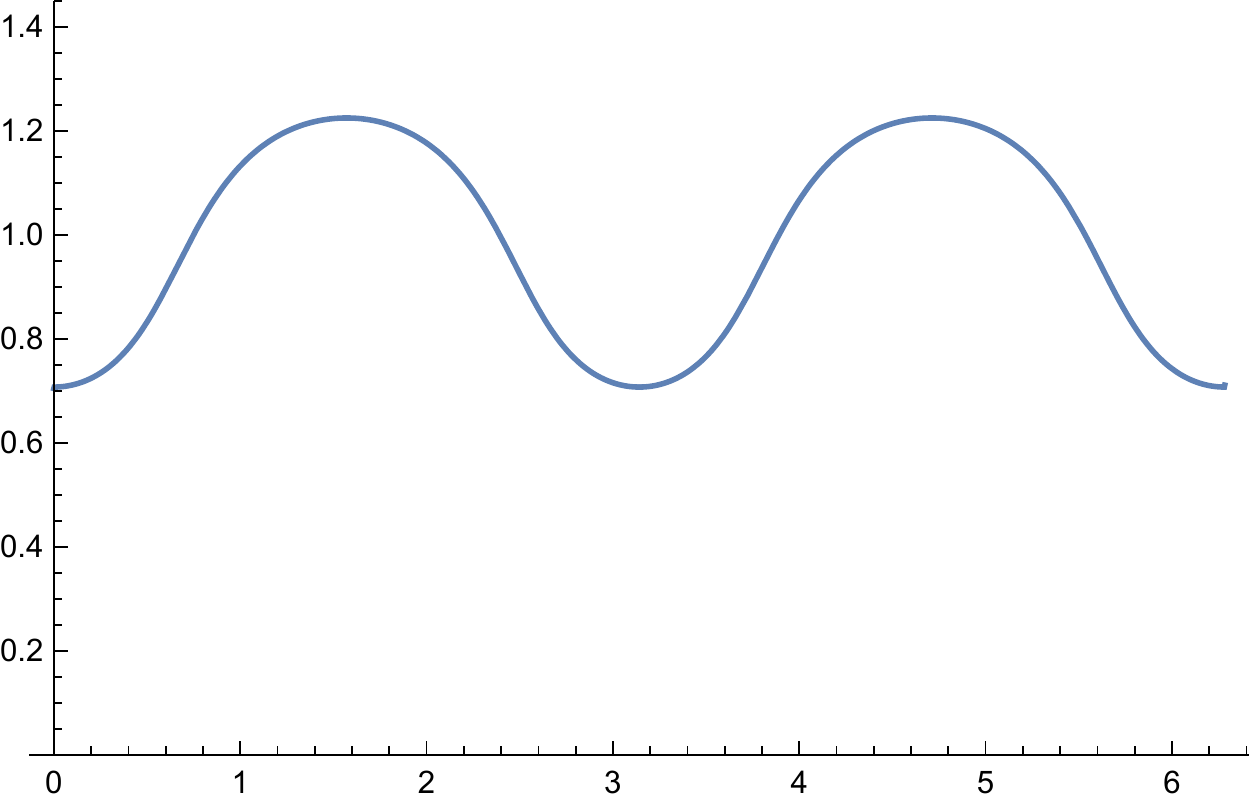}
         \\
     \\
      & \multicolumn{3}{c}{\small\em Phase-compensated dense scale
       selection without post-smoothing} \\ \\
     & {\small $\Gamma_s = 0$} 
     & {\small $\Gamma_s = \frac{1}{4}$} 
     & {\small $\Gamma_s = \frac{1}{2}$} \\ \\
     & \includegraphics[width=0.25\textwidth]{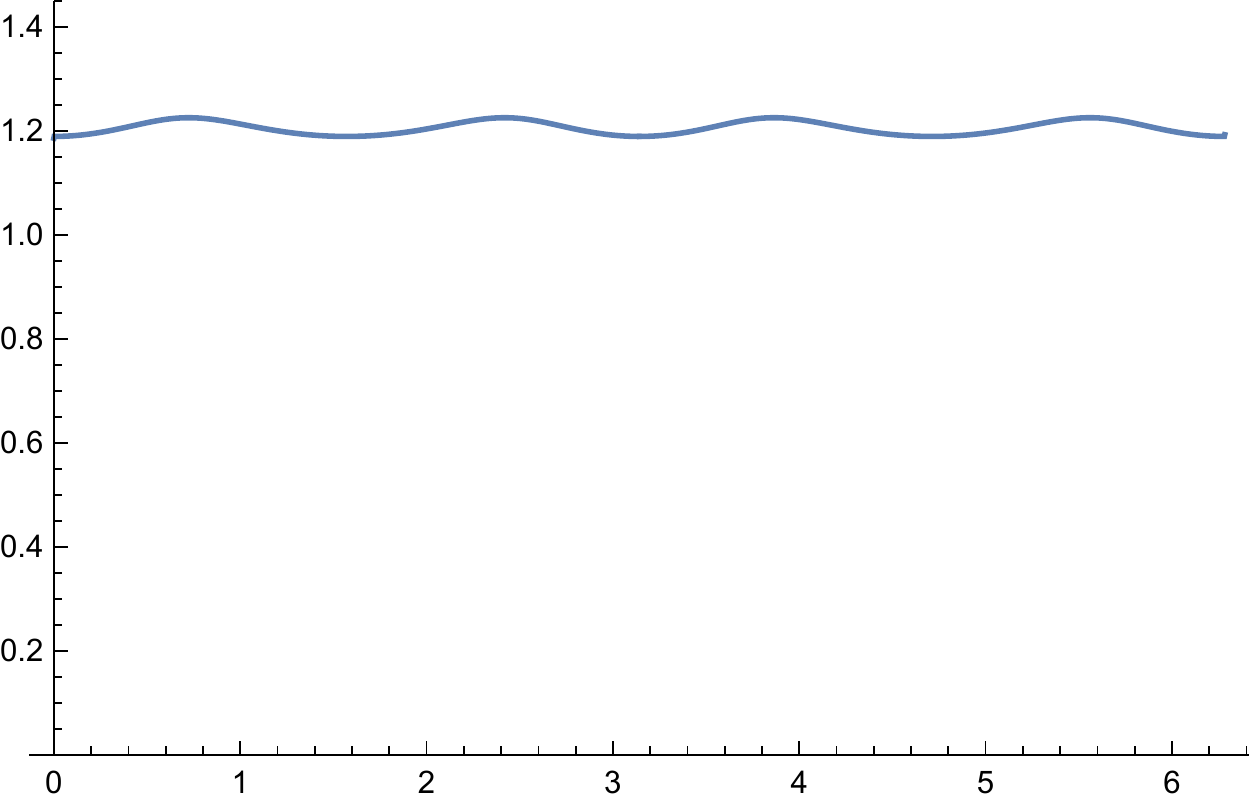}
       & \includegraphics[width=0.25\textwidth]{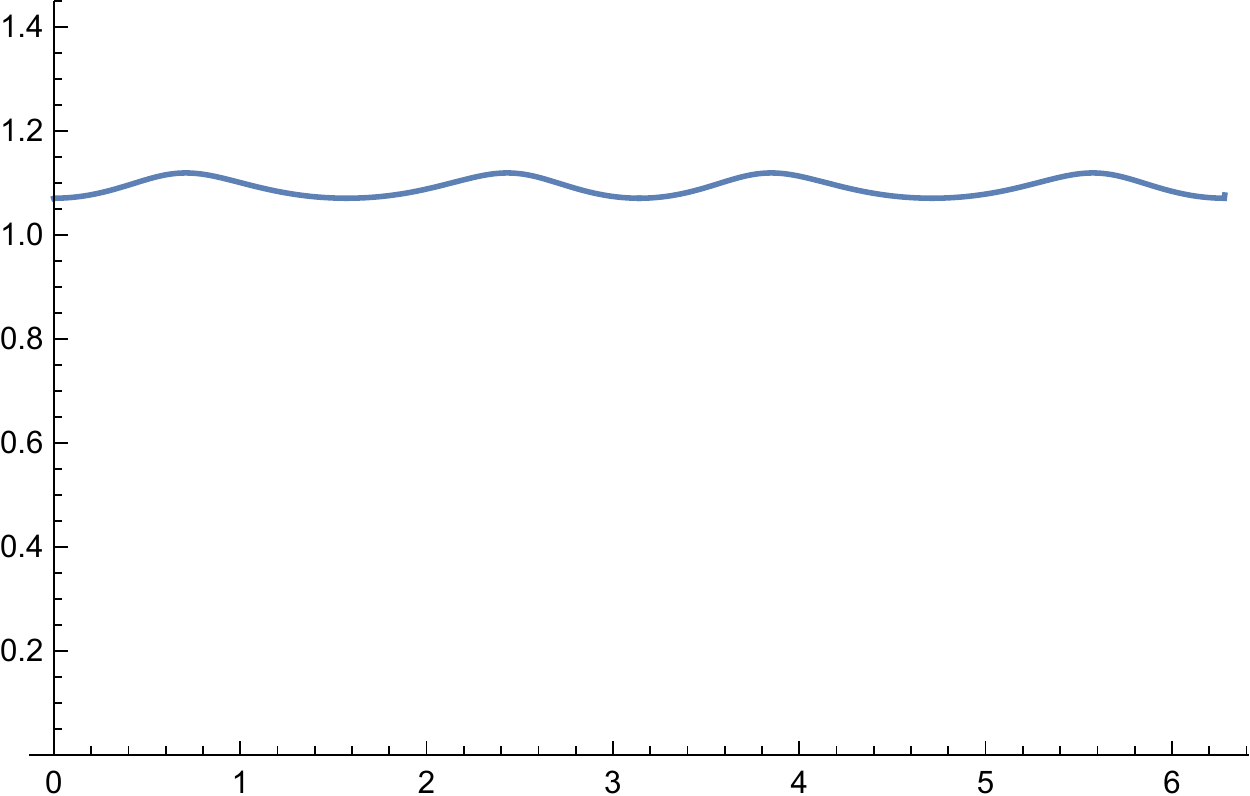}
       & \includegraphics[width=0.25\textwidth]{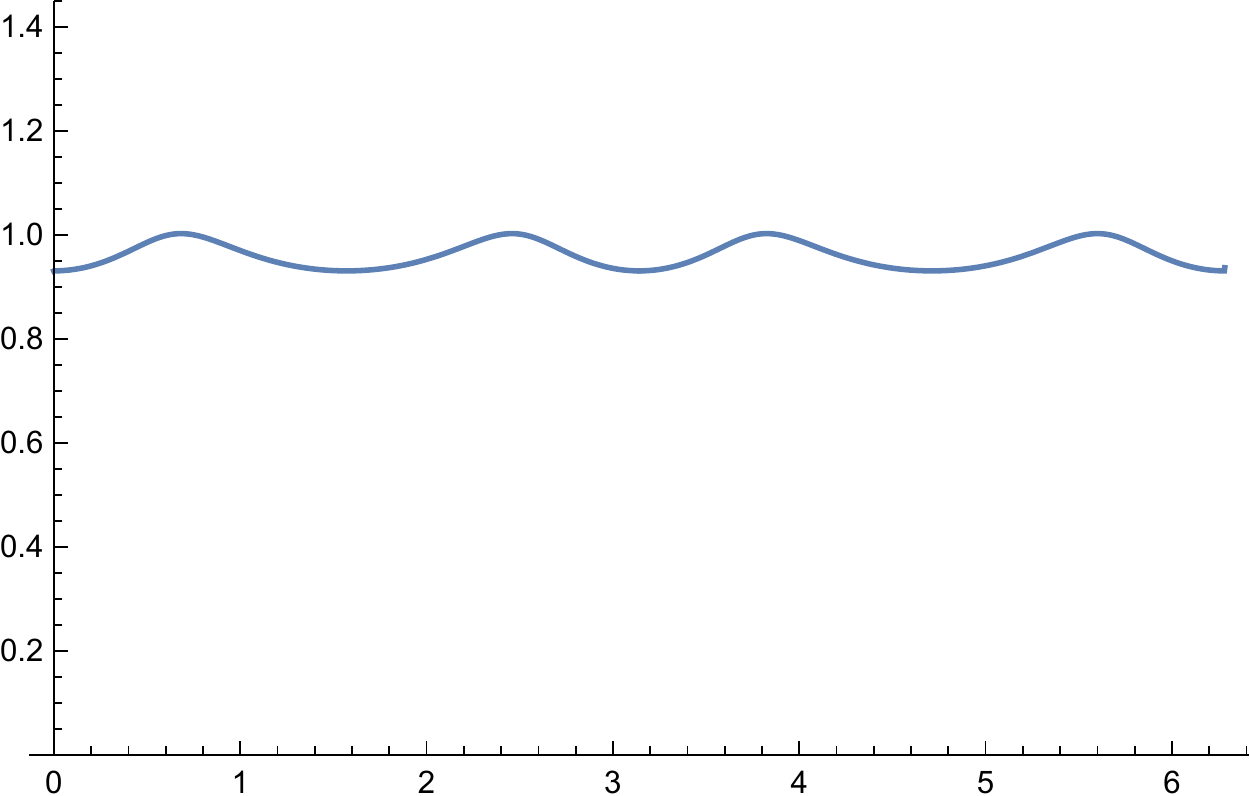}
      \\
    \\
    & \multicolumn{3}{c}{\small\em Dense scale selection with post-smoothing} \\ \\
     & {\small $\Gamma_s = 0$} 
     & {\small $\Gamma_s = \frac{1}{4}$} 
     & {\small $\Gamma_s = \frac{1}{2}$} \\ \\
    {\small $c=1/\sqrt{2}$}
       & \includegraphics[width=0.25\textwidth]{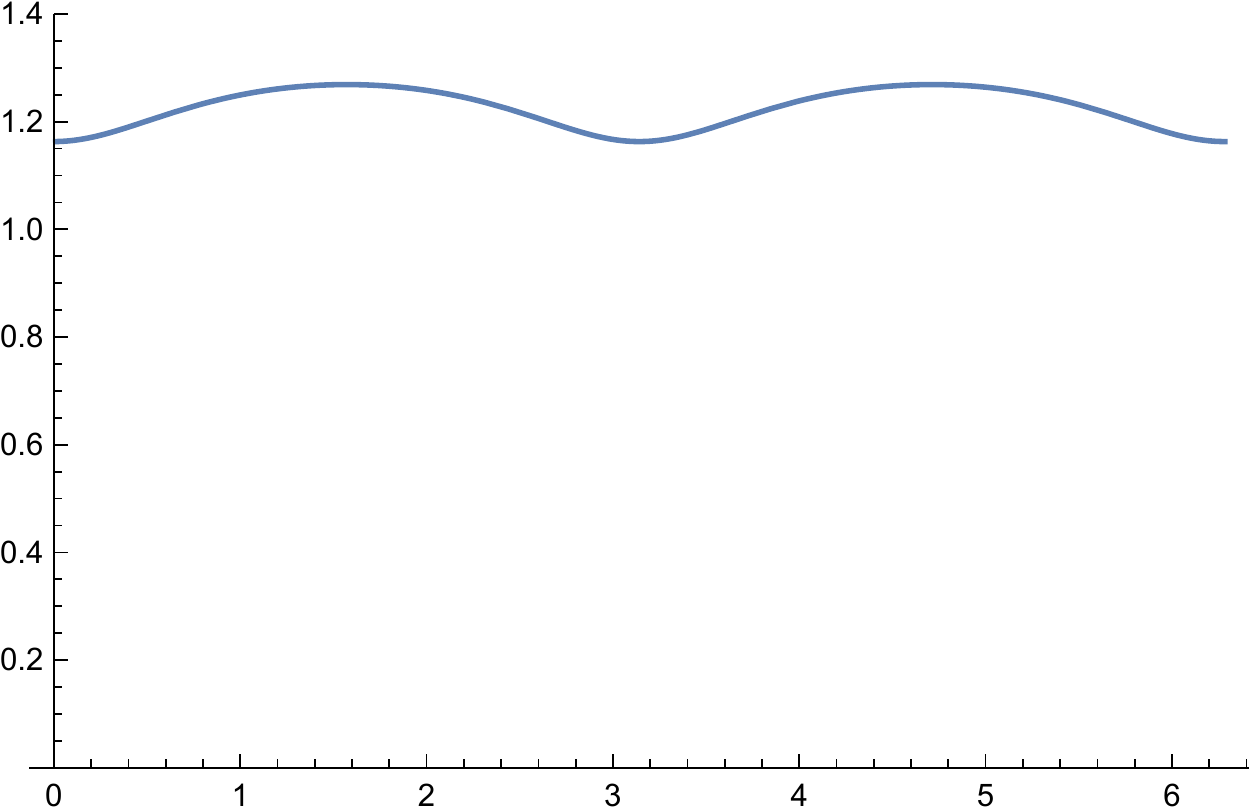} 
       & \includegraphics[width=0.25\textwidth]{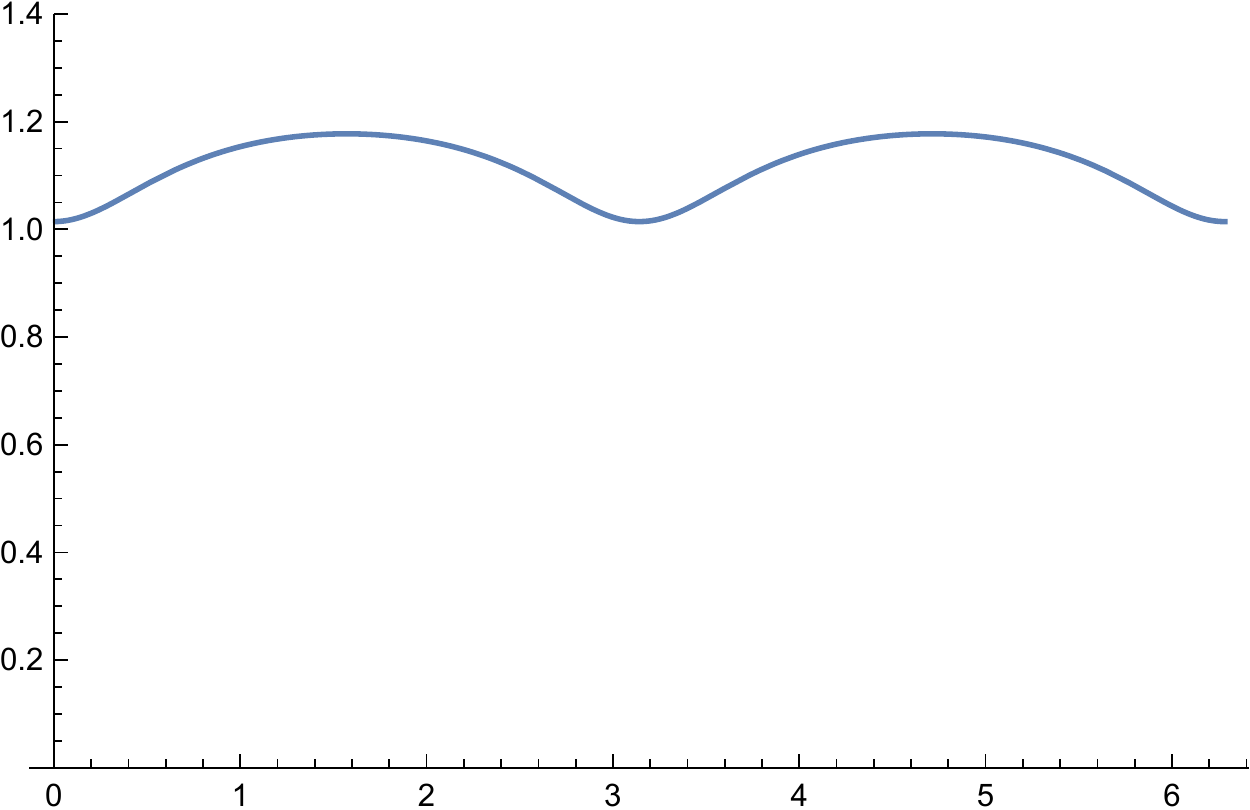} 
       & \includegraphics[width=0.25\textwidth]{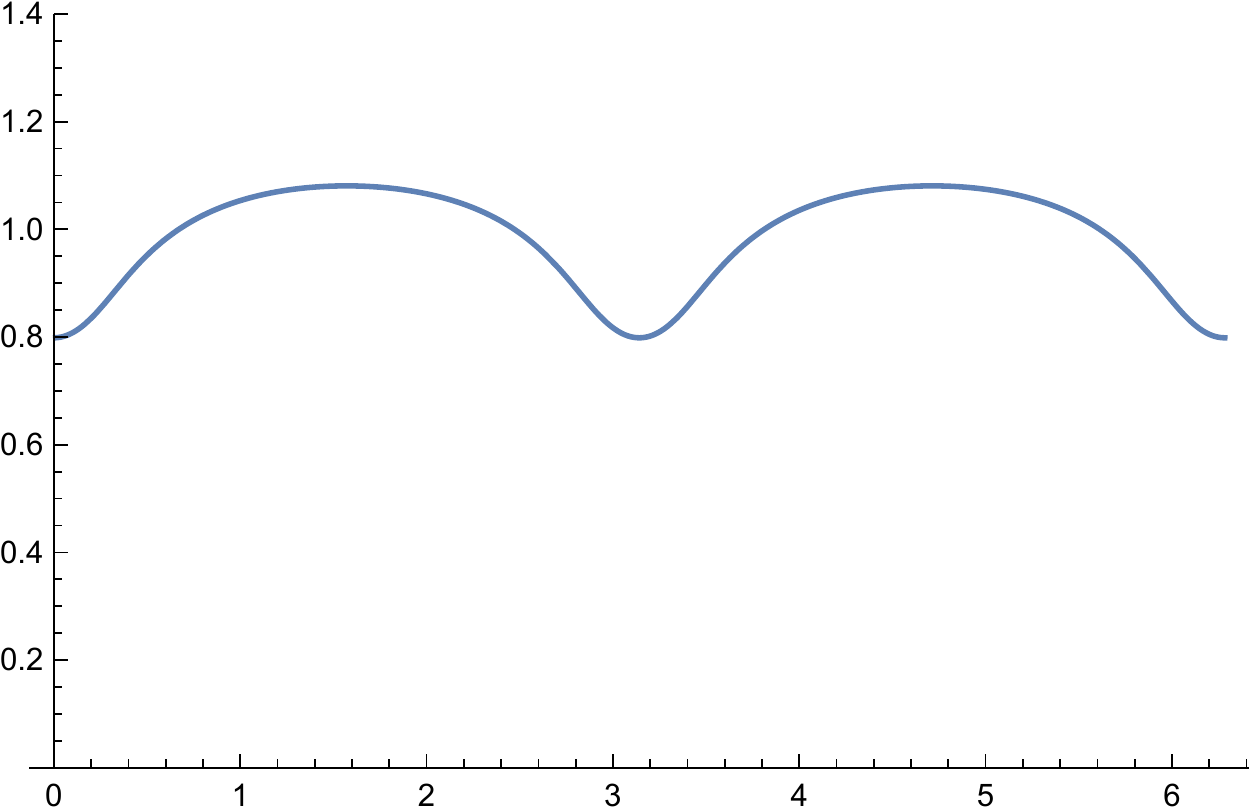} \\
      {\small $c=1$}
       & \includegraphics[width=0.25\textwidth]{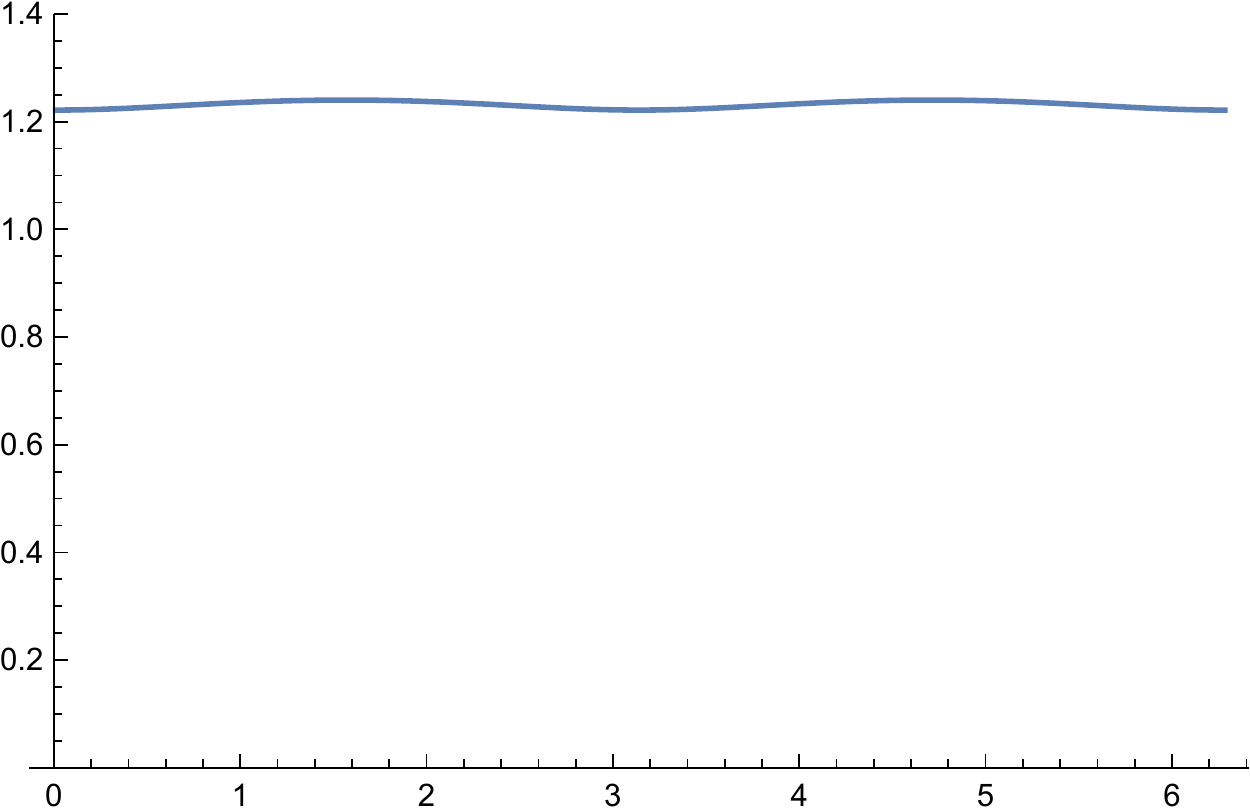} 
       & \includegraphics[width=0.25\textwidth]{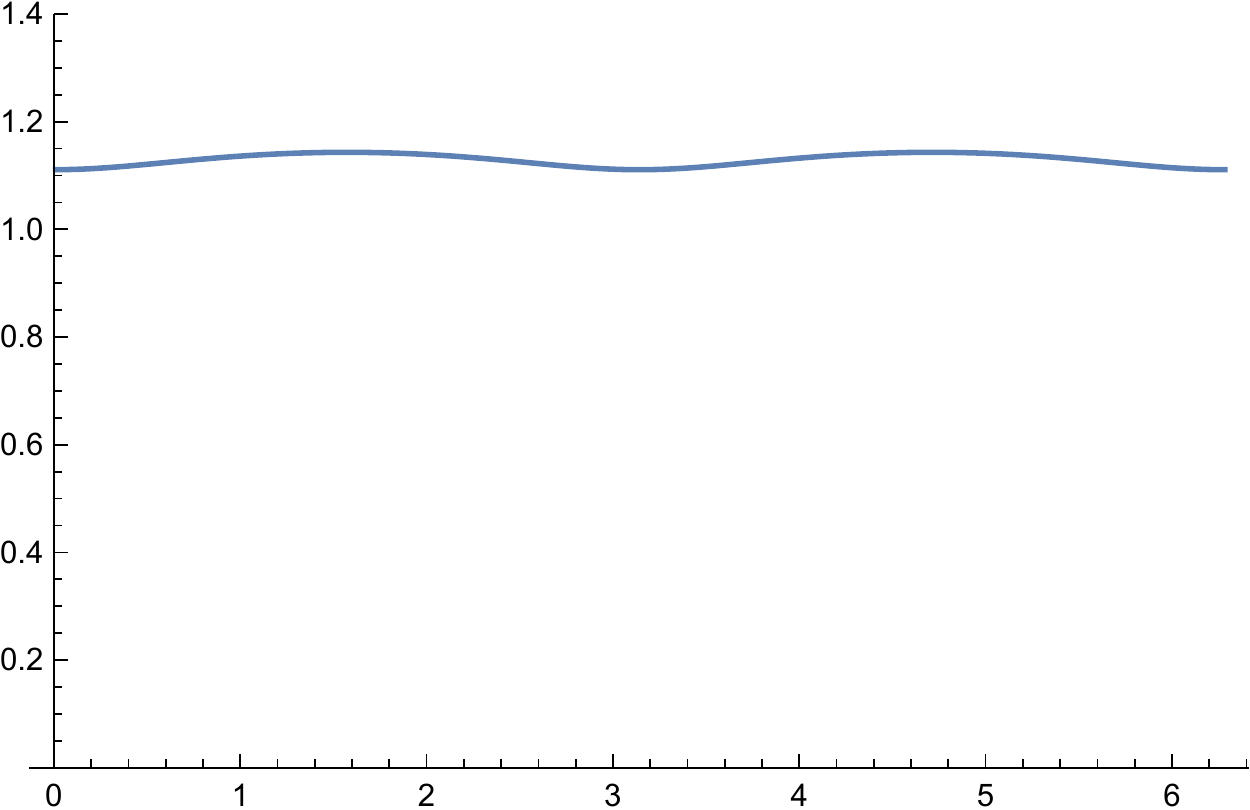} 
       & \includegraphics[width=0.25\textwidth]{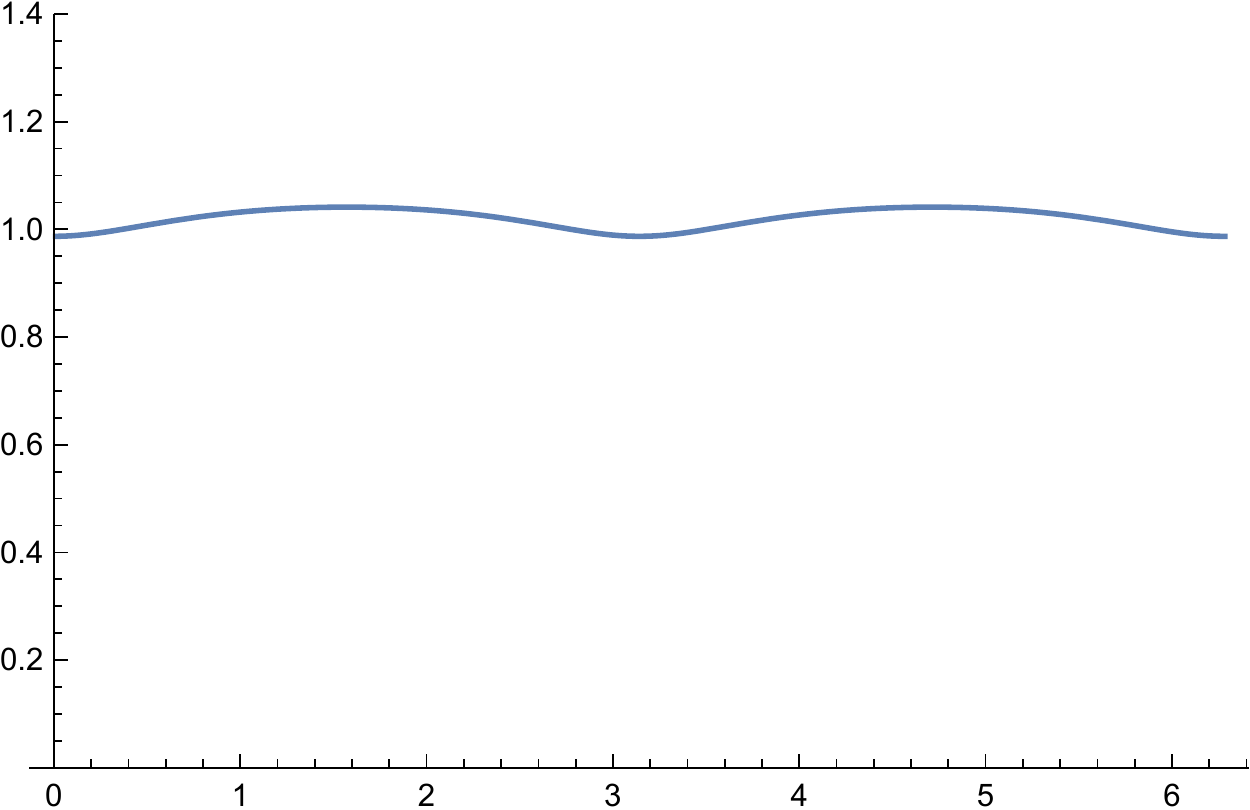} \\
     {\small $c=\sqrt{2}$}
       & \includegraphics[width=0.25\textwidth]{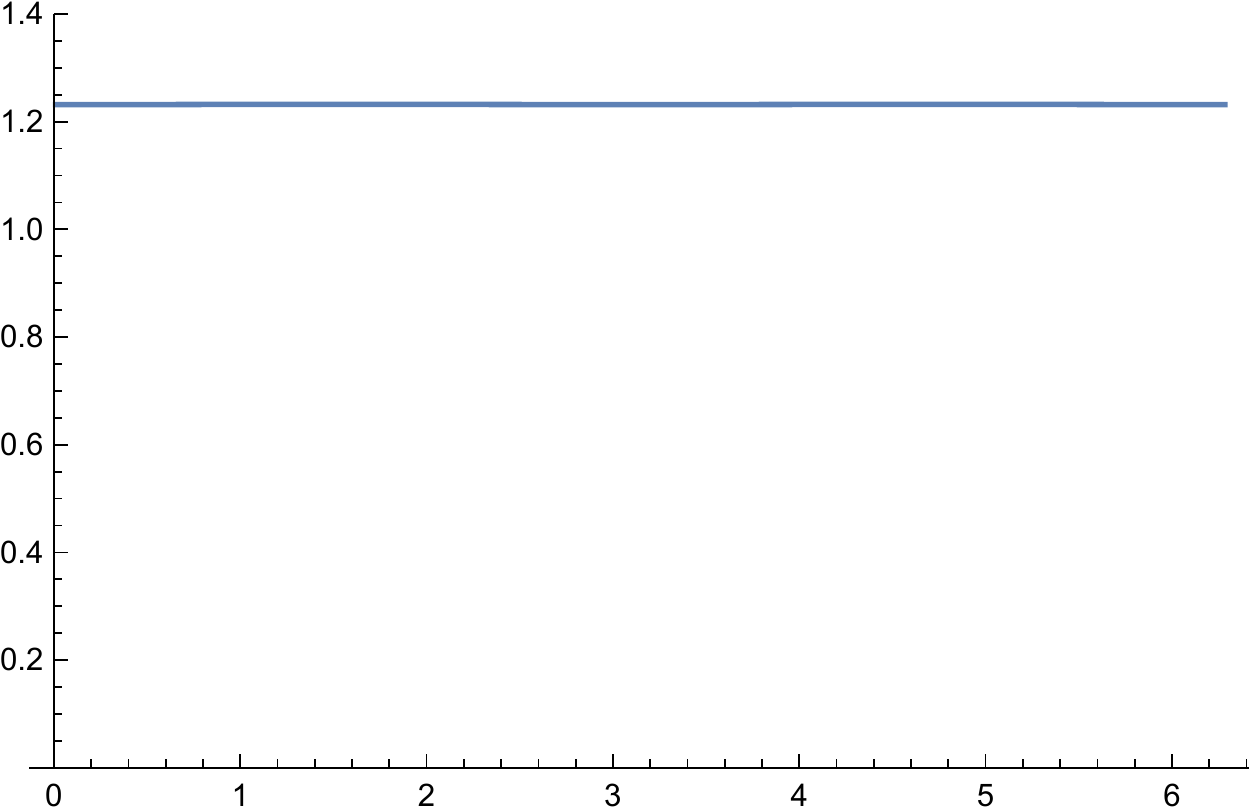} 
       & \includegraphics[width=0.25\textwidth]{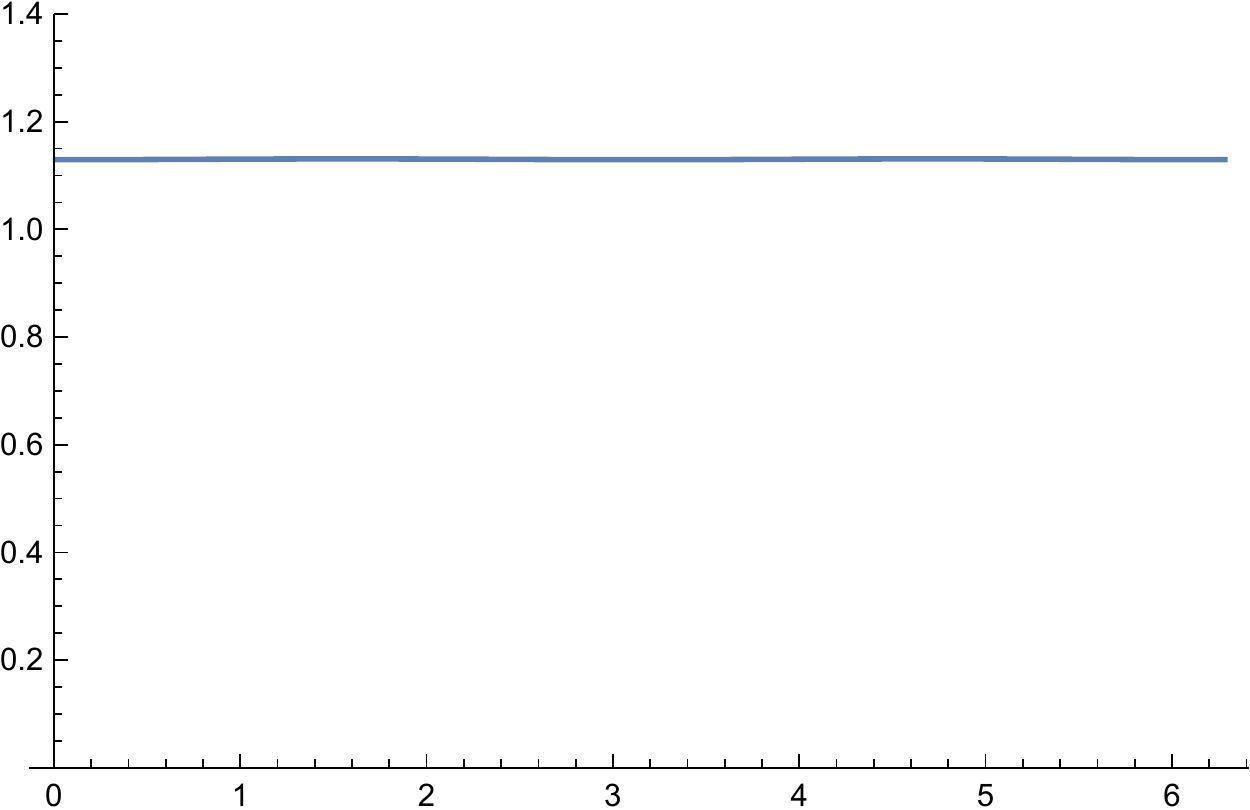} 
       & \includegraphics[width=0.25\textwidth]{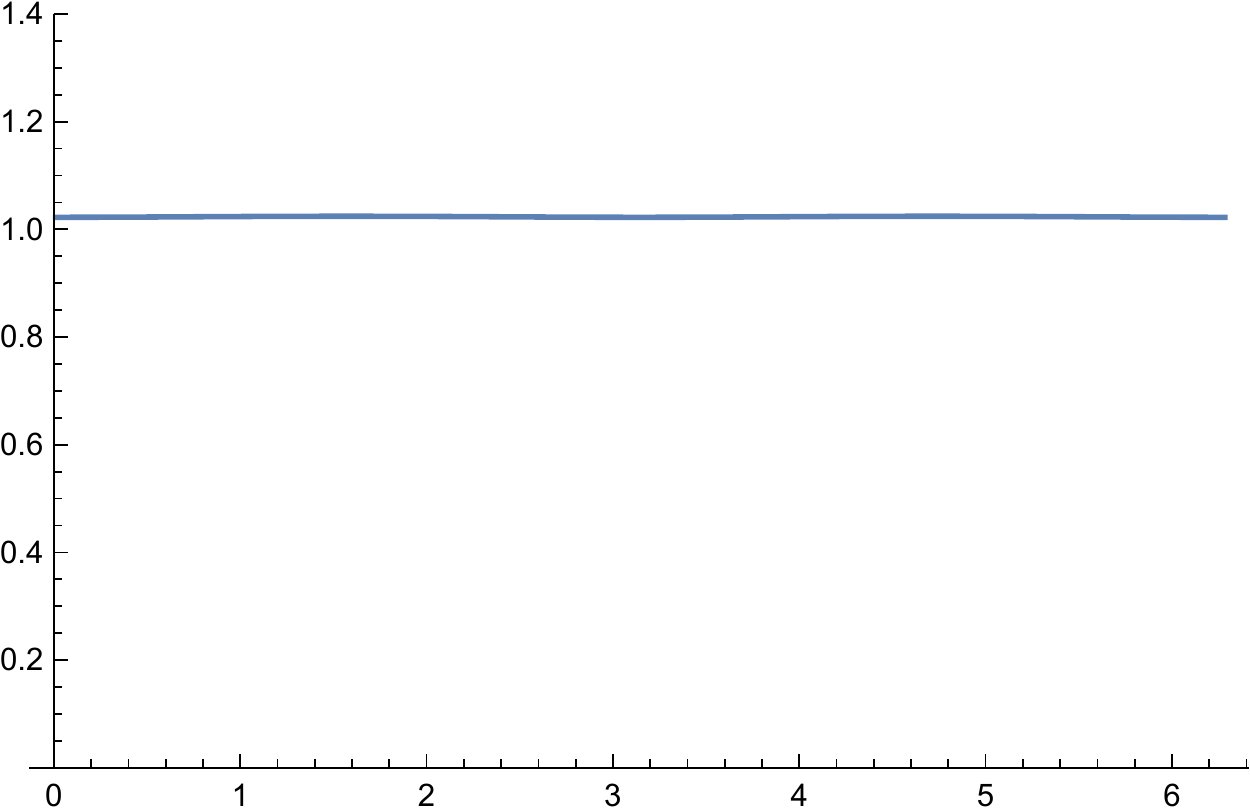} \\ 
     \end{tabular}
  \end{center}
\caption{Spatial variability of the local scale estimates in units of
  $\sigma_s = \sqrt{s}$
   for a 1-D sine wave $f(x) = \sin \omega_0 x$ with angular frequency $\omega_0=1$:
  (top row) regular scale estimates without post-smoothing according
  to (\ref{eq-scale-sel-quasi-quad}),
 (second row) phase-compensated scale estimates on a logarithmic scale
  and without
  post-smoothing according to (\ref{eq-scale-phasecorr-geom-sine-tau}),
  (third, fourth and bottom rows) scale estimates computed from local extrema
  over scale of the post-smoothed quasi quadrature
    entity $\overline{\cal Q}_{(x,y),\Gamma-norm} L$ according to
    (\ref{eq-scale-sel-quasi-quad-post-smooth}) for relative
    post-smoothing scales $c = 1/\sqrt{2}$, $c = 1$ and $c = \sqrt{2}$, respectively.
  (All results have been computed using $C_s =1/\sqrt{(1 - \Gamma_s)(2 - \Gamma_s)}$.)}
  \label{fig-quasi-quad-1D-sinewave-scsel}
\end{figure}

\subsection{Phase-compensated scale estimate}
\label{sec-phase-comp-sc-est-pure-temp-scsp}

Given this understanding of how the scale estimates depend on the local
phase of a sine wave, we can define a {\em phase-compensated scale estimate\/} 
according to either
\begin{align}
  \begin{split}
     \label{eq-scale-corr-sine-2-tau}
     \hat{s}_{{\cal Q} L, comp} = 
     & \frac{\sqrt{(1 - \Gamma_s)(2 - \Gamma_s)}}
                 {\left({\cal Q}_{(x,y),1,\Gamma-norm} L + {\cal Q}_{(x,y),2,\Gamma-norm} L\right)} \times
 \end{split}\nonumber\\
 \begin{split}
     & 
        \left( 
           \frac{{\cal Q}_{(x,y),1,\Gamma-norm} L}{1 - \Gamma_s}  
           + \frac{{\cal Q}_{(x,y),2,\Gamma-norm} L}{2 -\Gamma_s} 
         \right) \,
          \hat{s}_{{\cal Q} L}
  \end{split}
\end{align}
or
\begin{align}
  \begin{split}
     \label{eq-scale-phasecorr-geom-sine-tau}
     & \hat{s}_{{\cal Q} L, comp} = 
\end{split}\nonumber\\
 \begin{split}
      & \frac{\sqrt{(1 - \Gamma_s)(2 - \Gamma_s)}}
                {(1-\Gamma_s)^{\frac{{\cal Q}_{(x,y),1,\Gamma-norm} L}{{\cal Q}_{(x,y),1,\Gamma-norm} L + {\cal Q}_{(x,y),2,\Gamma-norm} L}}
                (2-\Gamma_s)^{\frac{{\cal Q}_{(x,y),2,\Gamma-norm} L}{{\cal Q}_{(x,y),1,\Gamma-norm} L + {\cal Q}_{(x,y),2,\Gamma-norm} L}}}
        \, \hat{s}_{{\cal Q} L}.
  \end{split}\nonumber\\
\end{align}
\vspace{-8mm}

\noindent
These expressions are
defined to be equal to the geometric average  
\begin{equation}
  \label{eq-center-scale-spat-phase-comp}
  \hat{s}_{geom} = \sqrt{\hat{s}_1 \, \hat{s}_2} = \frac{\sqrt{(1-\Gamma_s)(2-\Gamma_s)}}{\omega_0^2}
\end{equation}
of the extreme values when
only one of the first- or second-order components in ${\cal
  Q}_{\Gamma-norm} L$ responds and using blending
of these responses
by transfinite interpolation \cite{DykFlo09-CAGD} using the relative
strengths of the first- and second-order responses, respectively, to
achieve a much lower variability of the scale estimates in between (see 
Figure~\ref{fig-quasi-quad-1D-sinewave-scsel}).

The motivation to the definitions of equations 
 (\ref{eq-scale-corr-sine-2-tau}) and (\ref{eq-scale-phasecorr-geom-sine-tau}) is to
   express interpolation functions on a simple form that compensate
   for the phase
   dependency of the scale estimates depending on the relative
   strengths of the first- and second-order components. 
   When only the first-order component responds and the second-order
   component is zero, the scale estimate is $\hat{s}_1 = (1 - \Gamma_s)/\omega_0^2$.
   When only the second-order component responds and the first-order
   component is zero, the scale estimate is $\hat{s}_2 = (2 - \Gamma_s)/\omega_0^2$.
   
   In the first expression (\ref{eq-scale-corr-sine-2-tau}), the ratios 
   $w_1 = {\cal Q}_{(x,y),1,\Gamma-norm} L/({\cal Q}_{(x,y),1,\Gamma-norm} L
   + {\cal Q}_{(x,y),2,\Gamma-norm} )$ and
   $w_2 = {\cal Q}_{(x,y),2,\Gamma-norm} L/({\cal Q}_{(x,y),1,\Gamma-norm} L
   + {\cal Q}_{(x,y),2,\Gamma-norm} )$, which are positive and sum to
   one $w_1 + w_2 = 1$, are used as relative weights in a 
   a linear convex combination 
   \begin{equation}
      \hat{s}_{comp} 
      = s_{geom} 
          \left( w_1 \, \frac{\hat{s}}{\hat{s}_1} + w_2 \, \frac{\hat{s}}{\hat{s}_2} 
          \right)
   \end{equation}
   defined such that $\hat{s}_{comp} = s_{geom}$ if either 
   $(w_1 = 1, w_2 = 0, \hat{s} = \hat{s}_1)$ or $(w_1 = 0, w_2 = 1,
   \hat{s} = \hat{s}_2)$.
   In the second expression (\ref{eq-scale-phasecorr-geom-sine-tau}), the same ratios 
   $w_1$ and
   $w_2$ are used as relative weights in a 
   a geometric convex combination 
   \begin{equation}
     \hat{s}_{comp} 
     = s_{geom} 
         \left( \frac{\hat{s}}{\hat{s}_1} \right)^{w_1} 
         \left( \frac{\hat{s}}{\hat{s}_2} \right)^{w_2}
   \end{equation}
    again defined such that $\hat{s}_{comp} = s_{geom}$ if either 
   $(w_1 = 1, w_2 = 0, \hat{s} = \hat{s}_1)$ or $(w_1 = 0, w_2 = 1, \hat{s} = \hat{s}_2)$.

For the
first expression (\ref{eq-scale-corr-sine-2-tau}), the blending is thus
performed on a linear scale with respect to the spatial scale
parameter, whereas the blending is performed on a more natural
logarithmic scale in the second expression (\ref{eq-scale-phasecorr-geom-sine-tau}).

From these spatial scale estimates, we can in turn estimate the
temporal wavelength of the sine wave according to
\begin{equation}
  \label{eq-lambda-from-sigma}
  \hat{\lambda} = \frac{2 \pi \sqrt{\hat{s}_{{\cal Q}_L, comp}}}{\sqrt[4]{(1 - \Gamma_s)(2 - \Gamma_s)}}.
\end{equation}
Whereas the interpolation functions (\ref{eq-scale-corr-sine-2-tau})
and (\ref{eq-scale-phasecorr-geom-sine-tau}) in combination with
(\ref{eq-lambda-from-sigma}) will not lead to exact wavelength
estimates for all phases of a sine wave, these functions compensate for the gross behaviour of the phase
dependency, which substantially decreases the otherwise much higher spatial variability in
the spatial scale estimates.

%A more detailed analysis of the case when spatial smoothing is
%combined with local phase compensation is given in the supplement. 

\subsection{Scale calibration}
\label{sec-sc-calib-spat-dense-scsel}

In
Sections~\ref{eq-sc-sel-props}--\ref{sec-phase-comp-sc-est-pure-temp-scsp}
as well as a more detailed analysis in the appendix, it is shown how the scale estimates
$\hat{s}_{{\cal Q} L}$ and $\hat{s}_{\overline{\cal Q} L}$ according
to (\ref{eq-scale-sel-quasi-quad}) and
(\ref{eq-scale-sel-quasi-quad-post-smooth}) are influenced by the parameters
$\Gamma_s$, $c$ and $C_s$ in the quasi quadrature
measure. 
To decouple this dependency from the later stage visual modules for
which the dense scale selection methodology is intended to be used as an initial
pre-processing stage, we introduce the notion of 
{\em scale calibration\/}, which implies that the scale estimates are
to be multiplied by uniform scaling factors such that they are either:
\begin{itemize}
\item[(i)]
  equal to the scale estimate $\hat{s} = s_0$ obtained by applying the regular
  scale-normalized Laplacian $\nabla_{norm}^2 L = s \, (L_{xx} + L_{yy})$ 
  or the scale-normalized determinant of the Hessian 
  $\det {\cal H}_{norm} L = s^2 \, (L_{xx} L_{yy} - L_{xy}^2)$ at the
  center of a Gaussian blob of any spatial extent $s_0$ or
\item[(ii)]
  equal to the scale estimate $\hat{s} = \sqrt{2}/\omega_0^2$
  corresponding to the geometric average of the scale
  estimates obtained for a sine wave of any angular frequency
  $\omega_0$ when $\Gamma_s = 0$.
\end{itemize}
The first method, which aims at similarity with previous
scale selection methods at sparse image features, will be referred
to as {\em Gaussian scale calibration\/}, whereas the second method,
which aims at similarity for dense texture patterns,
will be referred to as {\em sine wave scale calibration\/}.

The necessary calibration factors can for the cases of either (i)~no
phase compensation and no post-smoothing or (ii)~phase compensation
without post-smoothing be obtained from the
theoretical results in section~\ref{eq-sc-sel-props}.
Ways of deriving the scale calibration factors when using spatial
post-smoothing are described in the appendix.

\subsection{Composed dense scale selection algorithms}

Given the above treatment, we can define four types of dense scale selection
algorithms:
\begin{description}
\item[Algorithm~I:] 
  Without post-smoothing or phase compensation, with local scale estimates
  at every image point computed according to (\ref{eq-scale-sel-quasi-quad}).
\item[Algorithm~II:] 
  With phase compensation and without post-smoothing, with local
  scale estimates at every image point computed according to
  (\ref{eq-scale-corr-sine-2-tau}) or
  (\ref{eq-scale-phasecorr-geom-sine-tau}) based on uncompensated local scale
  estimates according to (\ref{eq-scale-sel-quasi-quad}).
\item[Algorithm~III:]
  With post-smoothing and without phase compensation, with local scale estimates
  at every image point computed according to (\ref{eq-scale-sel-quasi-quad-post-smooth}).
\item[Algorithm~IV:]
  With both post-smoothing and phase compensation,  with local 
  phase-compensated scale estimates at every image point computed in
  an analogous way as (\ref{eq-scale-corr-sine-2-tau}) or
  (\ref{eq-scale-phasecorr-geom-sine-tau}) although based
  on post-smoothing according
  (\ref{eq-scale-sel-quasi-quad-post-smooth}) and with the factors $(1 - \Gamma_s)$
  and $(2 - \Gamma_s)$ in the expressions for phase compensation, which originate from the local extrema over scale
  when $c = 0$, replaced by $S_{sine,1}(\Gamma_s, c, C_s)$ and
  $S_{sine,2}(\Gamma_s, c, C_s)$ according to
  (\ref{eq-smin-smax-postsmooth}) in the appendix.
\end{description}
The scale estimates from each algorithm can in turn be calibrated using
either Gaussian scale calibration or sine wave calibration according
to Section~\ref{sec-sc-calib-spat-dense-scsel}.

\subsection{Scale covariance of the spatial scale estimates under
  spatial scaling transformations}
\label{sec-cov-prop-spat-sc-sel}

Consider a scaling transformation of the spatial image domain
\begin{equation}
f'(x_1', x_2') = f(x_1, x_2)
%\end{equation}
\quad\quad \mbox{for} \quad\quad
%\begin{equation}
(x_1', x_2') = (S_s \, x_1, S_s \, x_2),
\end{equation}
where $S_s$ denotes the spatial scaling factor.
Define the spatial scale-space representations $L$ and $L'$ 
of $f$ and $f'$, respectively, according to
\begin{align}
  \begin{split}
      & L(x_1, x_2;\; s) =
%  \end{split}\nonumber\\
%  \begin{split}
       %\quad\quad
         \left( T(\cdot, \cdot;\; s) * f(\cdot, \cdot,) \right)(x_1, x_2;\; s),
   \end{split}\\
  \begin{split}
      & L'(x'_1, x'_2;\; s')  = 
%  \end{split}\nonumber\\
 % \begin{split}
     %\quad\quad
     \left( T(\cdot, \cdot;\; s') * f'(\cdot, \cdot) \right)(x'_1, x'_2;\; s').
  \end{split}
\end{align}
Consider a spatial differential expression of the form
\begin{equation}
  \label{eq-sc-norm-hom-spat-diff-expr}
   {\cal D} L 
   = \sum_{i=1}^I \prod_{j=1}^J c_i \, L_{x^{\alpha_{ij}}}
   = \sum_{i=1}^I \prod_{j=1}^J c_i \, L_{x_1^{\alpha_{1ij}}  x_2^{\alpha_{2ij}}},
\end{equation}
required to be {\em homogeneous\/} in the sense that the sum of the orders of
differentiation in each term does not depend on the index of that term
\begin{equation}
   \sum_{j=1}^J |\alpha_{ij}| = \sum_{j=1}^J \alpha_{1ij} + \alpha_{2ij} = M.
\end{equation}
Then, the corresponding homogeneous differential expression ${\cal D}_{norm} L$ 
with the spatial
derivatives $L_{x_1^{m_1} x_2^{m_2}}$ replaced by scale-normalized
derivatives 
according to 
\begin{equation}
  \label{eq-sc-norm-part-der}
  L_{\xi_1^{m_1} \xi_2^{m_2}} 
  = s^{(m_1 + m_2) \gamma_s/2} \, L_{x_1^{m_1} x_2^{m_2}},
\end{equation}
transforms according to (Lindeberg \cite[Equation~(25)]{Lin97-IJCV})
\begin{equation}
  \label{eq-spat-scale-cov-hom-diff-inv}
   {\cal D'}_{norm} L' 
   = S_{s}^{M(\gamma_s - 1)} \, {\cal D}_{norm} L.
\end{equation}
Regarding the spatial quasi quadrature measure 
${\cal Q}_{(x, y),\Gamma-norm} L$ according to  (\ref{eq-quasi-quad-scale-mod})
that we use for dense spatial scale selection, this 
differential invariant is not homogeneous of the form
(\ref{eq-sc-norm-hom-spat-diff-expr}).
If we split this differential expression into two components based on
the orders of spatial differentiation
\begin{align}
  \begin{split}
     {\cal Q}_{(x, y),\Gamma-norm} L
     = & {\cal Q}_{(x, y), 1,\Gamma-norm} L
%  \end{split}\nonumber\\
%  \begin{split}
      + {\cal Q}_{(x, y), 2,\Gamma-norm} L
  \end{split}
\end{align}
where
\begin{align}
 \begin{split}
     {\cal Q}_{(x, y), 1,\Gamma-norm} L 
     & = \frac{s \,  (L_{x}^2 + L_{y}^2)}{s^{\Gamma_s}},
 \end{split}\\
  \begin{split}
     {\cal Q}_{(x, y), 2,\Gamma-norm} L 
     & = \frac{C_s \, s^2 \, \left( L_{xx}^2 + 2 L_{xy}^2 + L_{yy}^2 \right) }{s^{\Gamma_s}},
 \end{split}
\end{align}
 we can note that each one of these expressions is of the homogeneous
form 
(\ref{eq-sc-norm-hom-spat-diff-expr}) and corresponds to
$\gamma$-normalized scale-space derivatives in the respective cases:
\begin{equation}
     \gamma_{1} = 1 - \Gamma_s, \quad\quad
     \gamma_{2} = 1 - \frac{\Gamma_s}{2}.
\end{equation}
Applying the transformation property
(\ref{eq-spat-scale-cov-hom-diff-inv}) to each of the two
components of the spatial quasi quadrature measure, then gives
that they transform according to
\begin{align}
  \begin{split}
     {\cal Q}_{(x', y'), 1,\Gamma-norm} L'
%  \end{split}\nonumber\\
%  \begin{split}
     &  = S_s^{2(\gamma_1-1)} \, 
            {\cal Q}_{(x, y), 1,\Gamma-norm} L
%  \end{split}\nonumber\\
%  \begin{split}
      = S_s^{-2 \Gamma_s} \, 
            {\cal Q}_{(x, y), 1,\Gamma-norm} L,
  \end{split}\\
  \begin{split}
     {\cal Q}_{(x', y'), 2,\Gamma-norm} L'
%  \end{split}\nonumber\\
%  \begin{split}
     & = S_s^{2 \times 2(\gamma_2-2)} \, 
            {\cal Q}_{(x, y), 2,\Gamma-norm} L
%  \end{split}\nonumber\\
% \begin{split}
      = S_s^{-2 \Gamma_s} \, 
            {\cal Q}_{(x, y), 2,\Gamma-norm} L.
  \end{split}
\end{align}
In other words, because of the deliberate adding of %spatial
differential expressions corresponding to different orders of spatial 
differentiation for the maximally scale invariant case of 
$\gamma_s = 1$ prior to post-normalization by
the post-normalization power $\Gamma_s$, it
follows that
the two components transform in the same way under spatial
scaling transformations, implying that the composed %spatial
quasi quadrature measure transforms as
\begin{equation}
   ({\cal Q}_{(x', y'), \Gamma-norm} L')(x', y';\; s')
   = S_s^{-2 \Gamma_s} \, 
       ({\cal Q}_{(x, y), \Gamma-norm} L) (x, y;\; s)
\end{equation}
under uniform scaling transformations of the spatial image domain.

This covariance property under spatial scaling transformations does specifically imply that local extrema over
spatial scales are preserved under uniform scaling
transformations of the spatial image domain and are
transformed in a scale-covariant way
\begin{equation}
   \hat{s}' = S_s^2 \, \hat{s}
\end{equation}
or in units of the standard deviation 
$\sigma_s = \sqrt{s}$ of the spatial
scale-space kernel
\begin{equation}
   \hat{\sigma}'_s
  = S_s \, \hat{\sigma}_s.
\end{equation}
This property constitutes the theoretical foundation for dense
spatial scale selection and implies that the local
spatial scale estimates will automatically adapt to local
variations in the dominant spatial scales in the image data.

This scale covariance of the spatial scale estimates does also
extend to phase-compensated scale estimates according
to (\ref{eq-scale-corr-sine-2-tau}) or (\ref{eq-scale-phasecorr-geom-sine-tau}).
This property is straightforward to prove, since the underlying
uncompensated spatial scale estimates
$\hat{s}_{{\cal Q} L}$ in (\ref{eq-scale-corr-sine-2-tau}) or (\ref{eq-scale-phasecorr-geom-sine-tau})
are provably scale
covariant and additionally the ratio that determines the scale
compensation factor is invariant under independent scaling
transformations of the spatial domain provided that
the spatial scale levels are appropriately matched,
which they are if the phase compensation factors are computed at scale
levels corresponding to the spatial scale estimates.

Correspondingly, the scale covariance of the spatial scale
estimates also extends to spatial post-smoothing prior to
the detection of local extrema over spatial scales.
This property follows from the fact that the amount of spatial 
post-smoothing is proportional to the spatial scale level
at which the non-linear quasi quadrature measure
is computed.

Under affine intensity transformations 
\begin{equation}
  f'(x, y) = a \, f(x, y) + b, 
\end{equation}
the Gaussian derivatives are multiplied by a uniform scaling factor
$L'_{x^\alpha y^\beta}(x, y;\; t) = a \, L_{x^\alpha y^\beta}(x, y;\; t)$ 
and the quasi quadrature measure transforms
according to
\begin{equation}
  {\cal Q}'_{(x,y),\Gamma-norm}(x, y;\; s) = a^2 \, {\cal Q}_{(x,y),\Gamma-norm}(x, y;\; s).
\end{equation}
The scale estimates are therefore unaffected by 
illumination variations whose effects
can be well approximated by local affine transformations over the
intensity domain.

% If the scale-space representation of the underlying grey-level image
% is in turn defined relative to a logarithmic brightness scale 
% $f \sim \log I$, where $I$ represents the incoming energy, then the
% receptive field responses of the first- and second-order Gaussian
% derivatives will be invariant under local multiplicative intensity
% transformations \cite[Section~2.3]{Lin13-BICY} implying that the quasi
% quadrature measures will also be invariant under local multiplicative
% intensity transformations.

The spatial quasi quadrature entities used for scale selection are
based on the rotationally
invariant differential invariants $| \nabla L |^2$ and $\| {\cal H} L \|^2$ 
and are therefore rotationally invariant. This implies that the
resulting scale spatial estimates are covariant under rotations of the
spatial image domain.

\begin{figure*}[hbtp]
  \begin{center}
    \begin{tabular}{cc}
       {\footnotesize\em Original image} 
       & {\footnotesize\em Dense scale estimates using Algorithm II} \\
      \includegraphics[width=0.40\textwidth]{tiger2-vert-eps-converted-to} 
     & \includegraphics[width=0.40\textwidth]{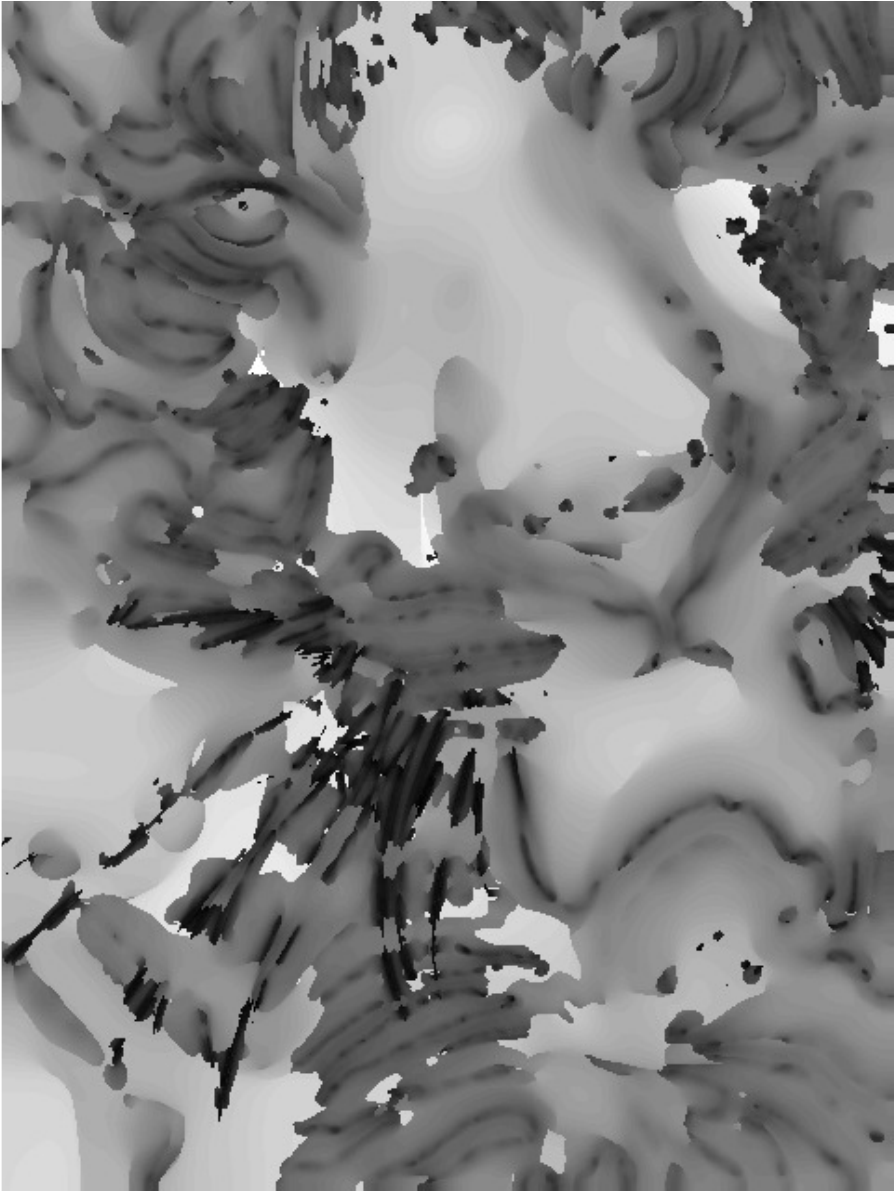} \\
        \includegraphics[width=0.46\textwidth]{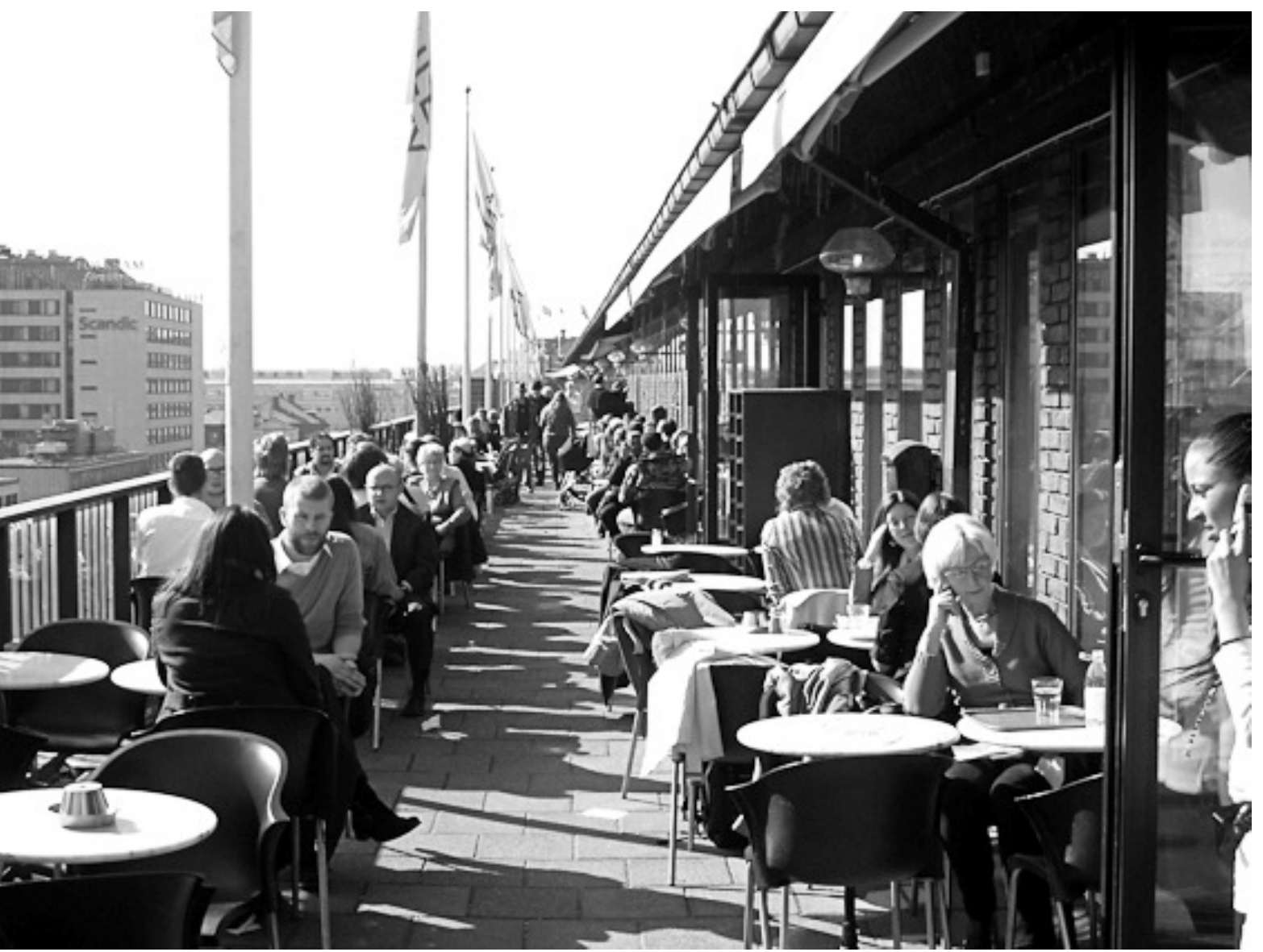} 
       & \includegraphics[width=0.46\textwidth]{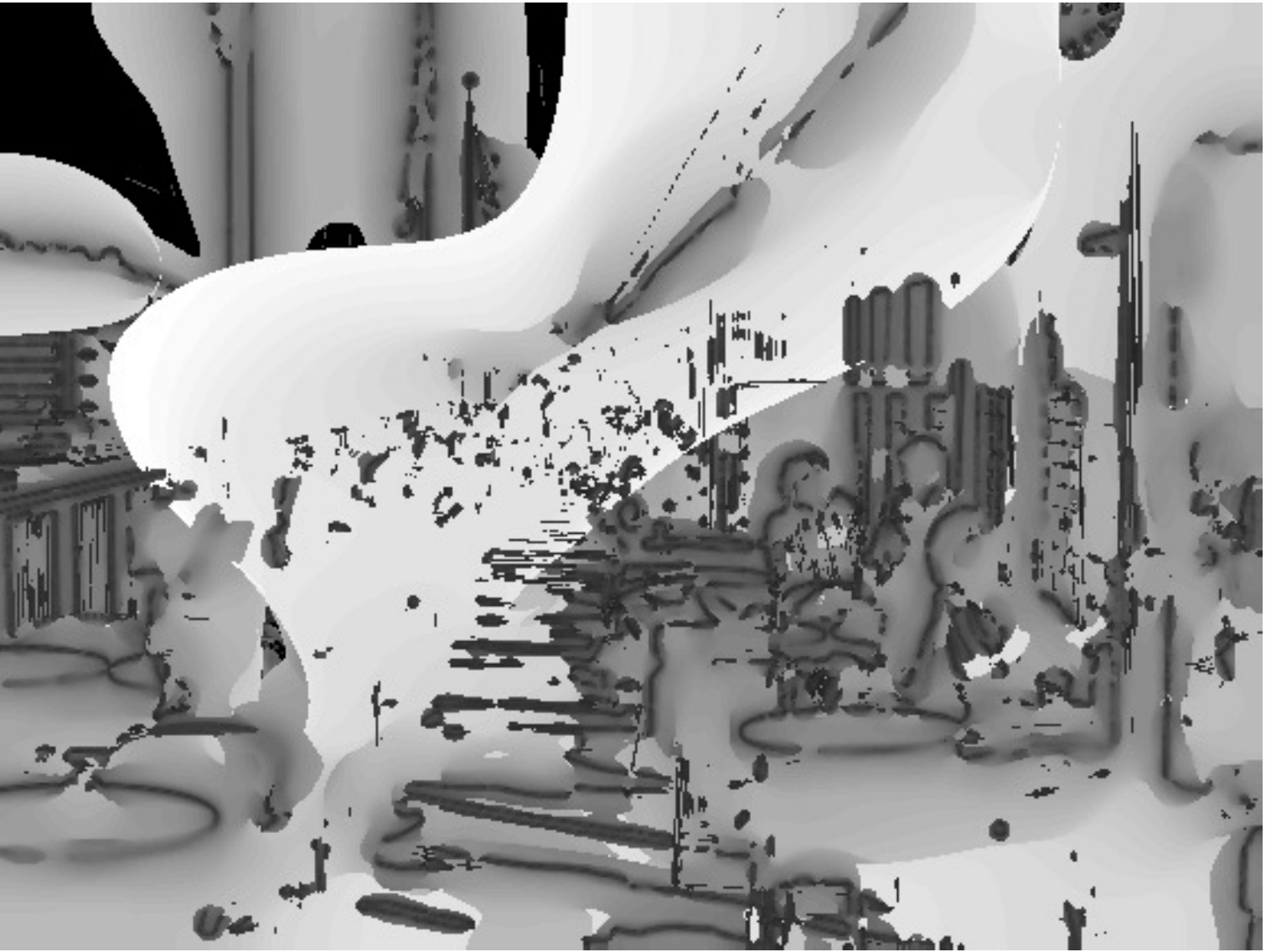} \\
       \includegraphics[width=0.46\textwidth]{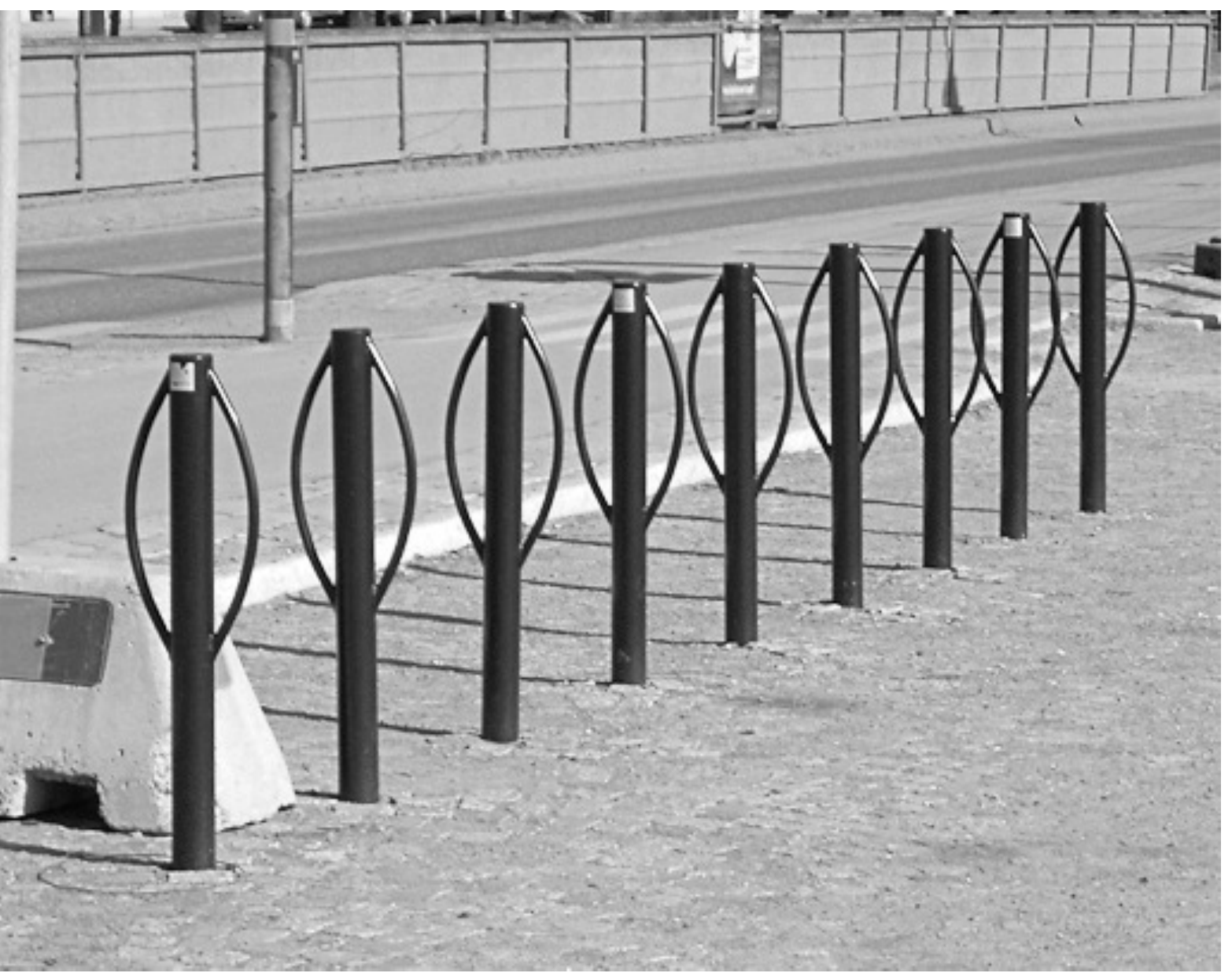}  
       & \includegraphics[width=0.46\textwidth]{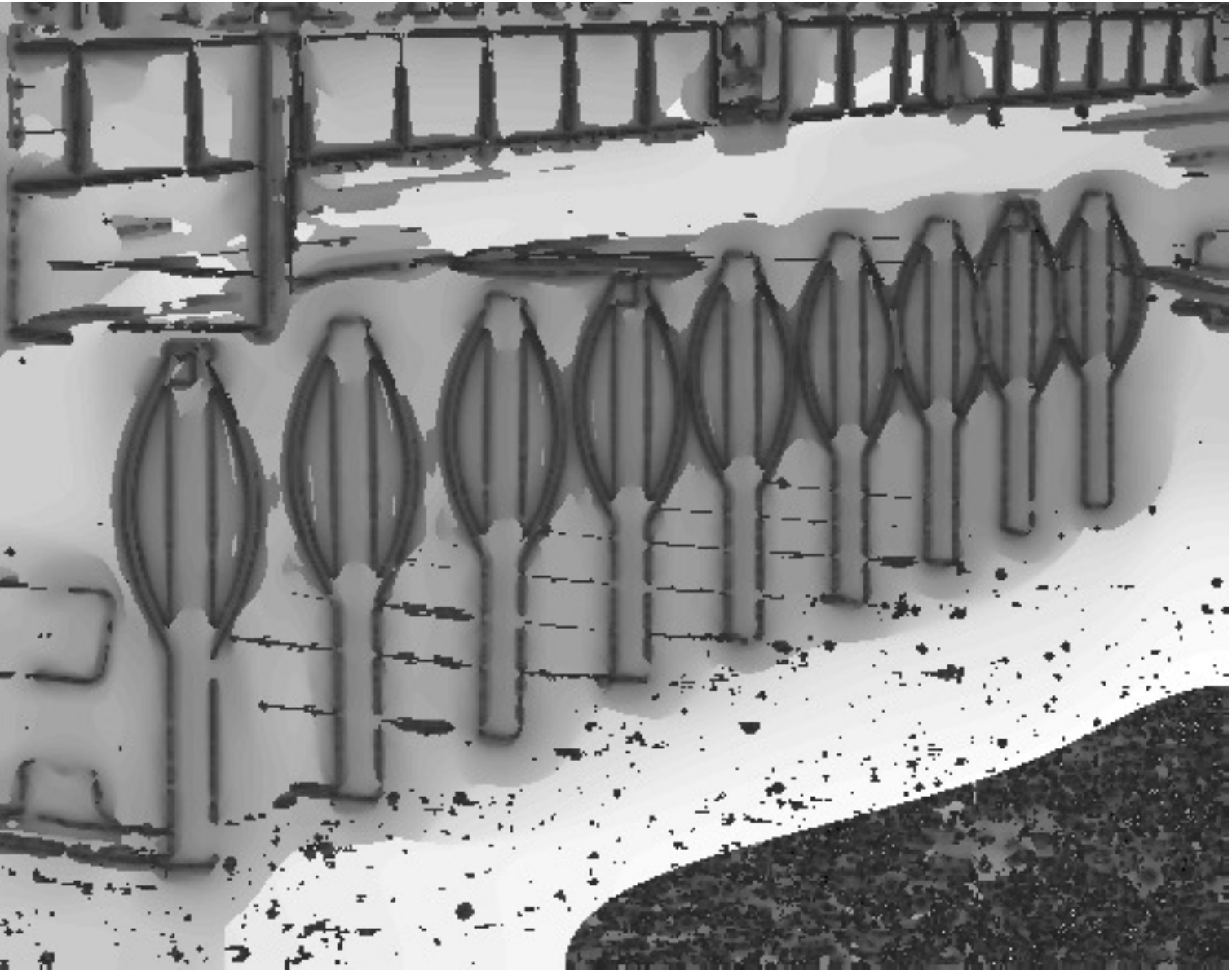} \\
  \end{tabular}
  \end{center}
\caption{Dense spatial scale maps computed using Algorithm~II
   (dense scale selection with phase compensation)
   for three different images using $\Gamma_s = 1/4$ and
   $C_s = 1/\sqrt{(1 - \Gamma_s)(2 - \Gamma_s)}$.
    The grey-levels code for effective scale
    approximated by $s_{eff} = \log_2(s_0 + s)$ for $s_0 = 1/8$ in
    such a way that darker means finer spatial scales and brighter
    indicates coarser spatial scales. See
    Section~\ref{sec-exp-dense-spat-scsel} for more detailed explanations.
    Observe, however, that for the quite common phenomenon when there are multiple local extrema
    over scale corresponding to different types of dominant structures
    at different scales, this visualization only shows the one of the scale estimates
    that has the strongest maximum response. Thereby, 
    situations where the maximum value over scales switches between
    two scale selection surfaces at difference scales appear as
    discontinuities in this simplified form of visualization.
    Such layer discontinuities are therefore artefacts of the
    visualization method --- not the scale selection method.
    A more appropriate form of visualization is in terms of a 3-D visualization
    of the scale selection surfaces as shown in
    Figure~\ref{fig-terass-scale-sel-surfaces}, where multiple scale
    estimates may be displayed at every image point.}
    %  instead of
    % only the scale estimate that has the strongest magnitude response).
  % (In this figure,)}
  \label{fig-terass-cykelstall-scmaps-maxresp}
\end{figure*}

\begin{figure*}[hbtp]
  \begin{center}
    \begin{tabular}{c}
     {\small\em Original image} \\
     \includegraphics[width=0.54\textwidth]{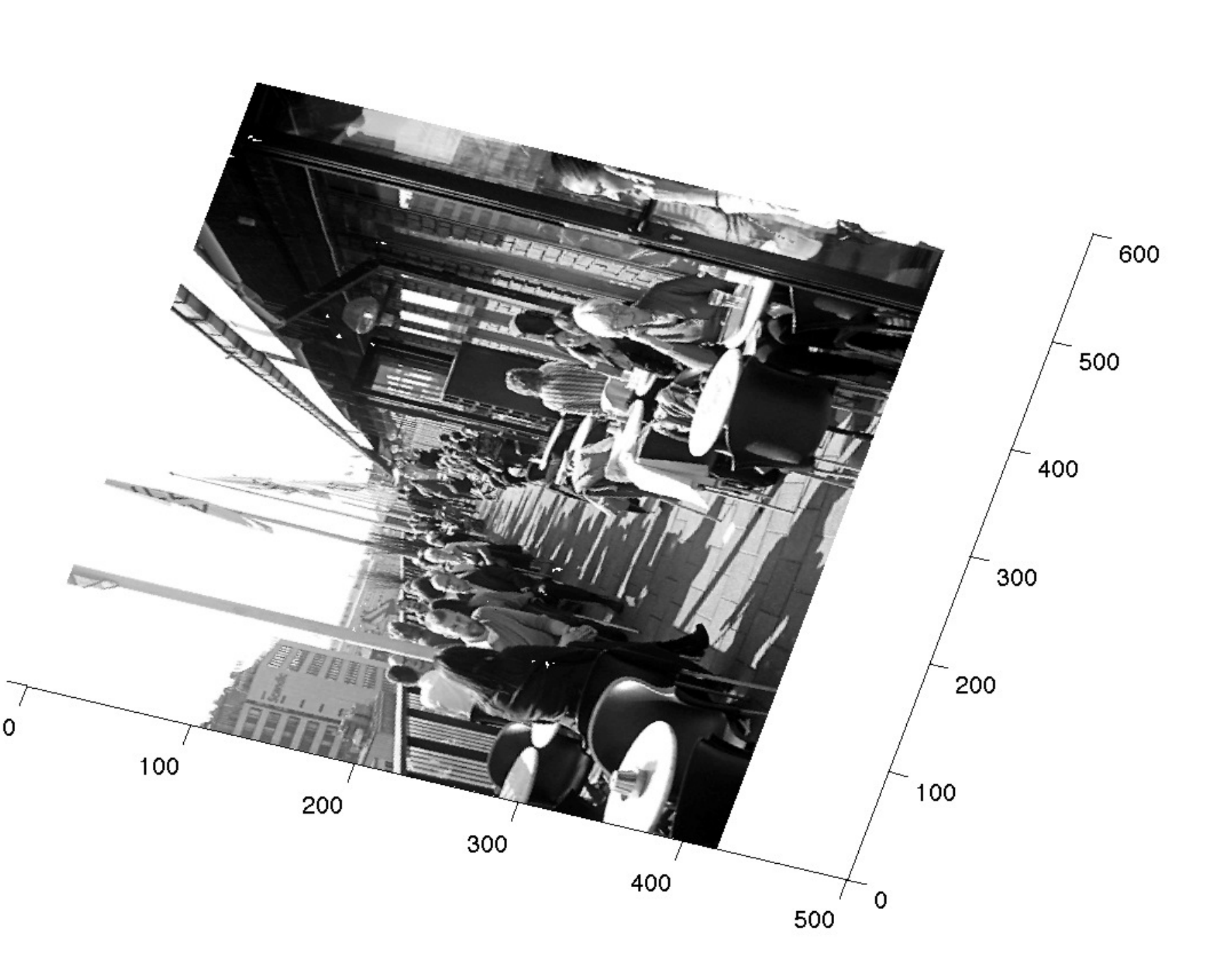} \\
     {\small\em Scale selection surfaces from $\partial_s ({\cal Q}_{(x,y),\Gamma-norm} L) = 0$
       painted with $L$.}  \\
     \includegraphics[width=0.60\textwidth]{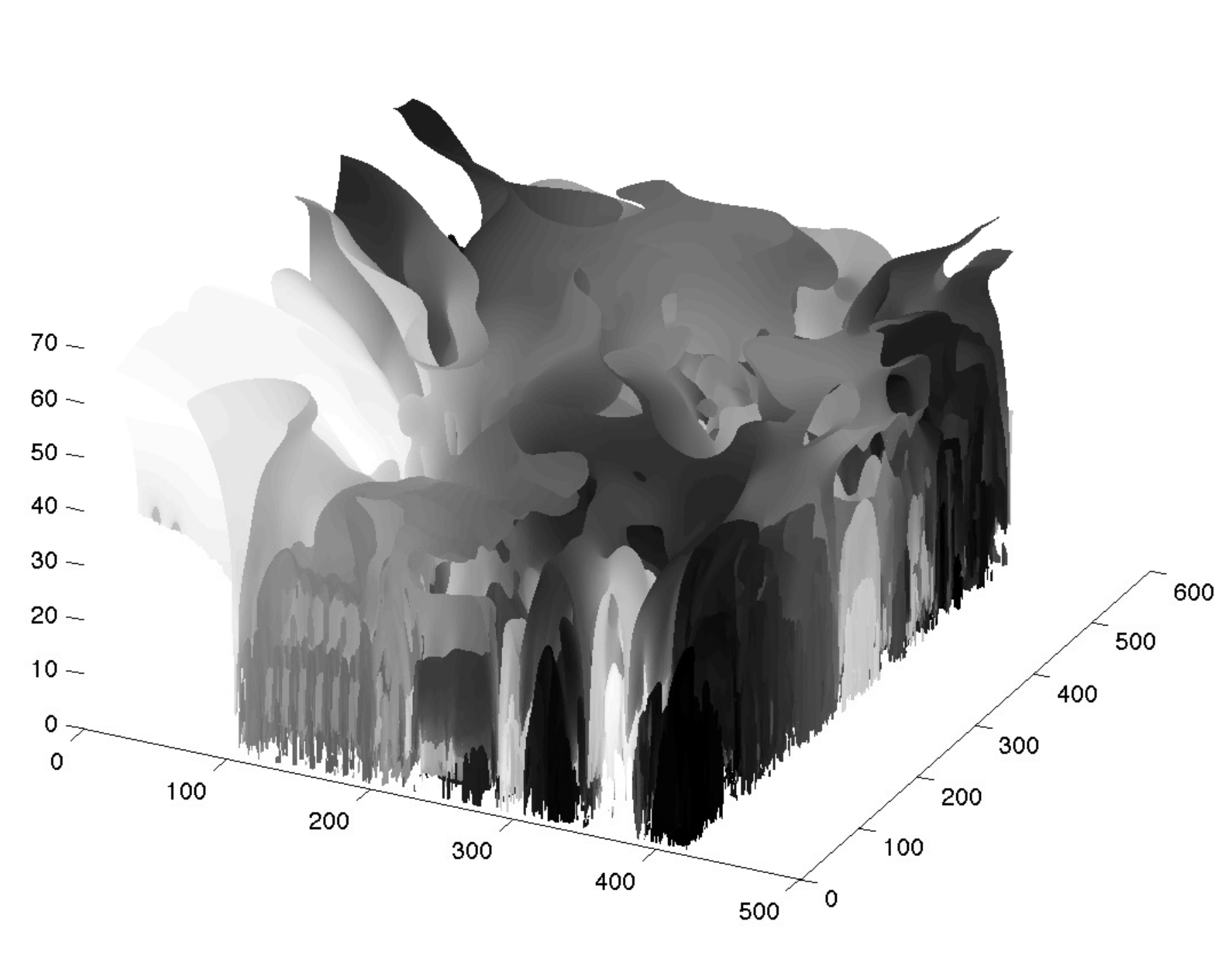} \\
     {\small\em Scale selection surfaces from $\partial_s ({\cal Q}_{(x,y),\Gamma-norm} L) = 0$
       painted with $f$.}   \\
     \includegraphics[width=0.60\textwidth]{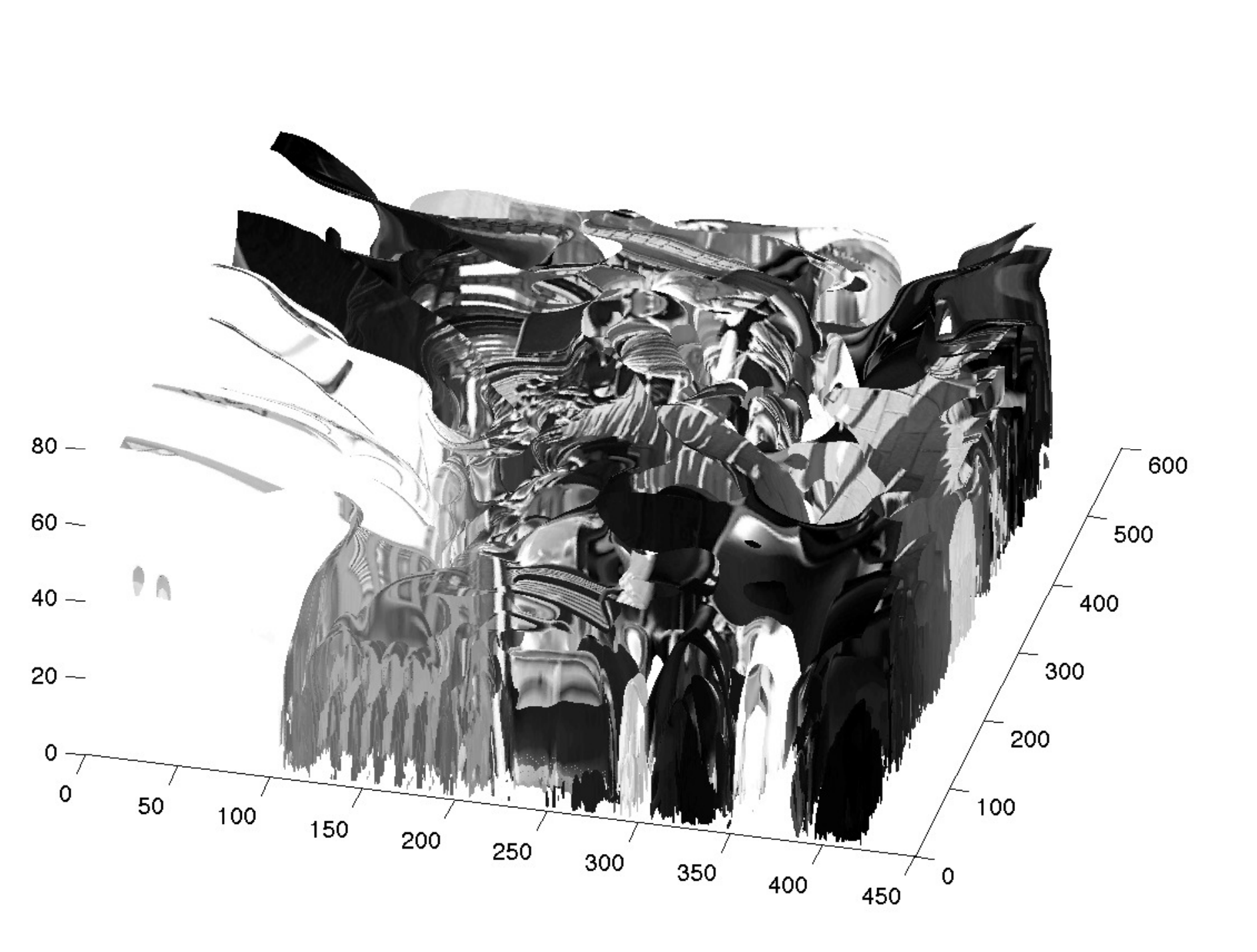} \\
   \end{tabular}
  \end{center}
  \caption{3-D visualization of local scale estimates
    underlying the dense scale maps shown in
    Figure~\ref{fig-terass-cykelstall-scmaps-maxresp}(right) and
    shown as scale selection surfaces in 3-D scale space over space $(x, y)$
    and scale $s$, here also visualizing multiple local extrema over
    scales at every image point.
    (Scale values in units of effective scale
     approximated by $s_{eff} = \log_2(s_0 + s)$ for $s_0 = 1/8$.)}
  \label{fig-terass-scale-sel-surfaces}
\end{figure*}

\subsection{Experimental results}
\label{sec-exp-dense-spat-scsel}

Figure~\ref{fig-terass-cykelstall-scmaps-maxresp} shows spatial scale
maps computed using Algorithm~II for three images.
All the scale maps have been computed using Gaussian scale calibration 
for $\Gamma_s = 1/4$.
% In the graphical illustrations, we have visualized the effective scale parameter
% approximated by $s_{eff} \approx \log (s_0+s)$ for $s_0 = 1/8$ for the
% scale estimate
% corresponding to the strongest local extremum over scale at any image
% point, measured
% in terms of the quasi quadrature entity $Q_{norm} L$ evaluated using
% $\Gamma_s = 0$ and corresponding to the perfectly scale invariant case
% for $\gamma$-normalized derivatives with $\gamma_s = 1$ such that also
% the magnitude measures become scale invariant.
%
For all images in this illustration, we can note how
finer scale estimates are selected near edges, leading to a
sketch-like representation of prominent edges. This behaviour is in
good agreement with the theory and does also imply that image features or
image descriptors that are computed with this type of scale selection methodology will
be well localized near edges.

In general, the detection of local extrema over scale of the quasi
quadrature entity ${\cal Q}_{(x,y),\Gamma-norm} L$ will sweep out
{\em scale selection surfaces\/}
defined by
\begin{equation}
  \left\{
     \begin{array}{l}
       \partial_s ({\cal Q}_{(x,y),\Gamma-norm} L) = 0 \\
       \partial_{ss} ({\cal Q}_{(x,y),\Gamma-norm} L) < 0
    \end{array}
  \right.
\end{equation}
in the 3-D scale space spanned by the spatial dimensions $(x, y)$ and
the scale parameter $s$. Specifically, when the local image structures
contain different types of structures at different scales, multiple
local extrema over scale may be detected, corresponding to multiple patches
of the scale selection surfaces at different scales, which may represent
qualitatively different types of image structures in the image domain
while at different scales.

In Figure~\ref{fig-terass-cykelstall-scmaps-maxresp}, a very much
simplified form of visualization has been used, by only showing the
scale value of the local maximum over scale that has the maximum
response among the possibly multiple local maxima over scale.
When moving between different points over the spatial domain, this
global maximum may at some places switch
between different patches of the scale selection surfaces at different
scales. The discontinuities in the scale maps that can be seen in the
right column correspond to such switching between multiple scale
selection surfaces and are artefacts of the visualization method, not
the scale selection method.

More generally, one should of course treat multiple local extrema
over scale as multiple scale hypotheses as done in the more appropriate 3-D
visualization of such multiple scale selection surfaces in Figure~\ref{fig-terass-scale-sel-surfaces}.
From a more detailed inspection of the scale selection surface patch corresponding to the edge
of the roof in the upper part of the image, one can also find that the
selected scale levels decrease with increasing distance from the
camera as caused by perspective effects.
The similarities between the scale maps for the repetitive
structures in the image in the lower part of the figure demonstrate the
stability of the dense scale estimates under natural imaging conditions,
whereas the relative differences in scale estimates between corresponding
parts of the different but similar looking pillars reveal the size
gradient caused by perspective scaling effects.

To quantify the numerical stability of the scale estimates for a
stimulus for which the scale estimates should be approximately
constant, we computed dense scale estimates for a set of 2-D sine
waves of the form
\begin{equation}
  f(x, y) = \sin \omega x + \sin \omega y
\end{equation}
with wavelengths $\lambda = 8, 16, 32$ and $64$ for each one
of the four types of
algorithms and compared the results with corresponding theoretical predictions based on 
the scale selection properties of the 1-D sine wave model (\ref{eq-mean-scale-est-from-t1-t2-geom})
\begin{equation}
  \hat{\omega} = \frac{\sqrt[4]{(1-\Gamma_s)(2-\Gamma_s)}}{\sqrt{\hat{s}}}.
\end{equation}
The mean and the standard deviation around the mean were computed 
for the scale estimates in terms of effective scale $s_{eff}$ and these
measures were transformed into relative measures in units the
scale parameter $\sigma_s = \sqrt{s}$ of dimension $[\mbox{length}]$.
As can be seen from the results in
Table~\ref{tab-perf-measures-2D-sine-wave}, the use of phase
compensation and complementary post-smoothing substantially decreases
the spatial variability in the scale estimates by an order of
magnitude in units of $\sigma_s$. Specifically, pure phase compensation achieves a
a reduction in the variability of the same order as pure post-smoothing for $c = 1$, while also allowing for a
higher resolution in the scale estimates near edge-like structures.

% For this reason, we shall in later parts of this presentation
% primarily focus on Algorithms of Type~II, although Algorithms of
% Type~IV have the potential of achieving dense scale maps with lower
% scale variability.

\begin{table}[!h]
\footnotesize
\begin{center}
\begin{tabular}{ccccc}
    \hline
    \multicolumn{5}{c}{\em Accuracy of scale estimates for a 2-D sine wave pattern} \\
    \hline
     Error measure & Alg.~I & Alg.~II & Alg.~III &  Alg.~IV \\
    \hline
        Offset of mean 
           & + 5.0~\% & - 0.6~\% & + 1.6~\% & + 1.5~\% \\
        Relative spread 
           & $\pm$ 11.8~\% & $\pm$ 1.3~\% & $\pm$ 0.6~\% & $\pm$ 0.1~\% \\
     \hline
  \end{tabular}
  \end{center}
  \medskip
  \caption{Measures of the accuracy of the scale estimates computed
    for a 2-D sine wave $f(x, y) = \sin \omega x + \sin \omega y$ and
    compared to corresponding theoretical predictions based on a 1-D
    sine wave model according to (\ref{eq-lambda-from-sigma})
    for Algorithms~I--IV using $\Gamma_s = 1/4$, $c = 1$ and 
    $C_s = 1/\sqrt{(1 - \Gamma_s)(2 - \Gamma_s)}$.}
  \label{tab-perf-measures-2D-sine-wave}
\end{table}

\section{Dense temporal scale selection over a purely temporal domain}
\label{sec-temp-dense-scsel}

In this section, we develop a corresponding approach for dense scale
selection over a purely temporal domain.

\subsection{A temporal quasi quadrature measure}
\label{sec-temp-quasi-quad}

Motivated by the fact that first-order derivatives respond
primarily to the locally odd component of a signal, whereas
second-order derivatives respond primarily to the locally even
component of a signal, it is for dense applications natural to aim at a feature detector that combines
such first- and second-order temporal derivative responses.
By specifically combining the squares of the
first- and second-order temporal derivative responses in an additive
way, we obtain a temporal quasi quadrature measure of the form
\begin{equation}
  \label{eq-quasi-quad-t-gamma1-gamma2}
  {\cal Q}_{t,\Gamma-norm} L = \frac{\tau L_t^2 + C_{\tau} \tau^2 L_{tt}^2}{\tau^{\Gamma_{\tau}}},
\end{equation}
which is a reduction of the 2-D spatial quasi quadrature measure
(\ref{eq-quasi-quad-scale-mod}) to a 1-D purely temporal domain,
where $t$ denotes time and $\tau$ temporal scale.

This construction is closely related to a proposal by 
Koenderink and van Doorn \cite{KoeDoo90-BC} of summing up the squares of
the first- and second-order derivative responses of receptive fields and an
observation by De~Valois {\em et al.\/}\ \cite{ValCotMahElfWil00-VR}
that first- and second-order biological receptive fields typically
occur in pairs that can be modelled as approximate Hilbert pairs,
while here instead 
formalized in terms of scale-normalized
temporal 
scale-space derivatives.

\subsection{Scale covariance of temporal scale estimates under
  temporal scaling transformations}
\label{sec-cov-prop-temp-sc-sel}

Consider a temporal scaling transformation of the form 
\begin{equation}
  f'(t') = f(t)
  \quad\quad\mbox{for}\quad\quad 
  t' = S_{\tau} \, t.
\end{equation}
From the temporal scaling transformation of scale-normalized temporal
derivatives defined from either the non-causal Gaussian temporal scale
space or the time-causal temporal scale space obtained by convolution
with the time-causal limit kernel \cite{Lin16-JMIV},
%(\ref{eq-FT-comp-kern-log-distr-limit}), 
it follows that scale-normalized
temporal derivatives of order $n$ are transformed according to
\cite[Equations~(10) and (104)]{Lin17-JMIV}
\begin{equation}
  \label{eq-temp-scale-cov-hom-diff-inv}
   \partial_{\zeta'^n, {norm}} L'(t';\; \tau') 
  = S_{\tau}^{n(\gamma_{\tau} - 1)} \, \partial_{\zeta^n, {norm}} L(t;\; \tau)
\end{equation}
provided that the temporal scale levels are correspondingly matched 
\begin{equation}
  \tau' = S_{\tau}^2 \, \tau.
\end{equation}
Applied to the temporal quasi quadrature measure
\begin{equation}
  {\cal Q}_{t,\Gamma-norm} L  %= \\
  = \frac{\tau L_t^2 + C_{\tau} \tau^2 L_{tt}^2}{\tau^{\Gamma_{\tau}}}
  = {\cal Q}_{t,1,\Gamma-norm} L + {\cal Q}_{t,2,\Gamma-norm} L
\end{equation}
with its first- and second-order components
\begin{equation}
  {\cal Q}_{t,1,\Gamma-norm} L = \frac{\tau L_t^2}{\tau^{\Gamma_{\tau}}},
  \quad\quad
  {\cal Q}_{t,2,\Gamma-norm} L = \frac{C_{\tau} \tau^2 L_{tt}^2}{\tau^{\Gamma_{\tau}}},
\end{equation}
and which correspond to $\gamma$-normalized temporal derivatives
with $\gamma_1 = 1 - \Gamma_{\tau}$ and $\gamma_2 = 1 - \Gamma_{\tau}/2$
for the first- and second-order components, respectively,
it follows that the first- and second-order components transform
according to
\begin{align}
  \begin{split}
    ({\cal Q}_{t,1,\Gamma-norm} L')(t';\; \tau') 
   & = S_{\tau}^{2 (\gamma_1 - 1)} \, ({\cal Q}_{t,1,\Gamma-norm} L)(t;\; \tau) 
  \end{split}\nonumber\\
 \begin{split}
    & = S_{\tau}^{-2 \Gamma_{\tau}} \, ({\cal Q}_{t,1,\Gamma-norm} L)(t;\; \tau),
  \end{split}\\
 \begin{split}
    ({\cal Q}_{t,2,\Gamma-norm} L')(t';\; \tau') 
   & = S_{\tau}^{2 \times 2 (\gamma_2 - 1)} \, ({\cal Q}_{t,2,\Gamma-norm} L)(t;\; \tau) 
  \end{split}\nonumber\\
 \begin{split}
    & = S_{\tau}^{-2 \Gamma_{\tau}} \, ({\cal Q}_{t,2,\Gamma-norm} L)(t;\; \tau).
  \end{split}
\end{align}
Since the first- and second-order components transform in a similar way
because of the deliberate adding of entities depending on temporal
derivatives of different orders for the maximally scale-invariant
choice of $\gamma_{\tau} = 1$, it follows that the temporal quasi quadrature
measure despite its inhomogeneity still transforms according to a power law
\begin{equation}
  \label{eq-temp-sc-transf-temp-quasiquad}
  ({\cal Q}_{t,\Gamma-norm} L')(t';\; \tau') = S_{\tau}^{-2 \Gamma_{\tau}} \, ({\cal Q}_{t,\Gamma-norm} L)(t;\; \tau).
\end{equation}
Specifically, this implies that temporal scale estimates computed from
local extrema over temporal scales are preserved and are transformed in a scale
covariant way
\begin{equation}
  \hat{\tau}' = S_{\tau}^2 \, \hat{\tau}
\end{equation}
or in units of the standard deviation $\sigma_{\tau} = \sqrt{\tau}$ of the
temporal scale-space kernel
\begin{equation}
  \hat{\sigma}'_{\tau} = S_{\tau} \, \hat{\sigma}_{\tau}.
\end{equation}
This does in turn imply that the temporal scale estimates will adapt
to local temporal scaling transformations of the input signal, and
constitutes the theoretical basis for the dense temporal scale selection
methodology.

This temporal scale covariance property does also extend to
phase-compensated 
temporal scale estimates according to (\ref{eq-scale-phasecorr-geom-sine-tau})
\begin{align}
  \begin{split}
     \label{eq-scale-phasecorr-geom-sine-tau-time}
     \hat{\tau}_{{\cal Q}_t, comp} % = 
  %\end{split}\nonumber\\
  %\begin{split}
     & = \frac{\sqrt{(1 - \Gamma_{\tau})(2 - \Gamma_{\tau})} \, \hat{\tau}_{{\cal
           Q}_t}}
                   {(1-\Gamma_{\tau})^{\frac{{\cal Q}_{t,1,\Gamma-norm} L}{{\cal Q}_{t,1,\Gamma-norm} L + {\cal Q}_{t,2,\Gamma-norm} L}} 
                   (2-\Gamma_{\tau})^{\frac{{\cal Q}_{t,2,\Gamma-norm} L}{{\cal Q}_{t,1,\Gamma-norm} L + {\cal Q}_{t,2,\Gamma-norm} L}}},
  \end{split}
\end{align}
since the underlying uncompensated temporal scale estimates $\hat{\tau}_{{\cal Q}_t}$
transform in a scale-covariant way and the ratios 
%\begin{align}
 % \begin{split}
$w_1 = {\cal Q}_{t,1,\Gamma-norm} L/({\cal Q}_{t,1,\Gamma-norm} L + {\cal Q}_{t,2,\Gamma-norm} L)$
%     \frac{{\cal Q}_{t,1,\Gamma-norm} L}{{\cal Q}_{t,1,\Gamma-norm} L + {\cal Q}_{t,2,\Gamma-norm} L}
%  \end{split}\\
%  \begin{split}
and $w_2 = {\cal Q}_{t,2,\Gamma-norm} L/({\cal Q}_{t,1,\Gamma-norm} L + {\cal Q}_{t,2,\Gamma-norm} L)$
%     \frac{{\cal Q}_{t,2,\Gamma-norm} L}{{\cal Q}_{t,1,\Gamma-norm} L + {\cal Q}_{t,2,\Gamma-norm} L}
%  \end{split}
%\end{align}
that determine the scale compensation factors
are invariant under temporal scaling transformations, provided that
the temporal scale levels are appropriately matched.

The temporal scale covariance of the temporal scale estimates is also
preserved under temporal post-smoothing, since the amount of temporal
post-smoothing is proportional to the local temporal scale for
computing the temporal derivatives.
%Since the unsmoothed temporal quasi quadrature entity transforms
%according to a scaling law, the post-smoothed temporal quasi
%quadrature transforms according to a similar scaling law.
% Notably, the temporal scale estimates are invariant under
% transformations of the magnitude of the signal of the
% form $f'(t') = a \, f(t) + b$ for real $a \neq 0$ and any real $b$.

\begin{figure*}
  \begin{center}
     \begin{tabular}{cccc}
        {\footnotesize\em original signal} 
       & {\footnotesize\em  $\sqrt{{\cal Q}_{t,norm} L}$}
       & {\footnotesize\em basic scale selection}
       & {\footnotesize\em phase-compensated} \\
       \includegraphics[height=0.125\textheight]{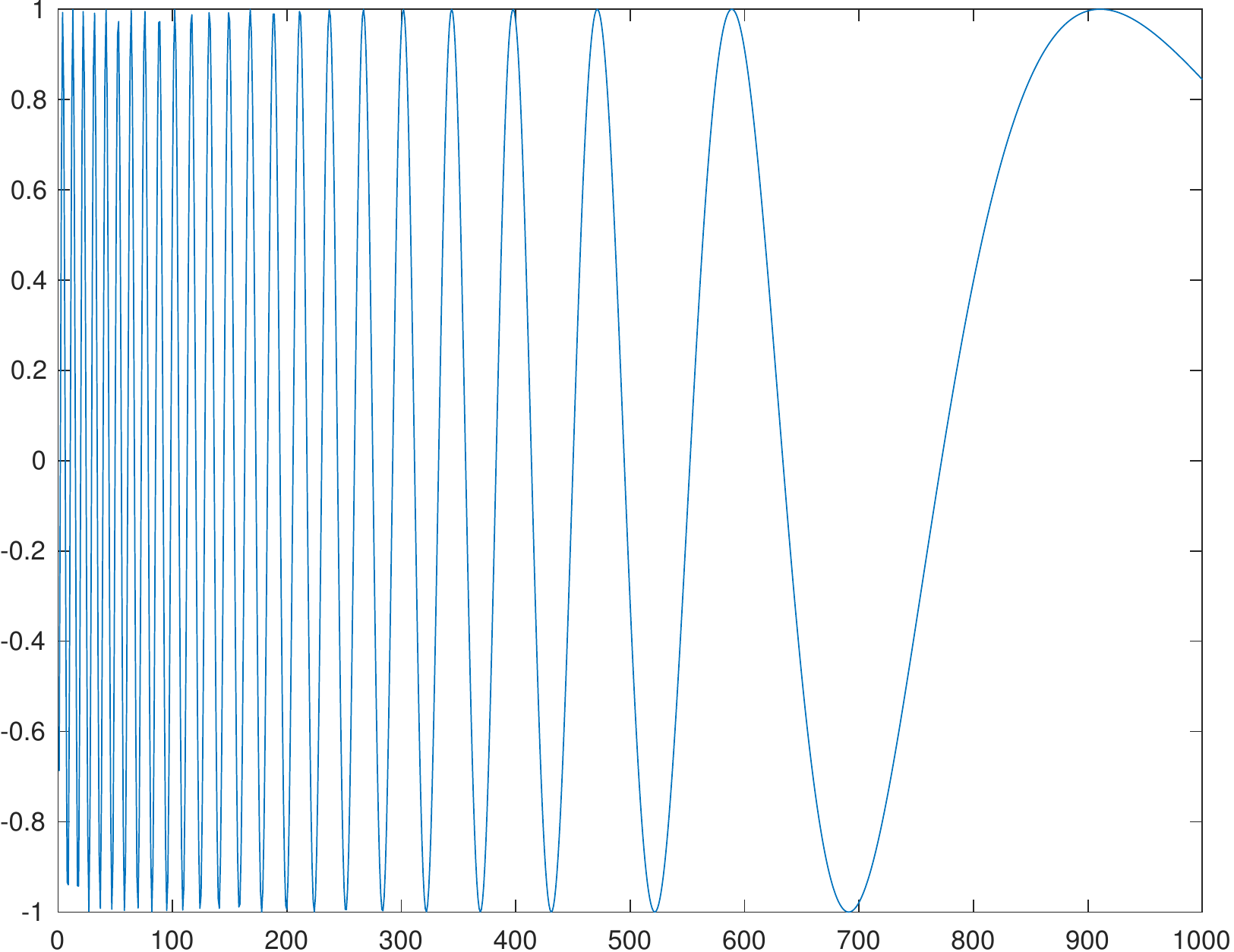} 
       & \includegraphics[height=0.125\textheight]{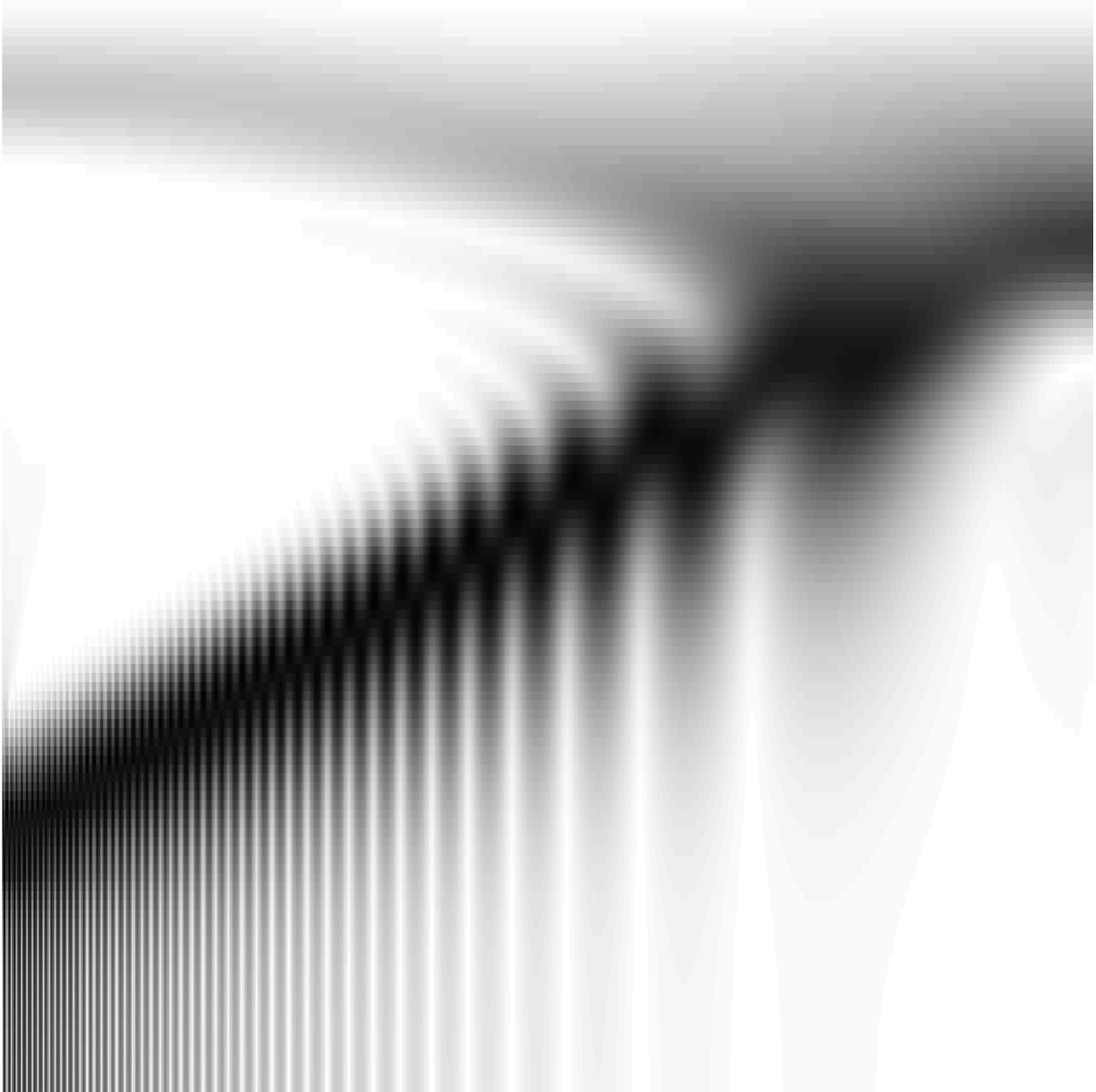} 
       & \includegraphics[height=0.125\textheight]{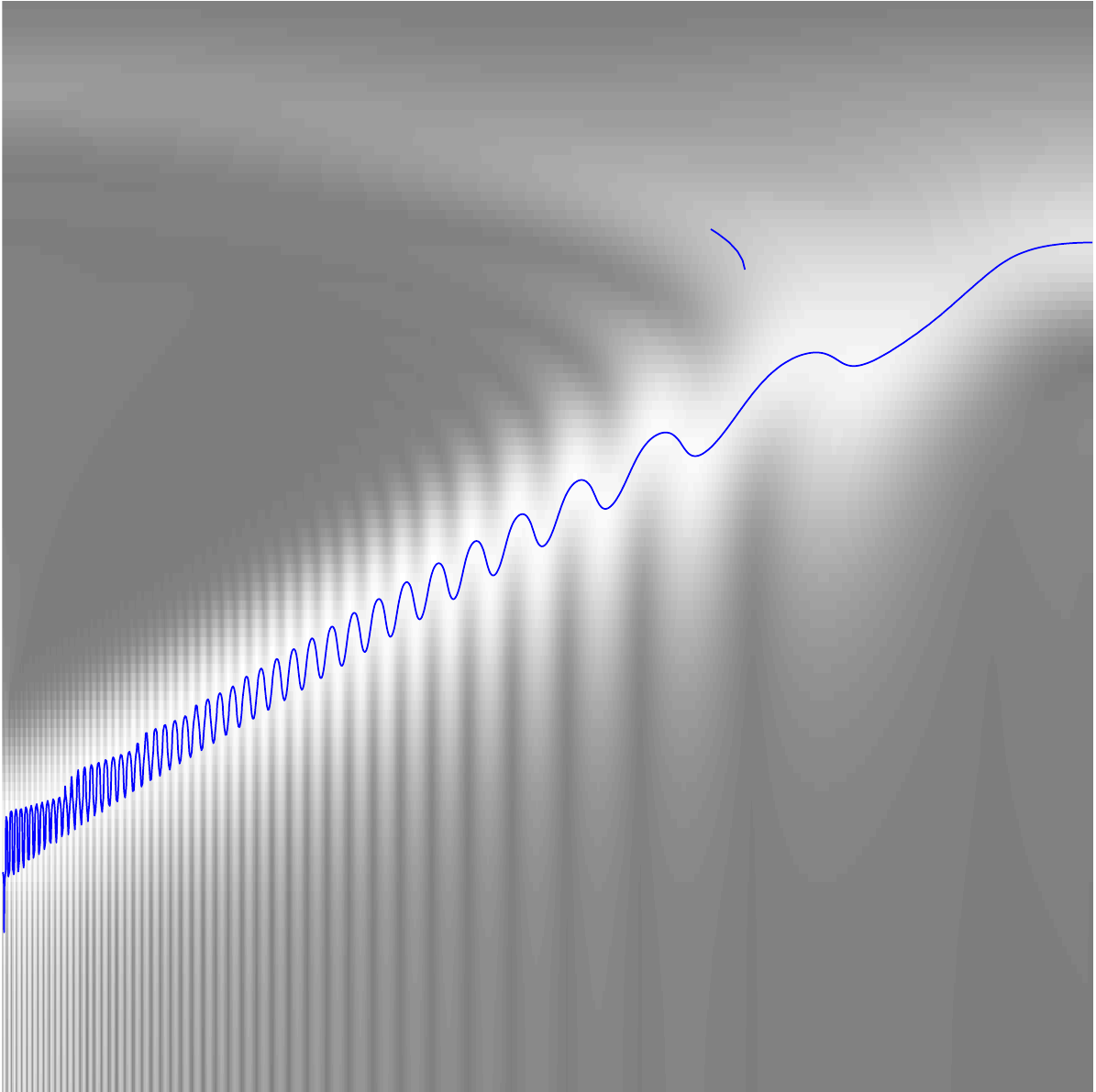}
       & \includegraphics[height=0.125\textheight]{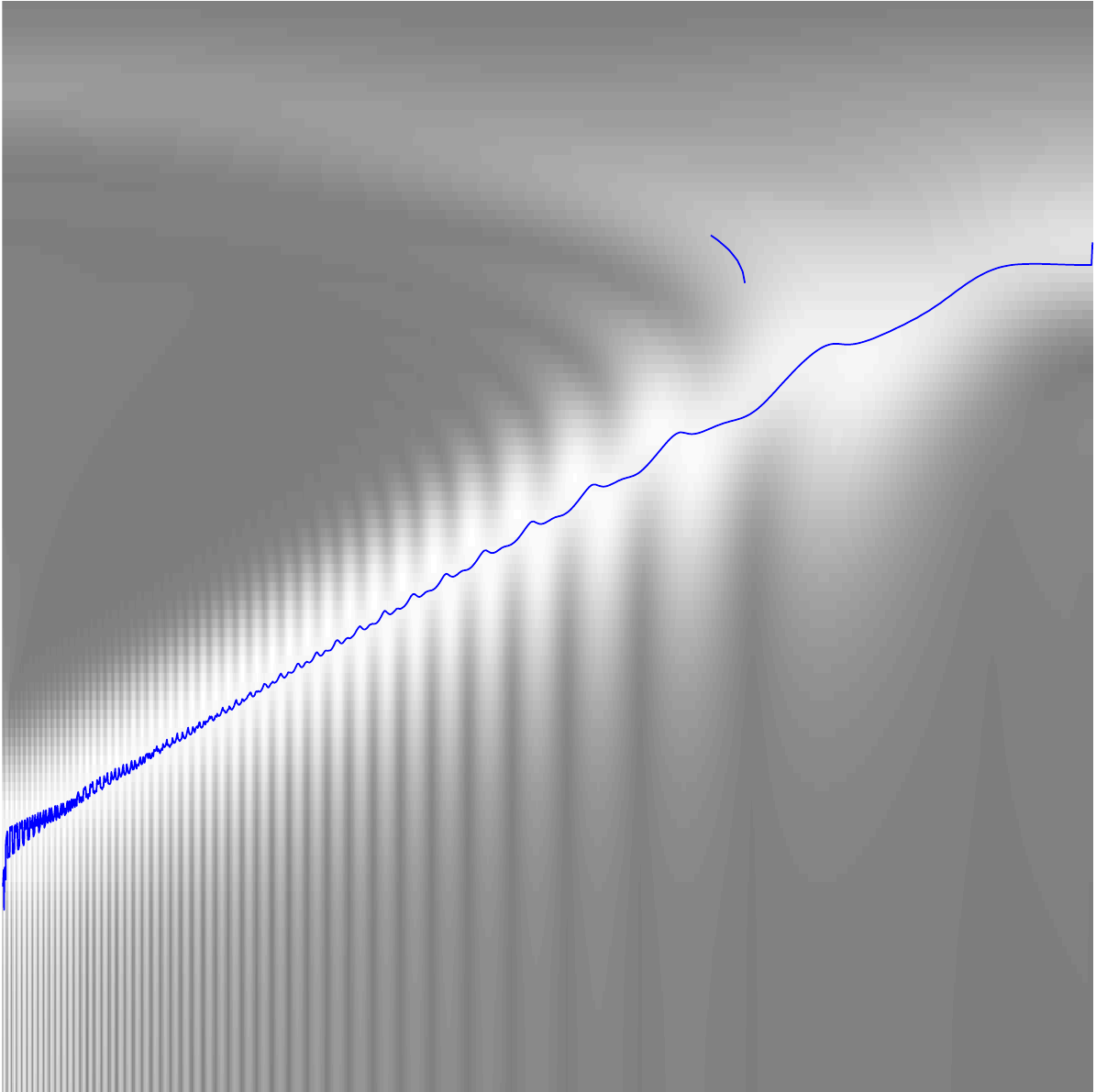} \\
     \end{tabular}
  \end{center}
\caption{Dense temporal scale selection from local extrema over scale of the temporal
               quasi quadrature measure ${\cal Q}_{t,\Gamma-norm} L$ applied to a synthetic sine
               wave signal $f(t) = \sin(\exp((b-t)/a))$ for $a = 200$ and $b = 1000$ with temporally
               varying frequency so that the wavelength increases with time $t$.
               (left) Original temporal signal.
               (middle left) The square root of the temporal
               quasi quadrature measure ${\cal Q}_{t,\Gamma-norm} L$
               computed for the most scale-invariant choice of the complementary scale normalization
               parameter $\Gamma_{\tau} = 0$.
               (middle right) Basic scale estimates $\hat{\tau}_{{\cal Q}_{t,\Gamma-norm}}$ 
               from local extrema over scale of the quasi quadrature measure according
               to (\ref{eq-scale-sel-quasi-quad}) reduced to 1-D for
               $\Gamma_{\tau} = 0$ and shown as blue curves overlaid
               on the magnitude map ${\cal Q}_{t,\Gamma-norm} L$ with
               reversed contrast.
               (right) Phase-compensated scale estimates
               $\hat{\tau}_{{\cal Q}_t, comp}$ according
               to (\ref{eq-scale-phasecorr-geom-sine-tau}) reduced to
               1-D for $\Gamma_{\tau} = 0$ and shown as blue curves overlaid
               on the magnitude map ${\cal Q}_{t,\Gamma-norm} L$ with
               reversed contrast.
               All results have been computed using $C_{\tau} =1/\sqrt{(1 - \Gamma_{\tau})(2 - \Gamma_{\tau})}$.
              (Horizontal axis: time $t \in [0, 1000]$) (Vertical axis in columns 2-4:
               effective temporal scale $\tau_{eff} = \log \tau$ over
               the range from $\sigma_{\tau,min} = 0.25$ to 
               $\sigma_{\tau,max} = 1000$ for $\sigma_{\tau} = \sqrt{\tau}$)}
  \label{fig-expsine-densetempscsel}
\end{figure*}

\begin{figure*}
  \begin{center}
      \begin{tabular}{cc}
        {\footnotesize\em post-smoothed scale selection (off med.)\/} 
          &  {\footnotesize\em post-smoothed scale selection (on med.)\/} \\
        \includegraphics[width=0.44\textwidth]{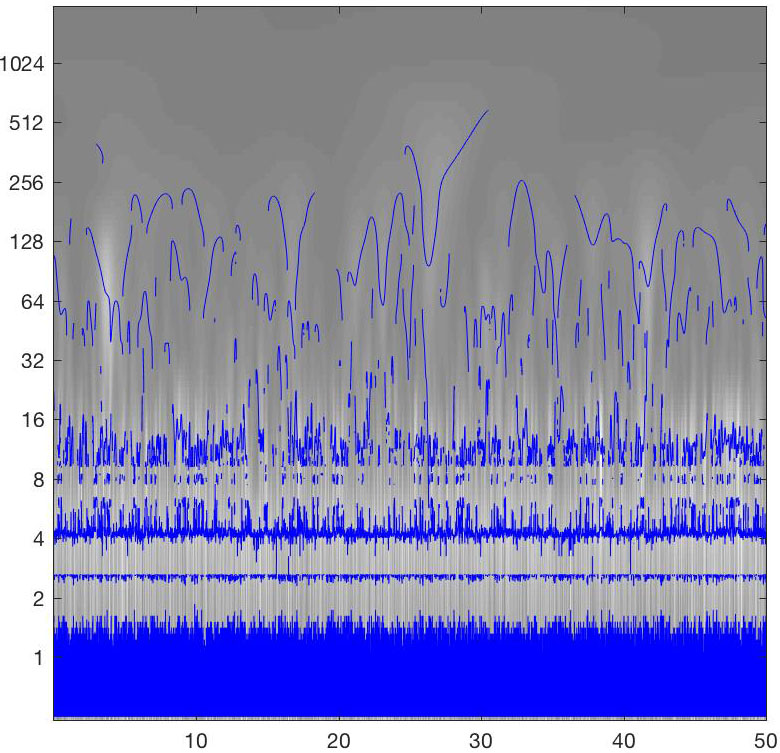} 
       & \includegraphics[width=0.44\textwidth]{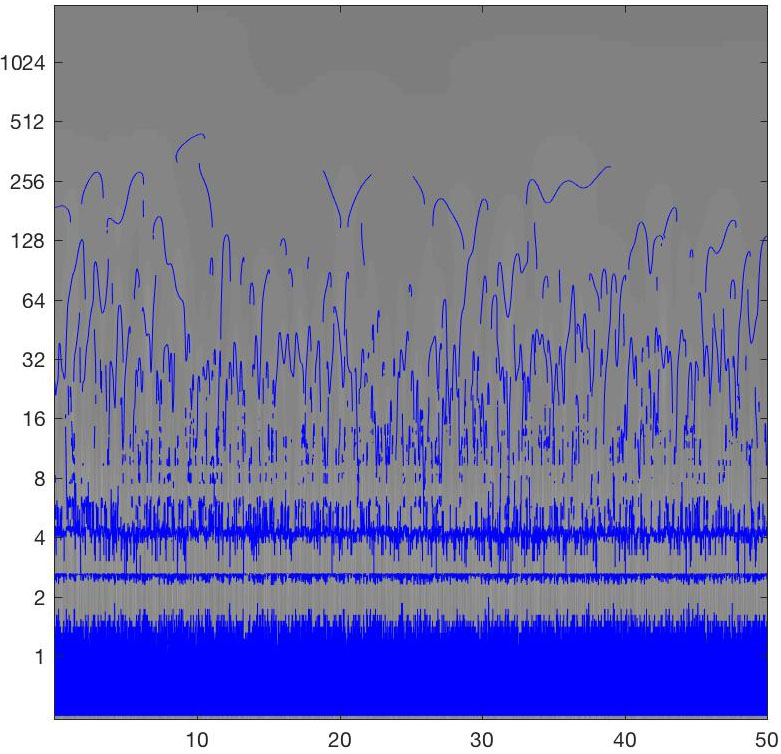}
        \\
         {\footnotesize\em $\sqrt{{\cal Q}_{t,norm} L}$ Gaussian
        kernel (off med.)\/} 
       & {\footnotesize\em $\sqrt{{\cal Q}_{t,norm} L}$ Gaussian
         kernel (on med.)\/} \\
         \includegraphics[width=0.44\textwidth]{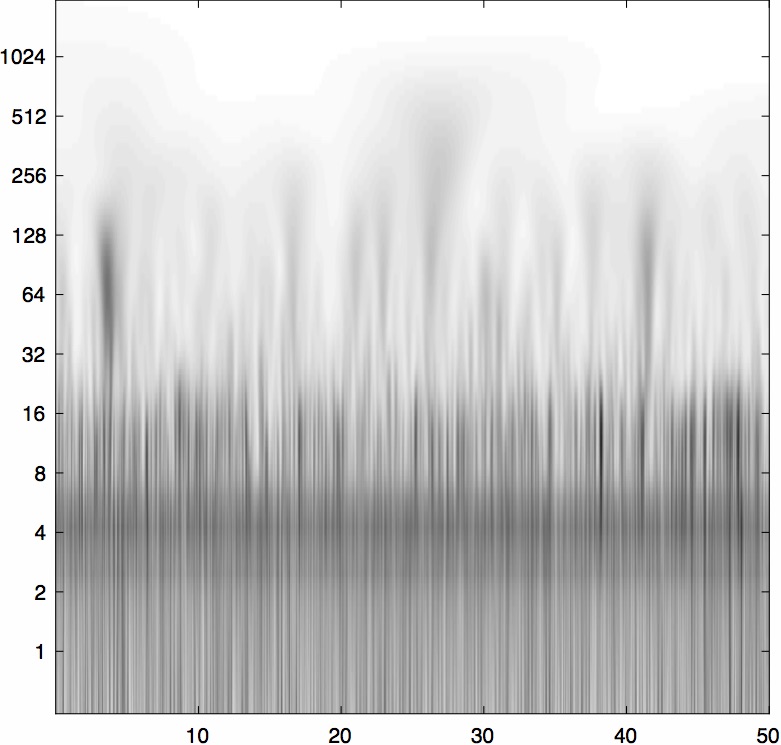} 
        &     \includegraphics[width=0.44\textwidth]{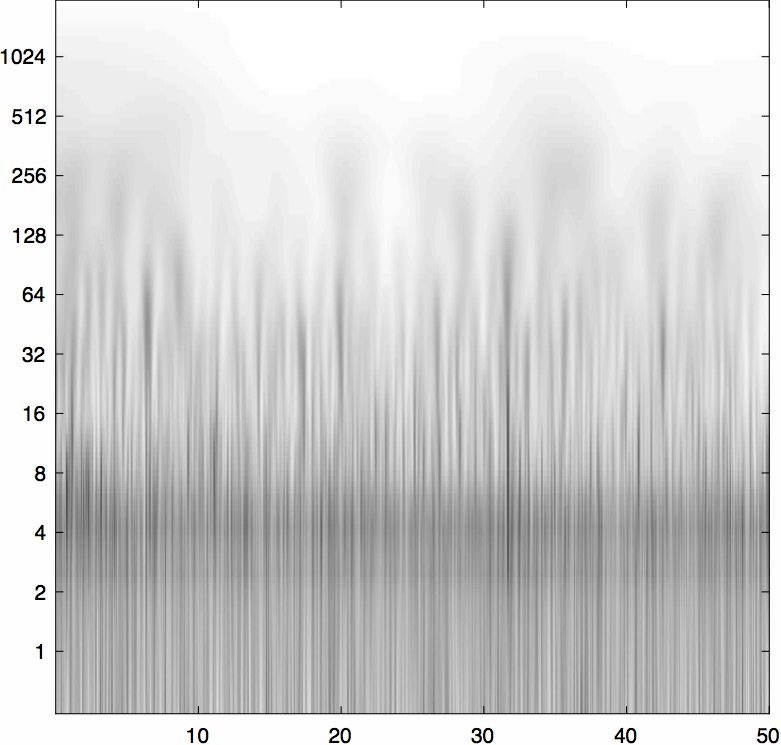} \\
        {\footnotesize\em  Original signal (off med.)\/}
            & {\footnotesize\em  Original signal (on med.)\/} \\
         \includegraphics[width=0.44\textwidth]{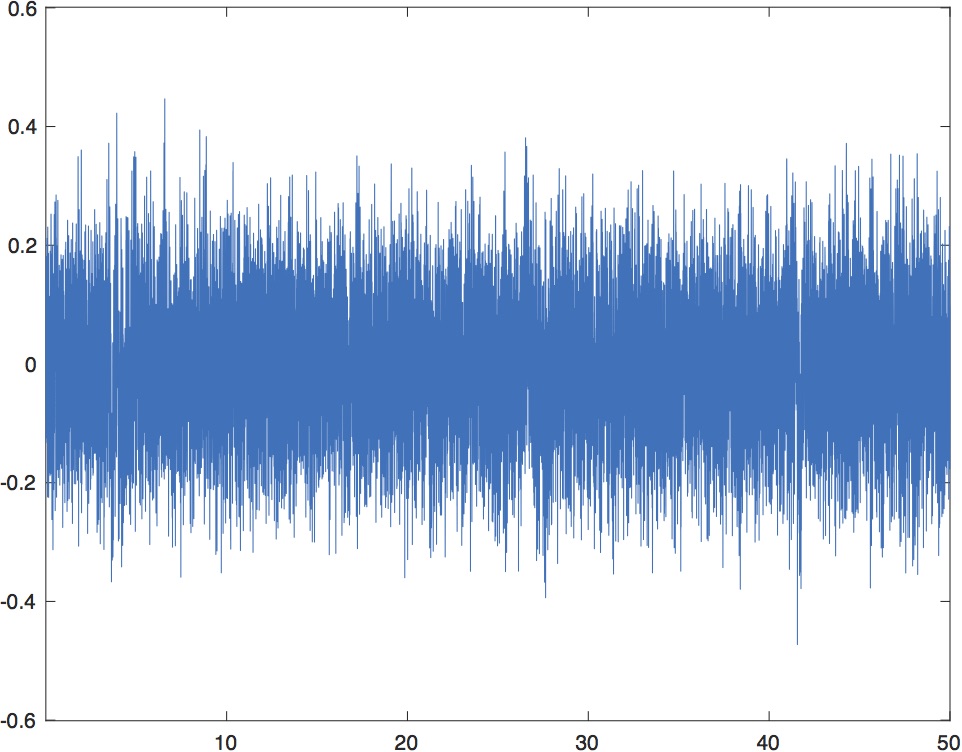} 
        &     \includegraphics[width=0.44\textwidth]{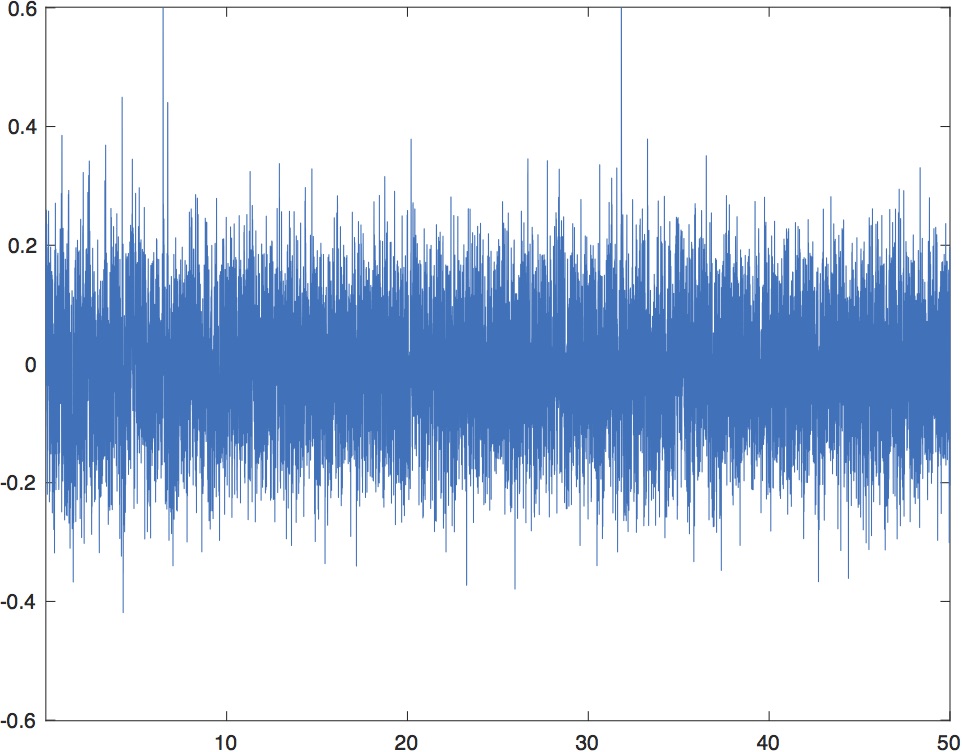} 
      \end{tabular}
  \end{center}
  \caption{Dense local scale analysis of a neurophysiological signal
    from Caghan {\em et al.\/} \cite{CagDufBro16-Brain}
    showing the local field potential sampled at $\nu = 512$~Hz during 50
    seconds for an electrode inserted in the
    sub-thalamic nucleus of an awake human subject with Parkinson's
    disease
   (left column) without medication or (right column) when the patient
   is on on
  medication by leva-dopa. 
  % In these experiments, we have included a temporal
  % post-smoothing stage with the post-smoothing scale proportional to
  % the local scale at which derivatives are computed with
  % proportionality constant = 3.
  Note how this local scale analysis method reveals that
  the uppermost stronger band of dense local scale estimates in the
  top left figure is spread out over a larger scale range in the top right
  figure as a result of the medication.
              (top row) Basic scale estimates $\hat{\tau}_{{\cal Q}_{t,\Gamma-norm}}$ 
               from local extrema over scale of the quasi quadrature measure according
               to (\ref{eq-scale-sel-quasi-quad-post-smooth}) reduced
               to 1-D for $c = 3$ and $\Gamma_{\tau} = 0$.
               (middle row) The square root of the temporal
               quasi quadrature measure ${\cal Q}_{t,\Gamma-norm} L$
               computed for the most scale-invariant choice of the complementary scale normalization
               parameter $\Gamma_{\tau} = 0$.
               (bottom row) Original temporal signal.
               All results have been computed using $C_{\tau} =1/\sqrt{(1 - \Gamma_{\tau})(2 - \Gamma_{\tau})}$.
               (Horizontal axis: time $t \in [0, 50]$ in seconds.) (Vertical axis
               in top and middle rows:
               temporal scale in milliseconds.)}
\label{fig-arvind-signal-densetempscsel}
\end{figure*}

\subsection{Experimental results}

\subsubsection{Sine wave with exponentially varying frequency}

Figure~\ref{fig-expsine-densetempscsel} shows the result of applying
this basic form of dense temporal scale selection to a sine wave with
exponentially varying frequency of the form
\begin{equation}
  f(t) = \sin(\exp(\frac{(b-t)}{a}
))
\end{equation}
for $a = 200$ and $b = 1000$. The left figure shows the
raw temporal signal. The middle left figure shows the
magnitude map over time and temporal scales of the temporal quasi quadrature measure 
${\cal Q}_{t,norm} L$ computed using a non-causal Gaussian temporal
scale-space representation.
% approximated by the discrete analogue of
%the Gaussian kernel \cite{Lin93-Dis}.
The middle right figure shows temporal scale estimates computed
as the zero-crossings of $\partial_{\tau}({\cal Q}_{t,\Gamma-norm} L) = 0$
that satisfy the sign condition $\partial_{\tau\tau}({\cal Q}_{t,\Gamma-norm}
L) < 0$.
These zero-crossings have been interpolated to higher accuracy along
the temporal scale dimension than the sampling density over the
temporal scales using
parabolic interpolation \cite[Equation~(115)]{Lin17-JMIV}.
In the rightmost figure, the basic temporal estimates from
the middle right column have been additionally phase-compensated according to
(\ref{eq-scale-phasecorr-geom-sine-tau-time}).

Note how: (i)~the temporal scale selection method is able to capture the
rapid variations in the temporal scales in the signal and
(ii)~the phase compensation method substantially suppresses the phase
dependency of the temporal scale estimates.

\subsubsection{Real measurement signals}

Figure~\ref{fig-arvind-signal-densetempscsel} shows an example of performing this
type of dense temporal scale selection analysis on two real measurements signals
using the non-causal Gaussian temporal scale-space concept.
The figures in the bottom row show measurements of the local
field potential recorded from the sub-thalamic nucleus of an awake human 
subject with Parkinson's disease with the patient either off or on
medication by leva-dopa (labelled ``off med.'' or ``on med.'' and shown in the
left and the right columns, respectively).

As can be seen from the dense local scale estimates in the top row,
the dense scale analysis does (beyond a wide band of responses at finer
scales up to temporal scale $\sigma_{\tau,0} = 1.4$~ms)
return three rather strong bands of coarser temporal scale
estimates when the patient is off medication.
These bands are assumed around temporal scales $\sigma_{\tau,1} = 2.6$~ms,
$\sigma_{\tau,2} = 4.5$~ms and $\sigma_{\tau,4} = 11$~ms with a weaker
additional band at temporal scale $\sigma_{\tau,3} = 7.9$~ms.
According to the approximate expression  (\ref{eq-lambda-from-sigma}) for the local scale estimate of a
sine wave,
these scale estimates correspond to frequencies $\nu = 1/\lambda$
around $\nu_0 = 136$~Hz, $\nu_1 = 73$~Hz, $\nu_2 = 42$~Hz, $\nu_3 = 24$~Hz and  $\nu_4 = 18$~Hz.

When the patient is on medication, the uppermost band of coarser scale
estimates around $\sigma_{\tau,4} = 11$~ms is replaced by a sparser 
set of dense local scale estimates over a wider scale range.
$[\sigma_{\tau,5}, \sigma_{\tau,6}] = [10, 42]$~ms and corresponding
to frequencies in the range $[\nu_{6}, \nu_{5}] = [5, 19]$~Hz.

Comparing the results for the patient off {\em vs.\/}\ on medication,
there is also a weaker band of responses over the scale range between
$\sigma_{\tau,7} = 44$~ms and $\sigma_{\tau,8} = 260$~ms
corresponding to a frequency range between 
$\nu_7 = 5$~Hz and  $\nu_8 = 0.75$~Hz when the patient is off medication and with not as
strong responses in this band when the patient is on medication.

%In this way, the proposed dense scale
%selection methodology is able to reveal how the medication affects the local temporal
%dynamics over multiple cales in the measured local field potential.

The biological background to this signal analysis problem is that in
Parkinson's disease (PD), several prominent rhythms appear in the
local field potentials. 
Among these, the low-frequency ($\sim$5~Hz) rhythms
are associated with tremors observed in PD patients. 
Next, the so-called beta band (15--30~Hz) rhythms are causally 
related to many motor deficits associated with PD 
(Hammond {\em et al.\/}\ \cite{HamBerBro07-TINS}). 
Recent analysis of local field potentials in the sub-thalamic nucleus using Fourier analysis revealed that beta
band oscillations are not persistent and instead occur in bursts
(Tinkhauser {\em et al.\/}\ \cite{TinPogTanHerKuhBro17-Brain}). 
Moreover, administration of L-dopa
medication was shown to reduce the frequency of beta bursts,
especially the long beta bursts are significantly reduced. Indeed,
quenching of the beta band oscillations is one the goals of PD
treatment. Specifically, modulation of beta-band oscillations can form
a basis for an event-triggered deep-brain-stimulation system 
(Rosin {\em et  al.\/}\ \cite{RosSloMitRIvHavIsrVaaBer11-Neuron}). 
To that end, however, it is important to correctly isolate
the occurrences of beta-band activity. Conventional methods based on
Fourier transforms have, however, been found to not be very precise for
this purpose. 

In relation to this biological background, our
temporal scale analysis thus reveals how medication by L-dopa affects
the temporal dynamics of neurons in bands at multiple scales that are related to the tremors
observed in Parkinson's disease patients.
Specifically, it shows how the responses in the band around $\nu_4 = 18$~Hz
related to pathology are reduced and spread out by the
L-dopa medication
and that the responses in the band with frequencies below $\nu_7 = 5$~Hz related to tremors
are weaker.

\section{Dense spatio-temporal scale selection over the joint
  spatio-temporal domain}
\label{sec-spattemp-dense-scsel}

In this section, we shall combine the mechanisms for dense spatial
scale selection and dense temporal scale selection developed in
Section~\ref{sec-spat-dense-scsel} and
Section~\ref{sec-temp-dense-scsel}
to design a mechanism for dense simultaneous selection of spatial and
temporal scales over the joint spatio-temporal domain.

\subsection{Spatio-temporal scale-space representation}

The context that we initially consider for dense spatio-temporal scale selection is a
space-time separable spatio-temporal scale-space representation $L(x, y, t;\; s, \tau)$
defined from any 2+1-D video sequence $f(x, y, t)$ by
convolution with space-time separable spatio-temporal Gaussian kernels
\begin{equation}
  T(x, y, t;\; s, \tau) %= \\
  = g(x, y;\; s) \, g(t;\; \tau) 
  = \frac{1}{2 \pi s} e^{-(x^2+y^2)/2s} \, \frac{1}{\sqrt{2 \pi \tau}} e^{-t^2/2\tau}
\end{equation}
at different spatio-temporal scales $(s, \tau)$
(Lindeberg \cite{Lin10-JMIV})
\begin{equation}
  L(\cdot, \cdot, \cdot;\; s, \tau) = T(\cdot, \cdot, \cdot;\; s, \tau) * f(\cdot, \cdot, \cdot) 
\end{equation}
and with $\gamma$-normalized spatio-temporal derivatives 
% at any scale
defined according to \cite{Lin97-IJCV,Lin16-JMIV}
\begin{equation}
   \partial_{\xi} = s^{\gamma_s/2} \, \partial_x, \quad
  \partial_{\eta} = s^{\gamma_s/2} \, \partial_y, \quad
  \partial_{\zeta} = \tau^{\gamma_{\tau}/2} \, \partial_t.
\end{equation}
Initially, we will develop the basic theory
based on a non-causal Gaussian temporal
scale-space model and then in the experiments for the purpose of also
being able to handle real-time image streams complement with a truly
time-causal spatio-temporal scale-space representation (Lindeberg
\cite{Lin16-JMIV}) defined based on temporal smoothing with the
time-causal limit kernel $\Psi(t;\; \tau, c)$ having a Fourier transform of
the form 
\begin{align}
  \begin{split}
     \hat{\Psi}(\omega;\; \tau, c) 
     %& = \lim_{K \rightarrow \infty} \hat{h}_{exp}(\omega;\; \tau, c, K) 
  %\end{split}\nonumber\\
  %\begin{split}
     \label{eq-FT-comp-kern-log-distr-limit}
     & = \prod_{k=1}^{\infty} \frac{1}{1 + i \, c^{-k} \sqrt{c^2-1} \sqrt{\tau} \, \omega},
  \end{split}
\end{align}
% thus 
% corresponding
% to space-time separable spatio-temp\-oral scale-space kernels of the form
% \begin{equation}
%  T(x, y, t;\; s, \tau, c) 
%  = g(x, y;\; s) \, \Psi(t;\; \tau, c)
% \end{equation}
and for which the discrete implementation of the temporal smoothing
operation is in turn approximated by a finite number of discrete
recursive filters coupled in cascade. 

\subsection{A spatio-temporal quasi quadrature measure}
\label{sec-spat-temp-quasi-quad}

In Lindeberg \cite{Lin16-JMIV}, the following
spatio-temporal quadrature was considered
%three spatio-temporal quadrature
%entities were considered out of which the %differential 
%entity
\begin{align}
  \begin{split}
     & {\cal Q}_{3,(x, y, t),norm} L 
  \end{split}\nonumber\\
  \begin{split}
     & = {\cal Q}_{(x, y),norm} L_t + C \, {\cal Q}_{(x, y),norm} L_{tt} 
 \end{split}\nonumber\\
  \begin{split}
      & = \tau
             \left( 
                 s \, (L_{xt}^2 + L_{yt}^2) 
             + C \, s^2 \left( L_{xxt}^2 + 2 L_{xyt}^2 + L_{yyt}^2 \right)
              \right)
 \end{split}\nonumber\\
  \begin{split}
      & \phantom{=}  \,
           + C \, \tau^2
             \left( 
                 s \, (L_{xtt}^2 + L_{ytt}^2)
             \right.
% \end{split}\nonumber\\
%  \begin{split}
      \label{eq-Q3-scalenorm-ders-gamma1}
% & \phantom{= + C \, \tau^2 \left( \right.}
            \left.
                 + C s^2 (L_{xxtt}^2 + 2 L_{xytt}^2 + L_{yytt}^2) 
            \right).
    \end{split}
\end{align}
%here expressed in terms of scale-normalized spatial and temporal
%derivatives for $\gamma_s = 1$ and $\gamma_{\tau} = 1$,
%was demonstrated to have 
%the best properties. 
This differential entity has been
constructed to constitute a simultaneous quasi quad\-rature measure over
both the spatial dimensions $(x, y)$ and the temporal dimension $t$,
implying that instead of combining a pair of first- and second-order
derivatives over a single dimension, here using an octuple of first- and
second-order derivatives over the three spatio-temporal dimensions and with additional
terms added to make the resulting differential expression rotationally
invariant over the spatial domain. 

Specifically, this differential
entity mimics some of the known properties of complex cells in the
primary visual cortex
as discovered by Hubel and Wiesel \cite{HubWie59-Phys,HubWie62-Phys,HubWie05-book}
in the sense of: (i)~being independent of the polarity of the stimuli,
(ii)~not obeying the superposition principle and 
(iii)~being rather insensitive to the phase of the visual stimuli.
The primitive components of the quasi quadrature measure (the partial
derivatives) do in turn mimic some of the known properties of simple cells in the
primary visual cortex in terms of: (i)~precisely localized ``on'' and ``off''
subregions with (ii)~spatial summation within each subregion,
(iii)~spatial antagonism between on- and off-subregions and
(iv)~whose visual responses to stationary or moving spots can be
predicted from the spatial subregions. This model is, however, also
simplified in the sense that the variability over different
orientations and eccentricities over the spatial domain as well as
over motion directions over joint space-time has been
replaced by primitive components in terms of partial derivatives based
on an isotropic scaling parameter over all spatial orientations and
space-time separable receptive fields over the joint space-time domain.

This spatio-temporal quasi quadrature measure is intended
to measure the local energy of the local spatio-temporal derivatives
obtained by combining first- and second-order derivative operators
over both the spatial dimensions and the temporal dimension.
Specifically, it can be seen as
a combination of the previously considered spatial
quasi quadrature measure of the form (\ref{eq-def-quasi-quad})
%\begin{equation}
%  {\cal Q} _{(x, y),\Gamma-norm} L 
%  = \frac{s \, (L_x^2 + L_y^2) +C_s \, s^2 \left( L_{xx}^2 + 2 L_{xy}^2 +  L_{yy}^2 \right)}{s^{\Gamma_s}}
%\end{equation}
for $\Gamma_s = 0$ with the previously derived temporal quasi quadrature measure
(\ref{eq-quasi-quad-t-gamma1-gamma2}) 
% \begin{equation}
%  \label{eq-quasi-time-gamma1}
%  {\cal Q}_{t,\Gamma-norm} L = \frac{\tau L_t^2 + C_{\tau} \tau^2 L_{tt}^2}{\tau^{\Gamma_{\tau}}}
%\end{equation}
%and with
%potentially different relative weighting factors $C_s$ and $C_{\tau}$ 
%between the first- and second-order derivatives over the spatial
%dimensions {\em vs.\/}\ the temporal domain.
for $\Gamma_{\tau} = 0$.
By adding more general $\Gamma$-normalization with 
independent scale normalization parameters $\Gamma_s$ and
$\Gamma_{\tau}$ over space and time, respectively,
% \begin{align}
%   \begin{split}
%     \label{eq-quasi-quad-space-Gamma-norm}
%     {\cal Q} _{(x, y),\Gamma-norm} L 
%     & = \frac{s \, (L_x^2 + L_y^2) +C_s s^2 \left( L_{xx}^2 + 2 L_{xy}^2 +  L_{yy}^2 \right)}{s^{\Gamma_s}}
%   \end{split}\\
%   \begin{split}
%     {\cal Q}_{t,\Gamma-norm} L 
%     & = \frac{\tau L_t^2 + C_{\tau} \tau^2 L_{tt}^2}{\tau^{\Gamma_{\tau}}}
%   \end{split}
% \end{align}
we here extend the definition of the differential expression (\ref{eq-Q3-scalenorm-ders-gamma1})
into the following more general form
\begin{align}
   \begin{split}
     & {\cal Q}_{(x, y, t),\Gamma-norm} L 
  \end{split}\nonumber\\
  \begin{split}
     & = \frac{\tau \, {\cal Q}_{(x, y),\Gamma-norm} L_{t} 
            + C_{\tau} \tau^2 \, {\cal Q}_{(x, y),\Gamma-norm} L_{tt}}{\tau^{\Gamma_{\tau}}}
 \end{split}\nonumber\\
  \begin{split}
      & = \frac{1}{s^{\Gamma_s} \tau^{\Gamma_{\tau}}}
             \left( \tau
               \left( 
                   s \, (L_{xt}^2 + L_{yt}^2) 
%               \right.
%              \right.
%\end{split}\nonumber\\
%  \begin{split}
%        & \phantom{= \frac{1}{s^{\Gamma_s} \tau^{\Gamma_{\tau}}}}
%            \left.
%              \left.
               + C_s \, s^2 \left( L_{xxt}^2 + 2 L_{xyt}^2 + L_{yyt}^2 \right)
               \right)
             \right.
 \end{split}\nonumber\\
  \begin{split}
      & \phantom{= \frac{1}{s^{\Gamma_s} \tau^{\Gamma_{\tau}}} \left( \right.}  \,
           \left.
             + C_{\tau} \, \tau^2
               \left( 
                   s \, (L_{xtt}^2 + L_{ytt}^2)
%               \right.
%          \right.
% \end{split}\nonumber\\
%  \begin{split}
      \label{eq-Q3-scalenorm-ders-Gammanorm}
% & \phantom{= + C_{\tau} \, \tau^2 \left( \right.}
%            \left.
%              \left.
                   + C_{s} \, s^2 (L_{xxtt}^2 + 2 L_{xytt}^2 + L_{yytt}^2) 
              \right)
            \right).
    \end{split}
\end{align}
By the tight integration of the spatial quasi quadrature ${\cal Q} _{(x, y),\Gamma-norm} L$
with the temporal quasi quadrature measure ${\cal Q}_{t,\Gamma-norm} L$,
the intention with 
this combined spatio-temporal quasi quadrature is
to simultaneously allow for combined scale selective properties over 
joint space-time, to allow for joint spatio-temporal scale
selection. Specifically, the fact that all individual components of
this differential invariant (all the partial derivatives $L_{x^{m_1} y^{m_2} t^n}$)
are expressed in terms of non-zero orders of spatial differentiation
$m_1 + m_2 > 0$ and temporal differentiation $n > 0$ ensures that the
resulting expression is localized over both space-time and
spatio-temporal scales.

\subsection{Scale selection properties for a spatio-temporal sine wave}
\label{sc-sel-prop-sine-wave-spat-tem}

In the following, we shall investigate the scale selection properties
that this quasi quadrature measure gives rise to for a multi-dimensional
sine wave of the form
\begin{equation}
  \label{eq-2+1D-sine-wave}
  f(x, y, t) = \left( \sin (\omega_s x) + \sin (\omega_s y) \right) \sin (\omega_{\tau} t)
\end{equation}
taken as an idealized model of a dense spatio-temporal structure over
both space and time and with the spatio-temporal image structures
having spatial extent of size $\lambda_s = 2\pi/\omega_s$ and
temporal duration $\lambda_{\tau} = 2\pi/\omega_{\tau}$.
The spatio-temporal scale-space representation of
(\ref{eq-2+1D-sine-wave}) obtained by Gaussian smoothing 
will then be of the form
\begin{equation}
  \label{eq-spat-temp-scsp-2+1D-sine-wave}
   L(x, y, t;\; s, \tau) %= \\
   = e^{-\omega_s^2 s/2} e^{-\omega_{\tau}^2 \tau/2} 
      \left( \sin (\omega_s x) + \sin (\omega_s y) \right) \sin (\omega_{\tau} t).
\end{equation}
%For later purposes, we shall 
Let us decompose this %spatio-temporal 
quasi quadrature measure into
the following four components based on spatial and temporal
derivatives of either first or second order
\begin{align}
  \begin{split}
     {\cal Q}_{(x, y, t),\Gamma-norm} L = 
 \end{split}\nonumber\\
  \begin{split}
      & = {\cal Q}_{(x,y),1,\Gamma-norm} L_t + {\cal Q}_{(x,y),2,\Gamma-norm} L_t 
  \end{split}\nonumber\\
  \begin{split}
     & \phantom{=} 
+ C_{\tau} \left( {\cal Q}_{(x,y),1,\Gamma-norm} L_{tt} +  {\cal Q}_{(x,y),2,\Gamma-norm} L_{tt} \right),
  \end{split}
\end{align}
where
\begin{align}
  \begin{split}
     {\cal Q}_{(x,y),1,\Gamma-norm} L_t
     & = \frac{s \, \tau \, (L_{xt}^2 + L_{yt}^2) }{s^{\Gamma_s} \tau^{\Gamma_{\tau}}},
  \end{split}\\
  \begin{split}
     {\cal Q}_{(x,y),2,\Gamma-norm} L_t
     & = \frac{C_s \, s^2 \, \tau \, \left( L_{xxt}^2 + 2 L_{xyt}^2 + L_{yyt}^2 \right)}{s^{\Gamma_s} \tau^{\Gamma_{\tau}}},
  \end{split}\\
%\end{align}
%\begin{align}
 \begin{split}
     {\cal Q}_{(x,y),1,\Gamma-norm} L_{tt}
     & = \frac{s \, \tau^2 \, (L_{xtt}^2 + L_{ytt}^2) }{s^{\Gamma_s} \tau^{\Gamma_{\tau}}},
  \end{split}\\
 \begin{split}
     {\cal Q}_{(x,y),2,\Gamma-norm} L_{tt}
     & = \frac{C_s \, s^2 \, \tau^2 \, \left( L_{xxtt}^2 + 2 L_{xytt}^2 + L_{yytt}^2 \right)}{s^{\Gamma_s} \tau^{\Gamma_{\tau}}}.
  \end{split}
\end{align}
By selecting both spatial and temporal scales from local extrema of
the quasi quadrature measure over both spatial and temporal scales
\begin{equation}
     \label{eq-Q3-spat-temp-scaleest-regular-geom-sine-s-tau}
     (\hat{s}_{{\cal Q}_{(x,y,t),\Gamma-norm}}, \hat{\tau}_{{\cal Q}_{(x,y,t),\Gamma-norm}})
     % = \\ 
     = \argmaxlocal_{s,\tau} {\cal Q}_{(x,y,t),\Gamma-norm} L,
\end{equation}
it follows that
\begin{itemize}
\item
   at the spatial points $(x = n \pi/\omega_s, y = n \pi/\omega_s)$  
  at which only the first-order spatial derivatives respond,
  the selected spatial scale will be
  \begin{equation}
    \hat{s}_{11} = \frac{1-\Gamma_s}{\omega_s^2},
  \end{equation}
\item
   at the spatial points $(x = (\pi/2 + n \pi)/\omega_s, y = (\pi/2+n \pi)/\omega_s)$  
  at which only the second-order spatial derivatives respond,
  the selected spatial scale will be
  \begin{equation}
    \hat{s}_{22} = \frac{2-\Gamma_s}{\omega_s^2},
  \end{equation}
\item
   at the temporal moments $t = n \pi/\omega_{\tau}$  
  at which only the first-order temporal derivative responds,
  the selected temporal scale will be
  \begin{equation}
    \hat{\tau}_1 = \frac{1-\Gamma_{\tau}}{\omega_{\tau}^2}
  \end{equation}
\item
   and at the temporal moments $t = (\pi/2 + n \pi)/\omega_{\tau}$  
   at which only the second-order temporal derivative responds,
  the selected temporal scale will be
  \begin{equation}
    \hat{\tau}_2 = \frac{2-\Gamma_{\tau}}{\omega_{\tau}^2}.
  \end{equation}
\end{itemize}
Determining the weighting parameters $C_s$ and $C_{\tau}$ such that the
relative strengths of the first- and second-order components become
equal at the spatial and temporal midpoints 
$(x = (\pi/4 + n \pi/2)/\omega_s, y = (\pi/4+n \pi/2)/\omega_s)$
and $t = (\pi/4 + n \pi/2)/\omega_{\tau}$
between the extreme points and at the spatial and temporal scales
corresponding to the geometric averages $\sqrt{\hat{s}_1 \hat{s}_2}$
and $\sqrt{\hat{\tau}_1 \hat{\tau}_2}$ of the extreme values,
% \begin{align}
%   \begin{split}
%     & \left. 
%         \left( {\cal Q}_{(x,y),1,\Gamma-norm} L_t + C_{\tau} {\cal Q}_{(x,y),1,\Gamma-norm} L_{tt} \right)
%     \right|_P
%   \end{split}\nonumber\\
%   \begin{split}
%     \label{eq-Q3-determ-Cs-balanced-first-second-order-responses}
%     & = \left.  
%            C_s \left( {\cal Q}_{(x,y),2,\Gamma-norm} L_t + C_{\tau} {\cal Q}_{(x,y),2,\Gamma-norm} L_{tt} \right)
%        \right|_P
%   \end{split}\\
%   \begin{split}
%     & \left. 
%         \left( {\cal Q}_{(x,y),1,\Gamma-norm} L_t + C_s {\cal Q}_{(x,y),2,\Gamma-norm} L_t \right)
%     \right|_P
%   \end{split}\nonumber\\
%   \begin{split}
%     \label{eq-Q3-determ-Ctau-balanced-first-second-order-responses}
%     & = \left.  
%            C_{\tau} \left( {\cal Q}_{(x,y),1,\Gamma-norm} L_{tt} + C_s {\cal Q}_{(x,y),2,\Gamma-norm} L_{tt} \right)
%        \right|_P
%   \end{split}
% \end{align}
% at
% \begin{equation}
%   P = (x, y, t, s, \tau) = %\\ =
%   \left(
%      \frac{\pi}{4 \omega_s}, \frac{\pi}{4 \omega_s}, \frac{\pi}{4 \omega_{\tau}},
%      \sqrt{\hat{s}_{11} \, \hat{s}_{22}}, \sqrt{\hat{\tau}_1, \, \hat{\tau}_2}
%   \right),
% \end{equation}
then implies that the relative weighting factors $C_s$ and $C_{\tau}$
between the first- and second-order derivative responses should be
chosen as
\begin{align}
  \begin{split}
    C_s = \frac{1}{\sqrt{(1 - \Gamma_s)(2 - \Gamma_s)}},
  \end{split}\\
  \begin{split}
    C_{\tau} = \frac{1}{\sqrt{(1 - \Gamma_{\tau})(2 - \Gamma_{\tau})}}. 
  \end{split}
\end{align}
Note that structural similarities between these results and the
corresponding analysis for the purely spatial quasi quadrature
measure ${\cal Q}_{(x,y),\Gamma-norm} L$ studied in
section~\ref{sec-spat-dense-scsel}.

\subsection{Spatio-temporal scale covariance of the joint
  spatio-temporal scale estimates under independent scaling
  transformations of the spatial and the temporal domains}

Consider an independent scaling transformation of the spatial and
the temporal domains of a video sequence 
\begin{equation}
f'(x_1', x_2', t') = f(x_1, x_2, t)
%\end{equation}
\quad\quad \mbox{for} \quad\quad
%\begin{equation}
(x_1', x_2', t') = (S_s \, x_1, S_s \, x_2, S_{\tau} \, t),
\end{equation}
where $S_s$ and $S_{\tau}$ denote the spatial and temporal scaling
factors, respectively.
Then, corresponding spatio-temporal scale covariance of the
spatio-temporal scale estimates 
\begin{equation}
  \label{eq-scale-cov-spattemp-scsel}
   (\hat{s}', \hat{\tau}') = (S_s^2 \, \hat{s}, S_{\tau}^2 \, \hat{\tau})
\end{equation}
provided that the spatial positions $(x, y)$ and the temporal moments
$t$ are appropriately matched 
%\begin{equation}
  $(x_1', x_2', t') = (S_s \, x_1, S_s \, x_2, S_{\tau} \, t)$
%\end{equation}
can be proven by combining the ideas in the proof of spatial
scale covariance in Section~\ref{sec-cov-prop-spat-sc-sel} with the
ideas in the proof of temporal scale covariance in Section~\ref{sec-cov-prop-temp-sc-sel}.

\subsection{Experimental results}

\begin{figure*}[hbtp]
  \begin{center}
    \begin{tabular}{cc} 
 {\footnotesize\em grey-level frame}
      & {\footnotesize\em quasi quadrature measure at fixed s-t scale} \\
    \includegraphics[width=0.44\textwidth]{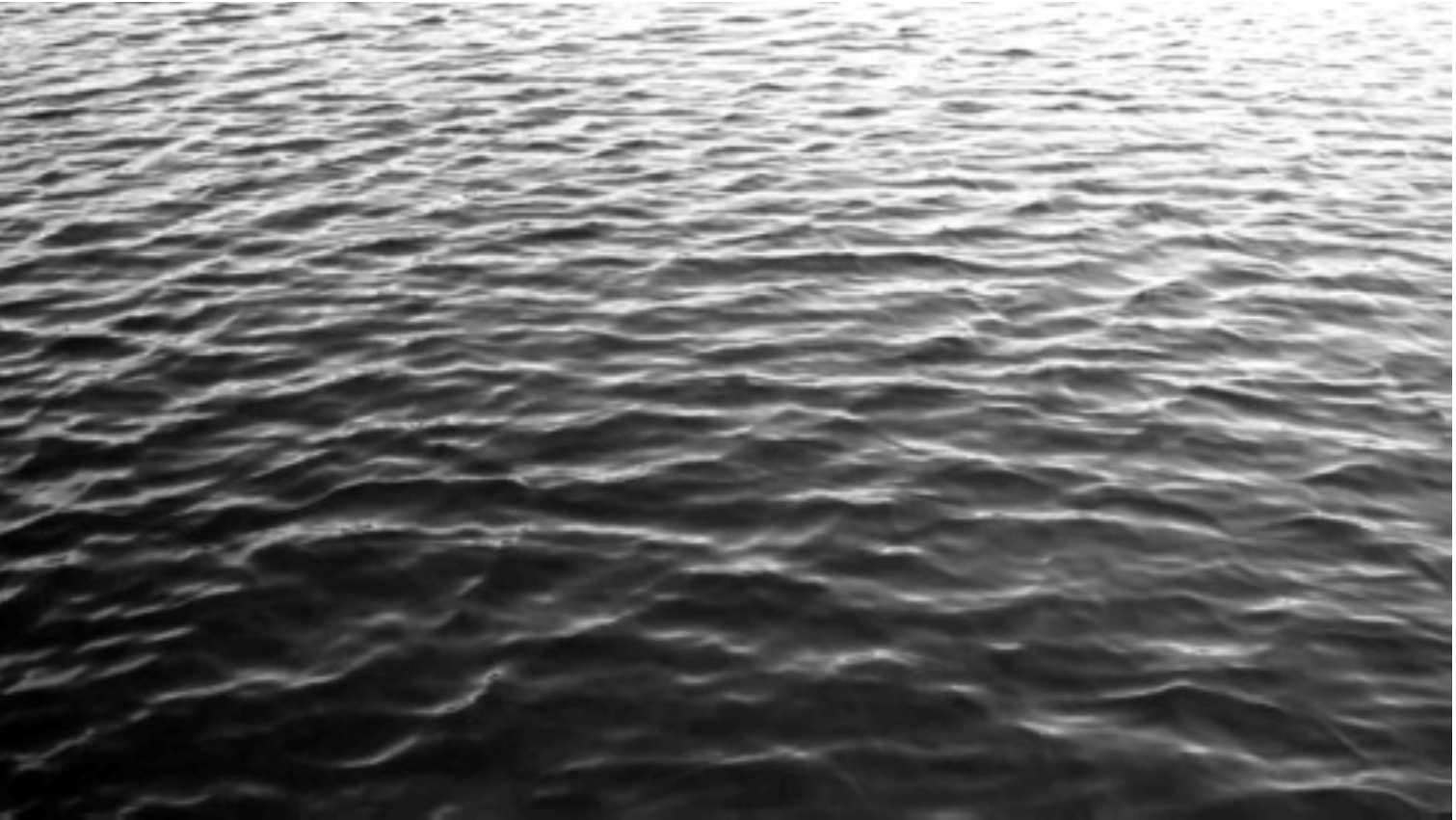} 
       & \includegraphics[width=0.44\textwidth]{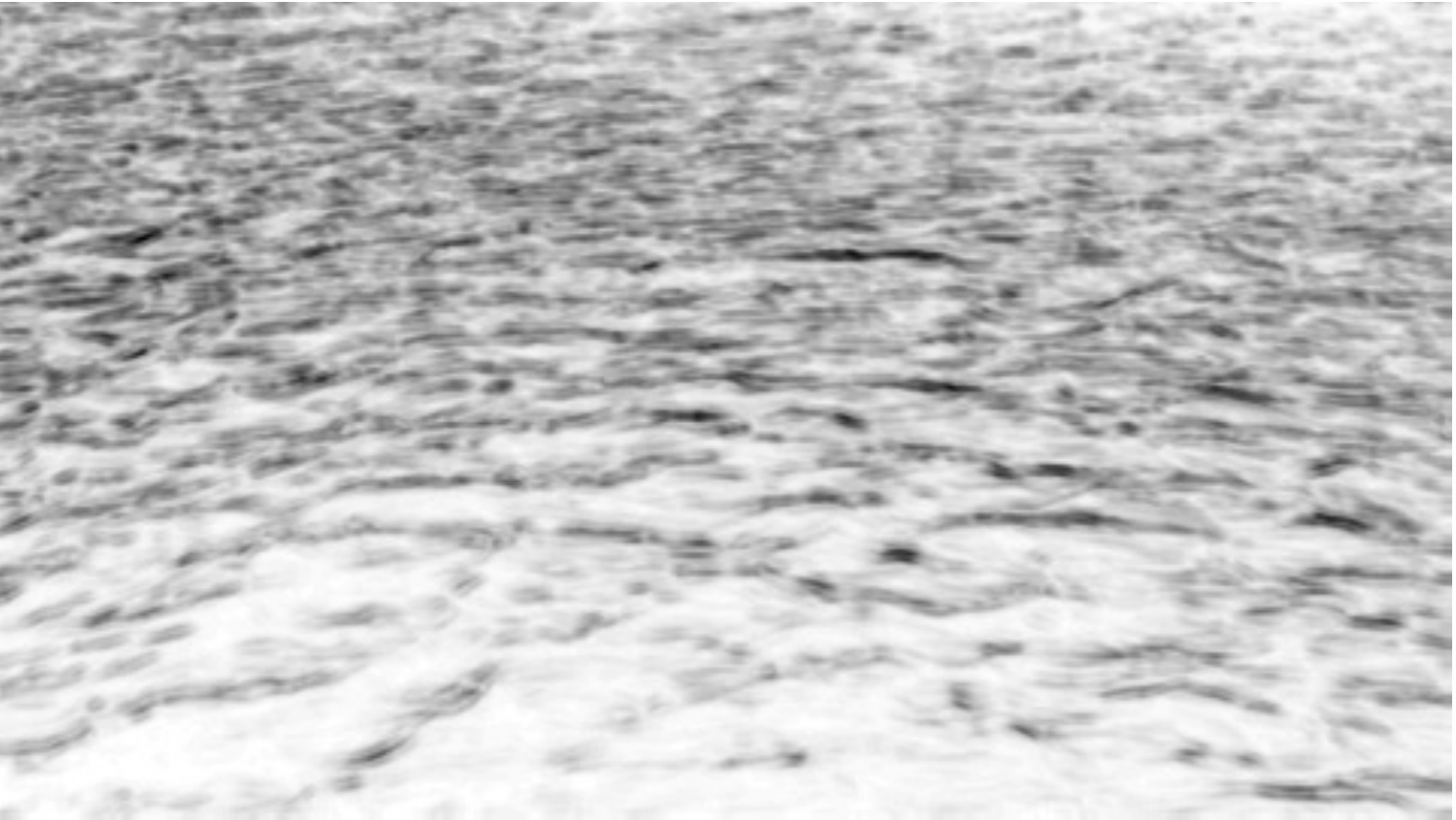} \\ 
     {\footnotesize\em scale map of effective spatial scales}
      & {\footnotesize\em scale map of effective temporal scales} \\
    \includegraphics[width=0.44\textwidth]{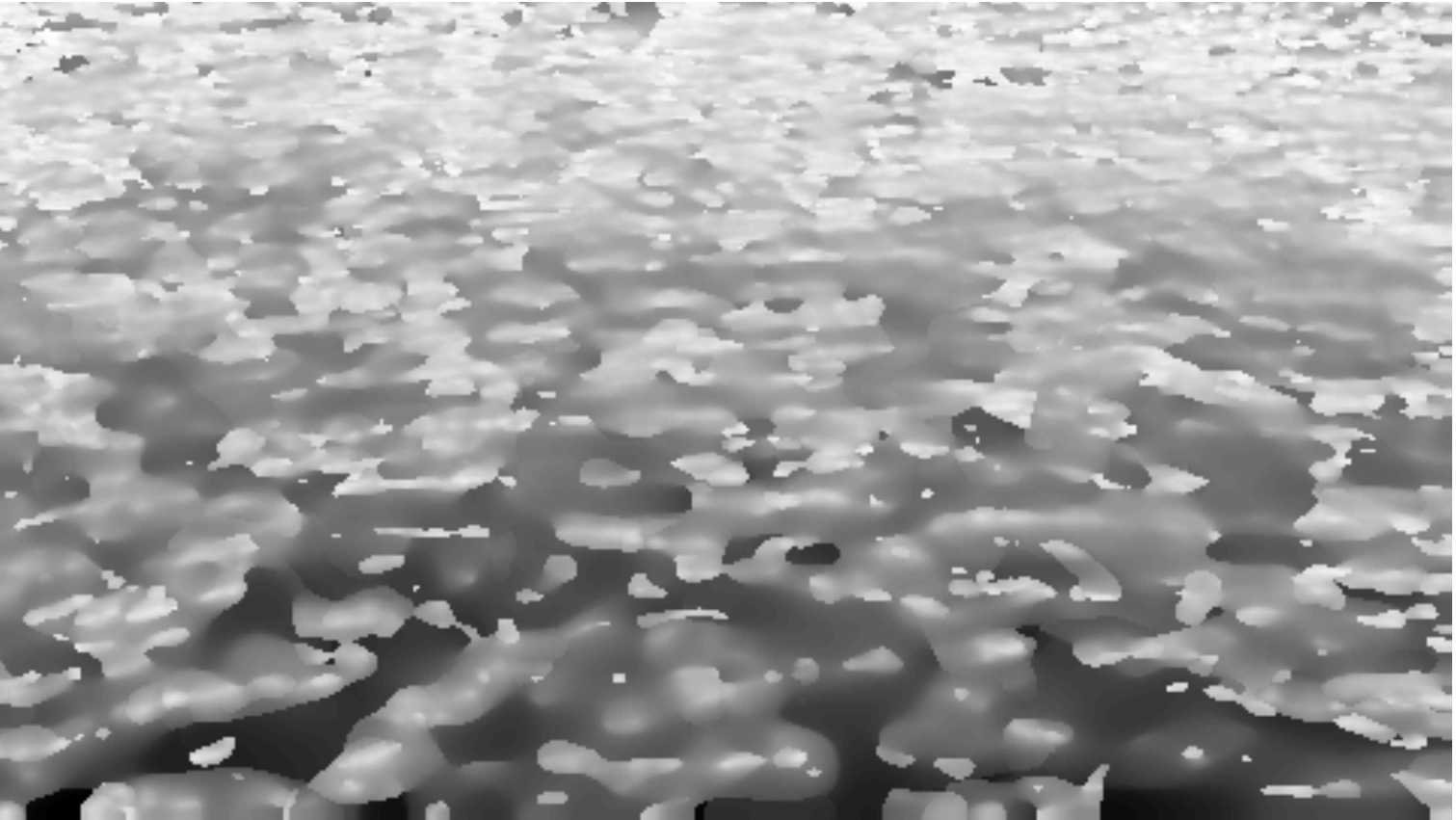}
       & \includegraphics[width=0.44\textwidth]{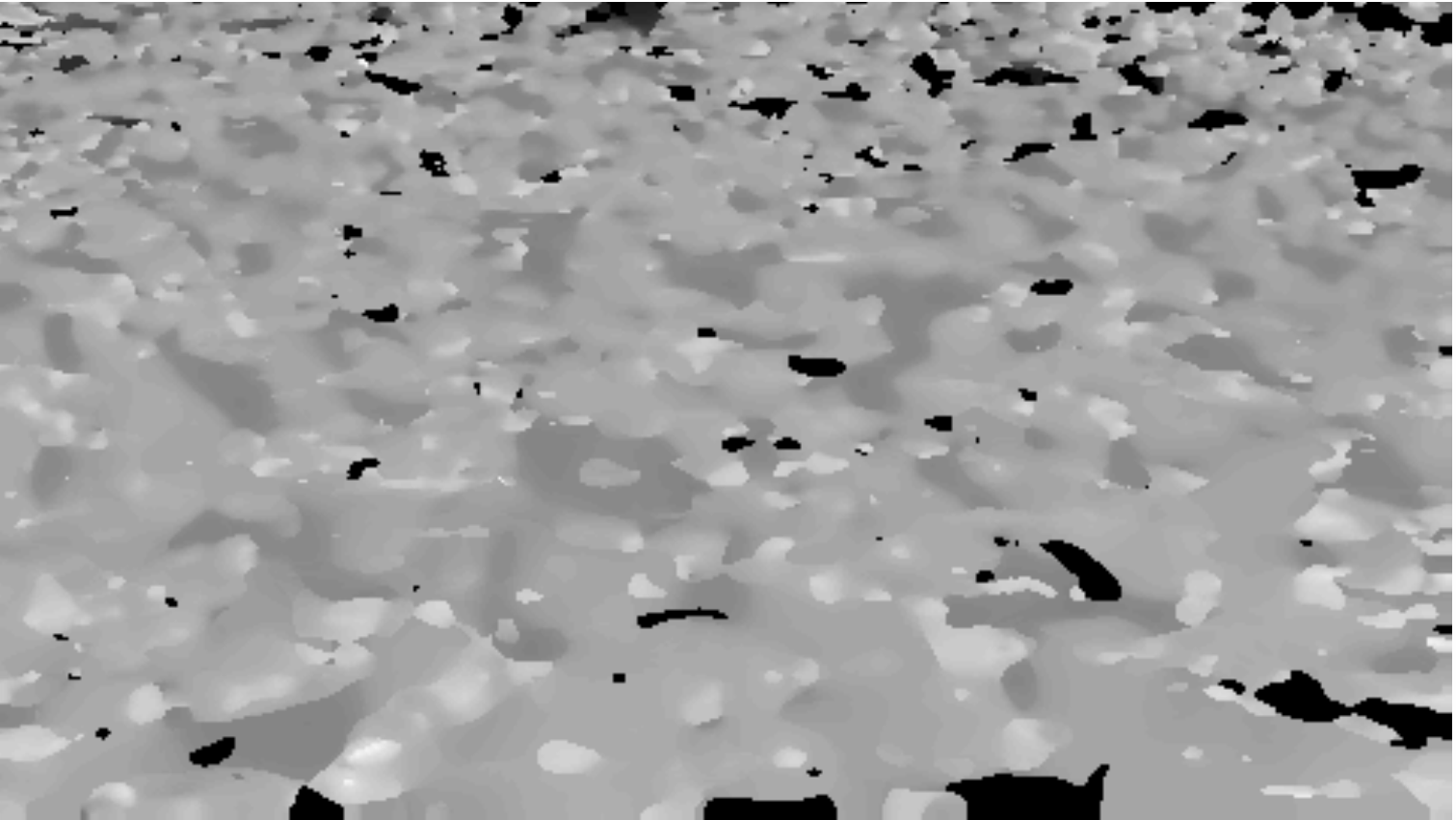} \\
     {\footnotesize\em  spatio-temporal scale-space signature}
      & {\footnotesize\em maximum magnitude response over all scales} \\
    \includegraphics[width=0.44\textwidth]{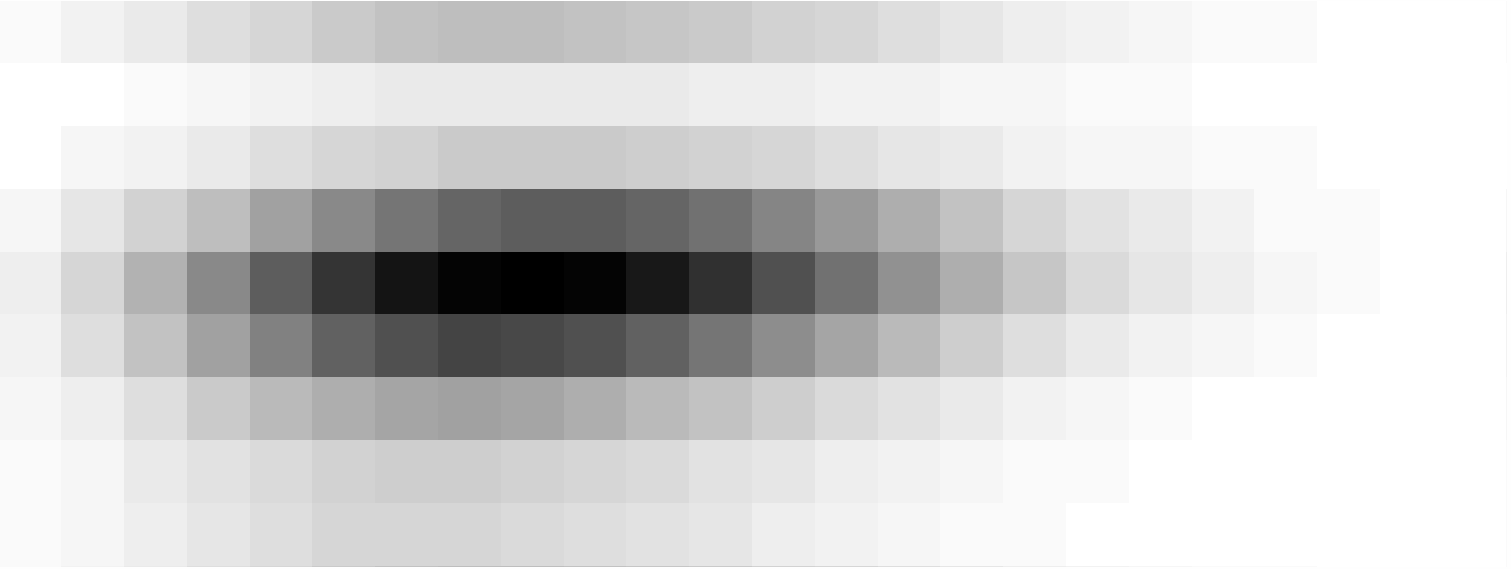}
       & \includegraphics[width=0.44\textwidth]{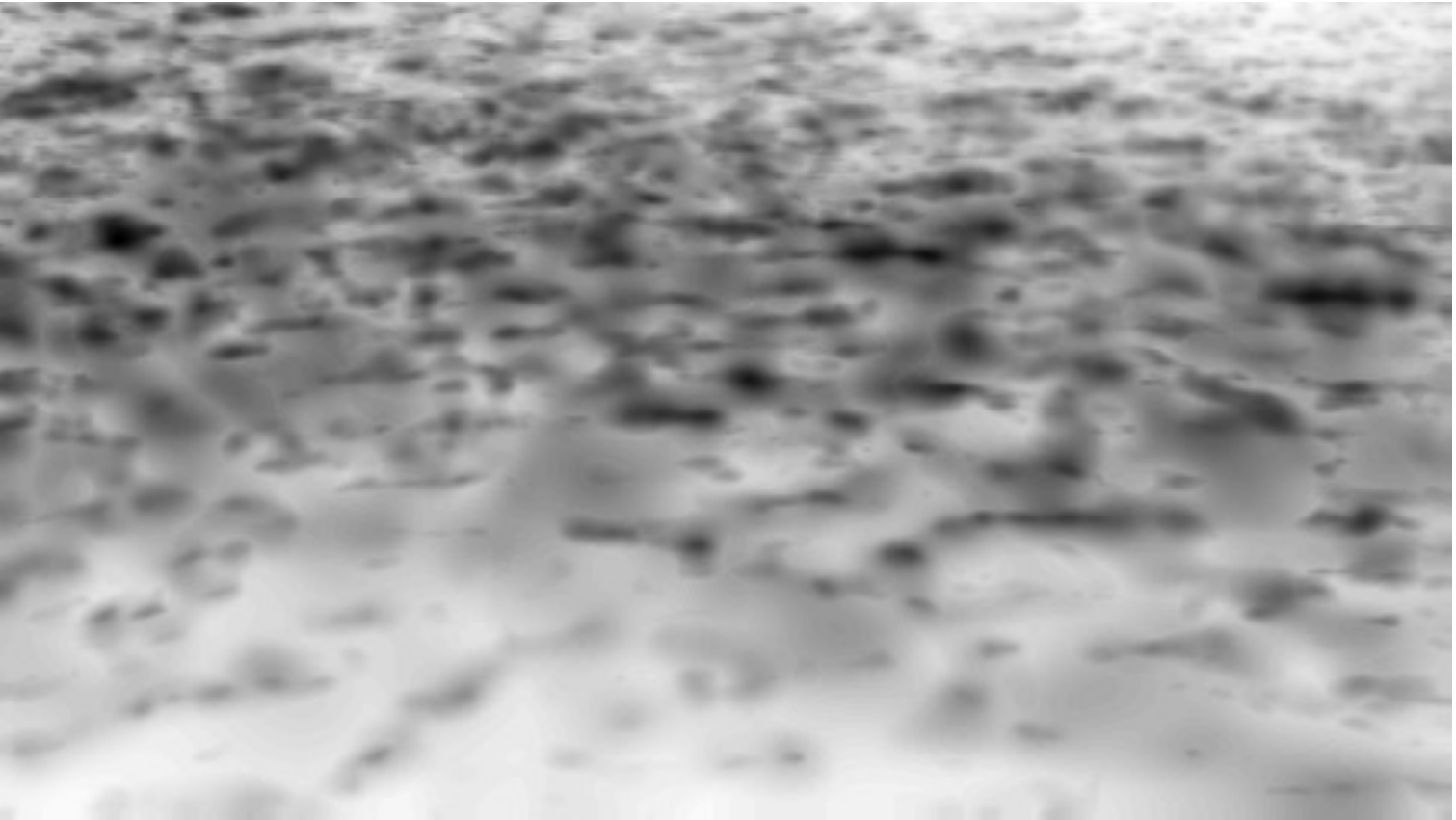} \\
     \end{tabular}
  \end{center}
\caption{Results of dense spatio-temporal scale selection applied to a
  video sequence with water waves. 
 The results have been computed with
  a time-causal and time-recursive spatio-temporal scale-space
  representation obtained by convolution with a Gaussian kernel over
  the spatial domain and the time-causal limit kernel over the
  temporal domain. 
  (top left) Grey-level image.
  (top right) The spatio-temporal quasi quadrature measure computed at
a fixed spatio-temporal scale.
  (middle left) Selected spatial scale levels in units of effective
  scale $s_{eff} = \log_2 \sigma_s$.
  (middle right) Selected temporal scale levels in units of effective
  scale $\tau_{eff} = \log_2 \sigma_{\tau}$.
  (bottom left) Spatio-temporal scale-space signature showing the
  magnitude variations of the spatio-temporal quasi quadrature measure
over both spatial and temporal scales, with effective spatial scale
increasing linearly from left to right and effective temporal scale
increasing linearly from bottom to top. While this illustration shows
the average over all the image points at the given image frame for the
purpose of suppressing the influence of local spatial variations, the general
dense scale selection method is otherwise local, based on individual
scale-space signatures at every image point and for every time moment.
  (bottom right) The maximum magnitude response over all spatial and
  temporal scales at every image point.
  Note from the maps of the selected spatial and temporal scales that
  there is a clear vertical size gradient for the the selected spatial
scales, whereas there is no size gradient for the selected temporal
scales. The reason for this is that there are size variations over the
spatial domain because of perspective depth effects affecting the
spatial scales, whereas the temporal scales are stationary over the
image domain, since temporal scale levels are not affected by the
perspective mapping. Note that the contrast of the maps showing
magnitude information has been set so that dark corresponds to larger
values and bright to lower values. In addition, the magnitude maps
have been stretched by a square root function. 
(Results computed using 24 logarithmically distributed spatial scale levels between $\sigma_{s,min} = 0.25$~pixels and 
 $\sigma_{s,max} = 24$~pixels and 9 logarithmically distributed temporal scale levels between
$\sigma_{\tau,min} = 10$~ms and $\sigma_{\tau,max} = 2.56$~seconds for
$C_s =1/\sqrt{(1 - \Gamma_s)(2 - \Gamma_s)}$
and $C_{\tau} =1/\sqrt{(1 - \Gamma_{\tau})(2 - \Gamma_{\tau})}$ using
complementary scale normalization parameters $\Gamma_s = 0$
and $\Gamma_{\tau} = 0$.)
(Image size: $480 \times 270$ pixels. Frame 100 of 250 frames at 25
frames per second.)}
  \label{fig-waves0007-denseSTscsel}
\end{figure*}

\begin{figure*}[hbtp]
  \begin{center}
    \begin{tabular}{cc} 
 {\footnotesize\em grey-level frame}
      & {\footnotesize\em quasi quadrature measure at fixed s-t scale} \\
    \includegraphics[width=0.44\textwidth]{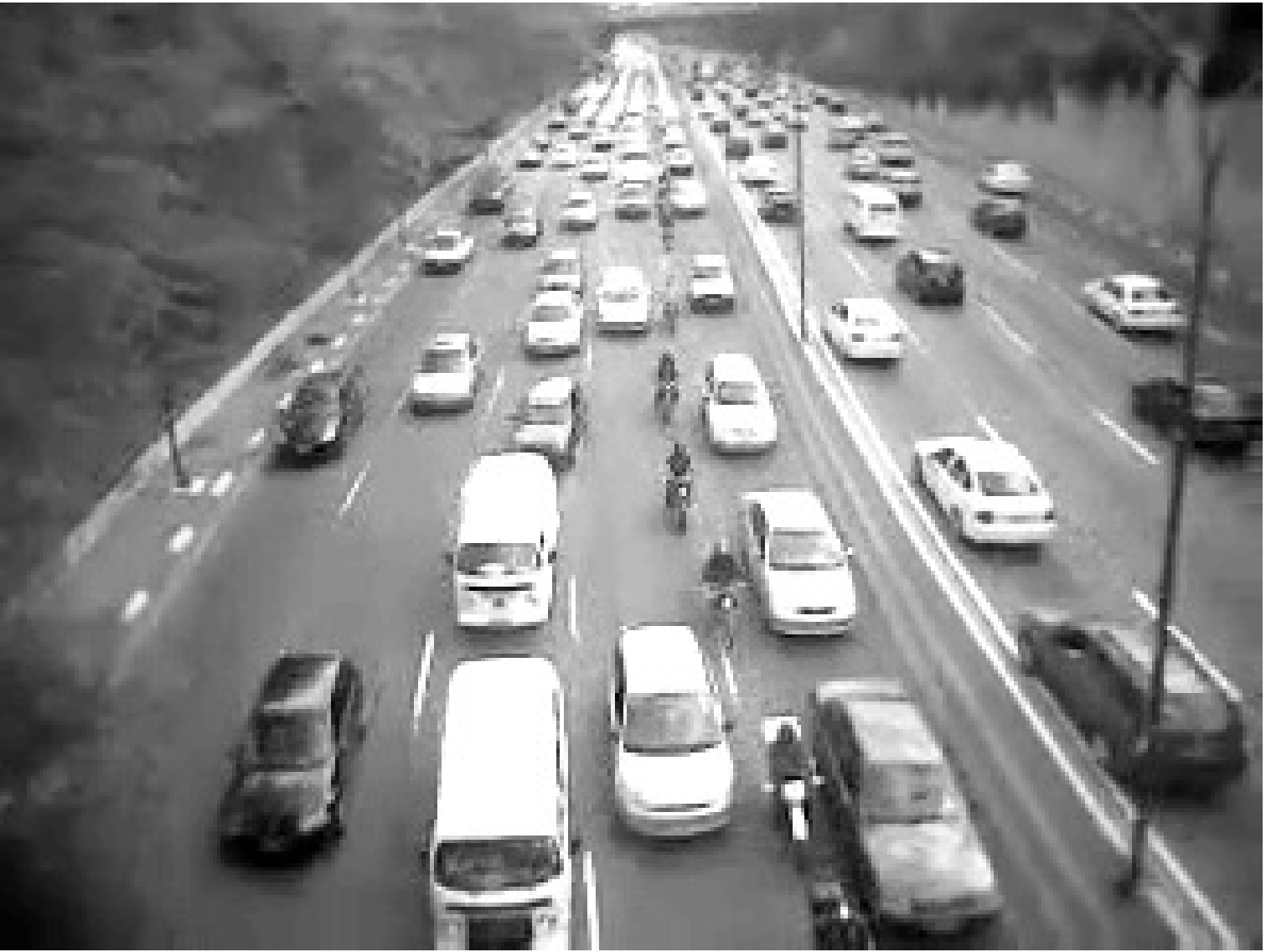} 
       & \includegraphics[width=0.44\textwidth]{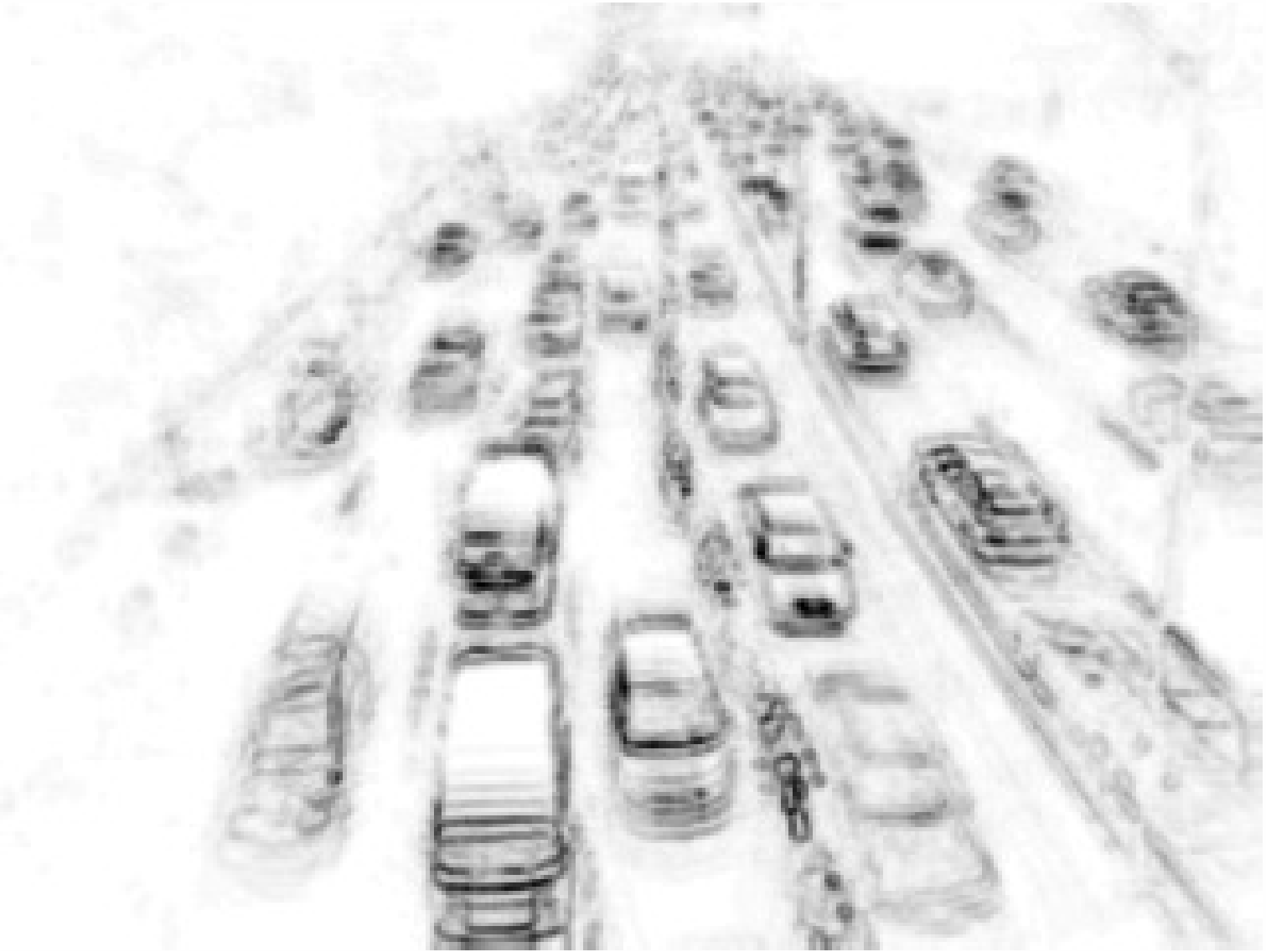} \\ 
     {\footnotesize\em scale map of effective spatial scales}
      & {\footnotesize\em scale map of effective temporal scales} \\
    \includegraphics[width=0.44\textwidth]{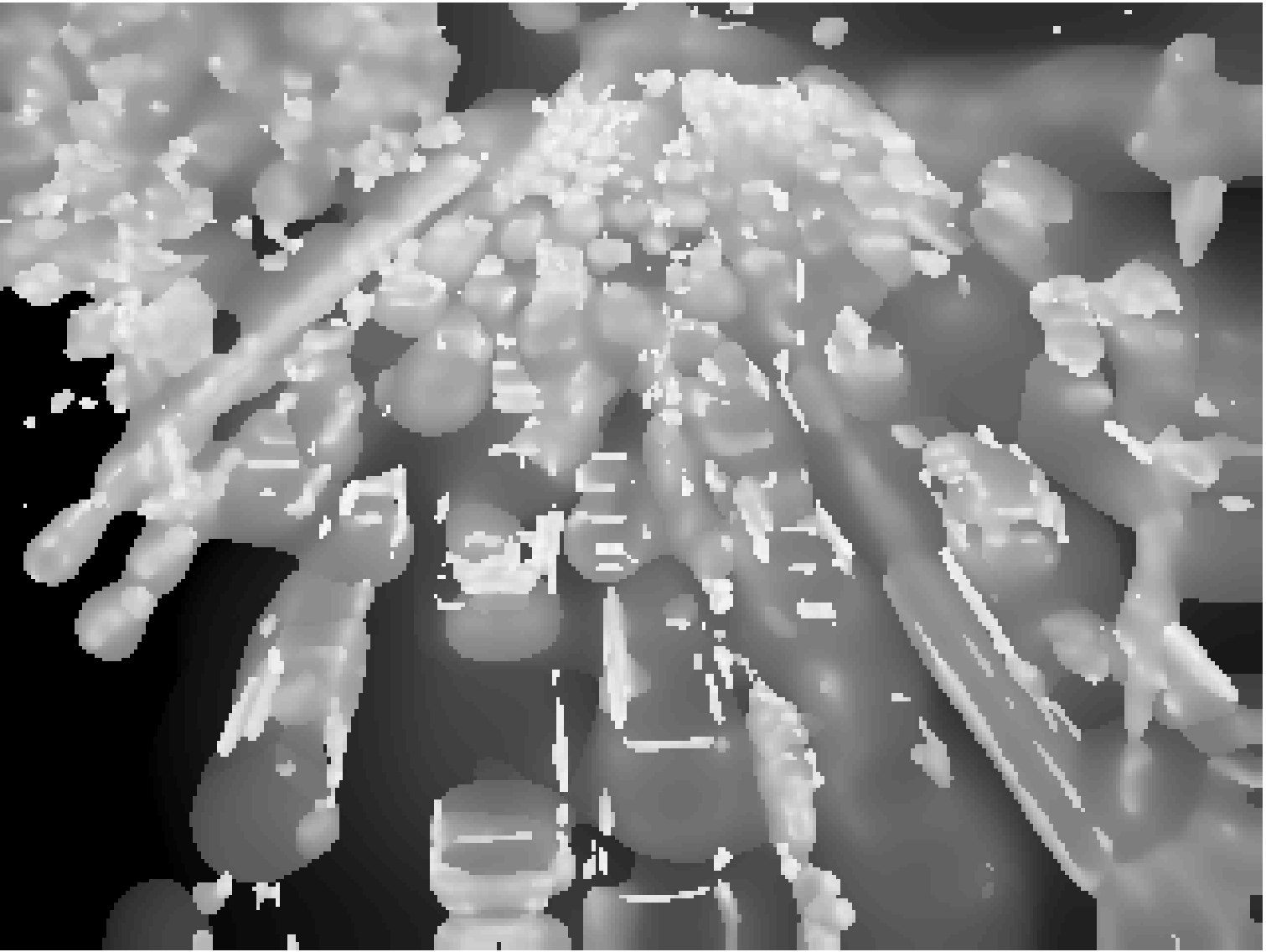}
       & \includegraphics[width=0.44\textwidth]{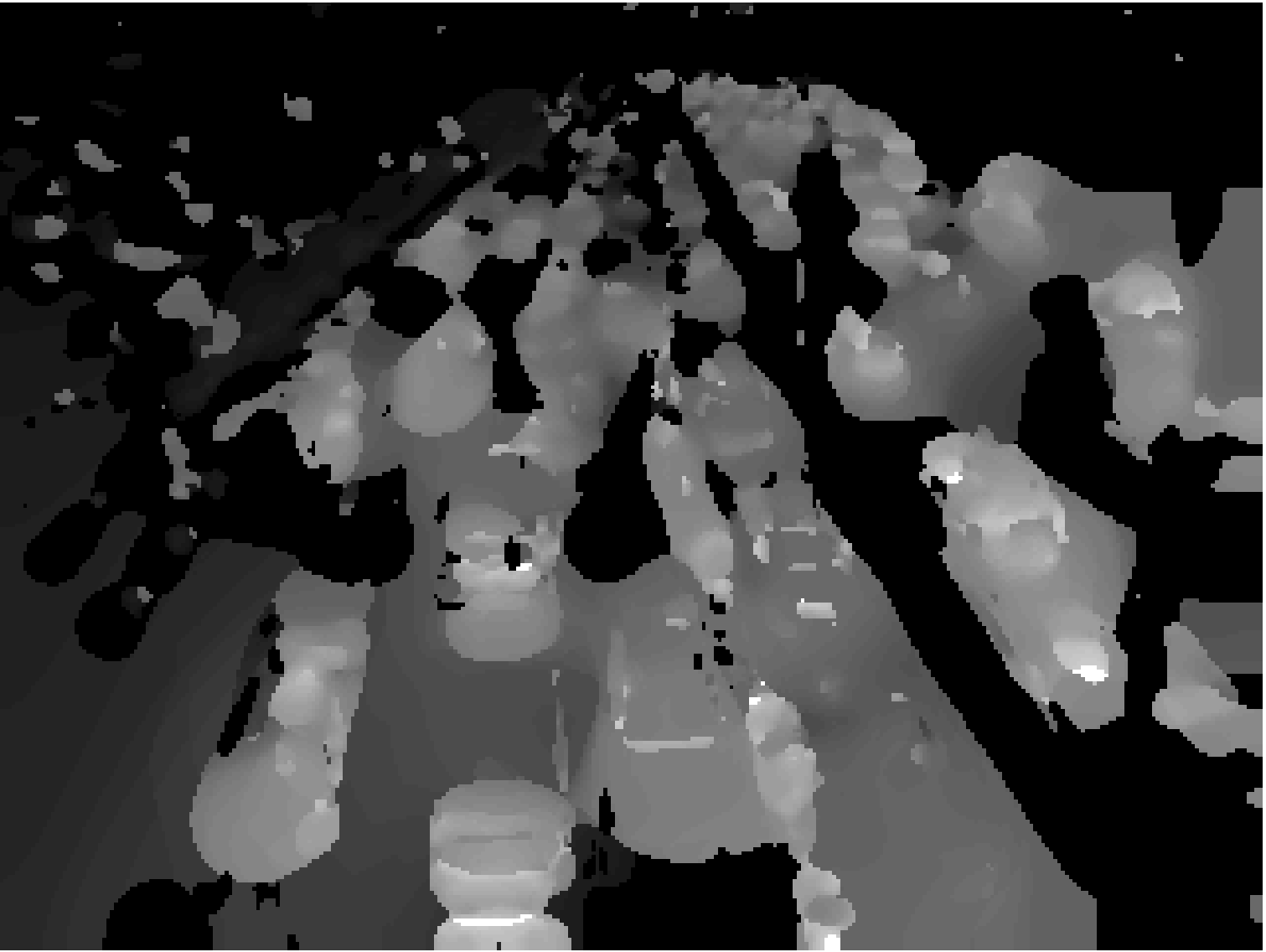} \\
     {\footnotesize\em  spatio-temporal scale-space signature}
      & {\footnotesize\em maximum magnitude response over all scales} \\
    \includegraphics[width=0.44\textwidth]{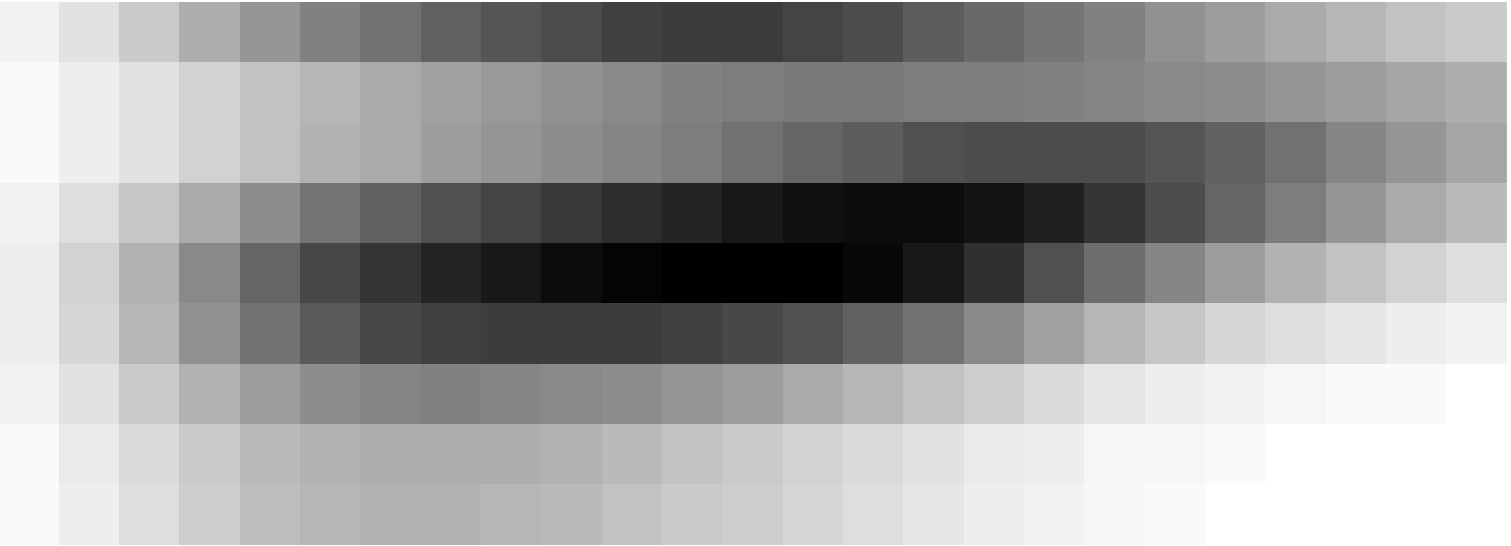}
       & \includegraphics[width=0.44\textwidth]{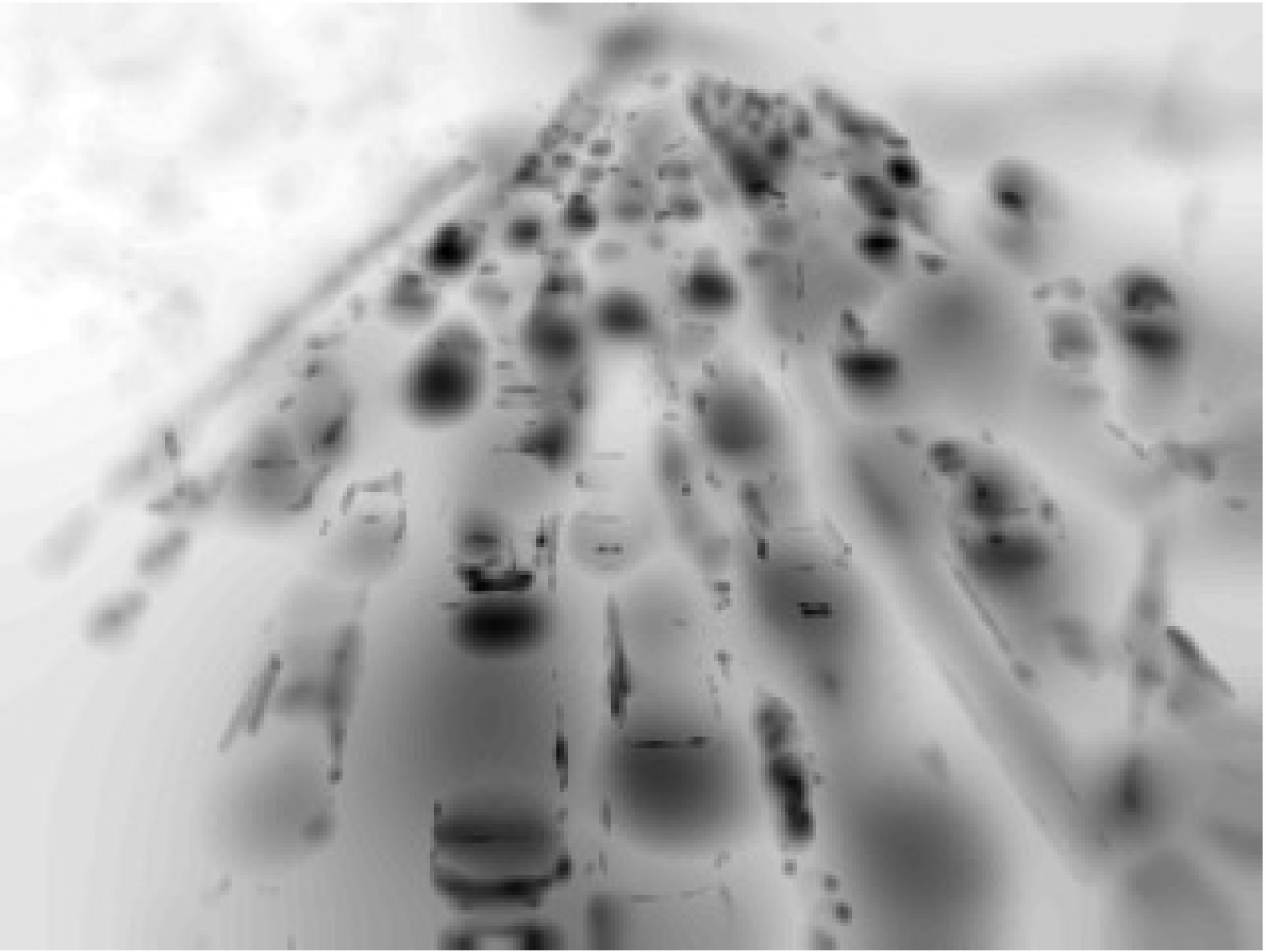} \\
     \end{tabular}
  \end{center}
\caption{Results of dense spatio-temporal scale selection applied to a
  traffic scene (video ``smooth\_traffic05'' from the Maryland dynamic scene
  dataset \cite{ShrTurChe10-CVPR}). 
%  The results have been computed with
%   a time-causal and time-recursive spatio-temporal scale-space
%   representation obtained by convolution with a Gaussian kernel over
%   the spatial domain and the time-causal limit kernel over the
%   temporal domain. 
%   (top left) Grey-level image.
%   (top right) The spatio-temporal quasi quadrature measure computed at
% a fixed spatio-temporal scale.
%   (middle left) Selected spatial scale levels in units of effective
%   scale $s_{eff} = \log_2 \sigma_s$.
%   (middle right) Selected spatial scale levels in units of effective
%   scale $\tau_{eff} = \log_2 \sigma_{\tau}$.
%   (bottom left) Spatio-temporal scale-space signature showing the
%   magnitude variations of the spatio-temporal quasi quadrature measure
% over both spatial and temporal scales, with effective spatial scale
% increasing linearly from left to right and effective temporal scale
% increasing linearly from bottom to top. While this illustration shows
% the average over all image points at the given image fram for the
% purpose of suppressing the influence of local variations, the general
% dense scale selection method is otherwise local, based on individual
% scale-space signatures at every image point and for every time moment.
%   (bottom right) The maximum magnitude response over all spatial and
%   temporal scales at every image point.
 Note that
distinct responses in the spatial and temporal scale maps are obtained
for the different moving cars, again with a size gradient in the
spatial scale estimates reflecting the perspective scaling effects,
whereas the temporal scale estimates are essentially unaffected by the
perspective transformation. Additionally, we can observe that large
spatial scales and long temporal scales are selected in the smooth
stationary regions on the road and in some parts of the background
%   Note that the contrast of the maps showing
% magnitude information has been set so that dark corresponds to larger
% values and bright to lower values. In addition, the magnitude maps
% have been stretched by a square root function. 
% (Results computed using 25 logarithmically distributed spatial scale levels between $\sigma_{s,min} = 0.25$~pixels and 
%  $\sigma_{s,max} = 30$~pixels and 9 logarithmically distributed temporal scale levels between
% $\sigma_{\tau,min} = 8.3$~ms and $\sigma_{\tau,max} = 2.1$~seconds.)
% (Complementary scale normalization parameters $\Gamma_s = 0$
% and $\Gamma_{\tau} = 0$.)
(Image size: $320 \times 240$ pixels. Frame 80 of 1217 frames at 30
frames per second.)}
  \label{fig-smoothtraff05-denseSTscsel}
\end{figure*}

\begin{figure*}[hbtp]
  \begin{center}
    \begin{tabular}{cc} 
 {\footnotesize\em grey-level frame}
      & {\footnotesize\em quasi quadrature measure at fixed s-t scale} \\
    \includegraphics[width=0.44\textwidth]{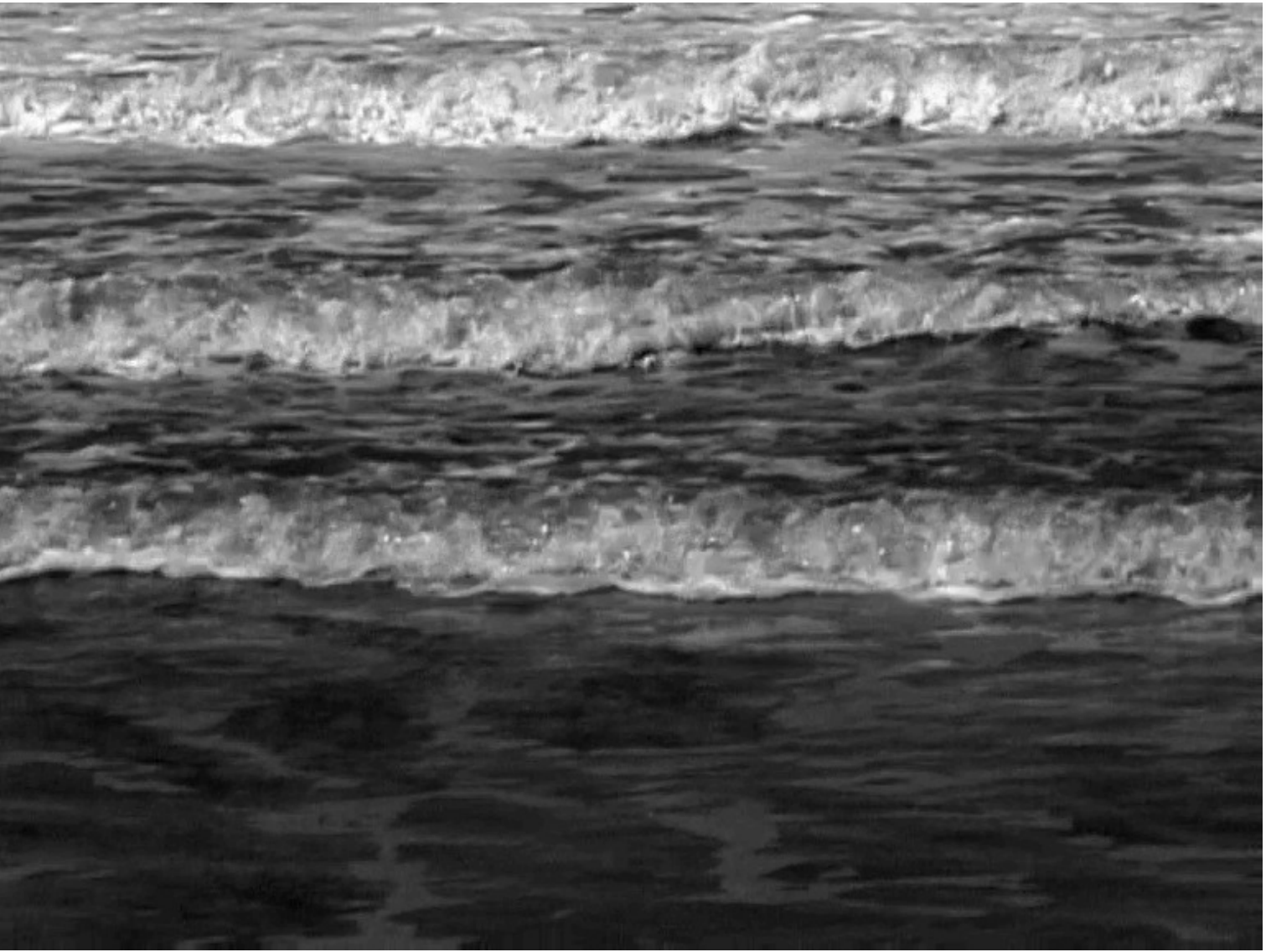} 
       & \includegraphics[width=0.44\textwidth]{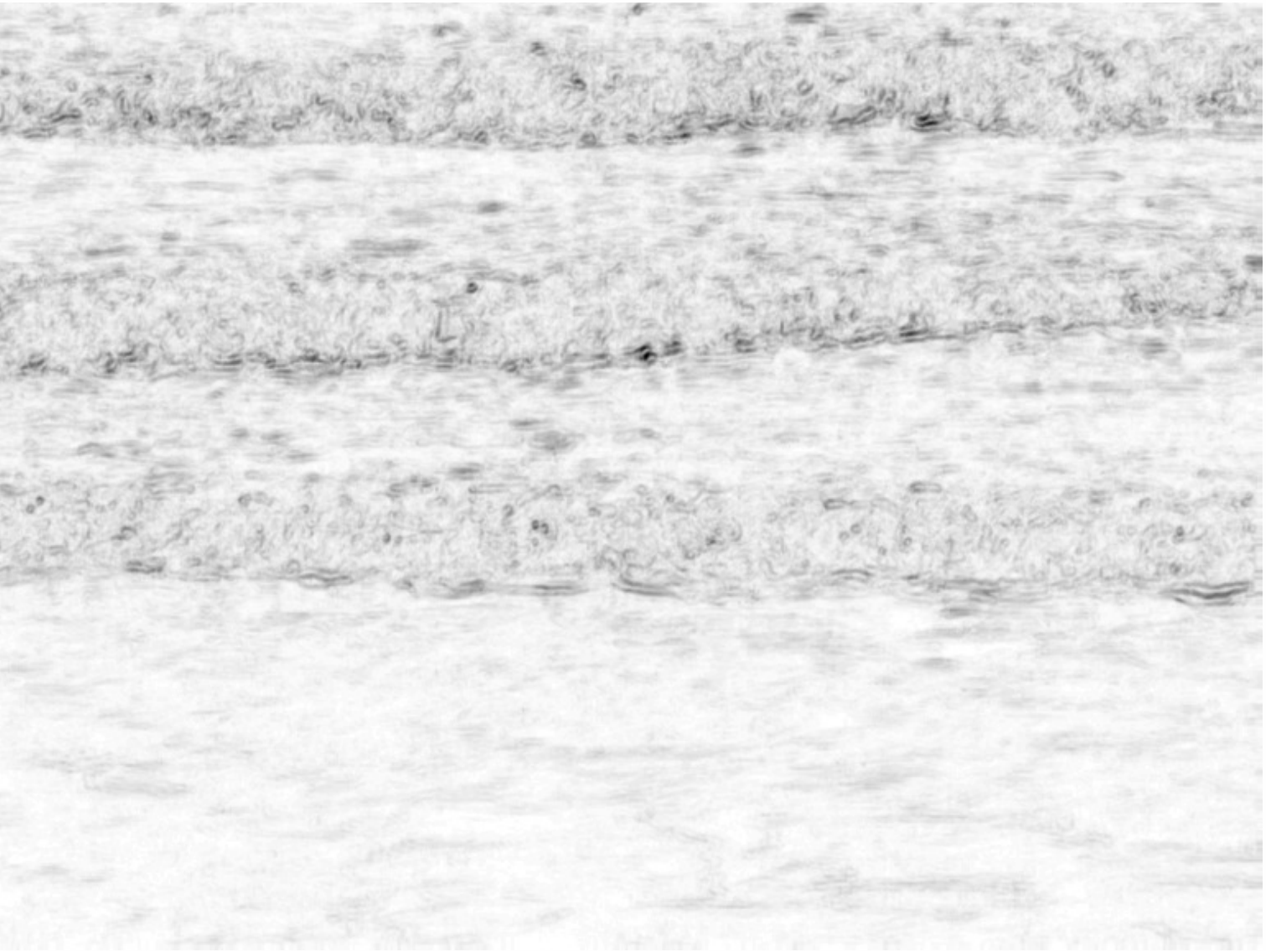} \\ 
     {\footnotesize\em scale map of effective spatial scales}
      & {\footnotesize\em scale map of effective temporal scales} \\
    \includegraphics[width=0.44\textwidth]{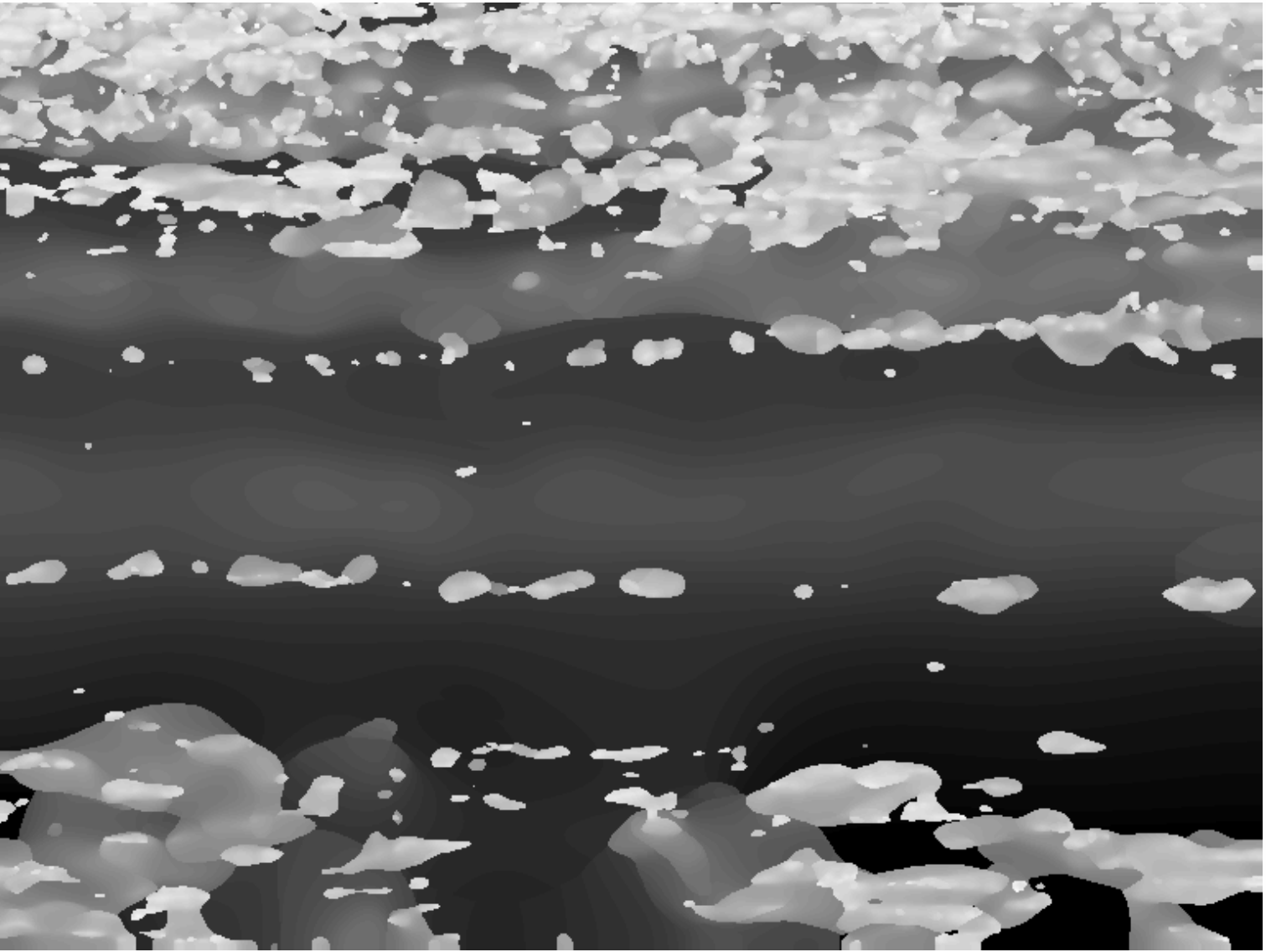}
       & \includegraphics[width=0.44\textwidth]{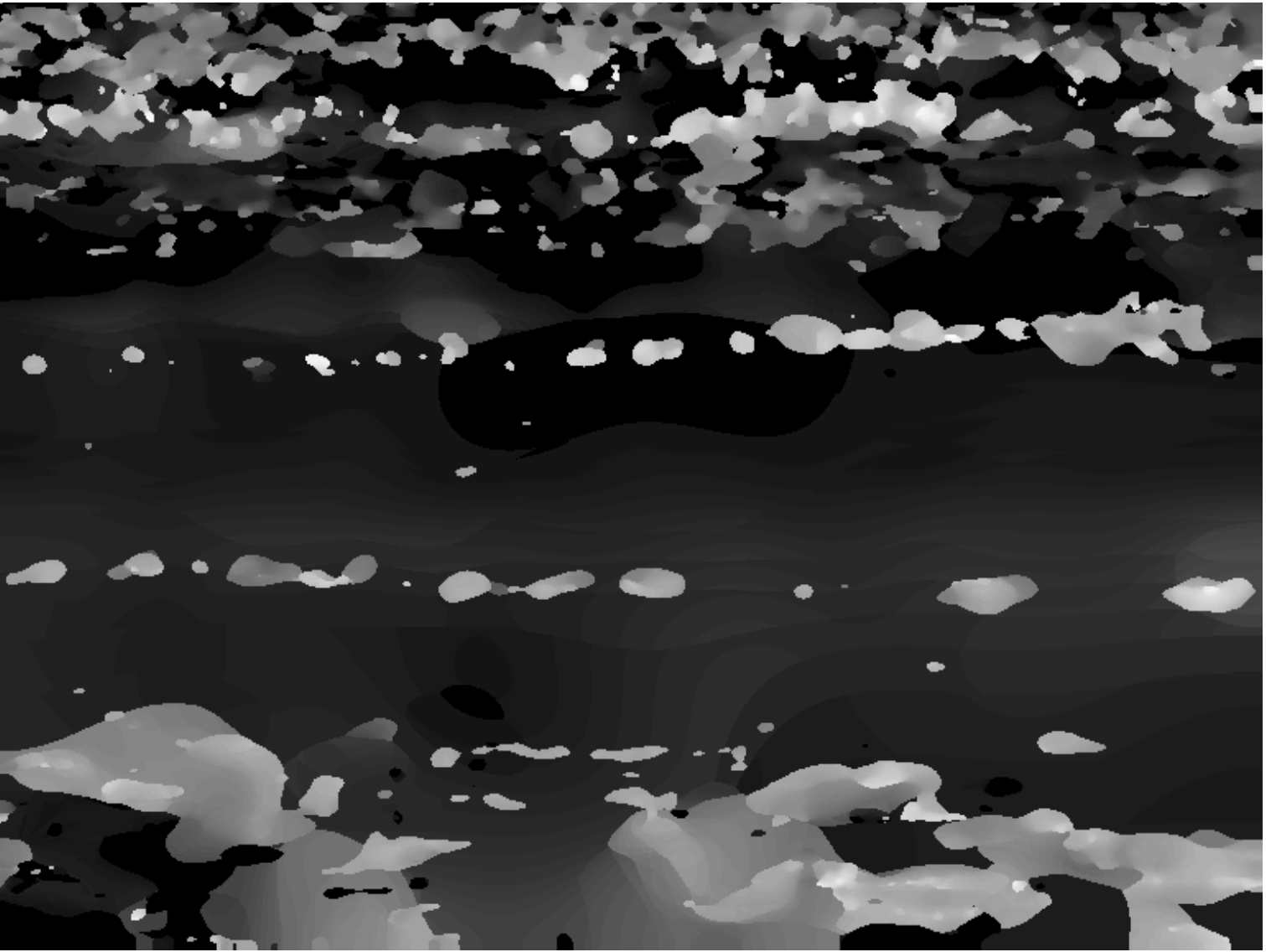} \\
     {\footnotesize\em  spatio-temporal scale-space signature}
      & {\footnotesize\em maximum magnitude response over all scales} \\
    \includegraphics[width=0.44\textwidth]{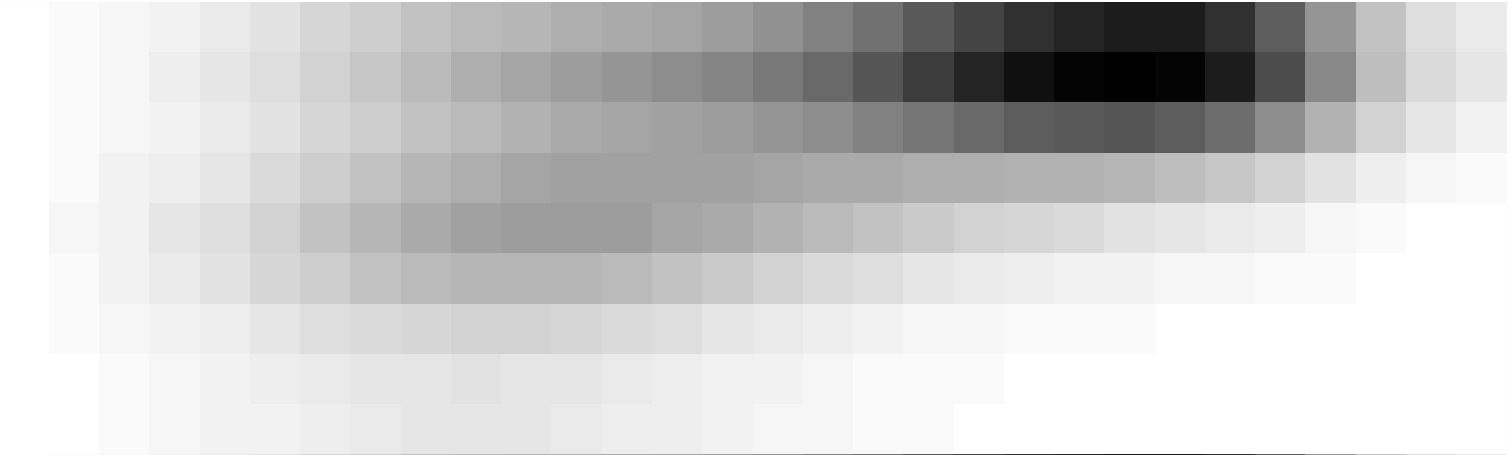}
       & \includegraphics[width=0.44\textwidth]{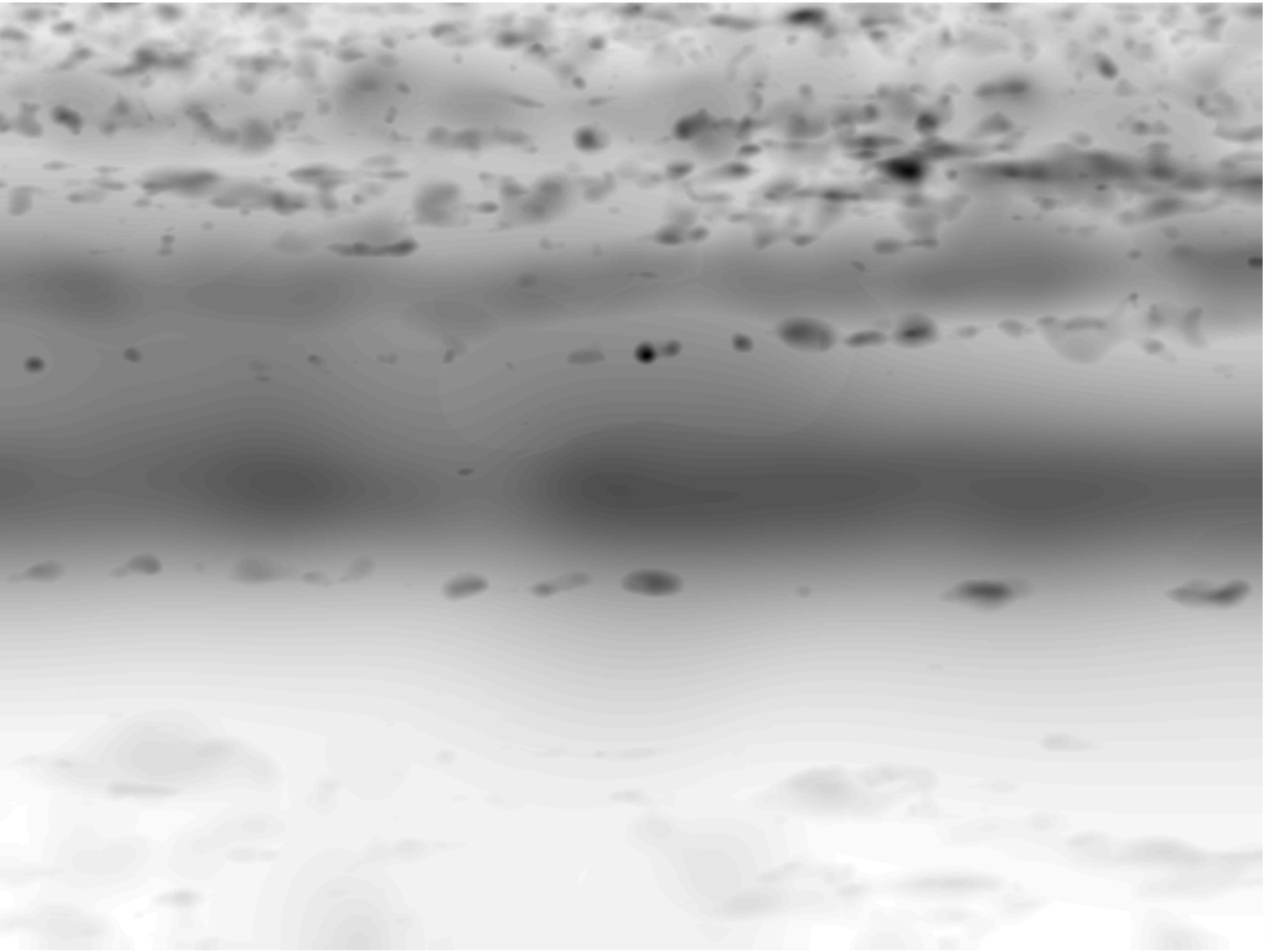} \\
     \end{tabular}
  \end{center}
\caption{Results of dense spatio-temporal scale selection applied to a
  scene with breaking waves (from the DynTex
  dataset \cite{PetRenFaz-PRL2010}). 
%  The results have been computed with
%   a time-causal and time-recursive spatio-temporal scale-space
%   representation obtained by convolution with a Gaussian kernel over
%   the spatial domain and the time-causal limit kernel over the
%   temporal domain. 
%   (top left) Grey-level image.
%   (top right) The spatio-temporal quasi quadrature measure computed at
% a fixed spatio-temporal scale.
%   (middle left) Selected spatial scale levels in units of effective
%   scale $s_{eff} = \log_2 \sigma_s$.
%   (middle right) Selected spatial scale levels in units of effective
%   scale $\tau_{eff} = \log_2 \sigma_{\tau}$.
%   (bottom left) Spatio-temporal scale-space signature showing the
%   magnitude variations of the spatio-temporal quasi quadrature measure
% over both spatial and temporal scales, with effective spatial scale
% increasing linearly from left to right and effective temporal scale
% increasing linearly from bottom to top. While this illustration shows
% the average over all image points at the given image fram for the
% purpose of suppressing the influence of local variations, the general
% dense scale selection method is otherwise local, based on individual
% scale-space signatures at every image point and for every time moment.
%   (bottom right) The maximum magnitude response over all spatial and
%   temporal scales at every image point.
 Notice how horizontal stripes of finer spatial and temporal scales
 are selected at the breaking waves.
Notice additionally that there are two dominant spatio-temporal scales
in the scene --- one for the larger scale overall waves and one for
fine-scale spatio-temporal structures where the waves break.
%   Observe that the contrast of the maps showing
% magnitude information has been set so that dark corresponds to larger
% values and bright to lower values. In addition, the magnitude maps
% have been stretched by a square root function. 
% (Results computed using 25 logarithmically distributed spatial scale levels between $\sigma_{s,min} = 0.25$~pixels and 
%  $\sigma_{s,max} = 30$~pixels and 9 logarithmically distributed temporal scale levels between
% $\sigma_{\tau,min} = 8.3$~ms and $\sigma_{\tau,max} = 2.1$~seconds.)
% (Complementary scale normalization parameters $\Gamma_s = 0$
% and $\Gamma_{\tau} = 0$.)
(Image size: $768 \times 576$ pixels. Frame 50 of 250 frames at 30
frames per second.)}
  \label{fig-brytandevagor-denseSTscsel}
\end{figure*}

\noindent
Figure~\ref{fig-waves0007-denseSTscsel} shows an example of applying
dense spatio-temporal scale selection to a real video sequence.
For reasons of computational efficiency, we only show results obtained
using a time-causal and time-recursive spatio-temporal scale-space
representation obtained by convolution with Gaussian kernels over the
spatial domain and convolution with the time-causal limit kernel over
the temporal domain.
Because of the time-recursive implementation of this scale-space
concept, it is not necessary to explicitly compute and build the
five-dimensional spatio-temporal scale-space representation over
space-time $(x, y, t)$ and spatio-temporal scales $(s, \tau)$.
Instead, the time-recursive implementation builds a four-dimensional
representation over the spatial domain $(x, y)$ and the spatio-temporal scale
parameters $(s, \tau)$ at every temporal image frame $t$.
Then, this representation is recursively updated to the next frame,
using only the temporal scale-space representation at the previous
frame as a sufficient temporal buffer of past information, using the
methodology of time-causal and time-recursive spatio-temporal
receptive fields developed in \cite{Lin16-JMIV}.

Because the notion of phase compensation is not yet fully developed for
the time-causal limit kernel, we did not use local phase compensation
in this experiment. Instead, we restricted ourselves to spatial
post-smoothing noting that the approach can in a straightforward
manner be extended to temporal post-smoothing by adding a second stage
of recursive temporal smoothing to the quasi quadrature measures
computed at every image frame.
To make the magnitude maps maximally scale invariant for purposes of
visualization, we used $\Gamma_s = \Gamma_{\tau} = 0$.

At every image frame, we computed a discrete approximation of the
spatio-temporal quasi quadrature measure at all spatial and temporal
scales and detected two-dimensional local extrema over spatial and
temporal scales as candidates for local spatio-temporal scale
levels. These local extrema were then interpolated to
higher resolution over spatial and temporal scales using
parabolic interpolation according to \cite[Equation~(115)]{Lin17-JMIV}.
For simplicity, the results shown in the figures display only
the global extremum over spatio-temporal scales at every image point.
When applying the scale selection methodology in practice, multiple
local extrema over spatio-temporal scales should, however, instead be
considered to make it possible to handle multiple characteristic
spatio-temporal scale levels at any image point.

From the scale maps in the middle row, we can note that the selected
spatial scale levels well reflect the perspective size gradient over
the vertical direction in the image domain, caused by the water waves
being assumed to have a stationary distribution of wavelengths over
the water surface, while these spatial lengths are shortened because
of the perspective scaling and foreshortening effects.
For the selected temporal scale levels, the distribution is more
stationary over the image domain, which can be understood from the
assumption that the temporal wavelengths of the waves should be
stationary over the water surface, while at the same time the temporal
scales are not affected by the perspective transformation (the temporal
periodicity of a wave remains the same under imaging transformations).

In the spatio-temporal scale-space signature, showing the average over
all the image points of the scale-normalized spatio-temporal quasi
quadrature measure as function of the spatial and temporal scales, we can see
that for this video sequence there is a narrow range of dominant spatial and
temporal scales. The spread over the spatial scale levels is, however, wider
than the spread over the temporal scales, caused by the additional
variabilities induced by the perspective scaling and foreshortening effects.

When comparing the maximum magnitude response over all spatio-temporal
scales to the quasi quadrature measure at a fixed
spatio-temporal scale, we can observe that the variability in the
maximum over all spatio-temporal scales is lower than the variability
in the response at a fixed scale.

Figures~\ref{fig-smoothtraff05-denseSTscsel}--\ref{fig-brytandevagor-denseSTscsel}
show results of applying corresponding dense spatio-temporal scale selection 
to videos of other dynamic scenes.
In the traffic scene in
figure~\ref{fig-smoothtraff05-denseSTscsel}, we can note that
distinct responses in the spatial and temporal scale maps are obtained
for the different moving cars, again with a vertical size gradient in the
spatial scale estimates reflecting the perspective scaling and
foreshortening effects,
whereas the temporal scale estimates are essentially unaffected by the
perspective transformation. %, as they should be. 
Additionally, we can observe that large
spatial scales and long temporal scales are selected in the smooth
stationary regions on the road and in some parts of the background.
For the video of breaking waves in
Figure~\ref{fig-brytandevagor-denseSTscsel}, we can note that there are
two dominant spatio-temporal scales
in the scene --- one for the larger scale overall waves and one for
fine-scale spatio-temporal structures where the waves break.
These two spatio-temporal scale levels are in turn reflected as
horizontal stripes in the maps of the selected spatial and temporal scales.

\section{Summary and discussion}
\label{sec-summ-disc}

We have presented a general methodology for performing dense scale
selection by detecting local extrema over scale of scale-normal\-ised
quasi quadrature entities, which constitute local energy measures of the
combined strength of first- and second-order scale-space derivatives.
Specifically, we have: (i)~analyzed how local scale estimates may in general be
strongly dependent on the relative strengths of first- 
{\em vs.\/}\ second-order image information at every image point and
(ii)~proposed two mechanisms to reduce this phase dependency substantially,
using post-smoothing and pointwise phase compensation.

Based on the presented theoretical analysis of scale selection
properties over a purely spatial image domain, we have in Section~\ref{sec-spat-dense-scsel}
%for the case of dense scale selection on a purely spatial domain 
presented four
types of algorithms for dense scale selection, depending on whether
the mechanisms of phase compensation and post-smoothing are included
or excluded. For Algorithms~II--IV that involve such mechanisms, we
have shown that these mechanisms substantially reduce the spatial variability of the
local scale estimates compared to the baseline Algorithm~I that neither
makes use of phase compensation nor post-smoothing.
These four methods do all lead to provable scale invariance in the sense
that the local scale estimates perfectly follow scaling
transformations over image space, and so do image features and image
descriptors that are computed at scales proportional to these local
scale estimates.

In Section~\ref{sec-temp-dense-scsel}, we developed corresponding
dense scale selection mechanisms over a purely temporal domain and
with corresponding mechanisms of post-smoothing and local phase
compensation to reduce the phase sensitivity of the local scale
estimates. 
%Additionally, we extended the applicability of the temporal
%scale selection mechanism to the time-causal temporal scale-space
%concept based on convolution with the time-causal limit kernel.
By experiments on a synthetic sine wave with exponentially varying
wavelength as function of time, we demonstrated that the local scale
estimates do well adapt to the variabilities of the time-dependent
characteristic temporal scales in the signal. By experiments on a
neurophysiological signal with approximate stationarity properties, we
demonstrated how the proposed dense scale selection methodology is
able to reflect multiple levels of characteristic scales in the signal
that are not as visible in a spectral analysis based on Fourier transforms.

In Section~\ref{sec-spattemp-dense-scsel}, we combined the above dense scale
selection mechanisms over spatial and temporal domains to joint dense
spatio-temporal scale selection in video data and demonstrated how
the resulting approach is able to generate hypotheses about joint
characteristic spatio-temporal scales for different types of dynamic scenes.

A common property of these spatial, temporal and spatio-temporal scale
selection methods is that the scale estimates are computed in a
bottom-up way from the data in such a way that the scale estimates
will be covariant under independent scaling transformations of
the spatial and the temporal domains. 
We propose these forms of dense
scale selection as a general mechanisms for estimating local spatial
and temporal scales in spatial images, temporal signals and
spatio-temporal video.

As a complement to previous scale selection methodologies, which have been primarily
applied spars\-ely at spatial or spatio-temporal interest
points, the proposed dense scale selection methodology is intended for
applications where spatial, temporal or spatio-temporal receptive
field responses are to be computed densely at every image point and
for every time moment. Potential applications of such dense
receptive field responses include texture analysis over a static spatial
domain and dynamic texture analysis over a spatio-temporal domain.
For example, if the application of dense spatio-temporal scale
selection presented in Figure~\ref{fig-waves0007-denseSTscsel} is
applied to videos of water waves taken under different wind
conditions, then the spatial scale estimates will reflect the spatial
extent of the water waves, whereas the temporal scale estimates will
reflect their temporal duration. In this way, dynamic parameters of
the water waves can be estimated directly, without using a generative physical model of
the wave patterns.

More generally, the proposed framework provides a theory for modelling
and measuring how dense receptive field measurements respond selectively at
different spatial and temporal scales. 
This theory should be relevant for a large sets of computer vision problems where
receptive field based image measurements in terms of spatial or
spatio-temporal $N$-jets are used as the basis for image analysis or
video analysis applications. The presented theory could also be
relevant for computational modelling of biological vision. If we regard the
spatio-temporal quasi quadrature measure (\ref{eq-Q3-scalenorm-ders-Gammanorm}) 
as modelling important properties of complex cells %in biological vision 
as detailed in Section~\ref{sec-spat-temp-quasi-quad}, then the proposed dense
spatio-temporal scale selection theory can explain how complex cells
having receptive fields over different ranges of spatial and temporal
scales respond selectively to stimuli of different spatial extent
and temporal duration.

The only free parameters are the 
complementary spatial and temporal scale normalization parameters $\Gamma_s$ and
$\Gamma_{\tau}$, the relative integration scales $c$ for optional post-smoothing and the scale
calibration factor by which the generated scale estimates are
proportional to the scales at which the local extrema over scales are
assumed. These parameters should be optimized to the specific
application domain, where the scale dense scale selection methodology
is to be combined with higher-level visual modules. 

If suitable
values of these parameters can be determined for a specific
application domain, then by the general scale covariance property 
of the scale estimates, the proposed dense scale
selection theory guarantees that the resulting spatial, temporal
or spatio-temporal scale estimates will automatically adapt to
and follow variabilities in the characteristic scales in the input images,
signals, videos or image streams. In this way, the resulting chain of
computer vision/image analysis/signal analysis/video analysis operations can be made provably scale invariant.

\section*{Acknowledgements}

I would like to thank Prof.\ Peter Brown at Oxford University for
providing the data for the experiments in
Figure~\ref{fig-arvind-signal-densetempscsel} and
Arvid Kumar at KTH Royal Institute of Technology for serving as a link
and discussion partner regarding this dataset, 
specifically regarding the biological background.

\bigskip

\appendix

\section{Detailed analysis of scale calibration for dense spatial
  scale selection and when using spatial post-smoothing}
\label{sec-suppl-dense-spat-scsel-calib}

In the treatment of dense spatial scale selection in
Section~\ref{sec-spat-dense-scsel} in the main article, we analysed
the scale selection properties obtained from detecting local extrema
over scale of the scale-normalized spatial quasi quadrature measure
(\ref{eq-quasi-quad-scale-mod})
\begin{align}
  \begin{split}
    {\cal Q}_{(x,y),\Gamma-norm} L 
   & = \frac{s \, (L_x^2 + L_y^2) + C_s \, s^2 \, (L_{xx}^2 + 2 L_{xy}^2 + L_{yy}^2)}{s^{\Gamma_s}}
   \end{split}
\end{align}
in Section~\ref{eq-sc-sel-props}. Specifically, for the case of using
only phase compensation while no post-smoothing to reduce the phase dependency of the spatial
scale estimates, the spatial scale estimates are for a 1-D sine wave
with angular frequency $\omega_0$ centered around the
spatial scale level (\ref{eq-center-scale-spat-phase-comp})
\begin{equation}
  \hat{s} = \sqrt{\hat{s}_1 \, \hat{s}_2} = \frac{\sqrt{(1-\Gamma_s)(2-\Gamma_s)}}{\omega_0^2}
\end{equation}
By the notion of scale calibration described in
Section~\ref{sec-sc-calib-spat-dense-scsel}, it was proposed that
these scale estimates can be calibrated by multiplication with a
uniform scale calibration factor to be either:
\begin{itemize}
\item[(i)]
  equal to the scale estimate $\hat{s} = s_0$ obtained by applying the regular
  scale normalized Laplacian $\nabla_{norm}^2 L = s \, (L_{xx} + L_{yy})$ 
  or the scale-normalized determinant of the Hessian 
  $\det {\cal H}_{norm} L = s^2 \, (L_{xx} L_{yy} - L_{xy}^2)$ at the
  center of a Gaussian blob of any spatial extent $s_0$.
\item[(ii)]
  equal to the scale estimate $\hat{s} = \sqrt{2}/\omega_0^2$
  corresponding to the geometric average of the scale
  estimates obtained for a sine wave of any angular frequency
  $\omega_0$ when $\Gamma_s = 0$.
\end{itemize}
In this appendix section, we describe in more detail how such scale calibration can
be performed when also using spatial post-smoothing.

\subsection{Influence of the post-smoothing scale}

When applying post-smooth\-ing, the variability in the spatial scale
estimates decreases with the relative integration scale parameter $c$
(see the third, fourth and fifth rows in Figure~\ref{fig-quasi-quad-1D-sinewave-scsel}).
The minimum scale estimates are assumed at the spatial points at which
${\cal Q}_{(x,y),\Gamma-norm} L$ only responds to first-order information,
whereas the maximum scale estimates are assumed at the points at which 
${\cal Q}_{(x,y),\Gamma-norm} L$ only responds to second-order information.

By differentiating the 1-D version of (\ref{eq-Qnormbar-2D-sine-wave-general}) with
respect to scale $s$ and setting $x = 0$ and $x = \omega_0 \pi/2$ respectively as well as
$\omega_0 = 1$ in the resulting equation, we obtain the following
algebraic equations for how the minimum and maximum scale values 
depend on the
relative post-smoothing scale $c$, $\Gamma_s$ and $C_s$ for a
1-D sine wave with angular frequency $\omega_0$:
\begin{align}
  \begin{split}
  \label{eq-for-defining-Ssine1}
C_s \left(2 c^2+1\right) s^2-e^{2 c^2 s} \left(C_s s^2-2 C_s s+C_s s
   \Gamma_s+s-1+\Gamma_s\right) 
  \end{split}\nonumber\\
  \begin{split}
   \quad -s \left(2 C_s+2
   c^2+1\right)+\Gamma_s (C_s s-1)+1 = 0,
\end{split}\\
\begin{split}
  \label{eq-for-defining-Ssine2}
 C_s -2 C_s c^2 s^2-e^{2 c^2 s} \left(C_s s^2-2 C_s s+C_s s
   \Gamma_s+s-1+\Gamma_s\right) 
  \end{split}\nonumber\\
  \begin{split}
     \quad
 -C_s s^2+2 C_s s-C_s s \Gamma_s+2
   c^2 s+s-1+\Gamma_s = 0.
  \end{split}
\end{align}
By defining functions $S_{sine,1}(\Gamma_s, c, C_s)$ and $S_{sine,1}(\Gamma_s, c, C_s)$
that represent the solutions $\hat{s}_{min}$ and $\hat{s}_{max}$ of these equations as function
of the parameters $\Gamma_s$, $c$ and $C_s$ for $\omega_0 = 1$, 
the minimum and maximum scale values can because of the
self-similarity over scale for an arbitrary angular frequency
$\omega_0$ of the sine wave be expressed as:
\begin{equation}
  \label{eq-smin-smax-postsmooth}
  \hat{s}_{min} = \frac{S_{sine,1}(\Gamma_s, c, C_s)}{\omega_0^2},
  \quad\quad
  \hat{s}_{max} = \frac{S_{sine,2}(\Gamma_s, c, C_s)}{\omega_0^2}.
\end{equation}
Table~\ref{table-Ssine1-Ssine2} shows numerical values of these
entities for different values of the %complementary 
scale normalization
parameter $\Gamma_s$ and the relative post-smoothing scale~$c$.

\begin{table}[hbtp]
     \footnotesize
  \begin{center}
      {\footnotesize\em Minimum relative scale estimate
        $\hat{s}_{min} = S_{sine,1}(\Gamma_s, c, C_s)$ for a
        1-D sine wave}

    \bigskip

    \begin{tabular}{cccc}
      \hline
      {\footnotesize $c$} & {\footnotesize $\Gamma_s = 0$} 
        & {\footnotesize $\Gamma_s = \frac{1}{4}$} & {\footnotesize $\Gamma_s = \frac{1}{2}$} \\
      \hline
        0                  & 1.000 & 0.750 & 0.500 \\
       $1/2$           & 1.033 & 0.741 & 0.466 \\
       $1/\sqrt{2}$ & 1.132 & 0.772 & 0.446 \\
        1                  & 1.329 & 0.963 & 0.451 \\
        $\sqrt{2}$    & 1.408 & 1.125 & 0.779 \\
        $2$              & 1.414 & 1.145 & 0.863 \\
       \hline
    \end{tabular}

  \bigskip

     {\footnotesize\em Maximum relative scale estimate 
       $\hat{s}_{min} = S_{sine,2}(\Gamma_s, c, C_s)$ for 1-D
       sine wave}

    \bigskip
   \begin{tabular}{cccc}
      \hline
      {\footnotesize $c$} & {\footnotesize $\Gamma_s = 0$} 
        & {\footnotesize $\Gamma_s = \frac{1}{4}$} & {\footnotesize $\Gamma_s = \frac{1}{2}$} \\
      \hline
        0                  & 2.000 & 1.750 & 1.500 \\
       $1/2$           & 1.701 & 1.485 & 1.264 \\
       $1/\sqrt{2}$ & 1.584 & 1.363 & 1.147 \\
        1                  & 1.474 & 1.242 & 1.021 \\
        $\sqrt{2}$    & 1.420 & 1.163 & 0.914 \\
        $2$              & 1.414 & 1.146 & 0.869 \\
       \hline
    \end{tabular}
  \end{center}

  \bigskip

  \caption{Minimum and maximum scale estimates 
                $\hat{s}_{min} =  S_{sine,1}(\Gamma_s, c, C_s)/\omega_0^2$ and
                $\hat{s}_{max} =  S_{sine,2}(\Gamma_s, c, C_s)/\omega_0^2$ 
                for dense scale selection based on local extrema over
                scale of the post-smoothed quasi quadrature entity 
                ${\overline {\cal Q}}_{(x,y),\Gamma-norm} L$ applied to a
                1-D sine wave with angular frequency
                $\omega_0 = 1$ for different values of the relative scale
                normalization parameter $\Gamma_s$ and the relative
                post-smoothing scale $c$ with $C_s$ according to 
                (\protect\ref{eq-C-from-Gamma-when-c-is-zero-geom}).
                These entities are defined as the solutions of
                equations (\protect\ref{eq-for-defining-Ssine1})
                and (\protect\ref{eq-for-defining-Ssine2}), and
                for increasing values of $c$ they
                approach $\hat{s}_{intermed} = \sqrt{(1- \Gamma_s) \, (2- \Gamma_s)}/\omega_0^2$
                in (\protect\ref{eq-mean-scale-est-from-t1-t2-geom}).}
   \label{table-Ssine1-Ssine2}
\end{table}

Whereas the functions $S_{sine,1}(\Gamma_s, c, C_s)$ and
$S_{sine,1}(\Gamma_s, c, C_s)$ are not expressed in terms of
elementary functions, it is straightforward to
implement these functions using standard numerical methods for
computing the solutions of a 1-D equation.

\subsection{Phase-compensated scale estimates with post-smoothing}

Given these expressions for the minimum and maximum scale estimates
for a sine wave, we can also define a corresponding notion of
phase compensation in the presence of spatial post-smoothing:
\begin{align}
  \begin{split}
     \label{eq-scale-corr-sine-4-tau}
     \hat{s}_{{\overline {\cal Q}} L, comp} = 
     & \frac{\sqrt{S_{sine,1}(\Gamma_s, c, C_s) \,  S_{sine,2}(\Gamma_s, c, C_s)}}
          {\left(\overline{\cal Q}_{(x,y),1,\Gamma-norm} L + \overline{\cal Q}_{(x,y),2,\Gamma-norm} L\right)} \times
  \end{split}\nonumber\\
  \begin{split}
     & \left( 
           \frac{\overline{\cal Q}_{(x,y),1,\Gamma-norm} L}{S_{sine,1}(\Gamma_s, c, C_s)}  
     + \frac{\overline{\cal Q}_{(x,y),2,\Gamma-norm} L}{S_{sine,2}(\Gamma_s, c, C_s)} 
         \right) \,
          \hat{s}_{{\overline {\cal Q}} L}
  \end{split}
\end{align}
or
\begin{align}
  \begin{split}
     & \hat{s}_{{\cal Q} L, comp} = 
%\end{split}\nonumber\\
 %\begin{split}
       \sqrt{S_{sine,1}(\Gamma_s, c, C_s) \, S_{sine,2}(\Gamma_s, c, C_s)} \, \hat{s}_{{\cal Q} L}/
\end{split}\nonumber\\
 \begin{split}
          (S_{sine,1}(\Gamma_s, c, C_s))^{\frac{\overline{\cal Q}_{(x,y),1,\Gamma-norm} L}{\overline{\cal
                      Q}_{(x,y),1,\Gamma-norm} L + \overline{\cal Q}_{(x,y),2,\Gamma-norm} L}} /
\end{split}\nonumber\\
 \begin{split}
          (S_{sine,2}(\Gamma_s, c, C_s))^{\frac{\overline{\cal Q}_{(x,y),2,\Gamma-norm} L}{\overline{\cal Q}_{(x,y),1,\Gamma-norm} L + \overline{\cal Q}_{(x,y),2,\Gamma-norm} L}}
  \end{split}
\end{align}
with $\overline{\cal Q}_{(x,y),1,\Gamma-norm} L$ and $\overline{\cal Q}_{(x,y),2,\Gamma-norm} L$
according to (\ref{eq-Q1bar-spatial}) and (\ref{eq-Q2bar-spatial}) and
with the normalization chosen such that the scale values should aim
towards the geometric mean of the extreme values $S_{sine,1}(\Gamma_s, c, C_s)$ 
and $S_{sine,2}(\Gamma_s, c, C_s)$ according to (\ref{eq-for-defining-Ssine1}) 
and (\ref{eq-for-defining-Ssine2}).

\subsubsection{Two-dimensional blob}
\label{sec-scsel-props-2D-gauss-blob}

For a two-dimensional Gaussian blob
\begin{equation}
  f(x, y) = g(x, y;\; s_0) = \frac{1}{2 \pi s_0} e^{-(x^2+y^2)/2 s_0},
\end{equation}
if follows from the semi-group property of the Gaussian that 
the scale-space representation is given by
\begin{equation}
  L(x, y;\; s) = g(x, y;\; s_0 + s) 
               = \frac{1}{2 \pi (s_0 + s)} e^{-(x^2+y^2)/2 (s_0 + s)}
\end{equation}
and the unsmoothed quasi quadrature entity at the origin is of the form
\begin{equation}
    {\cal Q}_{(x,y),\Gamma-norm} L 
    = \frac{C_s s^{2-\Gamma_s}}{2 \pi ^2 (s+s_0)^4}.
\end{equation}
Differentiating this expression with respect to the scale parameter $s$,
shows that the maximum value over scales is assumed at scale
\begin{equation}
  \label{eq-scsel-quasquad-Gammanorm-Gaussblob}
  \hat{s}_{Gauss} = \frac{2-\Gamma_s}{2 + \Gamma_s} \, s_0.
\end{equation}
With complementary post-smoothing with relative integration scale $c$,
corresponding computation of the post-smoothed quasi quadrature and
differentiation of the resulting expression gives an algebraic
equation of the form
\begin{align}
  \begin{split}
4 c^6 s^2 \left((2 C_s+1) s^2 (2+\Gamma_s) \right.
%  \end{split}\nonumber\\
%  \begin{split}
 \left. 
+2 s s_0 (C_s
   (\Gamma_s-1)+1+\Gamma_s)+s_0^2 \Gamma_s\right)+
  \end{split}\nonumber\\
  \begin{split}
+ 4 c^4
   s (s+s_0) \left((2 C_s+1) s^2 (2+\Gamma_s) \right.
%  \end{split}\nonumber\\
%  \begin{split}
  \left.
+s s_0 (2 C_s
   (\Gamma_s-2)+1+2 \Gamma_s)+s_0^2
   (\Gamma_s-1)\right)+
  \end{split}\nonumber\\
  \begin{split}
+c^2 (s+s_0)^2 \left((4 C_s+1) s^2
   (2+\Gamma_s)
\right.
%  \end{split}\nonumber\\
%  \begin{split}
  \left.
+2 s s_0 (2 C_s
   (\Gamma_s-1)+\Gamma_s)+s_0^2 (\Gamma_s-2)\right)
  \end{split}\nonumber\\
  \begin{split}
 +C_s
   (s+s_0)^3 (s (2+\Gamma_s)+s_0 (\Gamma_s-2)) =0
  \end{split}%\nonumber\\
   \label{eq-def-sel-sc-Gauss-blob-post-smooth-quasi-quad}
\end{align}
for the selected scale level $\hat{s}$ as function of the complementary scale
normalization parameter $\Gamma_s$, the relative post-smoothing scale
$c$ and the relative weighting factor $C_s$ between first- and
second-order information. Let us define the following function for
representing the solution of this equation:
\begin{equation}
  \label{eq-scale-comp-factor-S-gauss-blob}
  \hat{s} = S_{Gauss}(\Gamma_s, c, C_s) \, s_0.
\end{equation}
Table~\ref{table-scale-estimates-gauss-blob-post-smooth-quasi-quad}
shows numerical values of this entity for different values of
$\Gamma_s$ and $c$.

\begin{table}[!h]
  \footnotesize
  \begin{center}
      {\footnotesize\em Scale estimates $\hat{s} = S_{Gauss}(\Gamma_s, c) \, s_0$ based on 
                         $\overline{\cal Q}_{(x,y),\Gamma-norm} L$ \\at the center of a Gaussian blob}

    \bigskip

    \begin{tabular}{cccc}
      \hline
      {\footnotesize $c$} & {\footnotesize $\Gamma_s = 0$} 
        & {\footnotesize $\Gamma_s = \frac{1}{4}$} & {\footnotesize $\Gamma_s = \frac{1}{2}$} \\
      \hline
        0                  & 1.000 & 0.778 & 0.600 \\
       $1/2$           & 0.839 & 0.650 & 0.498 \\
       $1/\sqrt{2}$ & 0.751 & 0.578 & 0.440 \\
        1                  & 0.641 & 0.487 & 0.367 \\
        $\sqrt{2}$    & 0.519 & 0.385 & 0.283 \\
        $2$              & 0.402 & 0.285 & 0.199 \\
       \hline
    \end{tabular}
  \end{center}

  \bigskip

  \caption{Numerical values of the ratio $\hat{s}/s_0 = S_{Gauss}(\Gamma_s, c, C_s)$ for which 
   the post-smoothed quasi quadrature entity $\overline{\cal Q}_{(x,y),\Gamma-norm} L$ 
   assumes its maximum over scale at the center of a Gaussian blob
    with scale parameter $s_0$ for different values of the
    complementary scale normalization parameter $\Gamma_s$ and the
    relative post-smoothing scale parameter $c$
    with $C_s$ according to (\protect\ref{eq-C-from-Gamma-when-c-is-zero-geom}).}
   \label{table-scale-estimates-gauss-blob-post-smooth-quasi-quad}
\end{table}

The difference in net effect between the Gaussian scale calibration
model and the sine wave scale calibration model 
under variations of $\Gamma_s$ and $c$ is essentially determined 
by the variation of the following ratio between the scale calibration
factors in units of $\sigma_s = \sqrt{s}$:
\begin{equation}
  \chi(\Gamma_s, c) = \sqrt{\frac{2 S_{Gauss}(\Gamma_s, c)}{S_{sine,2}(\Gamma_s, c)}}
\end{equation}
see Table~\ref{table-scale-calib-sine-gauss-ratio-post-smooth-quasi-quad}
for numerical values. For $c \leq 1$, it can be seen that the relative
differences in effects for scale calibration are within a range of 15~\% in units of $\sigma_s = \sqrt{s}$.

Given the similarity between the results
obtained from these qualitatively very different models, it seems
plausible that the results should also
generalize to wider classes of image structures.
% , which is supported by the experiments.
Choosing the parameter $\gamma$ for scale selection using $\gamma$-normalized derivatives
based on the behaviour for
Gaussian image models has also been demonstrated to lead to highly
useful results for a wide range of computer vision tasks 
(Lindeberg \cite{Lin97-IJCV,Lin98-IJCV,Lin99-CVHB,Lin14-EncCompVis}).

\begin{table}[!h]
  \footnotesize
  \begin{center}
      {\footnotesize\em Dependency of the ratio between the scale calibration  factors
                         $\sqrt{\frac{2 S_{Gauss}(\Gamma_s, c)}{S_{sine,2}(\Gamma_s, c)}}$ on $\Gamma_s$ and $c$}

    \bigskip

    \begin{tabular}{cccc}
      \hline
      {\footnotesize $c$} & {\footnotesize $\Gamma_s = 0$} 
        & {\footnotesize $\Gamma_s = \frac{1}{4}$} & {\footnotesize $\Gamma_s = \frac{1}{2}$} \\
      \hline
        0                  & 1.000 & 0.943 & 0.894 \\
       $1/2$           & 0.991 & 0.935 & 0.888 \\
       $1/\sqrt{2}$ & 0.974 & 0.921 & 0.876 \\
        1                  & 0.933 & 0.886 & 0.848 \\
        $\sqrt{2}$    & 0.855 & 0.814 & 0.786 \\
        $2$              & 0.754 & 0.705 & 0.677 \\
       \hline
    \end{tabular}
  \end{center}

  \bigskip

  \caption{Numerical values of the ratio between the scale calibration
  factors $S_{Gauss}(\Gamma_s, c)$ and $S_{sine,2}(\Gamma_s, c)$ in units
  of $\sigma_s = \sqrt{s}$ and normalized such that the ratio is equal
  to one for $\Gamma_s = 0$ and $c = 0$. This ratio provides an estimate
  of how much the calibrated scale estimates will depend on the choice of
  calibration model, based on either sparse image features or a dense
  texture pattern.}
   \label{table-scale-calib-sine-gauss-ratio-post-smooth-quasi-quad}
\end{table}

\section{Detailed analysis of phase compensation and scale calibration
  for dense spatio-temporal scale selection}

For dense spatio-temporal scale selection, we do according to
Section~\ref{sec-spattemp-dense-scsel} at every point $(x, y, t)$ in
space-time detect simultaneous local extrema over spatio-temporal
scales (\ref{eq-Q3-spat-temp-scaleest-regular-geom-sine-s-tau})
\begin{equation}
     (\hat{s}_{{\cal Q}_{(x,y,t),\Gamma-norm}}, \hat{\tau}_{{\cal Q}_{(x,y,t),\Gamma-norm}})
     % = \\ 
     = \argmaxlocal_{s,\tau} {\cal Q}_{(x,y,t),\Gamma-norm} L
\end{equation}
of the scale-normalized spatio-temporal quasi quadrature entity
(\ref{eq-Q3-scalenorm-ders-Gammanorm})
\begin{align}
   \begin{split}
     & {\cal Q}_{(x, y, t),\Gamma-norm} L 
  \end{split}\nonumber\\
  \begin{split}
     & = \frac{\tau \, {\cal Q}_{(x, y),\Gamma-norm} L_{t} 
            + C_{\tau} \tau^2 \, {\cal Q}_{(x, y),\Gamma-norm} L_{tt}}{\tau^{\Gamma_{\tau}}}
 \end{split}\nonumber\\
  \begin{split}
      & = \frac{1}{s^{\Gamma_s} \tau^{\Gamma_{\tau}}}
             \left( \tau
               \left( 
                   s \, (L_{xt}^2 + L_{yt}^2) 
%               \right.
%              \right.
%\end{split}\nonumber\\
%  \begin{split}
%        & \phantom{= \frac{1}{s^{\Gamma_s} \tau^{\Gamma_{\tau}}}}
%            \left.
%              \left.
               + C_s \, s^2 \left( L_{xxt}^2 + 2 L_{xyt}^2 + L_{yyt}^2 \right)
               \right)
             \right.
 \end{split}\nonumber\\
  \begin{split}
      & \phantom{= \frac{1}{s^{\Gamma_s} \tau^{\Gamma_{\tau}}} \left( \right.}  \,
           \left.
             + C_{\tau} \, \tau^2
               \left( 
                   s \, (L_{xtt}^2 + L_{ytt}^2)
%               \right.
%          \right.
% \end{split}\nonumber\\
%  \begin{split}
% & \phantom{= + C_{\tau} \, \tau^2 \left( \right.}
%            \left.
%              \left.
                   + C_{s} \, s^2 (L_{xxtt}^2 + 2 L_{xytt}^2 + L_{yytt}^2) 
              \right)
            \right).
    \end{split}
\end{align}

\begin{figure*}[bt]
  \begin{center}
    \begin{tabular}{ccc}
       {\small\em regular scale estimates $\Gamma_s = \Gamma_{\tau} = 0$} 
      & {\small\em phase-compensated $\Gamma_s = \Gamma_{\tau} = 0$} \\
      \includegraphics[width=0.40\textwidth]{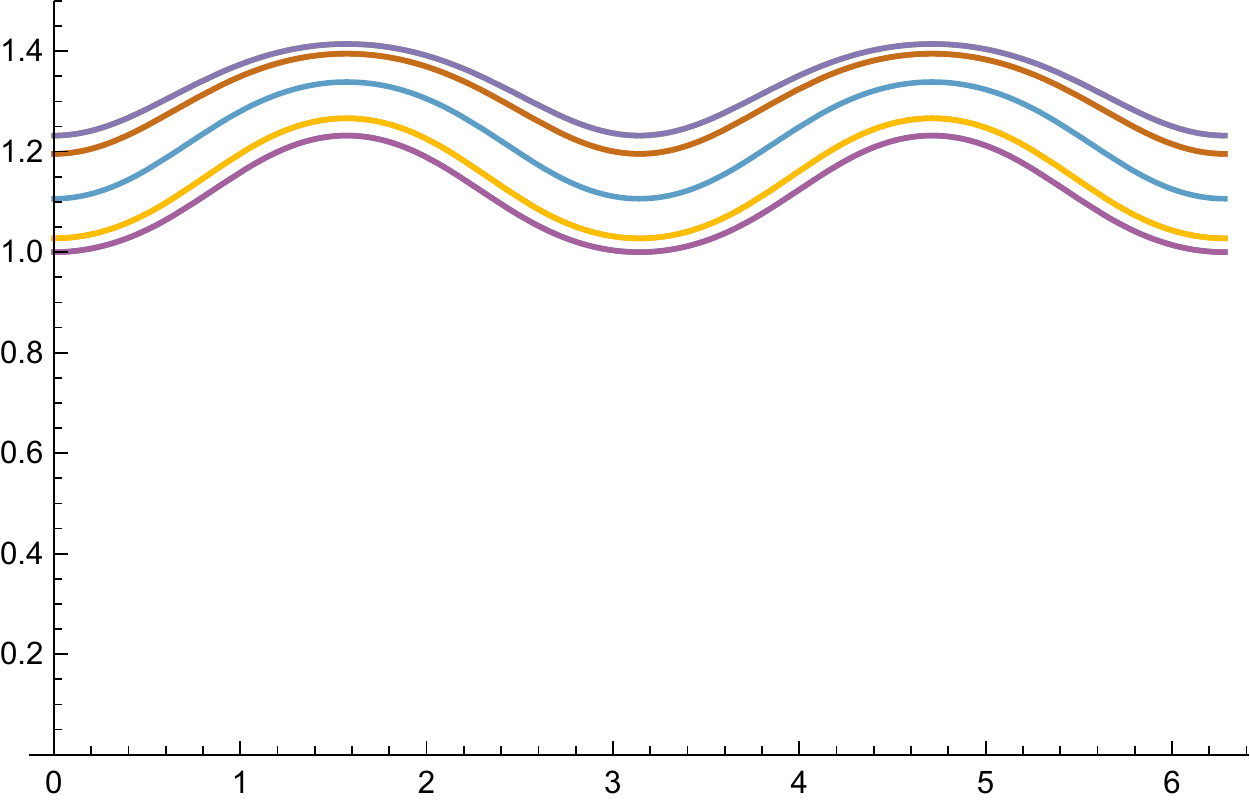} &
      \includegraphics[width=0.40\textwidth]{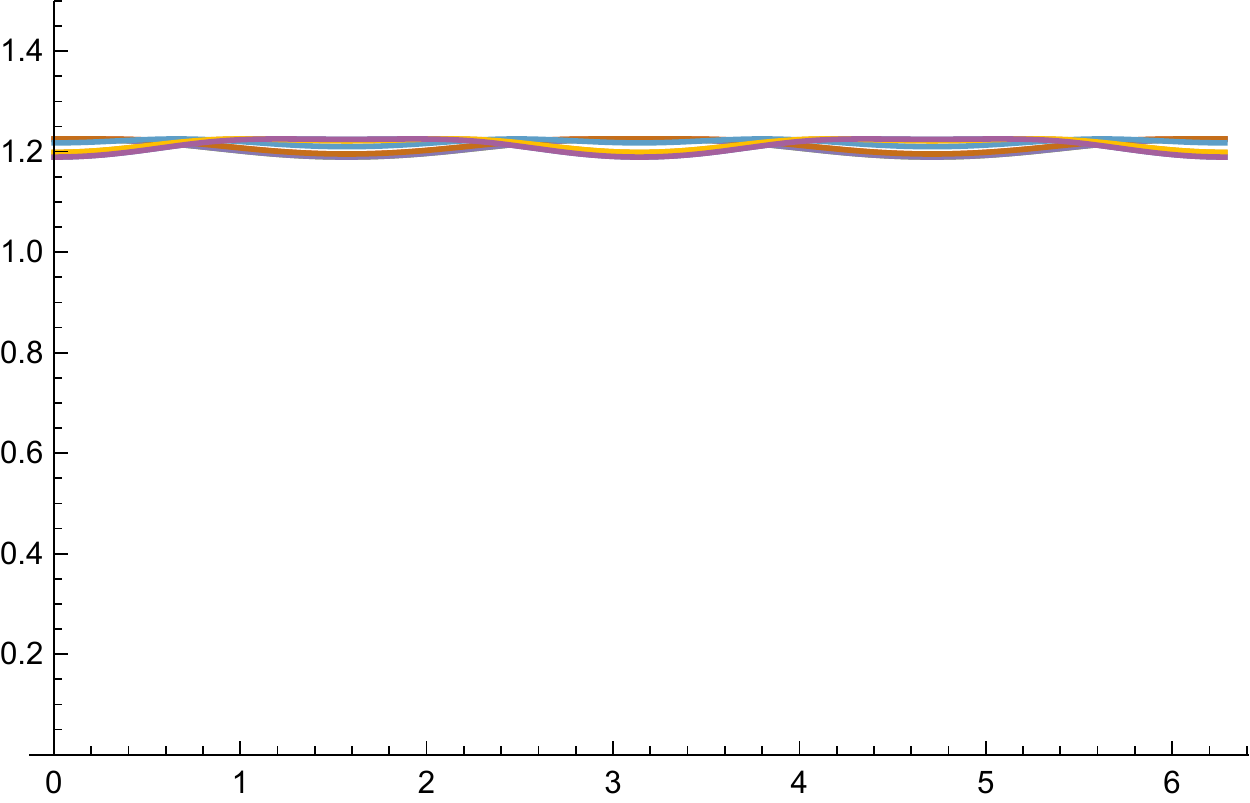} \\
  \\
       {\small\em regular scale estimates $\Gamma_s = \Gamma_{\tau} = 1/4$} 
      & {\small\em phase-compensated $\Gamma_s = \Gamma_{\tau} = 1/4$} \\
      \includegraphics[width=0.40\textwidth]{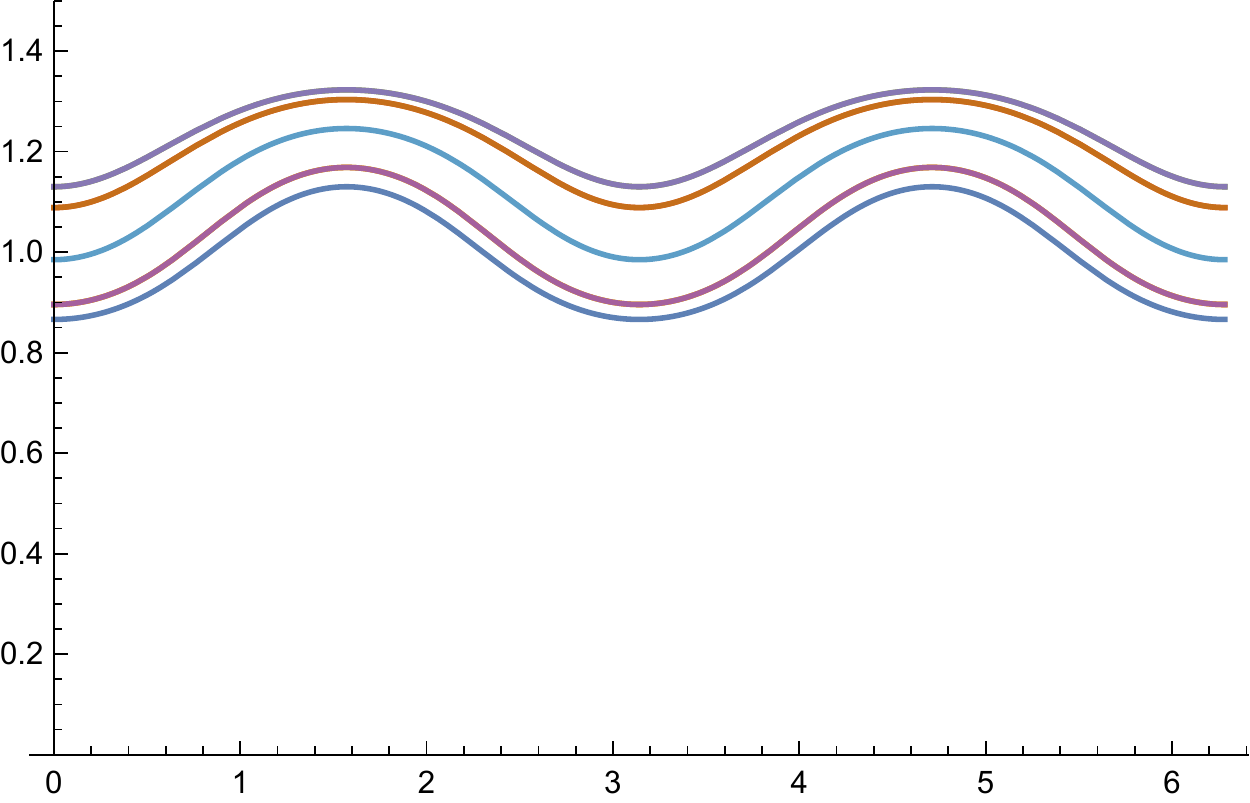} &
      \includegraphics[width=0.40\textwidth]{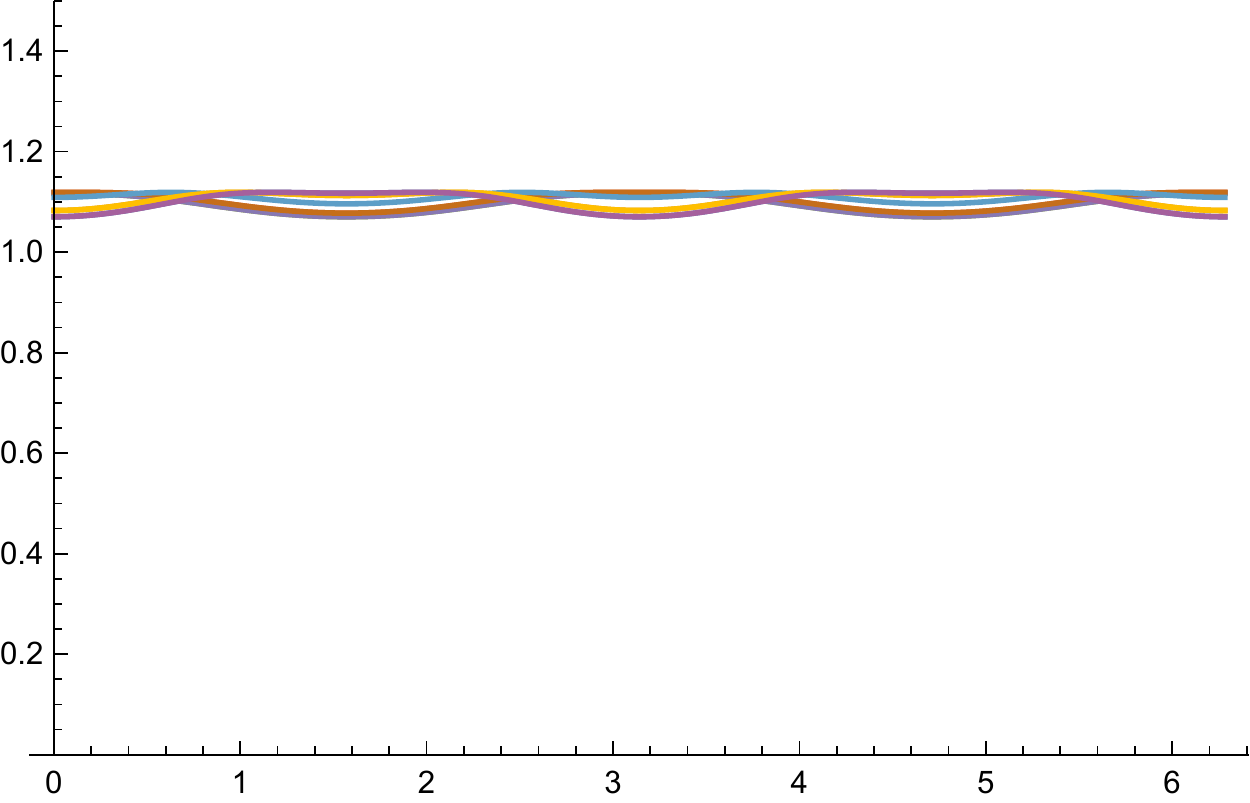} \\
 \\
       {\small\em regular scale estimates $\Gamma_s = \Gamma_{\tau} = 1/2$} 
      & {\small\em phase-compensated $\Gamma_s = \Gamma_{\tau} = 1/2$} \\
    \includegraphics[width=0.40\textwidth]{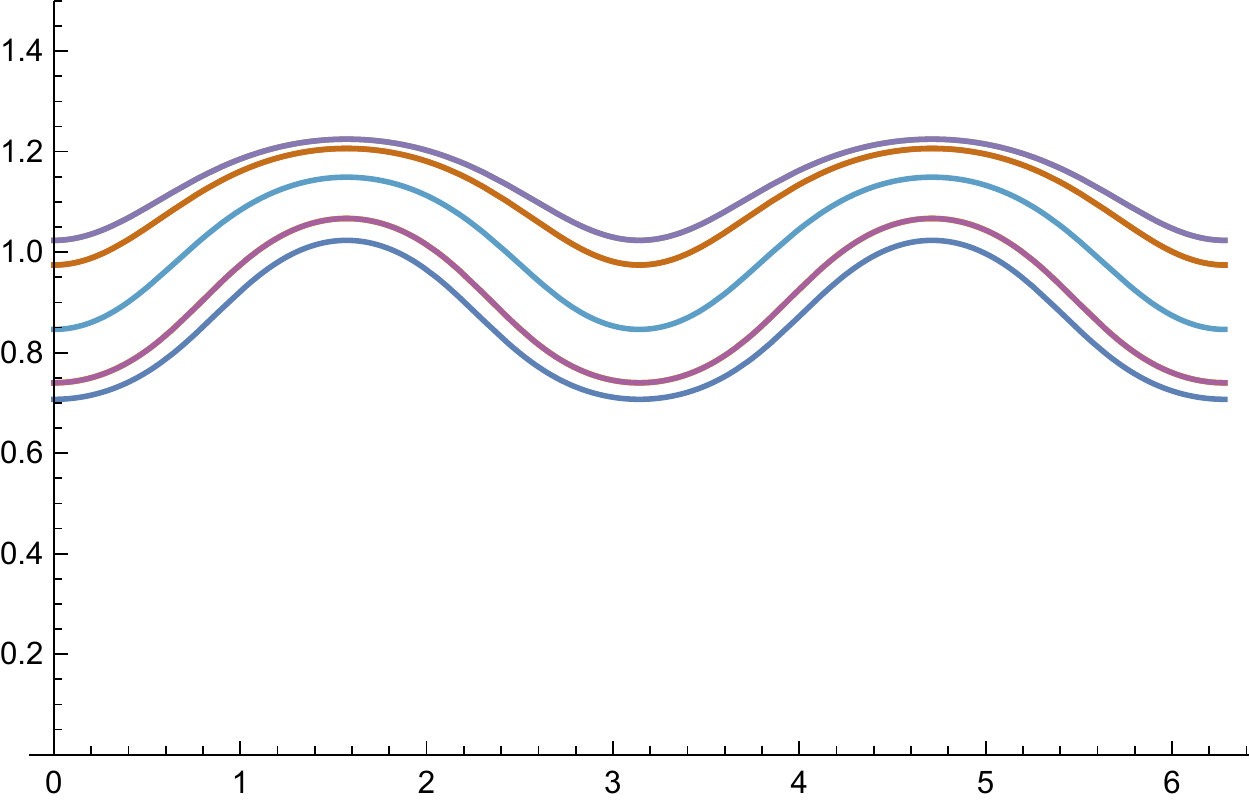} &
      \includegraphics[width=0.40\textwidth]{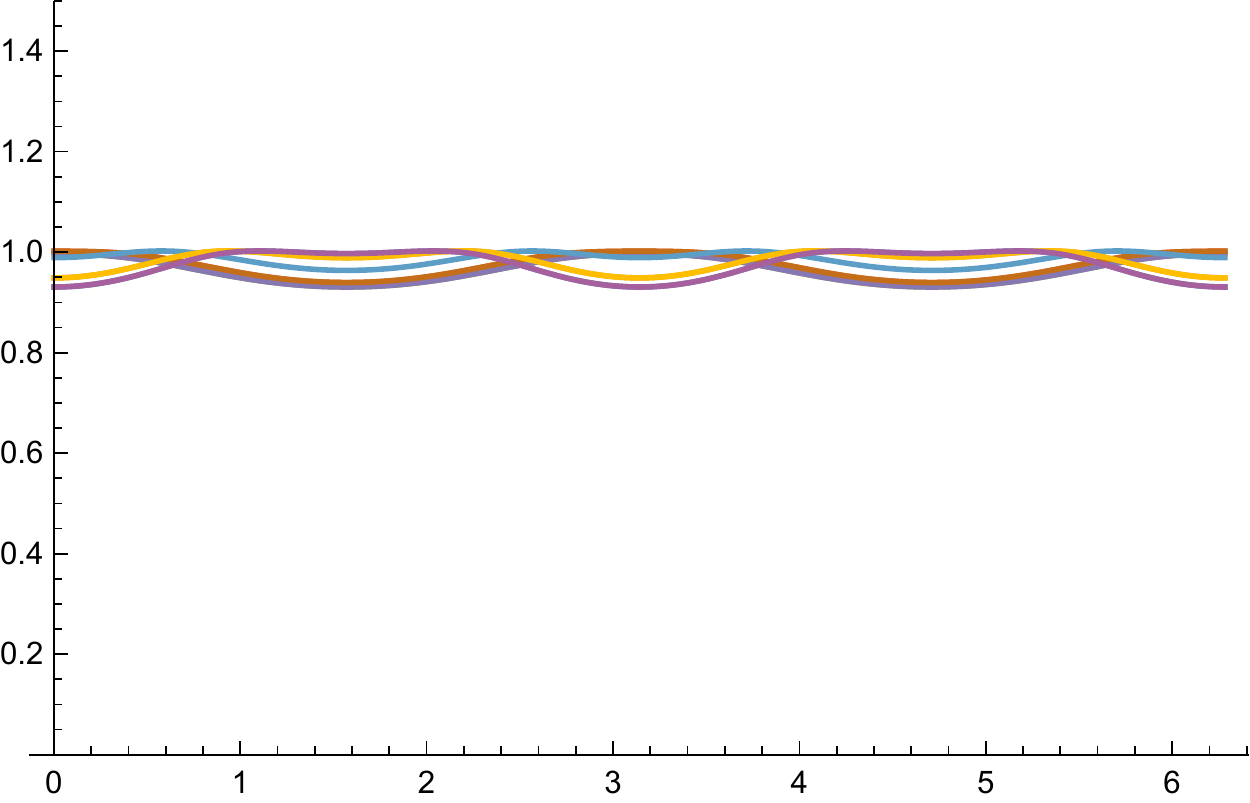} \\
   \end{tabular} 
  \end{center}
   \caption{Spatial variability of {\em spatial scale estimates\/}
     $\hat{\sigma}_s = \sqrt{s}$ computed
     (left column) from local extrema over spatial scale of the
     $\Gamma$-normalized quasi quadrature measure 
     ${\cal Q}_{(x, y, t),\Gamma-norm} L$
     according to (\ref{eq-Q3-spat-temp-scaleest-regular-geom-sine-s-tau}) for a 2+1-D spatio-temporal sine
     wave with spatial angular frequency $\omega_s = 1$ and 
     (right column) with phase compensation of the scale estimates
     according to (\ref{eq-Q3-spat-scaleest-phasecorr-geom-sine-s}).
(Horizontal axis: spatial position $t$) (The multiple graphs in each
diagram show the variability of the scale estimates for different
values of the complementary spatial coordinate $y = n \pi/8$.)}
  \label{fig-Q3-sine-spat-scale-estimates-regular-phasecomp}
\end{figure*}

\begin{figure*}[bt]
  \begin{center}
    \begin{tabular}{ccc}
       {\small\em regular scale estimates $\Gamma_s = \Gamma_{\tau} = 0$} 
      & {\small\em phase-compensated $\Gamma_s = \Gamma_{\tau} = 0$} \\
      \includegraphics[width=0.40\textwidth]{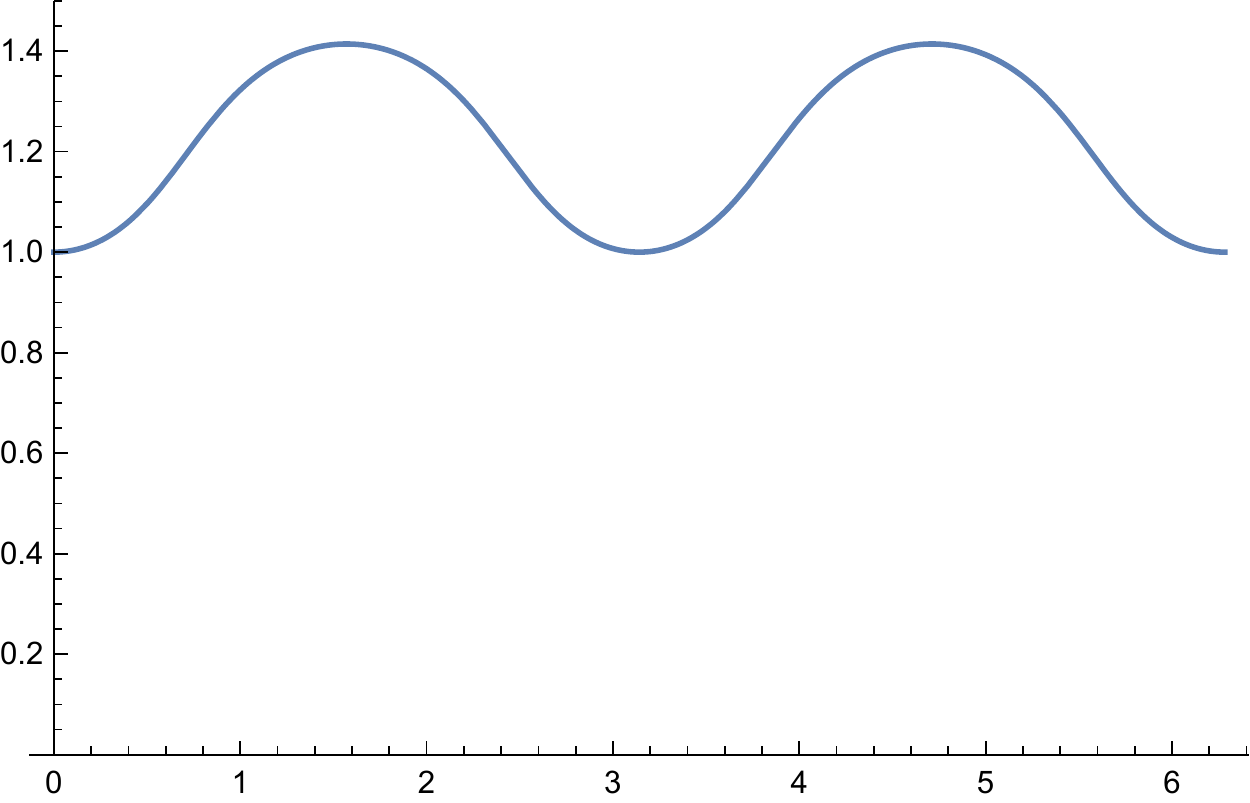} &
      \includegraphics[width=0.40\textwidth]{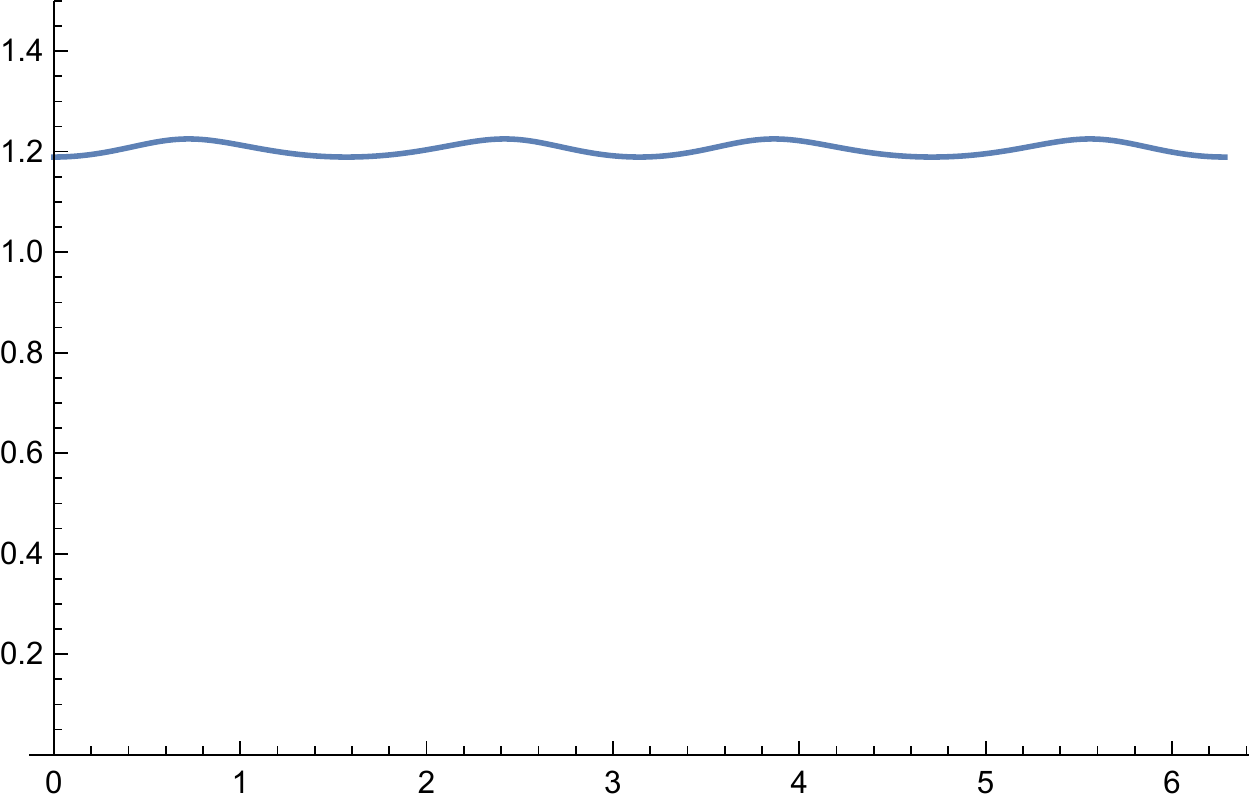} \\
  \\
       {\small\em regular scale estimates $\Gamma_s = \Gamma_{\tau} = 1/4$} 
      & {\small\em phase-compensated $\Gamma_s = \Gamma_{\tau} = 1/4$} \\
      \includegraphics[width=0.40\textwidth]{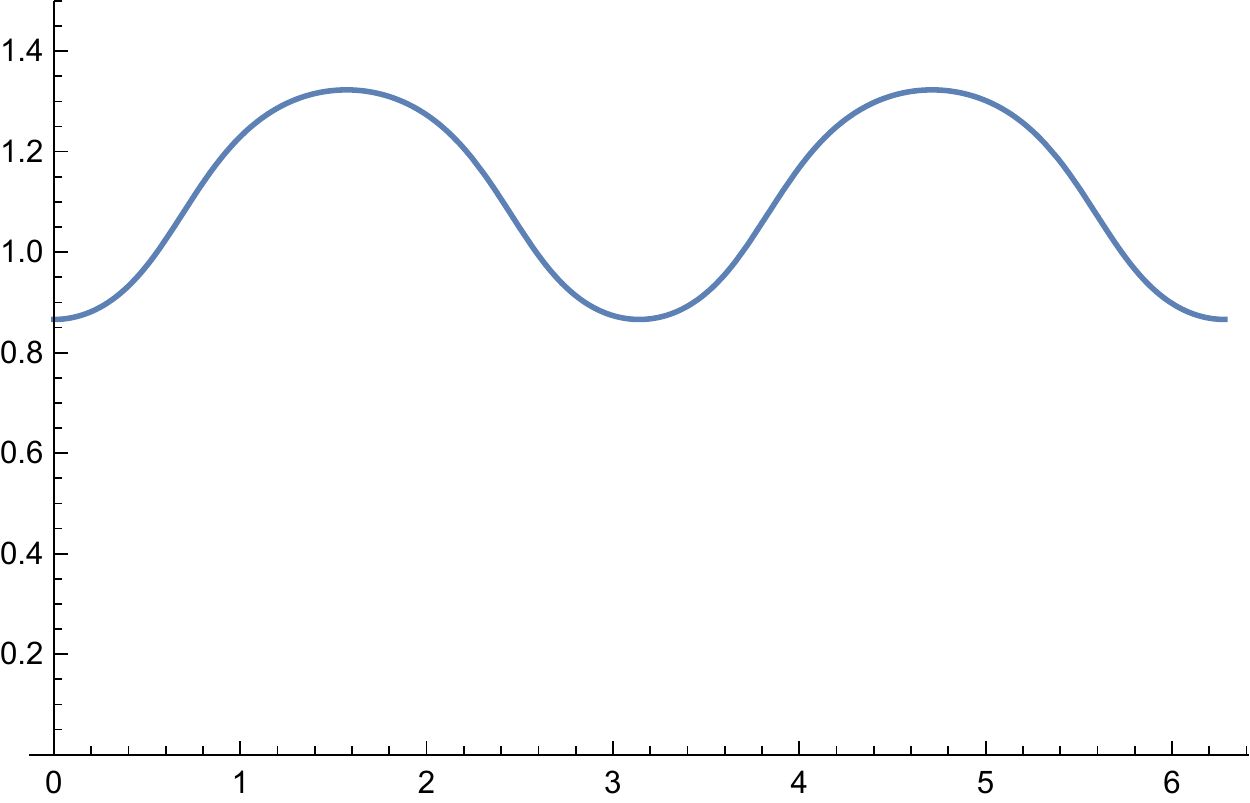} &
      \includegraphics[width=0.40\textwidth]{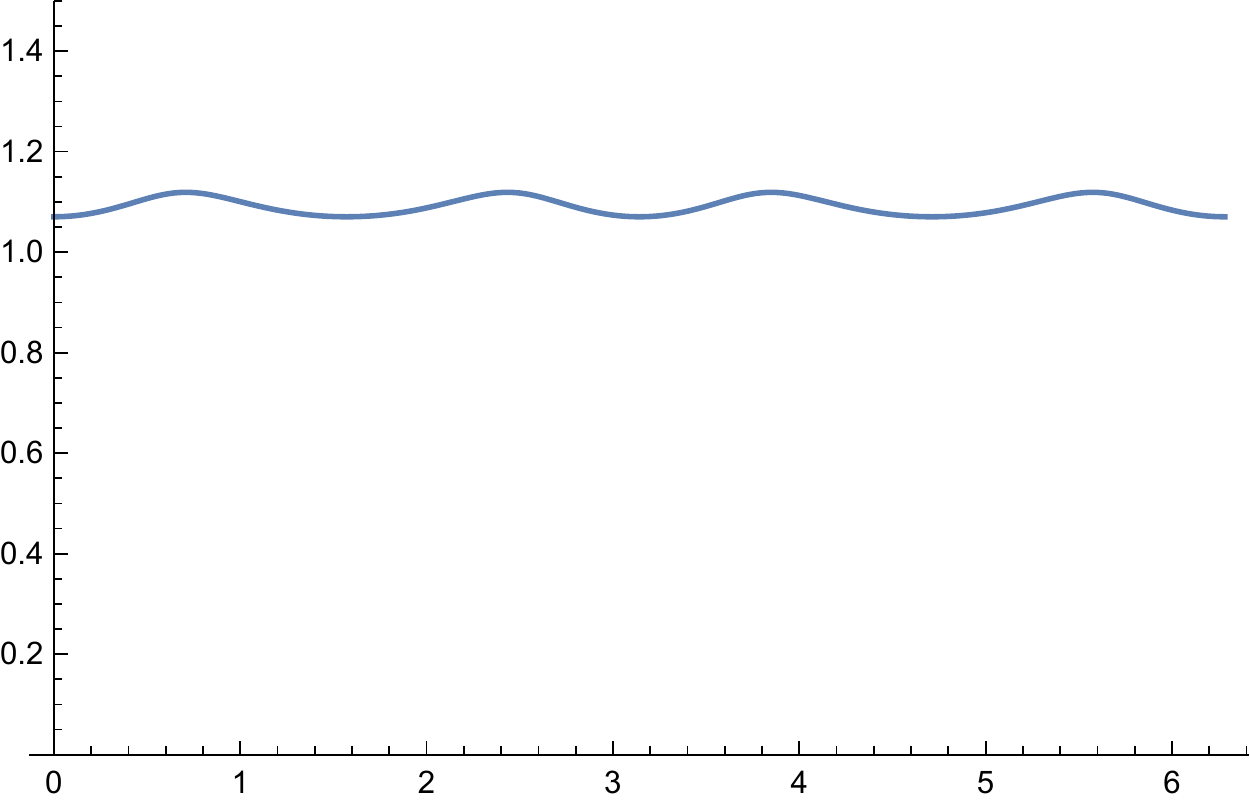} \\
 \\
       {\small\em regular scale estimates $\Gamma_s = \Gamma_{\tau} = 1/2$} 
      & {\small\em phase-compensated $\Gamma_s = \Gamma_{\tau} = 1/2$} \\
    \includegraphics[width=0.40\textwidth]{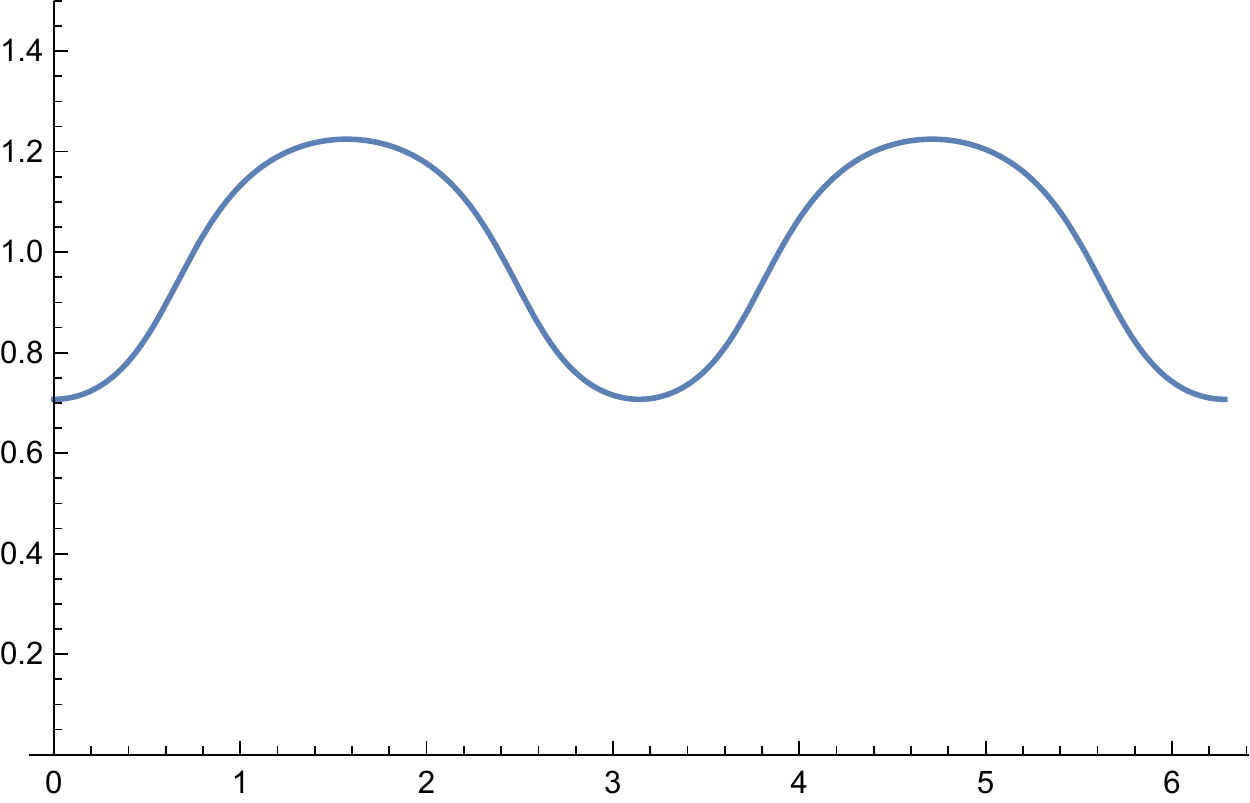} &
      \includegraphics[width=0.40\textwidth]{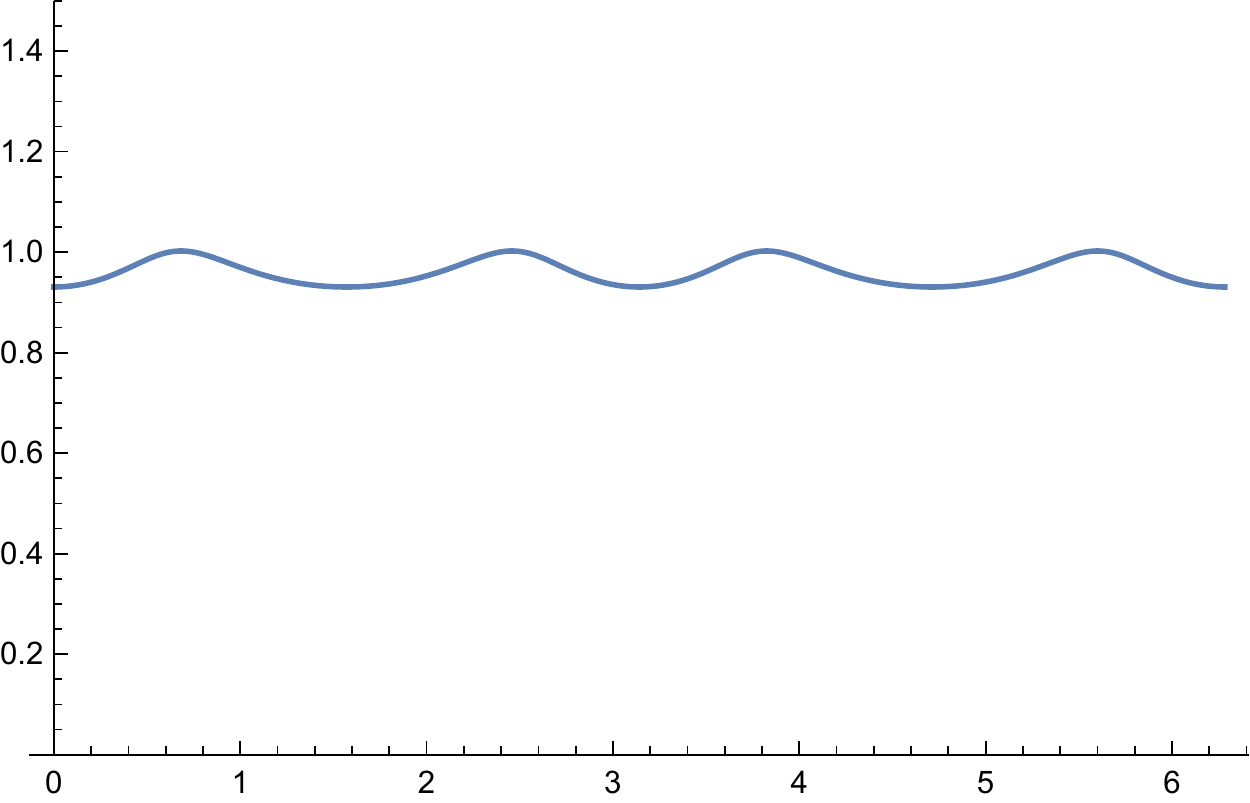} \\
   \end{tabular} 
  \end{center}
   \caption{Temporal variability of {\em temporal scale estimates\/}
     $\hat{\sigma}_{\tau} = \sqrt{\tau}$ computed
     (left column) from local extrema over temporal scale of the
     $\Gamma$-normalized quasi quadrature measure 
     ${\cal Q}_{(x, y, t),\Gamma-norm} L$
     according to (\ref{eq-Q3-spat-temp-scaleest-regular-geom-sine-s-tau}) for a 2+1-D spatio-temporal sine
     wave with spatial angular frequency $\omega_s = 1$ and 
     (right column) with phase compensation of the scale estimates
     according to (\ref{eq-Q3-temp-scaleest-phasecorr-geom-sine-tau}).
(Horizontal axis: time $t$)}
  \label{fig-Q3-sine-temp-scale-estimates-regular-phasecomp}
\end{figure*}

\subsection{Phase-compensated scale estimates}

Given the understanding from
Section~\ref{sc-sel-prop-sine-wave-spat-tem} 
of how the local spatio-temporal scale estimates depend on the local
phase of a sine wave, we can define {\em phase-compensated spatial
  and temporal scale estimates\/} according to
\begin{align}
 \begin{split}
     \label{eq-Q3-spat-scaleest-phasecorr-geom-sine-s}
     & \hat{s}_{{\cal Q}_{(x,y,t) L, comp}} = 
 \end{split}\nonumber\\
 \begin{split}
     & \frac{\sqrt{(1 - \Gamma_s)(2 - \Gamma_s)} \, \hat{s}_{{\cal Q}_{(x,y,t) L, \Gamma-norm}}}
                {(1-\Gamma_s)^{\frac{{\cal Q}_{1,(x,y),\Gamma-norm} L_t + C_{\tau} {\cal Q}_{1,(x,y),\Gamma-norm} L_{tt}}
                                                  {{\cal Q}_{(x,y,t),\Gamma-norm} L}}
                (2-\Gamma_s)^{\frac{{\cal Q}_{2,(x,y),\Gamma-norm} L_t + C_{\tau} {\cal Q}_{2,(x,y),\Gamma-norm} L_{tt}}
                                        {{\cal Q}_{(x,y,t),\Gamma-norm} L}}},
  \end{split}\\
  \begin{split}
     \label{eq-Q3-temp-scaleest-phasecorr-geom-sine-tau}
     & \hat{\tau}_{{\cal Q}_{(x,y,t) L, comp}} = 
 \end{split}\nonumber\\
 \begin{split}
     & \frac{\sqrt{(1 - \Gamma_{\tau})(2 - \Gamma_{\tau})} \, \hat{\tau}_{{\cal Q}_{(x,y,t) L, \Gamma-norm}}}
                {(1-\Gamma_{\tau})^{\frac{{\cal Q}_{1,(x,y),\Gamma-norm} L_t + {\cal Q}_{2,(x,y),\Gamma-norm} L_t}
                                               {{\cal Q}_{(x,y,t),\Gamma-norm} L}}
                (2-\Gamma_{\tau})^{\frac{C_{\tau} \left( {\cal Q}_{1,(x,y),\Gamma-norm} L_{tt} + {\cal Q}_{2,(x,y),\Gamma-norm} L_{tt} \right)}
                                              {{\cal Q}_{(x,y,t),\Gamma-norm} L}}},
  \end{split}
\end{align}
defined to be equal to the geometric averages of the extreme values
\begin{align}
  \begin{split}
    \hat{s} = \sqrt{\hat{s}_1 \, \hat{s}_2} 
    = \frac{\sqrt{(1-\Gamma_s)(2-\Gamma_s)}}{\omega_0^2}
  \end{split}\\
  \begin{split}
    \hat{\tau} = \sqrt{\hat{\tau}_1 \, \hat{\tau}_2} 
    = \frac{\sqrt{(1-\Gamma_{\tau})(2-\Gamma_{\tau})}}{\omega_{\tau}^2}
  \end{split}
\end{align}
in the extreme cases when only one of the first- or second-order
components in the components responds and with a much lower
variability in between because of the blending of the responses to the
first- {\em vs.\/}\ second-order spatial or temporal derivatives 
(see figures~\ref{fig-Q3-sine-spat-scale-estimates-regular-phasecomp}--\ref{fig-Q3-sine-temp-scale-estimates-regular-phasecomp}).

From these spatial and temporal scale estimates, we can in turn estimate the
spatial and temporal wavelengths of the sine wave according to
\begin{align}
  \begin{split}
     \hat{\lambda}_s
      & = \frac{2 \pi \sqrt{\hat{s}_{{\cal Q}_{(x,y,t) L, comp}}}}{\sqrt[4]{(1 - \Gamma_s)(2 - \Gamma_s)}},
  \end{split}\\
  \begin{split}
     \hat{\lambda}_{\tau}
      & = \frac{2 \pi \sqrt{\hat{\tau}_{{\cal Q}_{(x,y,t) L, comp}}}}{\sqrt[4]{(1 - \Gamma_{\tau})(2 - \Gamma_{\tau})}}.
  \end{split}
\end{align}

\subsection{Scale calibration} 

When applied to a Gaussian blink of spatial extent $s_0$
and temporal duration $\tau_0$
\begin{equation}
  \label{eq-gauss-blink-blob}
  f(x, y, t) 
  = g(x, y;\; s_0) \, g(t;\; \tau_0)
  = \frac{1}{(2 \pi)^{3/2} s_0 \sqrt{\tau_0}} \, e^{-(x^2+y^2)/2s_0} \, e^{-t^2/2\tau_0},
\end{equation}
for which the spatio-temporal scale-space representation is of the form
\begin{equation}
  \label{eq-gauss-blink-blob-spat-temp-scsp}
  L(x, y, t;\; s_0, \tau_0)
%  = \left( g(\cdot, \cdot;\; s) \, g(\cdot;\; \tau) \right) * \left( g(\cdot, \cdot;\; s_0) \, g(\cdot;\; \tau_0) \right)(x, y, t;\; s_0, \tau_0)
  = g(x, y;\; s_0+s) \, g(t;\; \tau_0+\tau),
\end{equation}
the spatial and temporal scale
estimates will according to the theoretical analysis in
(Lindeberg \cite{Lin17-SSVM,Lin17-JMIV-subm}) be given by
%(\ref{eq-spat-scale-est-blink-gauss-blob}) and
%(\ref{eq-temp-scale-est-blink-gauss-blob})
\begin{align}
  \begin{split}
     \hat{s} & = \frac{\gamma_s s_0}{2 - \gamma_s} \, s_0,
  \end{split}\\
 \begin{split}
     \hat{\tau} & = \frac{2 \gamma_{\tau} \tau_0}{3 - 2 \gamma_{\tau}} \, \tau_0.
  \end{split}
\end{align}
Here, for $\gamma_s = 1 - \Gamma_s/2$ and $\gamma_{\tau} = 1 - \Gamma_{\tau}/2$ 
this implies that the regular scale estimates for a Gaussian blink are
\begin{align}
  \begin{split}
     \hat{s} & = \frac{2-\Gamma_s}{2+\Gamma_s} \, s_0,
  \end{split}\\
 \begin{split}
     \hat{\tau} & = \frac{2 - \Gamma_{\tau}}{1 + \Gamma_{\tau}} \, \tau_0,
  \end{split}
\end{align}
and the corresponding phase-compensated scale estimates
\begin{align}
  \begin{split}
     \hat{s}_{{\cal Q}_{(x,y,t) L, comp}} 
     = \frac{\sqrt{(2-\Gamma_s) (1-\Gamma_s)}}{2+\Gamma_s} \, s_0,
  \end{split}\\
  \begin{split}
     \hat{\tau}_{{\cal Q}_{(x,y,t) L, comp}} 
     = \frac{\sqrt{(2-\Gamma) (1-\Gamma)}}{1+\Gamma_{\tau}} \, \tau_0.
  \end{split}
\end{align}
If we want to calibrate the spatial and temporal scale estimates such that the
spatial and temporal scale estimates are equal to $\hat{s} = s_0$ and
$\hat{\tau} = \tau_0$ for a Gaussian blink of spatial extent $s_0$ and
temporal duration $\tau_0$, we should therefore
calibrate the phase compensated scale estimates 
$\hat{s}_{{\cal Q}_{(x,y,t) L, comp}}$ and $\hat{\tau}_{{\cal Q}_{(x,y,t) L, comp}}$
according to
\begin{align}
  \begin{split}
    \label{eq-calib-sc-est-Q3-2+1D-sine-spat}
      \hat{s}_{{\cal Q}_{(x,y,t) L, calib}}
      = \frac{2+\Gamma_s}{\sqrt{(2-\Gamma_s) (1-\Gamma_s)}} \, \hat{s}_{{\cal Q}_{(x,y,t) L, comp}},
  \end{split}\\
  \begin{split}
    \label{eq-calib-sc-est-Q3-2+1D-sine-temp}
      \hat{\tau}_{{\cal Q}_{(x,y,t) L, calib}} 
      = \frac{1+\Gamma_{\tau}}{\sqrt{(2-\Gamma_{\tau}) (1-\Gamma_{\tau})}} \, \hat{\tau}_{{\cal Q}_{(x,y,t) L, comp}}.
  \end{split}
\end{align}
With this scale calibration, since the scale estimate for a Gaussian
temporal onset ramp, which for regular $\gamma$-normalized temporal
derivatives assumes the form (Lindeberg
\cite[equation~(23)]{Lin98-IJCV}) 
\begin{equation}
  \label{eq-sc-sel-ramp-1D-general-gamma-spat}
  \hat{s} 
  = \frac{\gamma}{1 - \gamma} \, s_0
  = \left\{ \gamma = 1 - \Gamma \right\}
  = \frac{1-\Gamma}{\Gamma} \, s_0,
\end{equation}
the spatial scale estimate for a diffuse Gaussian
edge will by combination of
% \cite[equation~(23)]{Lin98-IJCV}) and
%(\ref{eq-sc-sel-ramp-1D-general-gamma-spat}) and
(\ref{eq-Q3-spat-scaleest-phasecorr-geom-sine-s})
with (\ref{eq-calib-sc-est-Q3-2+1D-sine-spat}) be given by
\begin{equation}
  \label{eq-effect-of-Gammas-on-scale-at-edges}
  \hat{s} 
%   = \frac{2+\Gamma}{\sqrt{(2-\Gamma) (1-\Gamma)}} 
%       \, \frac{\sqrt{(1 - \Gamma)(2 - \Gamma)}}{1-\Gamma}
%       \, \frac{1-\Gamma}{\Gamma} \, \tau_0
  = \frac{2+\Gamma_s}{\Gamma_s} \, s_0,
\end{equation}
whereas the temporal scale estimate for a Gaussian onset ramp will by
combination of %(\ref{eq-sc-sel-ramp-1D-general-gamma}) and 
(\ref{eq-Q3-temp-scaleest-phasecorr-geom-sine-tau})
with (\ref{eq-calib-sc-est-Q3-2+1D-sine-temp}) be
given by
\begin{equation}
  \label{eq-effect-of-Gammatau-on-scale-at-ramps}
  \hat{\tau} 
%   = \frac{1+\Gamma}{\sqrt{(2-\Gamma) (1-\Gamma)}} 
%       \, \frac{\sqrt{(1 - \Gamma)(2 - \Gamma)}}{1-\Gamma}
%       \, \frac{1-\Gamma}{\Gamma} \, \tau_0
  = \frac{1+\Gamma_{\tau}}{\Gamma_{\tau}} \, \tau_0.
\end{equation}
By varying the parameters $\Gamma_s$ and $\Gamma_{\tau}$
we can thereby regulate the factor by which the spatial scale estimate
for a diffuse Gaussian edge will be proportional to its diffuseness
and in a corresponding manner the factor by which the temporal scale
estimate for a Gaussian onset ramp will be proportional to its
temporal duration, while ensuring that the spatial and temporal scale
estimates for a Gaussian blink will still reflect the spatial
extent and the temporal duration of the Gaussian blink.

\begin{figure*}[hbt]
  \begin{center}
    \begin{tabular}{ccc}
       {\small\em regular magnitudes $\Gamma_s = \Gamma_{\tau} = 0$} 
      & {\small\em phase-compensated $\Gamma_s = \Gamma_{\tau} = 0$} \\
      \includegraphics[width=0.40\textwidth]{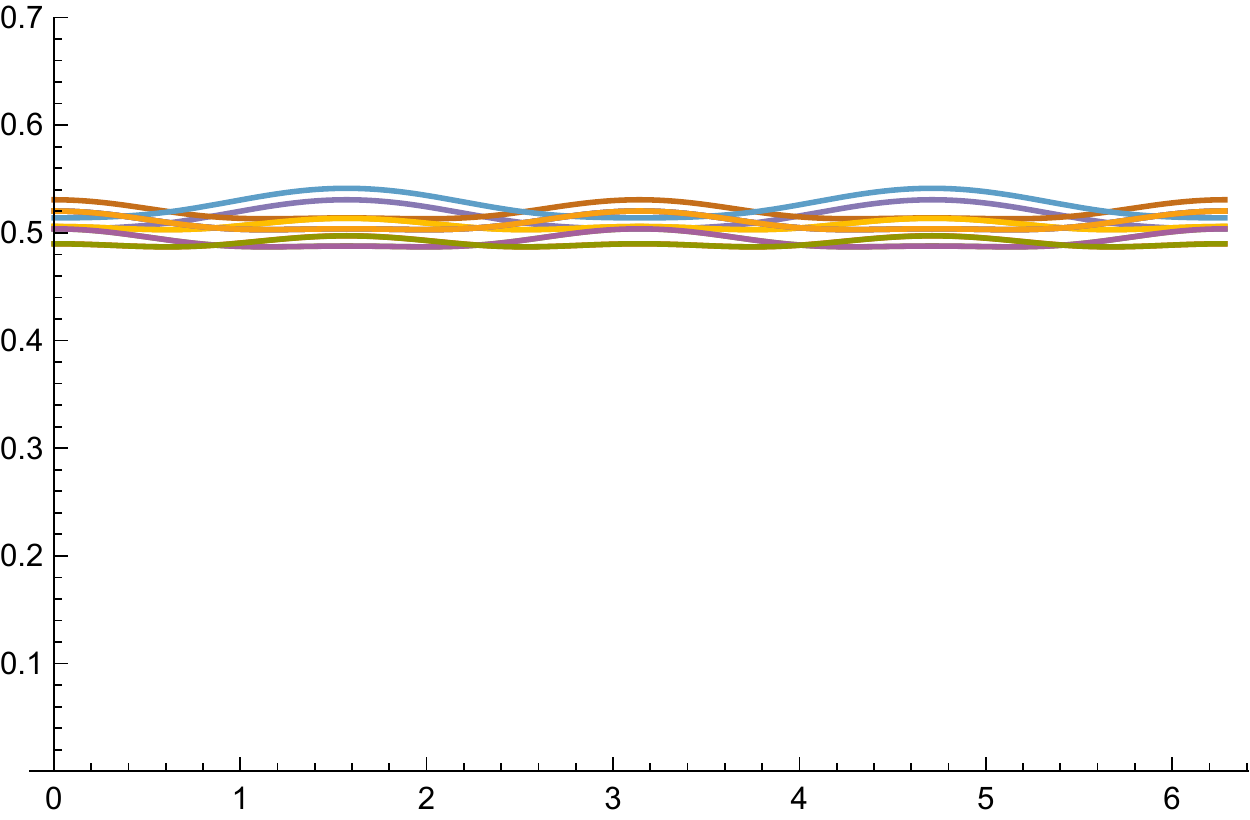} &
      \includegraphics[width=0.40\textwidth]{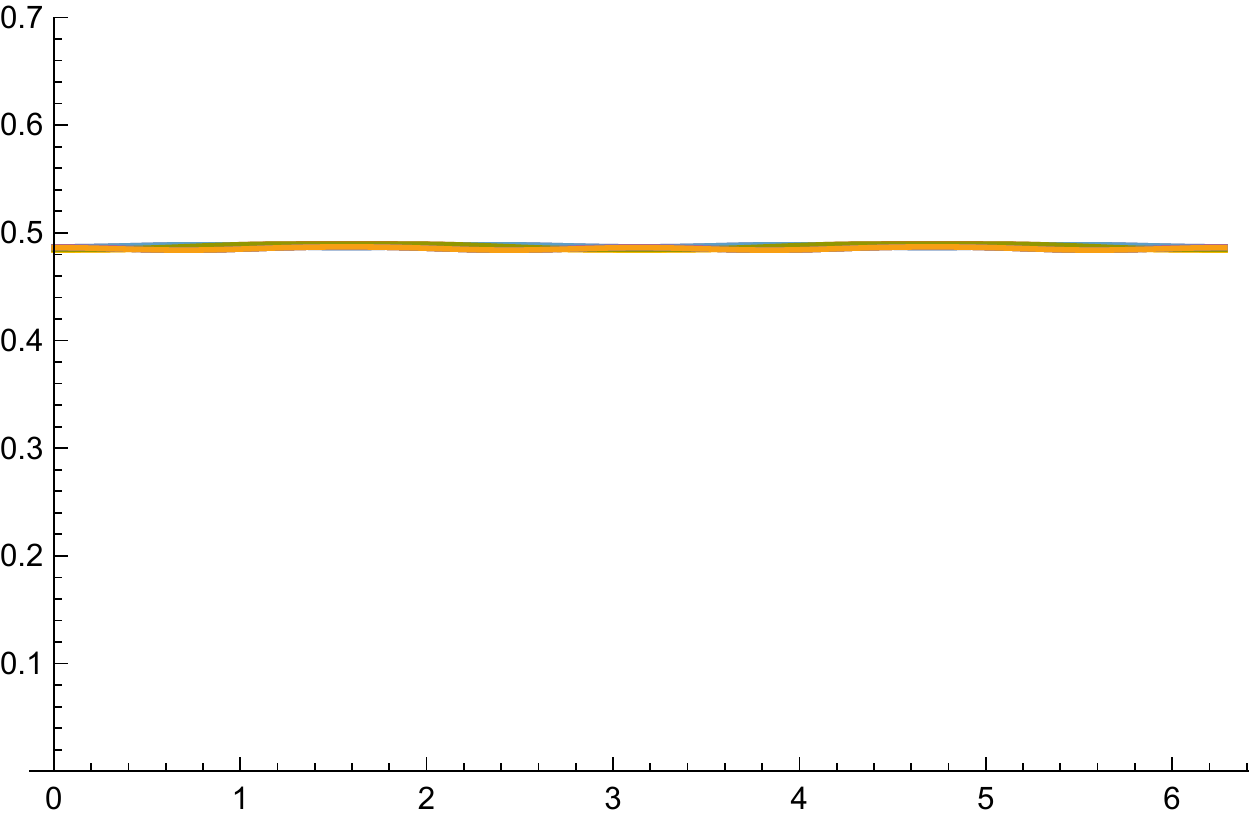} \\
  \\
       {\small\em regular magnitudes $\Gamma_s = \Gamma_{\tau} = 1/4$} 
      & {\small\em phase-compensated $\Gamma_s = \Gamma_{\tau} = 1/4$} \\
      \includegraphics[width=0.40\textwidth]{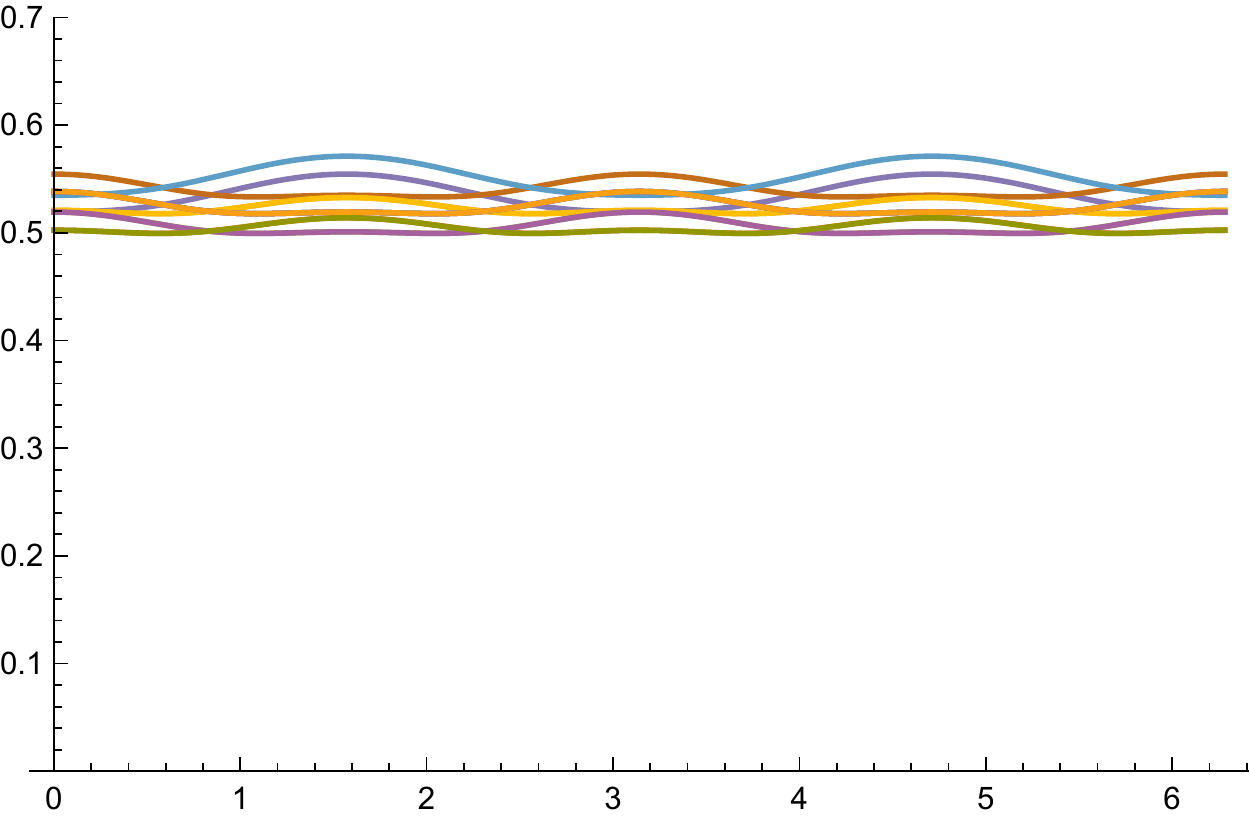} &
      \includegraphics[width=0.40\textwidth]{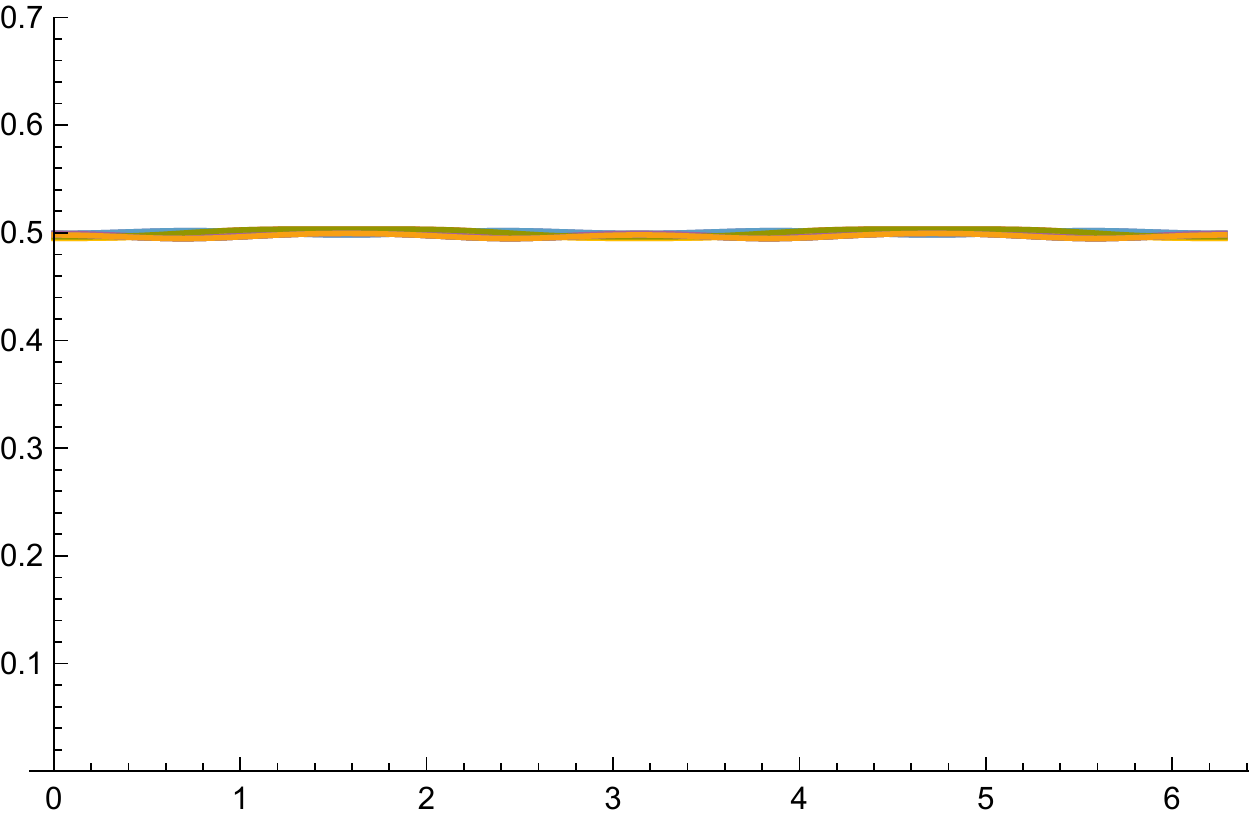} \\
  \\
       {\small\em regular magnitudes $\Gamma_s = \Gamma_{\tau} = 1/2$} 
      & {\small\em phase-compensated $\Gamma_s = \Gamma_{\tau} = 1/2$} \\
    \includegraphics[width=0.40\textwidth]{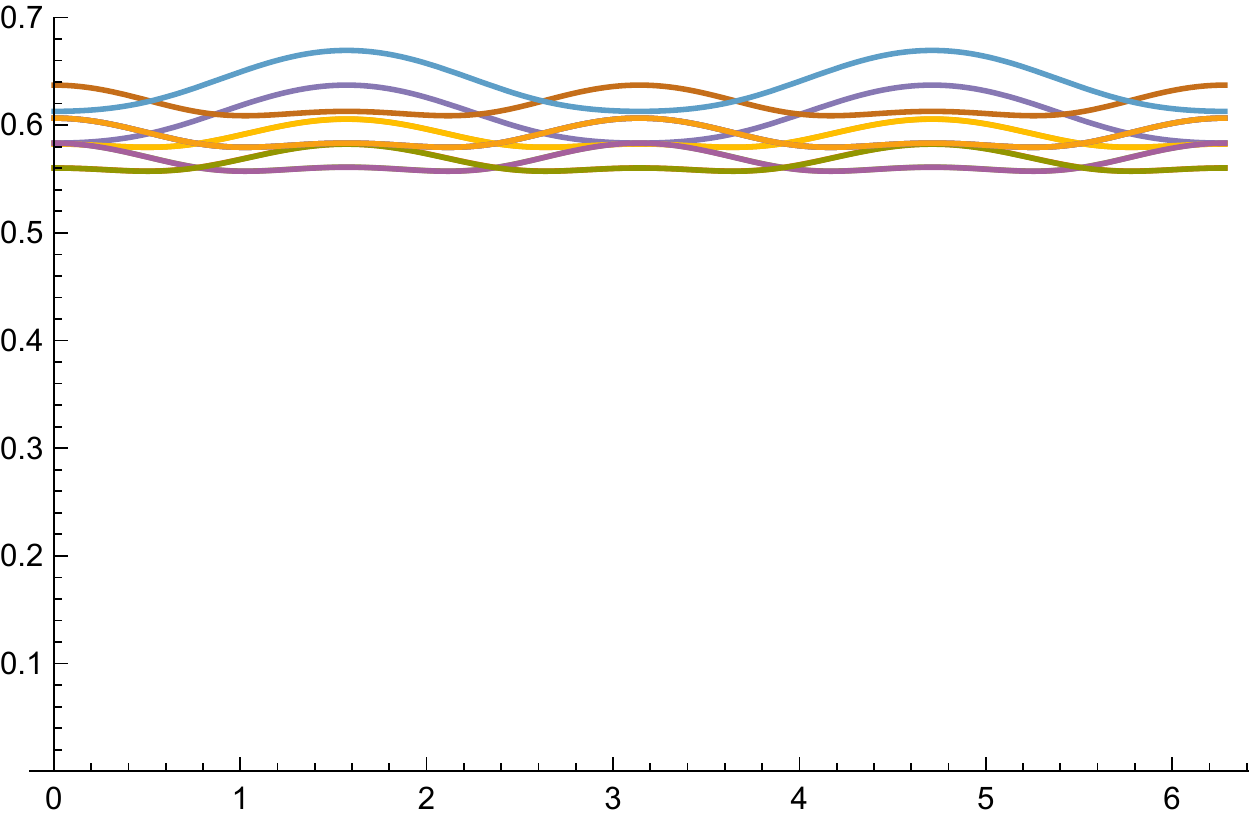} &
      \includegraphics[width=0.40\textwidth]{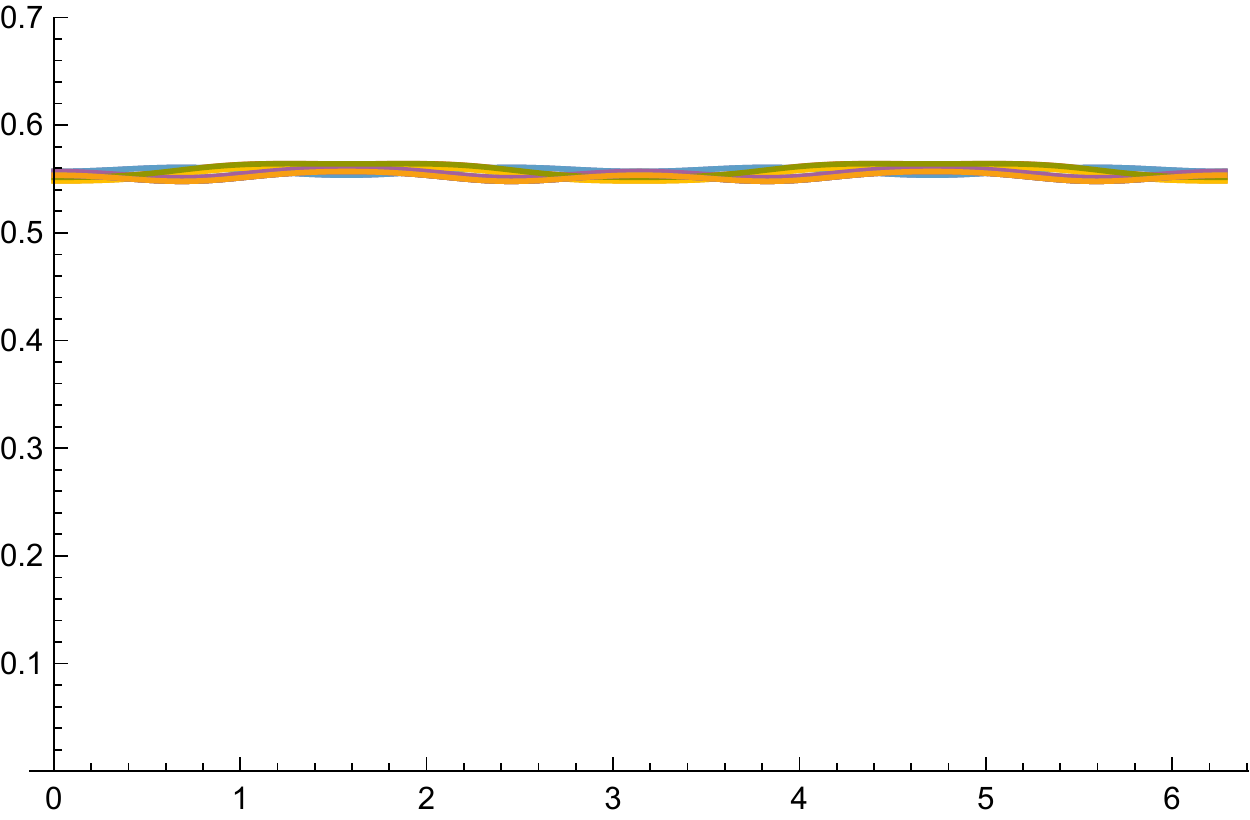} \\
   \end{tabular} 
  \end{center}
   \caption{Spatial variability of magnitude responses ${\cal Q}_{(x,y,t), \Gamma-norm}$  computed
     (left column) at the local extrema over both spatial and temporal scales of the
     $\Gamma$-normalized quasi quadrature measure ${\cal Q}_{(x,y,t), \Gamma-norm}$
     according to (\ref{eq-Q3-spat-temp-scaleest-regular-geom-sine-s-tau}) for a 2+1-D spatio-temporal sine
     wave pattern of spatial angular frequency $\omega_s = 1$ and
     temporal angular frequency $\omega_{\tau} = 1$
     (right column) at phase-compensated scale estimates
     according to (\ref{eq-Q3-spat-scaleest-phasecorr-geom-sine-s})
     and (\ref{eq-Q3-temp-scaleest-phasecorr-geom-sine-tau}).
    (Horizontal axis: spatial coordinate $x$)
   (The multiple graphs in each
  diagram show the variability of the scale estimates for different
   values of the complementary spatial coordinate $y = m \pi/4$
   and the temporal coordinate $t = n \pi/4$.)}
  \label{fig-Q3-sine-magn-estimates-regular-phasecomp-xvar}
\end{figure*}

\begin{figure*}[hbt]
  \begin{center}
    \begin{tabular}{ccc}
       {\small\em regular magnitudes $\Gamma_s = \Gamma_{\tau} = 0$} 
      & {\small\em phase-compensated $\Gamma_s = \Gamma_{\tau} = 0$} \\
      \includegraphics[width=0.40\textwidth]{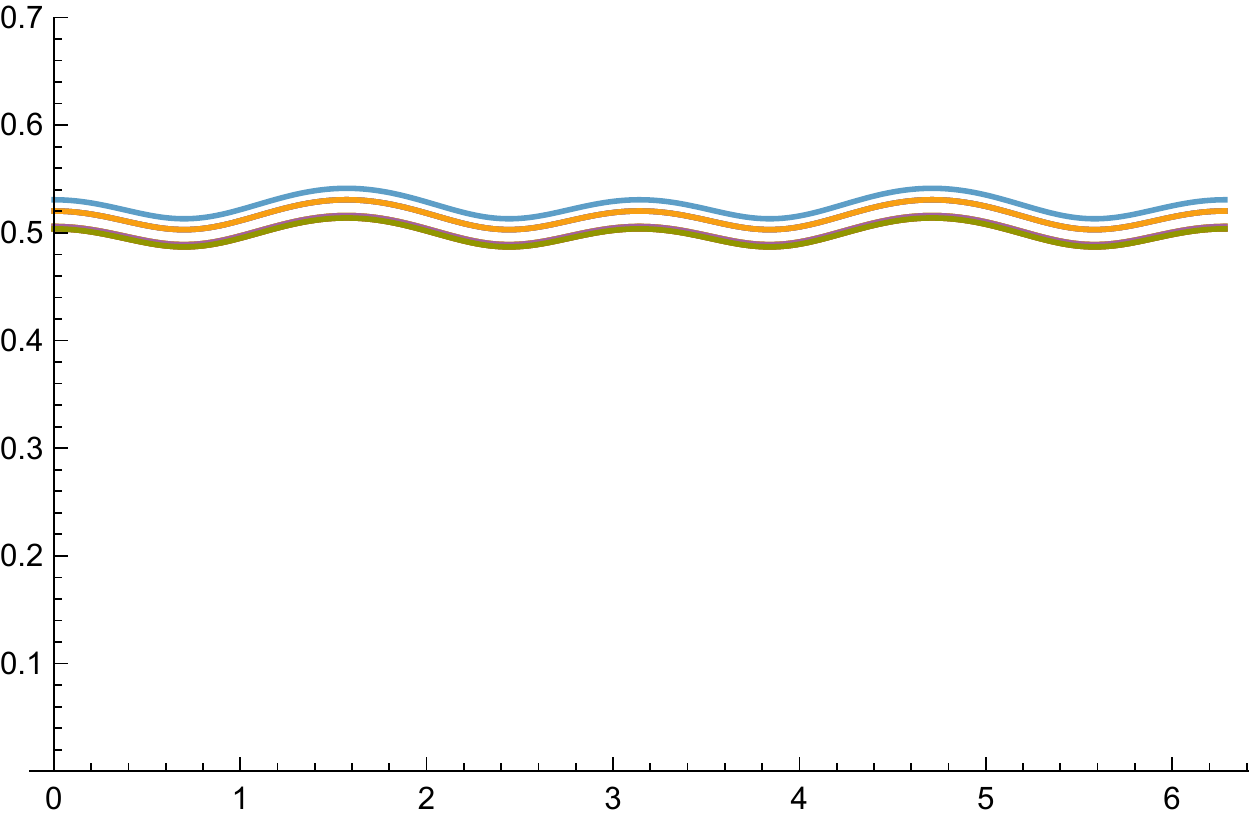} &
      \includegraphics[width=0.40\textwidth]{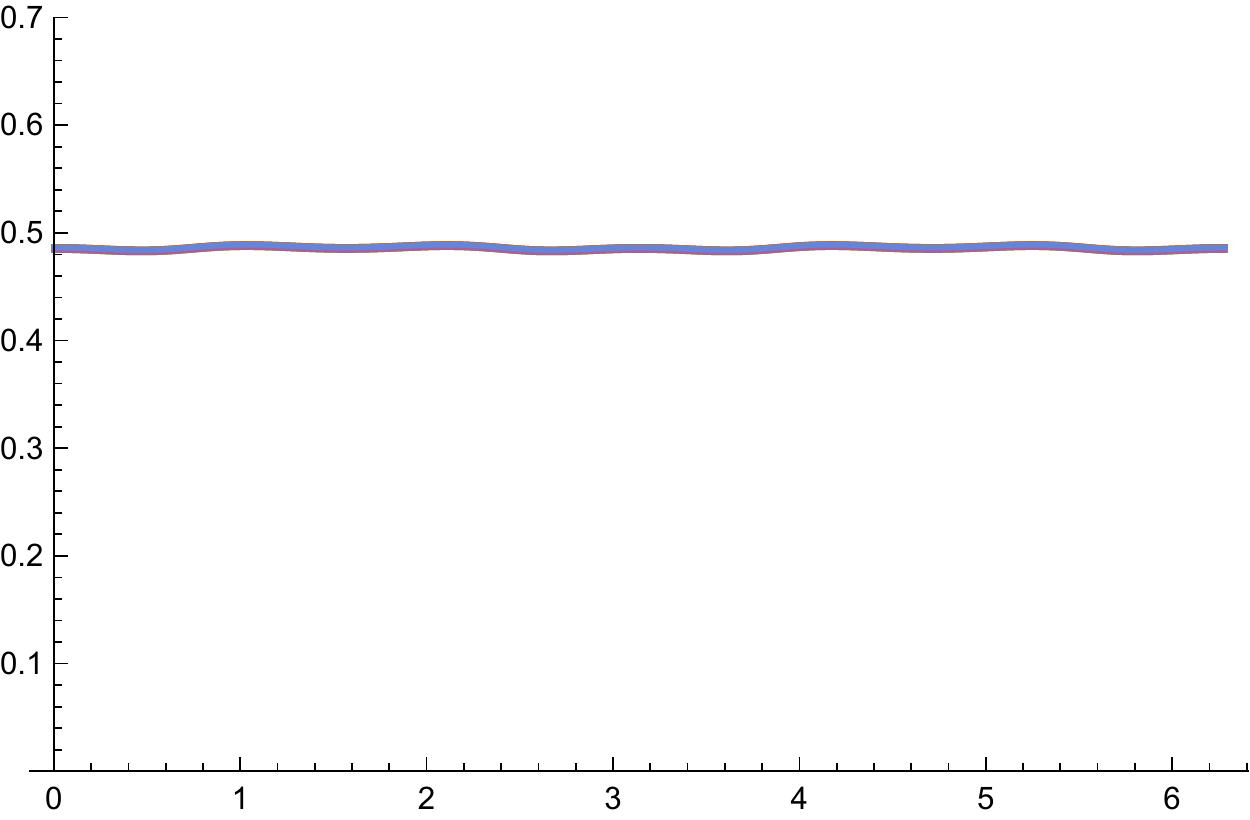} \\
  \\
       {\small\em regular magnitudes $\Gamma_s = \Gamma_{\tau} = 1/4$} 
      & {\small\em phase-compensated $\Gamma_s = \Gamma_{\tau} = 1/4$} \\
      \includegraphics[width=0.40\textwidth]{sinewave-2p1D-Q3-magn-regular-tvar-omega1-Gamma0} &
      \includegraphics[width=0.40\textwidth]{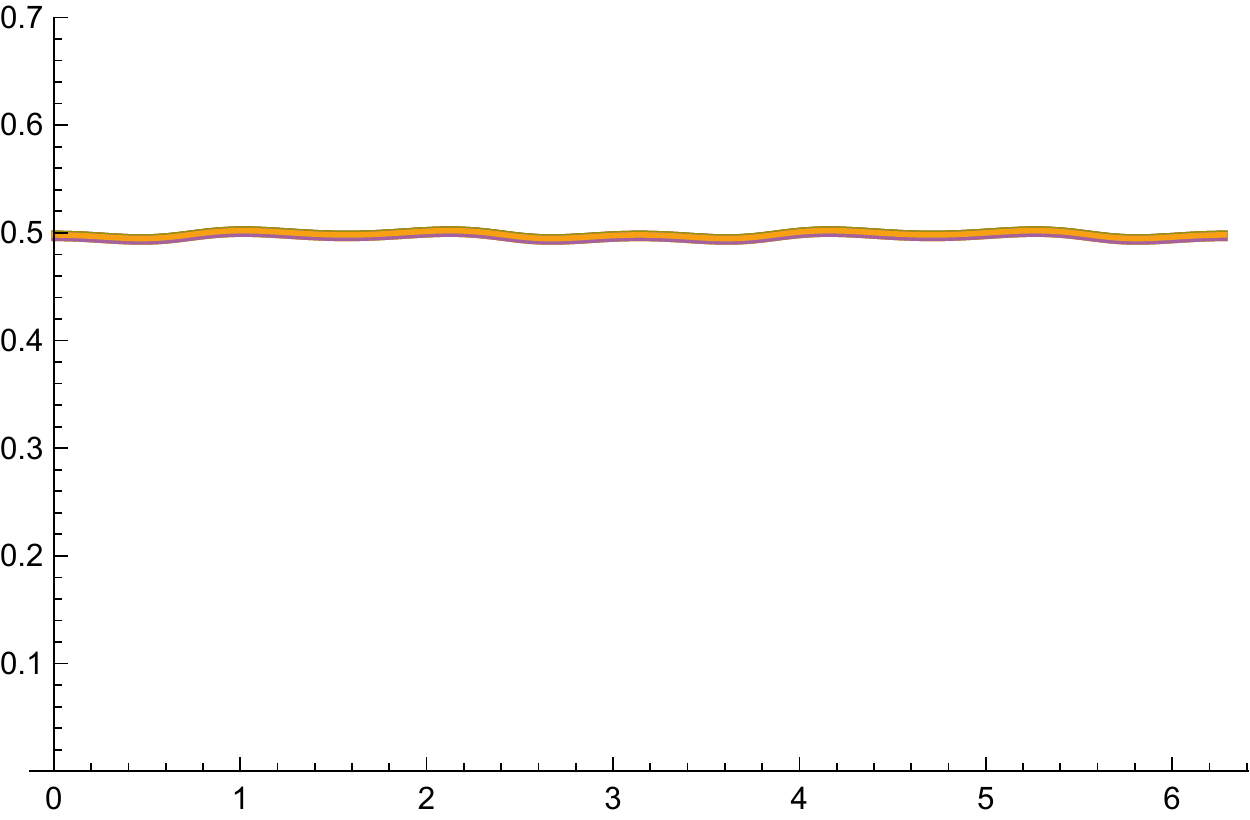} \\
  \\
       {\small\em regular magnitudes $\Gamma_s = \Gamma_{\tau} = 1/2$} 
      & {\small\em phase-compensated $\Gamma_s = \Gamma_{\tau} = 1/2$} \\
    \includegraphics[width=0.40\textwidth]{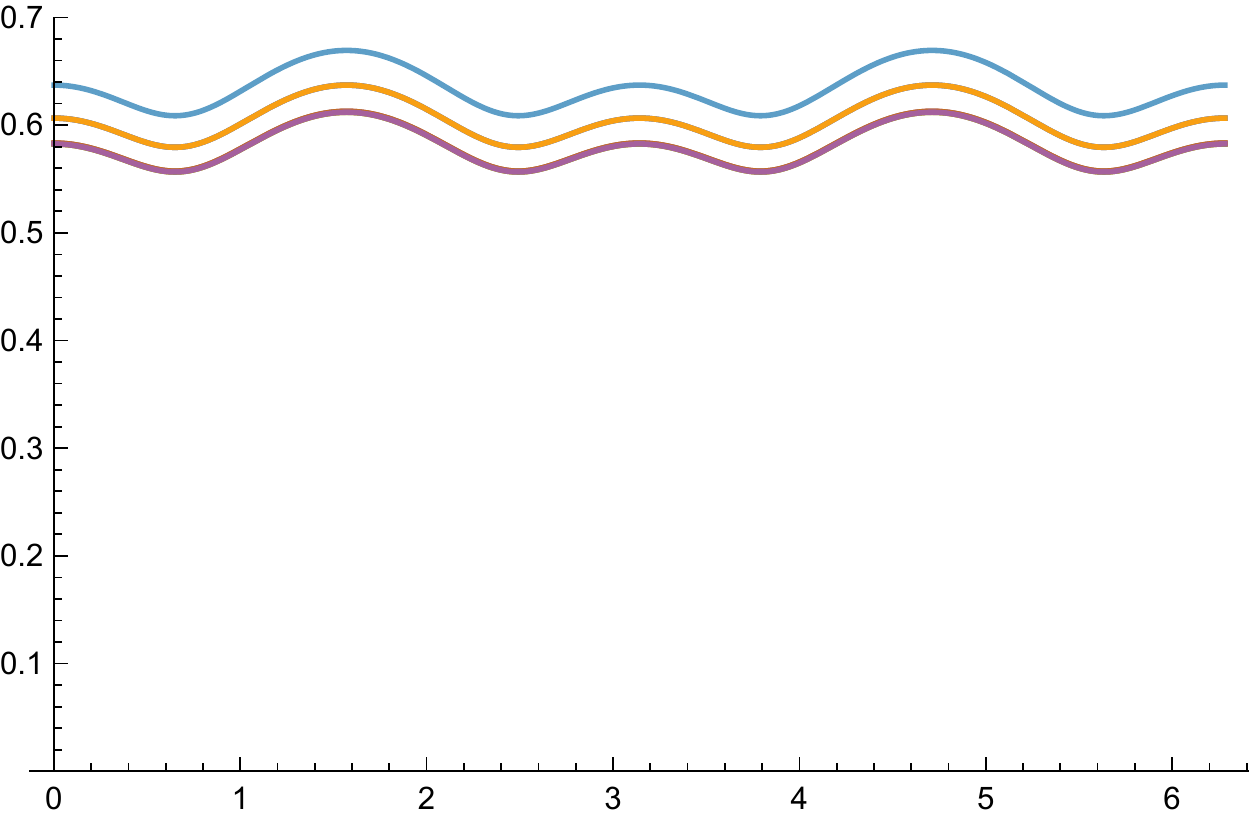} &
      \includegraphics[width=0.40\textwidth]{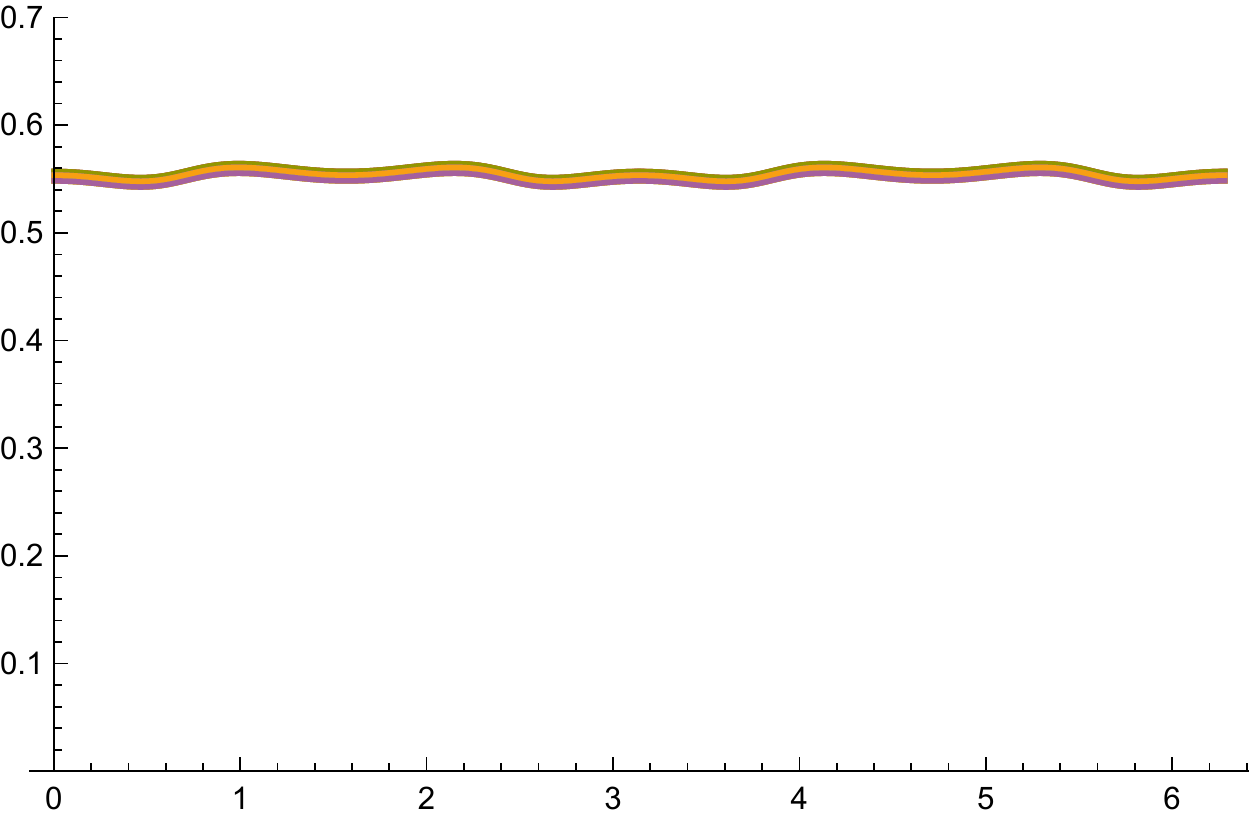} \\
   \end{tabular} 
  \end{center}
   \caption{Temporal variability of magnitude responses ${\cal Q}_{(x,y,t), \Gamma-norm}$  computed
     (left column) at the local extrema over both spatial and temporal scales of the
     $\Gamma$-normalized quasi quadrature measure ${\cal Q}_{(x,y,t), \Gamma-norm}$
     according to (\ref{eq-Q3-spat-temp-scaleest-regular-geom-sine-s-tau}) for a 2+1-D spatio-temporal sine
     wave pattern of spatial angular frequency $\omega_s = 1$ and
     temporal angular frequency $\omega_{\tau} = 1$
     (right column) at phase-compensated scale estimates
     according to (\ref{eq-Q3-spat-scaleest-phasecorr-geom-sine-s})
     and (\ref{eq-Q3-temp-scaleest-phasecorr-geom-sine-tau}).
    (Horizontal axis: time $t$)
   (The multiple graphs in each
  diagram show the variability of the scale estimates for different
   values of the complementary spatial coordinates $x = m \pi/4$
   and $y = n \pi/4$.)}
  \label{fig-Q3-sine-magn-estimates-regular-phasecomp-tvar}
\end{figure*}

\subsection{Phase-compensated magnitude estimates}

When performing temporal scale selection from the local extrema of the
scale-normalized quasi quadrature measure over scale
(\ref{eq-Q3-scalenorm-ders-Gammanorm}),
%(\ref{eq-scsel-quasi-quad-raw}),
the magnitude responses at the spatial points
$(x = n \pi/\omega_s, y = n \pi/\omega_s)$ and the temporal moments 
$t = n \pi/\omega_{\tau}$ at which only the first-order temporal
derivative responds are given by
\begin{equation}
   \label{eq-Q1-magn-sine-2+1D}
  {\cal Q}_{t, 1, \Gamma-norm} 
  = 2 (1-\Gamma_s)^{1-\Gamma_s} (1-\Gamma_{\tau})^{1-\Gamma_{\tau}}
     \, e^{\Gamma_s+\Gamma_{\tau}-2} \, \omega_s^{2 \Gamma_s} \, \omega_{\tau}^{2
   \Gamma_{\tau}},
\end{equation}
whereas the magnitude responses at the spatial points
$(x = (\pi/2 + n \pi)/\omega_s, y = (\pi/2 + n \pi)/\omega_s)$
and the temporal moments $t = (\pi/2 + n \pi)/\omega_0^2$ at which only the
second-order temporal derivative responds are given by
\begin{equation}
   \label{eq-Q2-magn-sine-2+1D}
  {\cal Q}_{t, 2, \Gamma-norm} 
  = \frac{2 (2-\Gamma_s)^{2-\Gamma_s} (2-\Gamma_{\tau})^{2-\Gamma_{\tau}}
              \, e^{\Gamma_s+\Gamma_{\tau}-4} \, \omega_s^{2 \Gamma_s} \, \omega_{\tau}^{2 \Gamma_{\tau}}}
  {\sqrt{(2-\Gamma_s)(1-\Gamma_s)} \sqrt{(2-\Gamma_{\tau})(1-\Gamma_{\tau})}}.
\end{equation}
Thus, the magnitude responses will have a certain phase dependency
because of the variability in the temporal scale estimates leading to
corresponding variability in the relative strengths of the first- {\em vs.\/}\
second-order responses (see the left columns in 
figures~\ref{fig-Q3-sine-magn-estimates-regular-phasecomp-xvar}--\ref{fig-Q3-sine-magn-estimates-regular-phasecomp-tvar}).
When performing phase compensation according to
(\ref{eq-Q3-spat-scaleest-phasecorr-geom-sine-s}) and
(\ref{eq-Q3-temp-scaleest-phasecorr-geom-sine-tau}), the temporal scale estimates
will on the other hand will be close to a temporal scale level where
the relative strengths of the first- and second-order responses are
balanced %(\ref{eq-Q3-determ-Cs-balanced-first-second-order-responses})--(\ref{eq-Q3-determ-Ctau-balanced-first-second-order-responses})
and leading to a much lower temporal variability in the magnitude
responses  (see the right columns in
figures~\ref{fig-Q3-sine-magn-estimates-regular-phasecomp-xvar}--\ref{fig-Q3-sine-magn-estimates-regular-phasecomp-tvar}).
If one additionally wants these magnitude estimates to be independent
of the wavelengths of the sine wave pattern, then this can be accomplished by
instead computing the corresponding post-normalized quasi quadrature entity
\begin{equation}
   {\cal Q}_{(x,y,t), post-norm} 
  = \hat{s}^{\Gamma_s} \, \hat{\tau}^{\Gamma_{\tau}} \, {\cal Q}_{(x,y,t), \Gamma-norm} 
\end{equation}
where $\hat{s}$ and $\hat{\tau}$ represent the phase-compensated
spatial and temporal scale estimates according to (\ref{eq-Q3-spat-scaleest-phasecorr-geom-sine-s}) and
(\ref{eq-Q3-temp-scaleest-phasecorr-geom-sine-tau}).

In these respects, the analysis in this appendix shows how the
notion of phase compensation also applies in a spatio-temporal setting
with independent variabilities in the spatial and the temporal scales
in the spatio-temporal image structures in video data.

When reduced to either a purely spatial or a purely temporal domain,
the analysis in this appendix also gives a more detailed treatment of
how the notion of scale calibration can be performed when applying
dense scale selection to either purely spatial image data or a purely
temporal signal. Specifically, the expressions
(\ref{eq-effect-of-Gammas-on-scale-at-edges}) and
(\ref{eq-effect-of-Gammatau-on-scale-at-ramps}) show how variations in
the complementary scale normalization parameters $\Gamma_s$ and
$\Gamma_{\tau}$ will influence the selection of spatial and temporal
scales at diffuse spatial edges and temporal ramps.

\clearpage

\bibliographystyle{siamplain}
\bibliography{defs,tlmac}

\end{document}